\documentclass[11pt]{article}

\usepackage[T1]{fontenc}
\usepackage[utf8]{inputenc}
\usepackage{lmodern}
\usepackage{textcomp}

\DeclareUnicodeCharacter{2013}{--}        
\DeclareUnicodeCharacter{2014}{---}       
\DeclareUnicodeCharacter{2192}{$\rightarrow$}
\DeclareUnicodeCharacter{2194}{$\leftrightarrow$}
\DeclareUnicodeCharacter{2208}{$\in$}
\DeclareUnicodeCharacter{2212}{$-$}       
\DeclareUnicodeCharacter{2248}{$\approx$}
\DeclareUnicodeCharacter{2264}{$\leq$}
\DeclareUnicodeCharacter{2265}{$\geq$}
\DeclareUnicodeCharacter{2713}{\checkmark}
\DeclareUnicodeCharacter{0394}{$\Delta$}
\DeclareUnicodeCharacter{03B1}{$\alpha$}
\DeclareUnicodeCharacter{03B2}{$\beta$}
\DeclareUnicodeCharacter{03C1}{$\rho$}
\DeclareUnicodeCharacter{0100}{\=A}        
\DeclareUnicodeCharacter{0101}{\=a}        
\DeclareUnicodeCharacter{012B}{\=\i}       
\DeclareUnicodeCharacter{2074}{$^{4}$}     
\DeclareUnicodeCharacter{2075}{$^{5}$}     
\DeclareUnicodeCharacter{2076}{$^{6}$}     
\DeclareUnicodeCharacter{2079}{$^{9}$}     
\DeclareUnicodeCharacter{207B}{$^{-}$}     

\usepackage[margin=1in]{geometry}

\usepackage{amsmath, amssymb}
\IfFileExists{amsthm.sty}{\usepackage{amsthm}}{}

\usepackage{booktabs, longtable, array, tabularx}
\usepackage{calc} 
\IfFileExists{makecell.sty}{\usepackage{makecell}}{}

\usepackage{lscape}
\IfFileExists{pdflscape.sty}{\usepackage{pdflscape}}{}

\usepackage{graphicx, float}
\graphicspath{{figures/}}

\IfFileExists{caption.sty}{%
  \usepackage{caption}%
  \captionsetup{singlelinecheck=false,labelformat=empty}%
}{}

\usepackage[colorlinks=false, hidelinks]{hyperref}
\hypersetup{
  pdftitle={Beyond Recall: Behavioral Specification as an Interpretive Layer for AI Personalization},
  pdfauthor={Aarik Gulaya},
  pdfsubject={Behavioral Specification as an Interpretive Layer for AI Personalization},
  pdfkeywords={Study repository: https://github.com/agulaya24/beyond-recall ; For AI agents: https://github.com/agulaya24/beyond-recall/blob/master/AGENTS.md ; Pipeline source: https://github.com/agulaya24/BaseLayer},
  pdfcreator={LaTeX via pandoc + arXiv preamble}}
\usepackage{url}
\usepackage[round, authoryear, sort&compress]{natbib}

\IfFileExists{listings.sty}{\usepackage{listings}}{}

\IfFileExists{framed.sty}{\usepackage{framed}}{}

\IfFileExists{upquote.sty}{\usepackage{upquote}}{}
\IfFileExists{microtype.sty}{%
  \usepackage[]{microtype}
  \UseMicrotypeSet[protrusion]{basicmath}%
}{}
\makeatletter
\@ifundefined{KOMAClassName}{%
  \IfFileExists{parskip.sty}{\usepackage{parskip}}{%
    \setlength{\parindent}{0pt}%
    \setlength{\parskip}{6pt plus 2pt minus 1pt}}%
}{\KOMAoptions{parskip=half}}
\makeatother
\usepackage{xcolor}
\IfFileExists{xurl.sty}{\usepackage{xurl}}{}
\usepackage{etoolbox}
\makeatletter
\patchcmd\longtable{\par}{\if@noskipsec\mbox{}\fi\par}{}{}
\makeatother
\IfFileExists{footnotehyper.sty}{\usepackage{footnotehyper}}{\usepackage{footnote}}
\makesavenoteenv{longtable}
\makeatletter
\def\maxwidth{\ifdim\Gin@nat@width>\linewidth\linewidth\else\Gin@nat@width\fi}
\def\maxheight{\ifdim\Gin@nat@height>\textheight\textheight\else\Gin@nat@height\fi}
\makeatother
\setkeys{Gin}{width=\maxwidth,height=\maxheight,keepaspectratio}
\makeatletter
\def\fps@figure{htbp}
\makeatother
\providecommand{\tightlist}{%
  \setlength{\itemsep}{0pt}\setlength{\parskip}{0pt}}

\title{Beyond Recall: Behavioral Specification as an Interpretive Layer for AI Personalization}

\begin{document}

\thispagestyle{empty}

\definecolor{titleblue}{HTML}{1F4E8C}
\begin{center}
  {\LARGE\bfseries\color{titleblue} Beyond Recall: Behavioral Specification\\[2pt]
  as an Interpretive Layer for AI Personalization\par}
\end{center}

\vspace{1.2em}

\begin{abstract}
\normalsize
\noindent If an AI agent makes more decisions on a person's behalf, those decisions must align with its user. We introduce \textbf{representational accuracy} to measure how faithfully a system captures a person's interpretation. An \textbf{interpretive layer} is operationalized as a \textbf{Behavioral Specification}. Our reference implementation aggressively compresses a person's data into interpretive patterns, served as context to a language model. We evaluate the Specification on a prototype benchmark of held-out behavioral predictions scored by a calibrated 5-judge LLM panel. We test it independently and in composition with a range of context conditions: full raw corpus, full extracted facts, and four commercial memory systems (Mem0, Letta, Supermemory, Zep).

\medskip
\noindent Across 14 public-domain autobiographical corpora, the Specification lifts representational accuracy in aggregate and nearly eliminates model hedging. It recovers most of what the raw corpus delivers, at \textasciitilde25$\times$ less context cost. Lift grows as the model's pretraining coverage of the subject decreases, suggesting the population of relevance is anyone not adequately represented in pretraining. Lift is greatest on interpretation-required questions, where providing an interpretive layer enables model behavior that extracted facts or raw corpus do not. Conversely, on recall-required questions, this layer can interfere rather than help.

\medskip
\noindent We conclude that representational accuracy is distinct from recall and that human-AI alignment is dependent on how accurately the user is represented. Representational accuracy makes that alignment testable.
\end{abstract}

\vspace{0.4em}

\begin{quote}
\raggedright
Study repository: \href{https://github.com/agulaya24/beyond-recall}{\nolinkurl{github.com/agulaya24/beyond-recall}}\\[4pt]
For AI agents: \href{https://github.com/agulaya24/beyond-recall/blob/master/AGENTS.md}{\nolinkurl{github.com/agulaya24/beyond-recall/blob/master/AGENTS.md}} (repository is fully indexed and vectorized for semantic search).\\[4pt]
Pipeline source: \href{https://github.com/agulaya24/BaseLayer}{\nolinkurl{github.com/agulaya24/BaseLayer}}
\end{quote}

\vspace{1.3em}
\begin{center}
\rule{0.5\textwidth}{0.4pt}
\end{center}
\vspace{1em}

\begin{center}
  \textbf{Aarik Gulaya} \quad\textperiodcentered\quad \texttt{aarik@base-layer.ai}\\[4pt]
  Preprint \quad\textperiodcentered\quad 2026-05-14 \quad\textperiodcentered\quad Manuscript CC-BY-4.0 \quad\textperiodcentered\quad Code Apache 2.0
\end{center}

\vfill

\begin{flushright}
{\footnotesize ORCID: \texttt{0009-0009-5902-9557}}
\end{flushright}

\clearpage

\setcounter{tocdepth}{2}
\tableofcontents
\clearpage

\hypertarget{introduction}{%
\section{1. Introduction}\label{introduction}}

This paper proceeds along a single path. \hyperref[recall-is-not-interpretation.-interpretation-can-be-measured.]{§1.1} states the problem: AI
memory optimizes for recall, and recall is not interpretation. \hyperref[what-we-tested]{§1.2}
defines the measurement target, representational accuracy, and the
held-out behavioral-prediction test that makes it measurable. \hyperref[study-design]{§3}
describes the artifact under test, a Behavioral Specification, and the
conditions it is evaluated against. \hyperref[results]{§4} reports what the experiment
found; \hyperref[discussion]{§5} develops what it implies for AI personalization. A reader who
wants the result before the machinery can read \hyperref[what-we-found]{§1.3} first.

\hypertarget{recall-is-not-interpretation.-interpretation-can-be-measured.}{%
\subsection{1.1 Recall is not interpretation. Interpretation can be
measured.}\label{recall-is-not-interpretation.-interpretation-can-be-measured.}}

AI is moving from a tool a person uses to an agent that acts on a
person's behalf, and that shift changes what ``memory'' must do for a
specific individual. State of the art AI memory has been optimizing for
recall as the success metric. The four prominent commercial systems
(Zep, Letta, Mem0, and Supermemory) compete on standard recall
benchmarks such as LOCOMO and LongMemEval, reporting accuracies in
roughly the 70\% to 93\% range depending on provider, model, and
benchmark variant (\hyperref[memory-systems-for-llm-agents]{§2.2}). Optimizing further on recall leaves something
more fundamental unmeasured. This paper examines how recall is one part
of memory, and how the function of memory is dictated by how an
individual processes the facts and experiences of their life.

We use \textbf{interpretation} to refer to this human-side property: the
way a specific person processes facts and experiences into judgments,
decisions, and reactions. Viewing situations from different lenses can
lead to entirely different interpretations of the same set of facts.
This principle holds across domains: the same set of facts can produce
different conclusions depending on the interpretive framework the reader
brings to them. Memory is therefore personal in a deeper sense than
recall: the same facts arrange differently inside different people. For
an AI to serve a specific person, it must be given context on the
framework that person uses to reason, not just the raw facts or
information itself. Throughout this paper we use the term
\textbf{Behavioral Specification} to refer to a static document that
extracts and encodes a person's behavioral patterns; the operational
definition is developed across \hyperref[pipeline-for-the-behavioral-specification]{§3.7}. A Behavioral Specification is an
artifact that captures this interpretive framework, and is provided to
an AI as context.

We introduce \textbf{representational accuracy} as the corresponding
AI-side property: how well a system's internal model of a specific
person captures their interpretive patterns. It is not recall,
preference matching, or persona consistency. It is a distinct property
of the AI system, and state of the art memory benchmarks do not isolate
it. Prior work closest to this axis (Twin-2K for scaled behavioral
prediction, PersonaGym for persona fidelity, AlpsBench for preference
alignment) measures related properties but not the transfer of a
person's interpretive patterns to new situations the system has never
seen. \hyperref[prior-measurement-targets-and-the-gap-representational-accuracy-fills]{§2.1} positions each benchmark against what this paper measures,
and Appendix F develops the scope differences in detail.

\textbf{The core hypothesis of this research is that representational
accuracy of a person's interpretation improves an AI system's behavioral
alignment with that person.} This is the operational primitive for any
AI system meant to act on a person's behalf: the system's behavior can
only match the user's reasoning to the extent the system represents that
reasoning accurately. The operational test in this paper is behavioral
prediction on held-out situations: given a situation drawn from text the
model has never seen, the model generates how the subject would respond;
the response is scored by a panel of calibrated large language model
(LLM) judges against the subject's own verbatim response in the held-out
text on a 1-5 interpretive rubric (\hyperref[scoring-rubric-with-calibrated-llm-judge-panel]{§3.3}). Accurate prediction on
held-out text is evidence that the representation captures the subject's
recurring patterns of reasoning, distinct from the facts and stylistic
surface that current extraction pipelines already produce. The design
also reduces the risk of sycophancy\footnote{Sycophancy: a model
  adjusting its answer to match what the user appears to want, often by
  agreeing with or flattering them. See \citet{sharma2023sycophancy}, Perez et
  al.~2022, Jain et al.~2025 (\hyperref[cognitive-and-representational-foundations]{§2.4}).}: the answer is checked against the
person's narrative, which the model has never seen, not against anything
the user says during the conversation. The held-out test is one
operationalization of the hypothesis.

We test this hypothesis on the leading state-of-the-art AI memory
systems and on a diverse set of 14 autobiographies from authors across
the world. For this initial examination we use baselined and calibrated
LLM judges to evaluate the performance of each memory system, on its own
and in combination with a \textbf{Behavioral Specification}: a static
document that extracts and encodes a stable representation of a corpus's
behavioral patterns. The Specification captures the recurring patterns
in how the subject reasons, drawn from the shape of judgments and
reactions across the corpus (for example: \emph{``spiritual integrity
over social cost\ldots{}''}, \emph{``reform through love\ldots{}''},
\emph{``hierarchical deference\ldots{}''}). A walked example of the
audit chain from such a pattern back to its grounding facts and source
passages appears in \hyperref[traceability-and-reasoning-traces]{§2.3}.

Defined terms used throughout the paper are collected in
\protect\hyperlink{appendix-h.-glossary}{\textbf{Appendix H}} for
reference.

\hypertarget{what-we-tested}{%
\subsection{1.2 What we tested}\label{what-we-tested}}

We tested a Behavioral Specification across 14 historical subjects, each
with a public domain autobiography sourced from Project Gutenberg or the
Internet Archive (per-subject sources in \hyperref[subjects]{§3.4} Table 3.1). For every
subject we split the source corpus in half: the training half was used
to generate the Specification, to seed each memory system, and to
provide the retrievable fact pool. The held-out half was used only to
produce behavioral prediction questions and was never shown to the
\textbf{response model}, the language model being asked to predict how
the subject would respond. The Behavioral Specification is one of the
context conditions the response model receives; the full set of
conditions is defined in \hyperref[experimental-conditions]{§3.2}. The set of held-out questions for each
subject is the \textbf{question battery} (size and composition per
subject in \hyperref[question-battery-formation]{§3.5}). The test was whether each system, under each tested
condition, could predict how that specific person would respond in
situations drawn from text it had never seen. The evaluation is a
prototype benchmark for representational accuracy, not a finished one;
\hyperref[future-work]{§7} flags the work needed to harden it into a standardized instrument.

The Behavioral Specification itself is built from the training-half
corpus through an extraction-and-authoring pipeline (\hyperref[pipeline-for-the-behavioral-specification]{§3.7}). The pipeline
distills the recurring patterns of how the subject reasons into a single
structured document, typically around 7,000 tokens (\textasciitilde5,000
words) long. That document is what the response model receives as
context when asked to predict how the subject would respond.

\textbf{Hypotheses.} The study tests five claims about how a
representation of a person shapes AI behavior on that person's behalf:

\begin{itemize}
\tightlist
\item
  \textbf{H1.} A response model given a Behavioral Specification
  produces responses that align with the person's documented behavior
  more closely than the same model given no context, facts retrieved by
  a memory system, the full extracted fact list, or the raw source
  corpus (\hyperref[the-cross-subject-gradient-and-its-per-question-mechanism]{§4.1}).
\item
  \textbf{H2.} The Specification's benefit is inversely proportional to
  the response model's pretraining coverage of the person. Its effect is
  largest on people the model does not already know (\hyperref[the-cross-subject-gradient-and-its-per-question-mechanism]{§4.1}).
\item
  \textbf{H3.} The benefit comes from the content of the correct
  Specification for the correct person, not from the mere presence of a
  structured prompt. A random other person's specification, applied in
  its place, produces a substantially smaller and content-specific
  effect than the matched Specification (\hyperref[mechanism-correct-content-not-format]{§4.3}).
\item
  \textbf{H4.} The Specification interacts with memory-system retrieval
  in a structured way that depends on the type of question being asked.
  Aggregate effects on each memory system reflect the balance of these
  per-question patterns and shift with retrieval architecture (\hyperref[memory-system-composition]{§4.4}).
\item
  \textbf{H5.} A Behavioral Specification's quality advantage is also a
  compression advantage: a \textasciitilde7,000-token
  (\textasciitilde5,000-word) Specification recovers most of the
  predictive accuracy of an 80-400K-token (\textasciitilde60-300K-word)
  raw corpus (\hyperref[compression-structure-vs.-raw-text]{§4.2}).
\end{itemize}

Post-hoc analyses surfaced during the work are reported alongside these
results.\footnote{Post-hoc analyses reported alongside H1--H5:
  retrieval-overlap divergence (\hyperref[cross-system-retrieval-providers-do-not-converge]{§4.4.1}, \hyperref[retrieval-overlap-sensitivity-semantic-similarity-matching-k-variation]{§4.6.6}), Letta case study (\hyperref[exploratory-case-study-letta-stateful-agent-n3-post-hoc]{§4.5};
  Appendix G), abstention-credit validity audit (\hyperref[rubric-handling-limitations-post-hoc-validity-audit]{§3.3.6}), per-subject
  wrong-Spec heterogeneity (\hyperref[wrong-spec-derangement-protocol-sensitivity]{§4.6.5}). Labeled exploratory where they
  appear; full breakdown in Appendix B.10.}

\textbf{Primary and secondary outcomes.} The \textbf{primary outcome} is
the mean prediction score on the 1-5 rubric across a 5-judge primary
panel (\hyperref[scoring-rubric-with-calibrated-llm-judge-panel]{§3.3}).\footnote{Aggregation rule (the ``locked rule'' referenced
  throughout): per-question scores within a (subject, condition) cell
  are first averaged across the 5 judges, then per-subject means are
  aggregated across the 14 subjects. Full mechanics in \hyperref[aggregation-and-statistical-analysis-plan]{§3.3.5}.}
Cross-subject claims are calculated subject-by-subject before averaging,
so they are not driven by subjects with larger question batteries. As a
\textbf{secondary outcome}, we report the per-question
\textbf{improvement rate}: how often a context condition helps relative
to the comparison baseline (\hyperref[per-question-improvement-rate]{§4.2.1}), not just by how much it helps when
averaged. The per-question secondary outcome is informative because each
context condition behaves differently across question types: aggregate
effects reflect the balance of interpretation-heavy items (where the
Specification lifts most) and literal-recall items (where retrieval
already suffices). The formal proposal and failure-mode analysis for the
secondary outcome are in \hyperref[per-question-improvement-rate]{§4.2.1}; full operational details for both
outcomes are in \hyperref[scoring-rubric-with-calibrated-llm-judge-panel]{§3.3}.

Each memory system is tested in both a controlled configuration
(identical pre-extracted fact pool) and a native configuration (the
provider's own ingestion pipeline); design detail in \hyperref[experimental-conditions]{§3.2}. Running in
parallel across both is the Behavioral Specification, tested alone and
layered on top of each configuration. Every meaningful combination of
inputs is evaluated as its own condition:

\begin{longtable}[]{@{}
  >{\raggedright\arraybackslash}p{(\columnwidth - 6\tabcolsep) * \real{0.25}}
  >{\raggedright\arraybackslash}p{(\columnwidth - 6\tabcolsep) * \real{0.15}}
  >{\raggedright\arraybackslash}p{(\columnwidth - 6\tabcolsep) * \real{0.30}}
  >{\raggedright\arraybackslash}p{(\columnwidth - 6\tabcolsep) * \real{0.30}}@{}}
\toprule
Group & Condition & Inputs given to the model & Purpose \\
\midrule
\endhead
Direct context manipulations & \textbf{No context} (C5) & Nothing. The
model answers from pretraining alone. & Pretraining baseline. Measures
what the model already knows about the subject from public sources. \\
& \textbf{Specification alone} (C2a) & The Behavioral Specification,
with no retrieval, no facts, and no corpus. & Tests whether structure
without retrieval is sufficient on its own. \\
& \textbf{Wrong-specification control} (C2c) & A different subject's
specification applied to this subject. Two variants: an adversarial
fixed pairing (v1) and a random derangement (v2). & Tests whether the
effect is driven by the content of the correct Specification, or by the
mere presence of structured prompting. \\
& \textbf{All facts, no specification} (C4) & Every extracted fact for
the subject, loaded into context at once. & Tests whether information
sufficiency alone drives prediction, independent of structure. \\
& \textbf{Facts + specification} (C4a) & Every extracted fact plus the
Specification. & Combines full information and structure to test the
upper bound of context-provided prediction. \\
& \textbf{Raw corpus, no specification} (C8) & The full training-half
corpus loaded into context. & Tests whether unstructured source text can
substitute for an interpretive representation. \\
& \textbf{Corpus + specification} (C9) & Raw training corpus plus the
Specification. & Tests whether structure is additive to unstructured
source text. \\
Memory-system configurations (controlled, all 5 systems) &
\textbf{Retrieval alone, controlled} (C1) & Top-k facts retrieved by
each memory system (Mem0, Letta, Supermemory, Zep, Base Layer) from the
shared fact pool. & Tests retrieval sufficiency, and whether providers
converge on which facts are most relevant given identical input. \\
& \textbf{Retrieval + specification, controlled} (C3) & Memory system
retrieval from the shared fact pool, plus the Specification. & Tests
whether the Specification layers cleanly on retrieval when the input is
held constant. \\
Memory-system configurations (native, 4 commercial systems) &
\textbf{Retrieval alone, native} (C1 native) & Top-k results from each
memory system's own ingestion pipeline operating over the raw training
corpus. & Real-world comparison of each memory system's full
ingestion-plus-retrieval stack. \\
& \textbf{Retrieval + specification, native} (C3 native) & Memory
system's own ingestion and retrieval, plus the Specification. & Tests
whether the Specification improves the real-world deployment of each
memory system. \\
\bottomrule
\end{longtable}

The 14 subjects span four continents and roughly two millennia of
written human experience. Ordered chronologically: Saint Augustine
(North Africa, 4th-5th c.), Bābur (Central Asia and India, 15th-16th
c.), Bernal Díaz del Castillo (Spain and Mexico, 15th-16th c.),
Benvenuto Cellini (Italy, 16th c.), Jean-Jacques Rousseau (France, 18th
c.), Olaudah Equiano (West Africa and Britain, 18th c.), Mary Seacole
(Jamaica and Britain, 19th c.), Elizabeth Keckley (United States, 19th
c.), Yung Wing (China and the United States, 19th c.), Philip Gilbert
Hamerton (Britain, 19th c.), Fukuzawa Yukichi (Japan, 19th c.), Georg
Ebers (Germany, 19th c.), Sunity Devee (India, late 19th-early 20th c.),
and Zitkala-Ša (Yankton Dakota, early 20th c.). Source corpora range
from 25,231 words (Hamerton) to 422,772 words (Bābur). Full source
references are in \hyperref[subjects]{§3.4}.

Predictions were scored on a 1-5 rubric where the integer anchors mark
categorical shifts in answer quality; the full rubric and verbatim judge
prompt are in \hyperref[scoring-rubric-with-calibrated-llm-judge-panel]{§3.3}. For example, a move from 1.8 to 2.4 crosses the 2.0
boundary: the model goes from refusing the question or producing an
off-base answer (anchor 1) to engaging with the question even when the
prediction is still wrong (anchor 2). Absolute point gains, not
percentages, are the informative metric for cross-subject comparison.

\begin{longtable}[]{@{}
  >{\raggedright\arraybackslash}p{(\columnwidth - 4\tabcolsep) * \real{0.33}}
  >{\raggedright\arraybackslash}p{(\columnwidth - 4\tabcolsep) * \real{0.33}}
  >{\raggedright\arraybackslash}p{(\columnwidth - 4\tabcolsep) * \real{0.33}}@{}}
\toprule
Score & Anchor (verbatim) & Shift from previous anchor \\
\midrule
\endhead
\textbf{1} & Refuses or off-base & (rubric floor) No usable answer: the
model either declined (93\% of score-1 responses) or engaged with the
wrong subject (7\%) \\
\textbf{2} & Wrong prediction & From no usable answer to engaging with
the right subject \\
\textbf{3} & Right domain wrong outcome & From wrong prediction to the
right domain \\
\textbf{4} & General direction correct & From right domain to the
general direction of the held-out \\
\textbf{5} & Predicts specific outcome & From general direction to the
specific outcome documented in the held-out \\
\bottomrule
\end{longtable}

Score interpretation, including the cross-anchor rule for fractional
scores (e.g., 2.5, 3.4), is in \hyperref[score-interpretation]{§3.3.1}. Example questions per subject and
panel composition are in \hyperref[judge-panel]{§3.3.2}.

The \textbf{baseline} we refer to throughout is the No-Context Baseline
(C5): the response model's score with no external information.
\textbf{Low-baseline} subjects are the \textbf{population of relevance}:
people the model has insignificant pretraining understanding of, even
when fragments of their digital footprint exist in training data.
\textbf{High-baseline} subjects are the opposite, people the model
already knows about from pretraining (e.g., Benjamin Franklin, included
in this study as a known-figure reference, \hyperref[the-gradient-at-the-high-baseline-end-franklin-reference]{§4.1.2}). Of the 14 main-study
subjects, 9 are low-baseline (C5 ≤ 2.0) and 5 are mid-baseline (2.0
\textless{} C5 ≤ 3.0); Franklin is the single high-baseline reference
(C5 \textgreater{} 3.0) and is not part of the 14. The low-baseline band
is plausibly the default case for most users of frontier systems: even
people with substantial public output captured in training corpora have
only fragments of their reasoning represented. The study suggests the
low-baseline band is the norm rather than the exception.\footnote{Operational
  thresholds: low-baseline C5 ≤ 2.0 on the 1-5 rubric, high-baseline C5
  \textgreater{} 3.0. Full distribution and band assignments in \hyperref[pretraining-coverage-variance-high-vs-low-baseline]{§3.4.1}.}
Results are reported separately on the low-baseline band (n=9) alongside
the full 14-subject analysis.

The study is structured into two tiers. Tier 1 (main study) uses Claude
Haiku 4.5 as the response model across all 14 subjects on every
condition. Tier 2 is a smaller cross-provider directional probe (\hyperref[response-models]{§3.6},
\hyperref[cross-provider-response-generation-tier-2-replication]{§4.6.1}). The 7-judge panel spans three providers; the 5-judge primary
aggregate is composed of the Anthropic (Claude) and OpenAI (GPT) judges,
and the 2 Gemini judges are reported as a sensitivity check on
calibration grounds (\hyperref[calibration]{§3.3.3}).

Together these hypotheses test whether a Behavioral Specification can
increase a language model's representational accuracy of a specific
person.

\hypertarget{what-we-found}{%
\subsection{1.3 What we found}\label{what-we-found}}

An interpretive layer, operationalized as a Behavioral Specification
(Spec), lifts representational accuracy. Its benefit is largest where
the model knows the person least, and the mechanism is per-question. On
questions where the model needs an interpretive frame and lacks one, a
Behavioral Specification categorically improves the answer produced. On
questions where the model already has the answer, a Behavioral
Specification adds nothing and sometimes hurts.\footnote{``Low
  baseline'' means C5 ≤ 2.0 on the 1-5 rubric; \hyperref[what-this-implies]{§1.4} and \hyperref[why-the-gradient-is-the-load-bearing-finding]{§5.2} develop why
  this band is the population of importance for AI personalization.}
What follows are seven findings, beginning with the
cross-subject gradient (primary outcome) and the per-question mechanism
beneath it. Prediction on held-out reasoning is what we measure; whether
higher representational accuracy translates into aligned action is the
downstream claim developed in \hyperref[discussion]{§5} and \hyperref[future-work]{§7}.

\textbf{Headline findings.}

\begin{itemize}
\tightlist
\item
  \textbf{Gradient.} \textbf{A Behavioral Specification's benefit is
  largest where the model knows the person least.} Every one of the 9
  low-baseline subjects improved when the Behavioral Specification was
  added on top of All Facts (C4a); per-subject mean lift \textbf{+0.89}
  points on the 1-5 rubric over the No-Context Baseline (n=9), with
  \textbf{78.6\%} of individual questions improving (351 paired
  questions).\footnote{Wilcoxon signed-rank \emph{W} = 11, \emph{N} =
    14, \emph{p} = 0.007. The \textbf{+0.89} is the cross-subject mean
    of per-subject Δ\_C4a (locked aggregation rule, \hyperref[what-we-tested]{§1.2}); the
    grand-mean alternative is +0.93. Full regression, the
    leveler-framing of the gradient, and the aggregation reconciliation
    are in \hyperref[the-cross-subject-gradient-and-its-per-question-mechanism]{§4.1} and Appendix B.9.}\footnote{The +0.89 mean lift is
    conservative against the \hyperref[rubric-handling-limitations-post-hoc-validity-audit-1]{§4.6.7} rubric-handling audit: residual
    refusal-anchor ambiguity pulls baseline scores upward more than
    post-Spec scores, so the true Δ\_C4a is likely modestly larger than
    +0.89.} The gradient comes from the
  per-question mechanism described next: low-baseline subjects have more
  questions where the model lacks an interpretive frame, and therefore
  more questions where the Behavioral Specification lifts. Detail in
  \hyperref[the-cross-subject-gradient-and-its-per-question-mechanism]{§4.1}.
\item
  \textbf{Per-question interpretive lift.} \textbf{The Behavioral
  Specification moves 55\% of low-baseline questions across at least one
  rubric anchor upward; 18\% cross two or more.} Crossing one
  rubric anchor moves a response from ``wrong prediction'' to ``general
  direction correct.'' Crossing two or more anchors is a bigger jump: a
  single question where the model moves from refusal or generic guessing
  to a recognizable, person-specific response. 5.7\% cross three or more
  anchors (20 of 351 paired low-baseline questions). The pattern holds
  across Spec Only (C2a), All Facts + Spec (C4a), and Corpus + Spec (C9)
  conditions on the low-baseline subjects. Detail in \hyperref[the-cross-subject-gradient-and-its-per-question-mechanism]{§4.1}, \hyperref[compression-structure-vs.-raw-text]{§4.2}.
\item
  \textbf{Compression.} \textbf{The Behavioral Specification recovers
  75\% of what the raw corpus delivers at \textasciitilde25× less
  context on average (per-subject range 7× to 79×).} A 7,000-token
  structured representation matches most of the predictive accuracy of
  a 163,000-token raw corpus. A Behavioral Specification selects and
  structures the behavioral signal; the interpretive layer drives the
  result, not the volume of context. Spec Only +0.68 vs.~raw corpus
  +0.91 over baseline.\footnote{The +0.68 and +0.91 figures are 9-row
    low-baseline-band per-subject-mean Δs. Earlier drafts reported +0.71
    and +0.93 on an asymmetric 8-row baseline; v12.1 standardizes to the
    symmetric 9-row computation. Full recompute, both grains, and the
    Bābur C9 carve-out are in Appendix B.9.} On Hamerton (smallest
  corpus tested), the
  Behavioral Specification scores higher than the raw corpus (2.63
  vs.~2.27). Detail in \hyperref[compression-structure-vs.-raw-text]{§4.2}.
\item
  \textbf{Content specificity.} \textbf{A wrong Behavioral Specification
  drops accuracy below the No-Context Baseline (Δ = −0.25); the correct
  Behavioral Specification lifts accuracy above (Δ = +0.35).} What
  produces the lift is the content of the correct
  Behavioral Specification for the correct person, not the presence of a
  structured prompt. Random pairings (a wrong Spec assigned by chance to
  a different subject) sometimes still produce predictions that align
  with the held-out text, suggesting some behavioral patterns transfer
  across subjects, but the correct Behavioral Specification consistently
  outperforms. Random-derangement Δ = +0.15. Detail in \hyperref[mechanism-correct-content-not-format]{§4.3}.
\item
  \textbf{Memory-system layering.} \textbf{A Behavioral Specification
  layers cleanly on top of commercial memory systems: it moves 20--36\%
  of individual questions up across a rubric anchor, excelling on
  interpretation-heavy questions and reducing refusals on questions
  where retrieved facts could not ground the model; conversely, a
  Specification negatively affects performance on recall-heavy questions
  where retrieval already supplied the answer.} This per-question
  structure is what the aggregate hides: some questions improve, some
  regress, and the balance shifts with retrieval architecture. On
  aggregate, the Behavioral Specification lifts 3 of 4 commercial memory
  systems; Mem0, Letta, and Zep show positive mean lift under at least
  one configuration, Supermemory does not. Detail in \hyperref[memory-system-composition]{§4.4}.
\item
  \textbf{Hedging reduction.} \textbf{The Behavioral Specification
  collapses baseline hedging from 41.2\% of responses to
  0.4\%.}\footnote{Headline number uses the broad rule (any
    refusal pattern anywhere in the response) under the All Facts + Spec
    (C4a) condition. The stricter rule (refusal pattern as the first
    non-whitespace text) gives 28.8\% → 0.0\% on the same condition.}
  The reduction is content-specific, not prompt-driven: under the
  wrong-Spec adversarial control, the model continues to hedge or
  explicitly flag the mismatch on 60.6\% of responses (\hyperref[mechanism-correct-content-not-format]{§4.3}). The
  pattern is consistent with the model carrying an implicit
  \emph{evidentiary bar} before committing to a behavioral prediction:
  where the matched Behavioral Specification supplies the interpretive
  scaffolding that clears the bar (combined with All Facts (C4) or
  retrieval), the model commits; where it does not, the model abstains.
  On low-baseline subjects, where the model would otherwise refuse to
  engage, the matched Behavioral Specification converts refusal into
  substantive response. This is the gradient operating at its floor.
  Detail in \hyperref[per-question-baseline-engagement-and-the-worked-rubric-example]{§4.1.1} (abstention versus non-abstention misalignment
  decomposition), \hyperref[mechanism-correct-content-not-format]{§4.3} (wrong-Spec content-specificity), and \hyperref[where-the-spec-helps-where-it-hurts-and-which-question-types-route-to-each]{§4.4.3}
  (evidentiary-bar pattern across memory systems).
\item
  \textbf{Retrieval divergence.} \textbf{Given identical input,
  memory-system providers share zero top-10 facts on 35.9\% of
  question-pairs; mean pairwise overlap is 8.3\%.} On standard
  recall benchmarks like LongMemEval and LOCOMO, the four commercial
  memory systems we tested perform within a few percentage points of
  each other. Yet on which facts to surface as most relevant, they
  substantially diverge. Convergence on top-K under identical input
  would have been evidence of a shared interpretive framework; the
  systems do not converge. On 65.6\% of (system pair, question)
  instances they share one or fewer facts. Detail in \hyperref[cross-system-retrieval-providers-do-not-converge]{§4.4.1}.
\end{itemize}

\textbf{Mechanism: three patterns of interaction with retrieval} (full
development in \hyperref[where-the-spec-helps-where-it-hurts-and-which-question-types-route-to-each]{§4.4.3}). Baseline runs suggest the model already attempts
shallow inference from a user's raw data on its own; the Specification
makes that inference inspectable and structured.

\begin{itemize}
\tightlist
\item
  \textbf{Pattern 1, Interpretation-heavy questions.} The Specification
  supplies a generalized pattern from the source that has to transfer to
  a new situation; retrieved facts alone are not enough (Fukuzawa Q26).
\item
  \textbf{Pattern 2, Literal-recall questions.} Retrieval already
  returns the plain answer; the Specification's interpretive framing
  drifts past the question and negatively impacts the response (Yung
  Wing Q5).
\item
  \textbf{Pattern 3, Refusal-triggering questions.} When the Spec
  supports refusing without enough information (not all specs do), the
  model produces principled refusals aligned with the Spec; the
  content-match rubric still scores them as off-base (Zitkala-Ša Q18).
\end{itemize}

\textbf{Robustness across providers.} We varied both the
question-battery generation model and the response model across
providers; the Spec direction reproduces. Detail in \hyperref[cross-provider-response-generation-tier-2-replication]{§4.6.1}.

\textbf{Exploratory note: Letta stateful-agent path.} Letta's
stateful-agent architecture self-edits a persistent memory block during
ingestion. Given that unique architecture, we ran an exploratory case
study on it: on 3 subjects (post-hoc), it scored above Base Layer's
full-stack Behavioral Specification at matched response model. The full
case study, the duplication audit, and the architectural-ceiling
analysis are in \hyperref[exploratory-case-study-letta-stateful-agent-n3-post-hoc]{§4.5} and Appendix G.

\hypertarget{what-this-implies}{%
\subsection{1.4 What this implies}\label{what-this-implies}}

Representational accuracy is distinct from recall, and human-AI
alignment is dependent on how accurately the user is represented. As AI
agents make more decisions on a person's behalf, the infrastructure
required to serve that alignment is user-held, portable, inspectable,
traceable, and representation-grade. A Behavioral Specification is one
implementation of this infrastructure, not the only one.

The gap it fills cannot be closed by training a larger model on more
public data. The private record does not exist in a form a training
corpus can capture; even where fragments exist, they are scattered
across formats and channels and cannot be reliably reassembled into how
a specific person reasons. AI is becoming a broadly used technology,
comparable to email or mobile phones in reach, but different in the
amount of cognition it carries out on a person's behalf. The population
of relevance (\hyperref[what-we-tested]{§1.2}) is anyone who uses or will use an AI system. Even
the subjects of the autobiographies in this study, people whose work is
in pretraining and who should technically be known to the model, score
near the rubric floor in the No-Context Baseline.

The structural options for what fills the gap are narrow:

\begin{itemize}
\tightlist
\item
  Each person supplies their own representation to whatever AI system
  serves them. A Behavioral Specification is one such implementation.
\item
  Personalization remains surface-level (style, voice, preference,
  demographic inference, observable behavior), addressing the layer
  current memory systems already cover but missing the interpretive
  framework that lets an agent act on a specific person's behalf.
\item
  AI systems infer a representation of the user from observed
  interactions, building it opaquely, without explicit input from the
  user or the ability for the user to inspect or correct it.
\end{itemize}

\hyperref[discussion]{§5} is an extended discussion of these implications; \hyperref[future-work]{§7} develops the
safety, alignment, and deployment implications.

\begin{center}\rule{0.5\linewidth}{0.5pt}\end{center}

\hypertarget{prior-work-industry-benchmarks-the-fifth-target}{%
\section{2. Prior Work, Industry Benchmarks, The Fifth
Target}\label{prior-work-industry-benchmarks-the-fifth-target}}

AI memory and personalization research today is organized around four
measurement targets: recall of stored facts, survey-response prediction,
persona fidelity, and preference alignment. Each is supported by its own
benchmark family and its own line of system design. They do not measure
whether an AI system has an accurate internal model of how a specific
person reasons. This paper proposes a fifth target,
\textbf{representational accuracy}, and uses behavioral prediction on
held-out reasoning situations as its operational test. \hyperref[prior-work-industry-benchmarks-the-fifth-target]{§2} walks the four
existing targets, names the benchmarks attached to each, and positions
the fifth alongside them.

Memory systems today optimize for recall. Recall-optimized efforts
include both \textbf{neural-memory-analogue systems}\footnote{Architectures
  that borrow from human memory engineering: episodic consolidation,
  working-memory slots, retrieval over embeddings.} and the broader
class of vector-retrieval and embeddings-based commercial memory
providers (Mem0, Zep, Supermemory, Letta). These systems do store and
retrieve information for a specific user, but they are designed and
benchmarked for recall accuracy on standard benchmarks, not how
accurately the system represents that user's reasoning. The optimization
target is general by construction; any individual user's interpretation
is not what these systems are measured against. A separate body of
research, \textbf{cognitive-representation research}, studies human
reasoning itself: how people form representations of others, how schemas
compress experience. The gap between these directions is the
translation: applying what we know about human reasoning to the direct
interaction between an AI system and a specific individual, and shaping
the system's internal model of that individual in a way that serves them
rather than serving an average.

Language models are trained to produce responses that are helpful on
average across a large population of users. That optimization target
produces outputs that no single user is the reference point for.
Personalization requires the opposite property: a system whose outputs
are tuned to a specific individual rather than to a population
aggregate. That kind of intentional individual-specificity, not ``bias''
in the negative sense but an explicit design target, is the missing
thread in current AI memory and human-AI interaction research.

\textbf{Personalization in this paper's sense.} ``Personalization'' in
current AI research typically means responsiveness to stated preferences
(dietary restrictions, communication style) or stored facts about the
user (location, occupation, history). Both are useful and both live at
the surface of the user. We use ``personalization'' in a stronger sense
throughout this paper. We mean representing the interpretive layer that
sits beneath stated preferences and biographical facts: how a specific
person organizes experience, what they treat as evidence, what reasoning
patterns they apply across new situations. Preferences and facts are
downstream artifacts of that interpretive layer; the layer itself is
what produces them. The behavioral prediction battery and Behavioral
Specification described in \hyperref[study-design]{§3} instantiate personalization in this deeper
sense, and \hyperref[discussion]{§5} returns to what this layer is and is not.

\hypertarget{prior-measurement-targets-and-the-gap-representational-accuracy-fills}{%
\subsection{2.1 Prior measurement targets and the gap representational
accuracy
fills}\label{prior-measurement-targets-and-the-gap-representational-accuracy-fills}}

This subsection walks each of the four existing targets, naming their
attached benchmarks and scopes. Representational accuracy is positioned
as the fifth target at the end of the walk. An extended
benchmark-by-benchmark analysis is in Appendix F.

\textbf{Recall measures retrievability of facts, not reasoning about
them.} LOCOMO \citep{maharana2024locomo} measures
conversational-memory quality: after a multi-session conversation, the
system is asked questions like ``what did the user say about their job
on day 3?'' and scored on fact retrieval. LongMemEval \citep{wu2025longmemeval} measures long-term memory across multiple
sessions on five capability dimensions (single-session, multi-session
reasoning, temporal reasoning, knowledge updates, abstention) and is
heavily recall-weighted. A system can saturate recall on such benchmarks
and still fail behavioral prediction, because retrieval answers the
question ``can the fact be found'' rather than ``does the system know
how the person reasons about the fact.'' Recall is a necessary property
for most downstream uses of memory but it is not sufficient for
representational accuracy.

\textbf{Survey-response prediction infers how a person would answer one
questionnaire item from how they answered others.} Twin-2K \citep{toubia2025twin2k} does this for 2,058 participants on a
17-task heuristics-and-biases battery; items share a common format
(multiple choice, Likert scale, numeric), scored by distance-based
accuracy. Twin-2K's stated target is \emph{prediction accuracy on survey
interpolation}: the model is scored on how well it predicts a held-out
questionnaire response, not on whether it represents the underlying
reasoning that produced the response. Our target is representational
accuracy on a cross-format task: autobiographical prose input,
open-ended behavioral prediction output, rubric-based scoring against a
verbatim held-out passage. The structured-questionnaire format and the
open-ended behavioral reasoning this paper studies measure different
properties. A system could perform well on Twin-2K and not on our
battery (survey interpolation does not require modeling reasoning
transfer to new contexts), and a system could perform well on our
battery and not on Twin-2K (accurate reasoning representation does not
guarantee survey-format numerical accuracy). The two benchmarks diagnose
different properties of the same general capability.

\textbf{Persona fidelity measures whether a model stays in character
across the back-and-forth of a conversation.}\footnote{A ``turn'' is one
  round of conversation, a single exchange of one user message and one
  model reply. Persona-fidelity benchmarks score whether the model stays
  in character across many such exchanges in sequence.} PersonaGym
\citep{samuel2025personagym} scores
consistency with a described persona during conversation: given a
one-line persona (``You are a 45-year-old skeptical accountant from
Toronto''), the model is scored on whether its multi-turn replies stay
in-character, graded against a held-out criterion set. In practice the
model is checked for consistency with the persona's surface attributes
(skeptical, accountant-flavored responses; not breaking character into a
different age or profession), not for whether it reproduces how a
specific person would reason. PersonaGym's one-line descriptor is a
substantially shallower input than this paper's
\textasciitilde7,000-token Behavioral Specification or Twin-2K's
full-text survey persona;\footnote{Twin-2K's full \texttt{persona\_text}
  runs \textasciitilde32,000 tokens; the \texttt{persona\_summary} runs
  \textasciitilde3,750 tokens. Both are substantially deeper than
  PersonaGym's one-line descriptor. Full breakdown of persona-input
  depth across benchmarks in Appendix F.} consistency with it does not
require modeling that person's reasoning on new situations. PersonaGym
measures a useful property (holding voice over a conversation); fidelity
to a one-line persona is a weaker condition than representational
accuracy.

\textbf{Preference alignment measures whether responses match user
preferences.} AlpsBench \citep{xiao2026alpsbench} evaluates
whether explicit memory mechanisms improve preference-aligned and
emotionally resonant responses: after ingesting a user profile, the
model is asked conversational questions (preferences, emotional support)
and responses are scored on preference alignment and emotional resonance
rubrics, not on predictive accuracy. Their central finding, \emph{that
recall improvement does not automatically carry into preference
alignment}, is arrived at independently and is complementary to this
paper. Both papers point at the same gap from different sides: solving
for recall is insufficient for what memory is ultimately for. Preference
alignment is an outcome property (whether a response matches what the
user prefers). Representational accuracy is an upstream property
(whether the AI's internal model of the user is correct). Preference
alignment is one downstream consequence of representational accuracy
being correct; it is not the same property.

\textbf{We propose behavioral prediction on held-out reasoning
situations as a test of a fifth target: representational accuracy.}

\textbf{Prediction is the test, not the goal.} We do not pursue
prediction accuracy as an end in itself. The target is representational
accuracy, the fidelity of an AI's internal model of a specific person,
and behavioral prediction on unseen situations is the instrument we use
to measure it. A prediction score tells us the representation captured
something that generalizes to new situations; a low score tells us it
did not. Prediction is a diagnostic; the Behavioral Specification is
what this paper is testing.

\textbf{The held-out design rests on a stability premise.} A person's
interpretive patterns must be stable enough within their own corpus that
what is captured from one half references what appears in the other.
Without that, held-out behavioral prediction is impossible in principle,
regardless of how good the representation is. The 14 main-study subjects
have coherent autobiographical narratives consistent with the premise;
\hyperref[the-cross-subject-gradient-and-its-per-question-mechanism]{§4.1} reports that the Behavioral Specification authored from training
text generalizes to held-out text at above-baseline rates. Subjects
whose reasoning shifts substantially across their corpus (across a major
career change, a profound life event, or a decades-long corpus with
distinct epochs) may not be well-represented by a single snapshot
specification, which is one reason temporality is a flagged follow-up in
\protect\hyperlink{future-work}{§7}.

\textbf{The missing axis is representational accuracy itself.} Each
existing benchmark family measures a real property of memory systems,
and each is useful for its own target. What is missing is an axis that
measures how accurately the memory system represents the person whose
behavior it is meant to anticipate. This paper's approach is a prototype
answer on that axis, not a finished benchmark.
\protect\hyperlink{future-work}{§7} flags a differentiated rubric (one
that separates interpretation-heavy from literal-recall questions, and
scores epistemic honesty as its own dimension) as the priority follow-up
for turning this prototype into a standardized benchmark.

\textbf{A single number does not capture a memory system's full
capability.} Recall, survey-response prediction, persona fidelity,
preference alignment, and representational accuracy are distinct axes. A
system that saturates one may do nothing on another. Production-grade
evaluation of memory systems should report results on multiple axes
rather than on any single one.

\hypertarget{memory-systems-for-llm-agents}{%
\subsection{2.2 Memory systems for LLM
agents}\label{memory-systems-for-llm-agents}}

The four commercial memory systems we evaluate (Mem0, Letta,
Supermemory, Zep) have converged on a shared set of capabilities:
semantic retrieval over embedded content, source attribution,
multi-level memory structures, and benchmark-validated recall
performance. They differ in how each of these is architected. None
positions representational accuracy or behavioral prediction of a
specific individual as a design target.

\textbf{Table 2.1. Memory system comparison.} Verified against primary
sources.

\begingroup\footnotesize
\begin{longtable}[]{@{}
  >{\raggedright\arraybackslash}p{(\columnwidth - 8\tabcolsep) * \real{0.10}}
  >{\raggedright\arraybackslash}p{(\columnwidth - 8\tabcolsep) * \real{0.30}}
  >{\raggedright\arraybackslash}p{(\columnwidth - 8\tabcolsep) * \real{0.25}}
  >{\raggedright\arraybackslash}p{(\columnwidth - 8\tabcolsep) * \real{0.20}}
  >{\raggedright\arraybackslash}p{(\columnwidth - 8\tabcolsep) * \real{0.15}}@{}}
\toprule
Provider & Core architecture & Retrieval method & Memory types &
Published recall score \\
\midrule
\endhead
\textbf{Mem0} & Extract → consolidate → retrieve pipeline; Mem0g graph
variant adds a directed labeled knowledge graph alongside the vector
store & Hybrid: semantic + keyword + entity & Conversation, session,
user, organizational & 91.6 LOCOMO, 93.4 LongMemEval (current
algorithm)\footnote{Vendor-reported; evaluation harness open-sourced at
  \texttt{github.\allowbreak{}com/\allowbreak{}mem0ai/\allowbreak{}memory-benchmarks}. The peer-reviewable
  paper \citep{chhikara2025mem0} reports 68.44 LOCOMO for the
  Mem0g variant with GPT-4o-mini.} \\
\textbf{Letta / MemGPT} & LLM-as-operating-system; virtual context
management with main context plus external context & Archival via
\texttt{archival\_memory\_search}; main-context memory blocks
self-edited via \texttt{core\_memory\_append},
\texttt{core\_memory\_replace} & \texttt{persona} and \texttt{human}
blocks in main context; archival and recall memory external & 74.0\% on
LOCOMO with GPT-4o-mini\footnote{Letta blog, 2025-08-12
  (\texttt{https:/\allowbreak{}/\allowbreak{}www.\allowbreak{}letta.\allowbreak{}com/\allowbreak{}blog/\allowbreak{}benchmarking-ai-agent-memory}).} \\
\textbf{Supermemory} & Five-component architecture: chunk-based
ingestion, relational versioning, temporal grounding, hybrid search,
session-based ingestion & Hybrid with reranking and query rewriting;
source chunks injected at retrieval & Contextual memories, relational
versions, session data & 81.6\% / 84.6\% / 85.2\% on LongMemEval\_s with
GPT-4o / GPT-5 / Gemini-3-Pro (self-reported) \\
\textbf{Zep} & Built on Graphiti (Apache 2.0, open source). Bi-temporal
knowledge graph & Hybrid: semantic + BM25 + graph traversal & Episodes
(ground-truth source), Entities, Facts-as-triplets with temporal
validity windows & 71.2\% on LongMemEval with GPT-4o\footnote{Rasmussen
  et al.~\citep{rasmussen2025zep}.} \\
\bottomrule
\end{longtable}
\endgroup

All four systems report recall scores in the 70-93\% range; on the
standard recall benchmarks, scores are localizing in the
\textgreater80\% range and recall is no longer as much a frontier issue
as it used to be.\footnote{The vendor-reported recall scores in this
  table are contested across providers and third-party reproductions;
  methodology varies significantly between evaluators. This paper does
  not adjudicate vendor scores; we measure on a different axis (\hyperref[memory-system-composition]{§4.4}).}
All four are sophisticated systems that solve real problems in memory
management. They optimize for storing, organizing, and retrieving what a
person said or did.

Of the four systems, Letta \citep{packer2023memgpt} is
architecturally distinct: it is the only one whose core architecture
treats memory as something an agent \emph{synthesizes} during
conversation rather than \emph{stores} for later retrieval.\footnote{Letta's
  main context holds structured memory blocks the agent edits during its
  inference loop; external context includes archival and recall memory.
  Full architectural detail in Appendix G. The other three systems
  (Mem0, Supermemory, Zep) follow extract-and-retrieve patterns
  characterized in Table 2.1.} This stateful-agent design is examined
separately as a post-hoc case study in \hyperref[exploratory-case-study-letta-stateful-agent-n3-post-hoc]{§4.5} (full case study in Appendix
G), distinct from the archival-retrieval path Letta exposes for the
main-study conditions. A Behavioral Specification targets the
interpretive layer that sits above retrieval, which three of the four
(Mem0, Supermemory, Zep) do not model at all, and which the fourth
(Letta) models implicitly through agent-initiated memory editing that
our main-study configuration did not exercise (see \hyperref[mechanism-correct-content-not-format]{§4.3} and \hyperref[exploratory-case-study-letta-stateful-agent-n3-post-hoc]{§4.5}).

\hypertarget{traceability-and-reasoning-traces}{%
\subsection{2.3 Traceability and reasoning
traces}\label{traceability-and-reasoning-traces}}

Traceability operates at two levels. \textbf{Fact-level traceability}
answers where a retrieved claim came from. \textbf{Reasoning-level
traceability} answers why the system believes this about this person.
The four memory systems we evaluate provide the first; representing how
a person reasons requires the second. Representational accuracy
operationalizes interpretation, and interpretation cannot be verified at
the fact level alone. A system that represents how a person reasons must
be auditable by that person, or the representation is a black box they
cannot verify.

Zep has the strongest explicit fact-level provenance of the four: every
entity and relationship traces back to the episode IDs that produced it.
Supermemory returns source chunks alongside retrieved memories. Mem0
tracks ingestion provenance through timestamps. Letta exposes agent
state and memory-block edit history rather than fact-level provenance.

The Behavioral Specification, as implemented in this study's reference
pipeline, is structured so that every claim it expresses is a piece of
reasoning, not just a piece of content. An interpretive pattern is
grounded in the facts that imply it, and each fact is grounded in the
source passage it was extracted from. Walking this chain backward gives
a user a reasoning trace: not only where a belief originated, but what
line of reasoning connects the source text to the interpretive claim.

A reasoning trace operates in four steps:

\begin{enumerate}
\def\labelenumi{\arabic{enumi}.}
\tightlist
\item
  The user reads the system's output (a prediction, an analysis, a
  stated pattern).
\item
  The user identifies the interpretive claim or phrase they want to
  audit.
\item
  They walk from that claim to the \textbf{pattern statement} in the
  system's representation of the person that licensed it.
\item
  They walk from the pattern statement to the \textbf{facts} that ground
  it, and from each fact to the \textbf{source passage} it was extracted
  from.
\end{enumerate}

The chain is bidirectional. From the response, the user walks down to
the source. From the source, they walk back up to the response. If a
fact misrepresents its source, correcting the fact propagates upward:
the pattern statement that depended on it changes, and the response that
depended on the pattern changes the next time the system is queried. A
worked example walking the chain on a single Sunity Devee question is in
Appendix D.6.

The four commercial memory systems can answer ``where is this fact
stored?'' but cannot answer ``why does the system believe this about
this person?'' --- because their architectures treat facts, not
interpretations, as the unit of storage. A representation that acts on a
person's behalf must be auditable at the interpretation level by that
person, or the representation is a black box. Reasoning-level
traceability is what makes that auditability operational.

A person should be able to inspect the system's model of them, challenge
any step in the reasoning, and correct it if it is wrong. A
fact-attribution memory system lets the person audit what the system
stores. A reasoning-trace specification lets the person audit what the
system believes. The first is a feature. The second is the minimum bar
for a representation that acts on someone's behalf.

\hypertarget{cognitive-and-representational-foundations}{%
\subsection{2.4 Cognitive and representational
foundations}\label{cognitive-and-representational-foundations}}

\textbf{Six prior research directions shaped how we designed this
paper's test.} Each motivates a specific choice about what to measure,
what to compare against, or what failure mode to expect.

\textbf{\citet{bartlett1932}} established that human memory is reconstructive
and schema-driven rather than literal playback. Reconstruction follows
the organizing structures a person has built up over time, not a record
of the original event. A Behavioral Specification is computationally
analogous: a structured compression meant to carry the signal of a
person's reasoning without storing every fact about them. We designed
the Specification with a schema-like architecture (anchors, core,
predictions) precisely so we could test whether it does the work a human
schema does: enable accurate anticipation of behavior in situations
never encountered in the source data. Our 50/50 train/held-out split is
the experimental realization of this question.

\textbf{\citet{hinton2015distilling}} showed that compressing a large neural
network into a smaller one preserves ``dark knowledge,'' the
relationships between outputs that carry more information than the
outputs themselves. This result motivates one of our central
experimental comparisons: on matched token budgets, does a compressed
interpretive artifact carry more predictive signal than the raw content
it was derived from? The Hamerton condition in \hyperref[compression-structure-vs.-raw-text]{§4.2} (4,500-token Spec
vs.~33,000-token training corpus at 2.63 vs.~2.27 on the 5-judge primary
panel) is a direct test of that question in the personal-representation
setting.

\textbf{\citet{chen2025personavectors}} show that the character a model takes on (its
``persona'') is encoded in specific directions inside the model's
internal numeric state (\emph{persona vectors}), and that those
directions can be identified, monitored, and nudged to shift the model's
behavior in predictable ways. Their approach modifies the model; ours
informs the model from outside via context. Both validate that persona
is a real, manipulable structure: one reachable through the model's
internals, the other through context. We chose the context route because
it produces a portable artifact users can own and audit, which internal
activation steering does not. This choice shows up in the experiment as
using a static response model (Haiku) served a variable context, rather
than a fine-tuned or activation-steered model.

\textbf{\citet{jiang2025knowme}} find that frontier
models achieve only \textasciitilde50\% accuracy on dynamic user
profiling tasks even with full conversation access. The paper documents
the failure empirically; our reading is that the cause is the gap
between having facts and having the interpretive structure to apply them
to new situations. Jiang's paper is the most direct existing evidence
for the gap this paper studies, and our test design inherits from it:
behavioral prediction on scenarios drawn from held-out text that the
model has not seen, with all relevant facts retrievable, measures
exactly the interpretive-application gap.

\textbf{\citet{jain2025sycophancy}} find that adding
conversation context to LLMs makes them more sycophantic: more likely to
agree with the user even when the user is wrong, more likely to adopt
the user's perspective on a question. Their result shows that context
without the right structure pushes the model toward what the user
appears to want rather than toward a grounded answer. This is why our
experiment includes a wrong-Spec control (\hyperref[what-we-found]{§1.3} Mechanism): we hand the
model a Behavioral Specification that does not match the actual subject.
If models drifted purely toward whatever context they are given, the
wrong-Spec should behave like any other structured prompt. Instead the
model flags the mismatch or attempts a low-quality application, neither
of which is sycophantic drift (per-condition rates in \hyperref[mechanism-correct-content-not-format]{§4.3}). Jain's
finding plus our wrong-Spec result bracket the question from both sides:
context shape matters (Jain), and content matters too.

\textbf{\citet{lu2026assistantaxis}} identify what they call the
Assistant Axis: a dominant internal direction that anchors assistant
models' default behavior toward generic helpfulness and harmlessness. A
Behavioral Specification can be read as an external override to the
Assistant Axis on a per-user basis: a structured anchor that shifts the
model from ``generic helpful assistant'' toward ``reasons as this
specific person would reason.'' This framing motivated our choice to
measure hedging as a primary outcome alongside accuracy: if the Spec
shifts the model off the generic Assistant Axis, the behavioral change
should show up both in what the model predicts and in what it is willing
to commit to. Our hedging-reduction finding (\hyperref[what-we-found]{§1.3} Mechanism, \hyperref[mechanism-correct-content-not-format]{§4.3}) is
consistent with this reading.

\hypertarget{study-design}{%
\section{3. Study Design}\label{study-design}}

The experimental strategy holds the response model constant and varies
the context conditions: nothing (pretraining only), retrieved facts, raw
corpus, a Behavioral Specification, or combinations of those. This lets
us evaluate each context condition for representational accuracy against
the No-Context Baseline, while isolating the contribution of the
interpretive layer across all of them, separate from model capability,
provider, or fine-tuning regime. All measurement choices are reflected
in \hyperref[results]{§4}, and the statistical commitments were pre-locked before final
analysis.

The apparatus is described in seven parts: \hyperref[operationalizing-representational-accuracy-via-the-behavioral-specification]{§3.1} establishes the property
being measured; \hyperref[experimental-conditions]{§3.2} specifies the experimental conditions; \hyperref[scoring-rubric-with-calibrated-llm-judge-panel]{§3.3} defines
the scoring rubric and the calibrated LLM judge panel; \hyperref[subjects]{§3.4} introduces
the subjects; \hyperref[question-battery-formation]{§3.5} covers the question batteries and circularity
controls; \hyperref[response-models]{§3.6} names the response models; \hyperref[pipeline-for-the-behavioral-specification]{§3.7} describes the pipeline
that produces the Behavioral Specification (the reference implementation
of the interpretive-layer artifact tested in this study).

\hypertarget{operationalizing-representational-accuracy-via-the-behavioral-specification}{%
\subsection{3.1 Operationalizing representational accuracy via the
Behavioral
Specification}\label{operationalizing-representational-accuracy-via-the-behavioral-specification}}

\textbf{Representational accuracy is the AI-side property: how
faithfully a model's internal model of a specific person captures that
person's reasoning patterns.} It is a property of the AI system, not of
any specific operationalization. Multiple routes can in principle
produce it (model fine-tuning, persona-vector steering, retrieval over
structured facts). This paper operationalizes one such route: a
\textbf{Behavioral Specification} served as context to a static response
model, tested for whether an interpretive layer can increase
representational accuracy relative to the other context conditions in
\hyperref[experimental-conditions]{§3.2}.

\textbf{The instrument we use to measure representational accuracy is
behavioral prediction on held-out situations.} Held-out passages from
the subject's autobiography serve as samples of situations the model has
not seen. The model is asked to predict how the subject would respond;
the response is scored against the verbatim held-out passage. Prediction
here is the test, not the goal: \hyperref[prior-measurement-targets-and-the-gap-representational-accuracy-fills]{§2.1} develops this distinction.

\textbf{Three things have to hold for a served Behavioral Specification
to register a positive score.} Each is a property of the Spec or the
serving step, not of representational accuracy itself:

\begin{enumerate}
\def\labelenumi{\arabic{enumi}.}
\tightlist
\item
  The person has behavioral patterns consistent enough to be captured in
  a Behavioral Specification.
\item
  The Behavioral Specification actually carries that signal.
\item
  A model given the Behavioral Specification can act on it.
\end{enumerate}

\textbf{Prediction on held-out situations tests all three at once.} When
the score is low, one of these three is failing: the patterns are not
consistent, the Spec is wrong, or the model is not using the Spec well.
Each failure mode is informative.

We do not claim to modify the model's internal parameters. Each
condition in \hyperref[experimental-conditions]{§3.2} varies what is served to the model at inference time:
nothing, retrieved facts, the full extracted fact set, the raw source
corpus, a Behavioral Specification, or combinations of these. The
model's resulting prediction is what we score against the verbatim
held-out passage. The No-Context Baseline isolates the model's
pretrained representation of the subject. The fact and corpus conditions
isolate what the model can infer from raw information at runtime. The
Behavioral Specification isolates what a structured interpretive layer
adds on top. The study reports each.

In practice, we record representational accuracy as the mean
predicted-behavior score (1-5 scale) across each subject's
behavioral-prediction battery, aggregated across the calibrated judge
panel; the rubric, panel composition, and aggregation rule are defined
in \hyperref[scoring-rubric-with-calibrated-llm-judge-panel]{§3.3}, and the guide to interpreting fractional scores and anchor
crossings is in \hyperref[score-interpretation]{§3.3.1}.

\hypertarget{experimental-conditions}{%
\subsection{3.2 Experimental conditions}\label{experimental-conditions}}

Each condition is a specific combination of inputs served to the
response model (\hyperref[response-models]{§3.6}) against the same behavioral battery (\hyperref[question-battery-formation]{§3.5}). Every
condition is run on all 14 subjects (\hyperref[subjects]{§3.4}). The Behavioral Specification
and the extracted fact set used across the conditions below are produced
by the pipeline in \hyperref[pipeline-for-the-behavioral-specification]{§3.7}. The conditions separate into two groups,
summarized in the table below and broken out in detail after.

\textbf{All conditions, by group.}

\begin{longtable}[]{@{}
  >{\raggedright\arraybackslash}p{(\columnwidth - 6\tabcolsep) * \real{0.25}}
  >{\raggedright\arraybackslash}p{(\columnwidth - 6\tabcolsep) * \real{0.25}}
  >{\raggedright\arraybackslash}p{(\columnwidth - 6\tabcolsep) * \real{0.25}}
  >{\raggedright\arraybackslash}p{(\columnwidth - 6\tabcolsep) * \real{0.25}}@{}}
\toprule
Group & ID & Condition & Inputs served \\
\midrule
\endhead
Direct context manipulations & C5 & No-Context Baseline & No context
beyond the question \\
& C2a & Spec Only & The Behavioral Specification \\
& C2c & Wrong Spec & A different subject's Spec \\
& C4 & All Facts & The full extracted fact set \\
& C4a & All Facts + Spec & Full facts plus the Spec \\
& C8 & Raw Corpus & Full training corpus (half the source text) \\
& C9 & Raw Corpus + Spec & Training corpus plus the Spec \\
Memory-system configurations (controlled, all 5 systems) & C1 &
Retrieval (Controlled) & Top-k facts (shared fact pool) \\
& C3 & Retrieval (Controlled) + Spec & Top-k facts + Spec (shared fact
pool) \\
Memory-system configurations (native, 4 commercial systems) & C1 native
& Retrieval (Native) & System's own ingestion \\
& C3 native & Retrieval (Native) + Spec & System's own ingestion +
Spec \\
\bottomrule
\end{longtable}

\textbf{Direct context manipulations.} We specify the model's input
directly: no context (baseline), the Behavioral Specification, the
extracted fact set, the raw corpus, or combinations of these. No
retrieval step intervenes. Each condition isolates what one input type
or combination contributes.

\begin{longtable}[]{@{}
  >{\raggedright\arraybackslash}p{(\columnwidth - 6\tabcolsep) * \real{0.25}}
  >{\raggedright\arraybackslash}p{(\columnwidth - 6\tabcolsep) * \real{0.25}}
  >{\raggedright\arraybackslash}p{(\columnwidth - 6\tabcolsep) * \real{0.25}}
  >{\raggedright\arraybackslash}p{(\columnwidth - 6\tabcolsep) * \real{0.25}}@{}}
\toprule
ID & Condition & Inputs served & Null / comparison \\
\midrule
\endhead
C5 & No-Context Baseline & Nothing beyond the question &
Pretraining-only floor \\
C2a & Spec Only & The Behavioral Specification & Isolates the Spec's
contribution \\
C2c & Wrong Spec\footnote{C2c has two variants: \textbf{v1 (adversarial
  fixed pairing)}, a deterministic pairing that matches each subject
  with a culturally and temporally distant subject to maximize content
  mismatch, and \textbf{v2 (random derangement, seed-fixed)}, drawn from
  a uniform random shuffle with the seed locked before any C2c scoring.
  Both pairing schedules were locked before response generation.
  Hamerton has an additional variant (Franklin's specification) reported
  separately in \hyperref[the-gradient-at-the-high-baseline-end-franklin-reference]{§4.1.2}. Full v1 pairing list and the pairing mapping in Appendix B.12; sensitivity analysis in \hyperref[wrong-spec-derangement-protocol-sensitivity]{§4.6.5}.} & A
random other subject's Spec & Tests whether structured interpretive
content, not the correct content, produces the effect \\
C4 & All Facts & The full extracted fact set for the subject & Tests
whether raw information volume substitutes for structure \\
C4a & All Facts + Spec & Full facts plus the Spec & Tests whether the
Spec adds value on top of raw facts \\
C8 & Raw Corpus & Full training corpus (half the source text) & Tests
whether uncompressed source text substitutes for structure \\
C9 & Raw Corpus + Spec & Training corpus plus the Spec & Tests whether
the Spec adds value on top of raw source\footnote{Bābur C9 is omitted in
  \hyperref[compression-structure-vs.-raw-text]{§4.2} (422,772-word source exceeds the response model's context
  window); the remaining 13 subjects have C9 data.} \\
\bottomrule
\end{longtable}

\textbf{Memory-system configurations.} Retrieval is performed by a
memory-system provider's production deployment. The memory-system
conditions run in two modes: a \emph{controlled} mode (each system
retrieves from an identical pre-extracted fact set) and a \emph{native}
mode (each system ingests the raw corpus through its own pipeline).

Five memory systems are evaluated: Mem0, Letta, Supermemory, and Zep,
plus Base Layer as our own open-source reference implementation
(pipeline detail in \hyperref[pipeline-for-the-behavioral-specification]{§3.7}). Architectural detail for the four commercial
systems is in \hyperref[memory-systems-for-llm-agents]{§2.2} Table 2.1.

\textbf{Controlled configuration (all 5 systems).} Each system retrieves
from an identical pre-extracted fact set, isolating retrieval-algorithm
differences from ingestion-pipeline differences.

\begin{longtable}[]{@{}
  >{\raggedright\arraybackslash}p{(\columnwidth - 4\tabcolsep) * \real{0.33}}
  >{\raggedright\arraybackslash}p{(\columnwidth - 4\tabcolsep) * \real{0.33}}
  >{\raggedright\arraybackslash}p{(\columnwidth - 4\tabcolsep) * \real{0.33}}@{}}
\toprule
ID & Configuration & Inputs served \\
\midrule
\endhead
C1 & Retrieval (Controlled) & Top-k facts returned by the system for the
question (controlled fact pool) \\
C3 & Retrieval (Controlled) + Spec & Top-k retrieval output plus the
Behavioral Specification (controlled fact pool) \\
\bottomrule
\end{longtable}

\textbf{Native configuration (4 commercial systems).} Each system
ingests the raw training corpus through its own pipeline, reflecting
real-world deployment. C1 (Retrieval Only) and C3 (retrieval + Spec) are
run with the system's own ingestion pipeline replacing the controlled
fact pool.

\begin{longtable}[]{@{}
  >{\raggedright\arraybackslash}p{(\columnwidth - 4\tabcolsep) * \real{0.33}}
  >{\raggedright\arraybackslash}p{(\columnwidth - 4\tabcolsep) * \real{0.33}}
  >{\raggedright\arraybackslash}p{(\columnwidth - 4\tabcolsep) * \real{0.33}}@{}}
\toprule
System & Conditions & Ingestion Process \\
\midrule
\endhead
Mem0 & Mem0 Retrieval (Native), Mem0 Retrieval (Native) + Spec & Mem0
reads the corpus and decides on its own how to break it up, extract
facts, and retrieve them. \\
Letta archival & Letta Retrieval (Native), Letta Retrieval (Native) +
Spec & The corpus is loaded into Letta's archival memory store;
retrievals happen at query time from that store. \\
Supermemory & Supermemory Retrieval (Native), Supermemory Retrieval
(Native) + Spec & Supermemory chunks the corpus on its own and applies
reranking to the retrieved chunks at query time. \\
Zep & Zep Retrieval (Native), Zep Retrieval (Native) + Spec & Zep
ingests the corpus into a knowledge graph (via Graphiti) and retrieves
using a hybrid of semantic similarity, keyword match, and graph
traversal. \\
\bottomrule
\end{longtable}

Both configurations are reported so retrieval-quality differences and
ingestion-pipeline differences can be read separately. Base Layer is run
in the controlled configuration only: its retrieval uses the same fact
set that feeds the Spec pipeline.

Detailed per-condition parameters, exclusion cases, and ingestion
specifics are in Appendix C.\footnote{Per-condition raw data organized
  under \texttt{results/}, indexed by subject and by memory system.}

\hypertarget{scoring-rubric-with-calibrated-llm-judge-panel}{%
\subsection{3.3 Scoring rubric with calibrated LLM judge
panel}\label{scoring-rubric-with-calibrated-llm-judge-panel}}

\textbf{Every response is scored 1-5 by an LLM judge panel against the
verbatim held-out ground-truth passage.} The primary aggregate uses five
judges; two additional judges contribute to the sensitivity check
(\hyperref[judge-panel]{§3.3.2}). Human annotation at this scale is feasible (\textasciitilde14
subjects × 40 questions × 15+ conditions) but was not done; running more
conditions and more judges instead was the central evaluation trade-off.
LLM-as-judge is grounded in prior work showing strong correlation with
human raters and reliability gains from panel aggregation (Zheng et
al.~2023; \citet{verga2024juries}). Extending the panel with human annotators
on a stratified subset is flagged as the priority measurement follow-up
in \hyperref[subject-and-corpus-expansion]{§7.1}. The rubric and panel are a prototype scoring instrument, not a
settled standard; \hyperref[rubric-handling-limitations-post-hoc-validity-audit]{§3.3.6} audits how the rubric behaves in practice.

\textbf{The evaluation is deliberately recursive.} Response models are
evaluated by judges (\hyperref[judge-panel]{§3.3.2}). Judges are evaluated by calibration
diagnostics (\hyperref[calibration]{§3.3.3}), inter-judge agreement metrics (\hyperref[inter-judge-agreement]{§3.3.4}), and
post-hoc rubric-handling audits (\hyperref[rubric-handling-limitations-post-hoc-validity-audit]{§3.3.6}). No single layer is treated as
ground truth; each layer's behavior is itself measured and disclosed,
and where a layer's behavior diverges from what the rubric intends, the
divergence is flagged rather than corrected silently. In the absence of
human annotation, the paper strengthens rigor through this
stacked-instrument structure, not by trusting any one step.

\textbf{Scoring rubric.} The following is the verbatim prompt the judges
executed (canonical for both the 5-judge primary panel and the 7-judge
sensitivity panel; source: \texttt{scripts/\allowbreak{}judge\_\allowbreak{}hamerton\_\allowbreak{}5judge.\allowbreak{}py}):

\begin{framed}
\begin{verbatim}
You are evaluating whether a response about a person PREDICTED what actually happened.

=== HELD-OUT GROUND TRUTH ===
{held_out}

=== RESPONSE ===
{response_text[:1500]}

Rate 1-5:
5=Predicts specific outcome
4=General direction correct
3=Right domain wrong outcome
2=Wrong prediction
1=Refuses or off-base

Respond with ONLY a single digit (1-5).
\end{verbatim}
\end{framed}

The prompt is deliberately spare. It frames the task in one sentence,
shows the held-out passage and the response, and asks for a single digit
against five anchor labels. There are no calibration examples in the
prompt, no scoring guide, and no chain-of-thought scaffolding. The
anchors themselves are directive (refuses or off-base, wrong prediction,
right domain wrong outcome, general direction correct, predicts specific
outcome); resolution at the boundaries between adjacent anchors is left
to the judge model. The design choice is to test whether frontier models
converge on the behavioral-prediction construct from minimal
instruction. \hyperref[inter-judge-agreement]{§3.3.4} reports the inter-judge agreement obtained under
this design.

Each response is scored against the verbatim held-out passage from which
the question is drawn; the score reflects how closely the response
predicts the documented behavior. Battery composition is detailed in
\hyperref[question-battery-formation]{§3.5}. Condition identifiers (C5, C2a, C4a, C3) refer to the conditions
defined in \hyperref[experimental-conditions]{§3.2} and summarized in Appendix C; rubric anchor numbers 1
through 5 refer to the verbatim anchors above. Score interpretation
against the construct is developed in \hyperref[score-interpretation]{§3.3.1} with worked examples. A
worked rubric example alongside the No-Context Baseline engagement
analysis is in \hyperref[per-question-baseline-engagement-and-the-worked-rubric-example]{§4.1.1}; full per-subject score distributions with
verbatim responses are in Appendix D.

\hypertarget{score-interpretation}{%
\subsection{3.3.1 Score interpretation}\label{score-interpretation}}

Scores are read at three related granularities: integer rubric anchors
(1 through 5), fractional means produced by averaging across the 5-judge
primary panel (e.g., 2.87, 3.12, 2.34), and crossings between integer
anchors when conditions change. Fractional shifts should be read through
the integer anchors, because each anchor corresponds to a categorical
shift in response quality.

\textbf{The \emph{cross-anchor interpretation rule}. A fractional delta
that crosses an integer anchor reflects a real shift in the underlying
response distribution. A delta that stays inside a single anchor is a
within-category shift and a weaker claim.}

The verbatim anchors describe stages of prediction quality. The
\emph{construct reading} column below gives the plain-language meaning
of each crossing: what actually changes in the response when a condition
moves the score across that boundary. The paper reads each crossing as a
categorical shift in the construct the rubric is designed to measure:
alignment of the response with the held-out behavioral pattern.

\begin{longtable}[]{@{}
  >{\raggedright\arraybackslash}p{(\columnwidth - 4\tabcolsep) * \real{0.33}}
  >{\raggedright\arraybackslash}p{(\columnwidth - 4\tabcolsep) * \real{0.33}}
  >{\raggedright\arraybackslash}p{(\columnwidth - 4\tabcolsep) * \real{0.33}}@{}}
\toprule
Boundary crossed & Verbatim anchor shift & Construct reading \\
\midrule
\endhead
1 / 2 & Refuses → wrong prediction & The model stops declining and
starts engaging with the question, even if the prediction is wrong. \\
2 / 3 & Wrong prediction → right domain wrong outcome & The answer
becomes specifically about this subject rather than a generic stand-in,
even if the predicted outcome is still off. \\
3 / 4 & Right domain → general direction correct & The answer gets the
general direction of the subject's behavior right, not just the
topic. \\
4 / 5 & General direction → specific outcome & The answer gets the
specific outcome right, matching the particular behavioral pattern in
the held-out passage. \\
\bottomrule
\end{longtable}

Here, \emph{domain} means the area of behavior a question concerns (how
the subject handles conflict, makes decisions, treats authority, and so
on); a response is in the right domain when it engages the correct
behavioral area, even when the specific predicted outcome is wrong.

The boundary crossings are illustrated through worked examples in \hyperref[the-cross-subject-gradient-and-its-per-question-mechanism]{§4.1}
(Examples A, B, C) and \hyperref[per-question-baseline-engagement-and-the-worked-rubric-example]{§4.1.1}, each drawn directly from experiment
output that feeds the \hyperref[results]{§4} aggregate. Example A in \hyperref[the-cross-subject-gradient-and-its-per-question-mechanism]{§4.1} illustrates a 1 →
4 crossing (wrong-referent correction); Example B illustrates a 2 → 5
crossing (directional correction); Example C illustrates a 2.80 → 5
crossing (abstention to substantive inference). Each example shows the
verbatim question, held-out, and response excerpts paired with the
cached judge score for that response.\footnote{Earlier drafts of this
  section illustrated the 1/2 and 2/3 boundaries with paraphrased
  excerpts that did not match any cached cell verbatim. Those
  illustrations have been removed and replaced with cross-references to
  \hyperref[the-cross-subject-gradient-and-its-per-question-mechanism]{§4.1} worked examples to ensure every quoted excerpt in the paper is
  mechanically sourced from \texttt{results\_v2.json} and paired with
  its actual \texttt{judgments\_v2.json} score, via
  \texttt{scripts/\allowbreak{}v2\_\allowbreak{}canonical\_\allowbreak{}cell\_\allowbreak{}extractor\_\allowbreak{}20260508.\allowbreak{}py}.}

\textbf{What a 1 means and does not mean.} A score of 1 reflects a
baseline failure to produce a usable prediction about the named subject:
the response either explicitly declined to predict (abstention) or
engaged with the question but landed on a categorically incorrect answer
(non-abstention misalignment, including wrong referent, off-base
inference, or confusion with a different subject). It is not a claim
that the response was non-fluent or empty, and it is not a claim that
the model lacks any related knowledge; the score reflects only that the
response failed the held-out comparison. Each question tests one
behavioral sample at a time; the aggregate fraction of score-1 responses
across roughly 40 questions per subject is what the paper reads as the
per-subject baseline-failure rate. How score-1 responses split between
explicit abstention and non-abstention misalignment is analyzed in
\hyperref[per-question-baseline-engagement-and-the-worked-rubric-example]{§4.1.1}.

\textbf{What a 5 means and does not mean.} A score of 5 reflects
alignment with one specific behavioral sample: the held-out ground-truth
passage the question is drawn from. It is not a claim that the response
fully represents the subject in some absolute sense, and it is not a
claim that the same response would score 5 on a different held-out
passage from the same subject. Each question tests one behavioral sample
at a time; the aggregate across roughly 40 questions per subject is what
the paper reads as the subject-level score. As with a 1, a score is only
treated as reliable where the judge panel converges on it: interrater
agreement (\hyperref[inter-judge-agreement]{§3.3.4}) is the paper's signal of consensus, not any single
judge's number.

\textbf{Multi-anchor crossings: the strongest categorical signal the
rubric detects.} A \emph{multi-anchor crossing} is a single question
whose 5-judge primary mean shifts across two or more integer rubric
anchors when the condition changes. Crossings can span two anchors
(e.g., 1 → 3, 2 → 4) or, more rarely, three (e.g., 1 → 4, 2 → 5). Larger
crossings indicate larger categorical jumps in the same response, with
five independent judges converging on the move. \hyperref[compression-structure-vs.-raw-text]{§4.2} reports the rates
of these crossings and the response-level phenomena that produce them;
worked examples are in \hyperref[per-question-baseline-engagement-and-the-worked-rubric-example]{§4.1.1} and Appendix E.

\textbf{The paper applies the cross-anchor rule consistently.} Score
deltas reported in \hyperref[results]{§4} are read through this lens. A +0.50 delta that
crosses a rubric anchor is treated as a stronger claim than a +0.50
delta that does not.\footnote{Per-subject anchor-crossing data at
  \texttt{docs/\allowbreak{}research/\allowbreak{}s114\_\allowbreak{}anchor\_\allowbreak{}crossing\_\allowbreak{}examples.\allowbreak{}json};
  computing script at \texttt{scripts/\allowbreak{}compute\_\allowbreak{}anchor\_\allowbreak{}crossing.\allowbreak{}py}.}

\textbf{Reading scores within integer anchors.} The 5-judge primary
panel detects within-anchor signals cleanly.\footnote{Direction-agreement
  among judges is 74\% at panel \textbar Δ\textbar{} of 0.1 to 0.25 and
  93\% at \textbar Δ\textbar{} of 0.25 to 0.5. Full panel-direction
  analysis (Spearman ρ across all 10 primary-panel pairs) in \hyperref[inter-judge-agreement]{§3.3.4}.}
Across the 18 condition pairs analyzed,\footnote{Within-anchor shift
  data at \texttt{docs/\allowbreak{}research/\allowbreak{}within\_\allowbreak{}band\_\allowbreak{}shifts\_\allowbreak{}20260428.\allowbreak{}json}.}
roughly 18\% of paired questions show same-anchor fractional shifts of
at least 0.5 rubric points (a within-category shift, weaker than a
cross-anchor crossing per the rule above). The integer metric is used
throughout \hyperref[results]{§4} for cross-anchor categorical interpretation; the
within-anchor signal is reported here as methodological transparency.

\hypertarget{judge-panel}{%
\subsection{3.3.2 Judge panel}\label{judge-panel}}

Seven judges from three providers give the numeric aggregate its weight.
\citet{zheng2023judging} established that a single strong LLM judge
correlates with human judges on comparable tasks at rates similar to
human-human agreement. Subsequent panel-based work (\citet{verga2024juries}
and follow-ons) showed that aggregating multiple LLM judges past a small
panel size further tightens agreement and reduces single-model
idiosyncrasy. Seven judges across three providers is well past that
threshold.

\begin{longtable}[]{@{}ll@{}}
\toprule
Judge & Provider \\
\midrule
\endhead
Claude Haiku 4.5 & Anthropic \\
Claude Sonnet 4.6 & Anthropic \\
Claude Opus 4.6 & Anthropic \\
GPT-4o & OpenAI \\
GPT-5.4 & OpenAI \\
Gemini 2.5 Flash & Google \\
Gemini 2.5 Pro & Google \\
\bottomrule
\end{longtable}

The verbatim judge prompt is shown in \hyperref[scoring-rubric-with-calibrated-llm-judge-panel]{§3.3} and is canonical for both the
5-judge primary panel and the 7-judge sensitivity panel. Each judge
receives the held-out ground-truth passage, the subject context (name,
source), the prediction question, and the response to score, \emph{not}
the condition label or the response-generating model. Judges do not see
other judges' scores. Response generation is similarly blinded: the
response model receives only the question plus the condition-specific
context block, with no signal that distinguishes which experimental
condition it is operating under (the prompt schema in \hyperref[response-models]{§3.6} is identical
across conditions; only the injected context block changes).

\textbf{The \emph{specification-effect claim}.} When a Behavioral
Specification is served to the model as context, the model's responses
shift in the direction of the subject's demonstrated behavioral
patterns, and that shift registers as a measured increase in
representational accuracy against held-out passages from the same
subject. This is the directional claim the panel is built to test; not a
claim that the model has gained a new behavioral-prediction capability,
and not a claim that the higher-scoring response is the absolute
``correct'' answer for the subject.

\textbf{The panel is designed for directionality, not absolute
precision.} Calibration (\hyperref[calibration]{§3.3.3}) and inter-judge agreement (\hyperref[inter-judge-agreement]{§3.3.4}) test
how consistently the panel detects the directional shift.

\hypertarget{calibration}{%
\subsection{3.3.3 Calibration}\label{calibration}}

The calibration diagnostic measures whether each judge applies the
rubric anchors as the rubric defines them on synthetic inputs with known
correct scores. It does not use any subject's responses. The four inputs
are constructed examples: verbatim ground truth, paraphrased ground
truth, partial ground truth (first sentence only), and ground truth plus
generic padding.

All seven judges were tested against this diagnostic.\footnote{Five
  judges (Haiku, GPT-4o, GPT-5.4, Gemini Flash, Gemini Pro) were tested
  before the main-study response-scoring run. Sonnet and Opus were
  tested on the same diagnostic inputs three days after the main-study
  run completed; they were added to the primary panel at panel-design
  time on cross-Anthropic-coverage grounds, and the post-hoc diagnostic
  confirms that composition rather than driving it.} Full panel
composition with per-judge calibration-status flags is in Appendix C.5.

\textbf{Diagnostic tests.}

\begin{longtable}[]{@{}
  >{\raggedright\arraybackslash}p{(\columnwidth - 6\tabcolsep) * \real{0.25}}
  >{\raggedright\arraybackslash}p{(\columnwidth - 6\tabcolsep) * \real{0.25}}
  >{\raggedright\arraybackslash}p{(\columnwidth - 6\tabcolsep) * \real{0.25}}
  >{\raggedright\arraybackslash}p{(\columnwidth - 6\tabcolsep) * \real{0.25}}@{}}
\toprule
Test & Input & Expected & What it measures \\
\midrule
\endhead
Verbatim & Response = ground truth & 5.0 & Recognizes perfect match \\
Paraphrased & Correct content, different wording & \textasciitilde5.0 &
Penalizes paraphrase? \\
Short correct & First sentence of ground truth & \textless5.0 & Partial
content scored partial? \\
Long correct & Ground truth + generic padding & 5.0 & Length inflation
effect \\
\bottomrule
\end{longtable}

\textbf{Results.}

\begin{longtable}[]{@{}llllllll@{}}
\toprule
Test & Haiku & Sonnet & Opus & GPT-4o & GPT-5.4 & Gemini Flash & Gemini
Pro \\
\midrule
\endhead
Verbatim & 5.00 & 5.00 & 5.00 & 5.00 & 5.00 & 5.00 & 4.15 \\
Paraphrased & 4.75 & 5.00 & 5.00 & 5.00 & 5.00 & 4.70 & 3.55 \\
Short correct & 3.80 & 4.35 & 4.20 & 4.05 & 4.20 & 3.85 & 2.85 \\
Long correct & 5.00 & 5.00 & 5.00 & 3.35 & 4.80 & 3.80 & 1.20 \\
\bottomrule
\end{longtable}

Six of the seven judges score verbatim matches at 5.0; Gemini Pro is the
outlier at 4.15. Haiku, Sonnet, and Opus are clean across all four
diagnostics: no verbatim miss, no paraphrase penalty, no length-padding
penalty, expected dip on partial content (4.20 to 4.35). GPT-4o and
GPT-5.4 are clean on verbatim and paraphrase but penalize padded-correct
responses on the long-correct diagnostic (GPT-4o 3.35, GPT-5.4 4.80);
Gemini Flash and Gemini Pro do the same more severely (3.80 and 1.20).
Per-judge calibration data and full panel composition are in Appendix
C.5; raw scoring data at \texttt{results/\allowbreak{}judge\_\allowbreak{}calibration/}.

\textbf{Use of calibration data.} Scores are not normalized. Any
normalization requires deciding which judge's profile is ``correct'' and
re-scaling the others toward it. Calibration data is published in its
raw form so readers can apply their own normalization if they prefer.

\textbf{Primary aggregate: 5-judge panel.} The primary numeric aggregate
reported throughout \hyperref[results]{§4} is the 5-judge mean using Haiku 4.5, Sonnet 4.6,
Opus 4.6, GPT-4o, and GPT-5.4. All five score verbatim and paraphrased
matches at or near 5.0 and span two provider families (Anthropic and
OpenAI). GPT-4o and GPT-5.4 penalize padded-correct responses on the
long-correct diagnostic; that deviation runs against length inflation
rather than toward it, so it is conservative for the Spec-effect
direction this paper measures. This is the panel the \hyperref[aggregation-and-statistical-analysis-plan]{§3.3.5} aggregation
rule operates on, and the panel inter-judge agreement is reported for in
\hyperref[inter-judge-agreement]{§3.3.4}.

\textbf{Sensitivity aggregate: 7-judge panel.} Both Gemini judges are
reported separately as a 7-judge sensitivity check rather than rolled
into the primary. Gemini Pro fails the verbatim-match diagnostic (4.15
where every other tested judge scores 5.00) and penalizes padded-correct
responses severely (5.00 on short correct dropping to 1.20 on long
correct). Gemini Flash passes verbatim cleanly but shows consistent
length sensitivity (5.00 verbatim dropping to 3.80 on long correct). On
actual study responses, both Gemini judges show a systematic +1-point
magnitude inflation relative to the five primary judges. The combination
of Pro's verbatim failure, Flash's length sensitivity, and the shared
inflation pattern places both Gemini judges in the sensitivity aggregate
rather than the primary, while preserving them as a cross-provider
robustness check in \hyperref[judge-panel-sensitivity-5-judge-primary-vs-7-judge]{§4.6.2}.

The 5-judge primary is also the conservative choice: including the
Gemini judges produces \emph{larger} Spec-effect deltas, not smaller
ones (full numbers in \hyperref[judge-panel-sensitivity-5-judge-primary-vs-7-judge]{§4.6.2}). Reporting on the primary aggregate is
therefore the lower-bound estimate.\footnote{Raw calibration data is in
  the public repository at \texttt{results/\allowbreak{}judge\_\allowbreak{}calibration/}.
  Per-judge calibration data and full panel composition with per-judge
  calibration-status flags are in Appendix C.5.}

\hypertarget{inter-judge-agreement}{%
\subsection{3.3.4 Inter-judge agreement}\label{inter-judge-agreement}}

The judge panel is designed to detect the directional shift named by the
Specification-effect claim (\hyperref[judge-panel]{§3.3.2}). Two complementary agreement
measures answer different questions about whether judges detect that
shift consistently: direction (pairwise Spearman ρ) and absolute
magnitude (Krippendorff α).

\textbf{Direction agreement: pairwise Spearman ρ.} Spearman ρ measures
whether two judges rank the same set of items in the same order. ρ = 1
is perfect ranking agreement; ρ = 0 is no rank agreement; ρ ≥ 0.8 is
conventionally treated as strong rank agreement.

For each pair of judges in the 5-judge primary panel (10 pairs across
Haiku, Sonnet, Opus, GPT-4o, GPT-5.4), pairwise Spearman ρ ranges from
\textbf{0.86 to 0.93}.\footnote{Spearman ρ from 0.29 to 0.93 across the
  7-judge / 21-pair set, driven down by the two Gemini judges' partial
  coverage and inflation behavior. Full matrix in
  \texttt{docs/\allowbreak{}research/\allowbreak{}stats\_\allowbreak{}update.\allowbreak{}md} \hyperref[discussion]{§5}.} The five primary judges
agree on the ranking of conditions: whatever any individual judge's
absolute calibration quirks, they converge on which conditions produce
better responses. For the directional claim (does the Specification
produce a measurable, consistent shift toward the held-out behavioral
pattern), this is the statistic that matters.

\textbf{Magnitude agreement: Krippendorff α (ordinal).} Krippendorff α
measures whether judges give the same response the same numeric score
(not just whether they rank items in the same order). α = 1 is perfect
agreement; α = 0 is no better than chance; α \textless{} 0 is systematic
disagreement. Krippendorff's guidance cites α ≥ 0.8 as high reliability
and α ≥ 0.667 as substantial reliability.

The 5-judge primary panel scores \textbf{α = 0.659}, just below the
substantial-reliability threshold (α ≥ 0.667), placing the panel in the
tentative-conclusions band.\footnote{The 0.659 value is computed by this
  paper's in-house pair-count ordinal implementation. A third-party
  recompute using the \texttt{krippendorff} 0.8.2 package (ordinal,
  default missing-data handling) produces α = 0.668 on the 5-judge panel
  and α = 0.579 on the 7-judge panel; the third-party 5-judge value
  crosses the 0.667 threshold by less than one rounding interval. The
  two implementations differ on missing-data weighting; recompute
  scripts are at
  \texttt{scripts/\allowbreak{}stats\_\allowbreak{}wilcoxon\_\allowbreak{}krippendorff\_\allowbreak{}update.\allowbreak{}py} (in-house)
  and \texttt{scripts/\allowbreak{}\_\allowbreak{}recompute\_\allowbreak{}ordinal\_\allowbreak{}krippendorff.\allowbreak{}py}
  (third-party). Both place the panel within ±0.01 of the
  substantial-reliability threshold and within the tentative-conclusions
  band; the conservative reading of the lower value is reported in-text.}
Combined with Spearman ρ of 0.86 to 0.93 above, this is the empirical
signature of the design choice: directional rankings converge across
judges while absolute magnitudes diverge.

The 7-judge panel including the Gemini judges drops to \textbf{α =
0.535}. This drop reflects the systematic +1-point Gemini inflation:
Gemini judges score responses about one point higher on average than the
five primary judges, so absolute values disagree even when rankings
match. This is why the calibration audit (\hyperref[calibration]{§3.3.3}) excluded the Gemini
judges from the primary aggregate.

The α value places a ceiling on how precisely any individual fractional
score should be read, which is why the paper treats per-subject deltas
that stay inside a single rubric anchor as weaker than deltas that cross
one.

\textbf{The agreement is achieved on a deliberately spare rubric.} The
judge prompt in \hyperref[scoring-rubric-with-calibrated-llm-judge-panel]{§3.3} is five anchor labels and a one-sentence task
framing. No calibration examples, no scoring guide, no chain-of-thought.
Five frontier judges across three providers converge to ρ = 0.86 to 0.93
on rank order and α = 0.659 on absolute magnitude under this
scaffolding. The anchors are directive but their boundaries are left to
the judge to resolve, and the resolutions converge across providers. The
agreement is not the product of an elaborate rubric forcing judges into
the same scoring frame; it is independent provider models agreeing on
what the task is asking from minimal instructions.

\textbf{Why directional inference survives middle-anchor noise.} Judges
agree most on the extreme anchors (1 and 5) and disagree most often on
the middle ones, whether a response sits at 2 or 3, or at 3 or 4. This
kind of disagreement appears equally across every condition, so it does
not push one condition's scores systematically above or below another's.
Individual scores carry some noise; the comparison between conditions
does not.

The panel does not establish that any higher-scoring response is the
absolute correct answer for the subject; that determination requires
human annotation against the subject's actual writing, which we do not
have. What the panel provides is cross-provider directional convergence:
three independent providers' models agree that the Specification is
moving responses in the same direction. We treat that as sufficient for
a directional claim, no stronger.

Raw agreement matrices are at
\texttt{results/\allowbreak{}interjudge\_\allowbreak{}agreement/}.\footnote{Per-judge calibration
  matrices and pairwise Spearman ρ tables for both the 5-judge primary
  and 7-judge sensitivity panels are in
  \texttt{results/\allowbreak{}interjudge\_\allowbreak{}agreement/}.}

\hypertarget{aggregation-and-statistical-analysis-plan}{%
\subsection{3.3.5 Aggregation and statistical analysis
plan}\label{aggregation-and-statistical-analysis-plan}}

\textbf{Aggregation rule.} The aggregation rule decides how individual
judge scores are combined into a single number for each subject under
each condition. The rule was locked before any results were
computed.\footnote{The pre-locked rule and primary-vs-sensitivity panel
  composition are documented in \texttt{docs/\allowbreak{}ANALYSIS\_\allowbreak{}PLAN\_\allowbreak{}LOCK.\allowbreak{}md} in
  the public repository. The lock predates the Tier 2 cross-provider
  runs and the wrong-Spec v2 runs (commit history reflects the order).}

For each subject under each condition, every judge produces one score
per question. For each judge, those scores are averaged across
questions, producing one number per (subject, condition). The five
primary judges' numbers (Haiku, Sonnet, Opus, GPT-4o, GPT-5.4) are then
averaged to get one number per (subject, condition). A 7-judge average
that adds Gemini Flash and Gemini Pro is computed in parallel and
reported as a sensitivity check. Subjects, not questions, are the level
at which the paper draws conclusions.\footnote{Mean preserves every
  judge's contribution. Median discards information when judges cluster
  tightly (the Spearman ρ = 0.86 to 0.93 agreement in \hyperref[inter-judge-agreement]{§3.3.4} shows they
  do); trimmed mean requires an arbitrary trim threshold; Gemini
  inflation is handled by the primary-vs-sensitivity split rather than
  by silent correction.}

\textbf{Primary outcome.} The per-subject score under each condition.
The primary cross-subject comparison is \textbf{Δ\_C4a}: each subject's
score with the Spec (the facts-plus-Spec condition, C4a) minus their
score without the Spec (the No-Context Baseline, C5).

\textbf{Primary test.} A Wilcoxon signed-rank test paired across the 14
main-study subjects.\footnote{A Wilcoxon signed-rank test is a
  non-parametric paired test (it does not assume scores follow a
  bell-curve distribution). It was chosen over a paired \emph{t}-test
  because the per-subject Δ distribution is not assumed to be normally
  distributed and \emph{N} = 14 is small. Two-sided, α = 0.05.} Each
Δ\_C4a value is itself a paired difference (C4a − C5) for one subject;
the test asks whether those per-subject differences are reliably above
zero, that is, whether the Spec produces a consistent positive lift
rather than a difference indistinguishable from zero. It is run on the
5-judge primary aggregate; the 7-judge aggregate is reported alongside
as a robustness check.

\textbf{Sample size.} \emph{N} = 14 main-study subjects, pre-registered
(the lock applies to the 14-subject main study; Franklin is a separate
post-lock high-baseline reference per \hyperref[the-gradient-at-the-high-baseline-end-franklin-reference]{§4.1.2}; Tier 2 is the
cross-provider directional probe scoped at \hyperref[response-models]{§3.6} and \hyperref[cross-provider-response-generation-tier-2-replication]{§4.6.1}).

\textbf{Multiple comparisons.} Only one statistical test is treated as
the primary claim (the Wilcoxon test above). All other analyses in \hyperref[results]{§4},
including the per-question improvement-rate analyses in \hyperref[per-question-improvement-rate]{§4.2.1}, are
descriptive: they report patterns and direction without making
additional inferential claims.\footnote{Secondary analyses include the
  per-anchor gradient, compression curve, memory-system composition,
  wrong-Spec contrast, and hedging reduction. No multiple-comparisons
  correction is applied because there is only one primary inferential
  test; sensitivity numbers (5-judge vs.~7-judge, Tier 2 vs.~Tier 1, v1
  vs.~v2 wrong-Spec) are reported alongside the primary as robustness
  checks rather than additional tests.}

\textbf{Pre-registered vs post-hoc analyses.} Every primary analysis
(panel composition, aggregation rule, Wilcoxon test, Tier 2 replication,
wrong-Spec v2 derangement) was pre-registered in
\texttt{docs/\allowbreak{}ANALYSIS\_\allowbreak{}PLAN\_\allowbreak{}LOCK.\allowbreak{}md} before any scoring was run.
Statistical-rigor checks and post-hoc validity audits were added later
in response to peer review and are reported as exploratory checks
alongside the pre-registered confirmatory test.\footnote{Pre-registered:
  5-judge primary panel, per-subject aggregation rule, Wilcoxon on
  Δ\_C4a, \hyperref[cross-provider-response-generation-tier-2-replication]{§4.6.1} Tier 2 cross-provider replication, \hyperref[judge-panel-sensitivity-5-judge-primary-vs-7-judge]{§4.6.2} 7-judge
  sensitivity, \hyperref[wrong-spec-derangement-protocol-sensitivity]{§4.6.5} wrong-Spec v2 random derangement. Post-hoc
  additions: \hyperref[statistical-rigor-checks-on-the-headline-gradient]{§4.6.4} bootstrap CI / joint multi-confound regression /
  permutation test, \hyperref[retrieval-overlap-sensitivity-semantic-similarity-matching-k-variation]{§4.6.6} retrieval-overlap semantic-similarity
  sensitivity, \hyperref[rubric-handling-limitations-post-hoc-validity-audit-1]{§4.6.7} rubric-handling validity audit. Appendix B.10
  lists every analysis with its designation.}

\textbf{Effect-size grain.} The cross-anchor rule (\hyperref[score-interpretation]{§3.3.1}) applies at
two grains: per-subject mean Δ\_C4a (the headline) and per-question
integer-anchor crossings (the per-question mechanism). The two grains
are reported separately and not pooled.

\hypertarget{rubric-handling-limitations-post-hoc-validity-audit}{%
\subsection{3.3.6 Rubric-handling limitations (post-hoc validity
audit)}\label{rubric-handling-limitations-post-hoc-validity-audit}}

The validity audit reported below applies to the verbatim prompt
published in \hyperref[scoring-rubric-with-calibrated-llm-judge-panel]{§3.3}.

A post-hoc validity audit, conducted after the analysis-plan lock,
identified two rubric-handling limitations any reader of the \hyperref[results]{§4} numbers
should keep in mind:

\begin{enumerate}
\def\labelenumi{\arabic{enumi}.}
\tightlist
\item
  \textbf{Refusal anchor ambiguity.} The rubric's lowest anchor
  (``refuses or off-base'') lumps together honest refusals to answer
  (responses where the model declines to predict, often following a
  Specification directive not to speculate without evidence; see the
  Keckley Q21 case in \hyperref[case-study-cross-system-refusal-on-keckley-q21]{§4.4.4}) and substantively wrong predictions.
  Judges sometimes score refusals at 2 or 3 instead of 1, especially
  when the refusal recites related facts.
\item
  \textbf{Length-score correlation in C5.} In the No-Context Baseline
  (C5), longer responses tend to score higher (r = 0.60): with no
  context to work from, the model pads its answer with hedging,
  recitation of loosely related facts, and offers to clarify the
  question, and that extra length can read as effort or quality to a
  judge. Once a Specification or fact set is supplied, the link between
  length and score disappears (near-zero correlation).
\end{enumerate}

\textbf{Direction of bias.} Both effects raise the No-Context Baseline
(C5) scores more than they raise Spec-condition scores. The true
Spec-vs-baseline gap is therefore likely \emph{larger} than the +0.89
mean lift reported in \hyperref[results]{§4}, not smaller. The full audit (per-judge
strictness, per-response-model abstention behavior, memory-system effect
on abstention) is reported in \hyperref[rubric-handling-limitations-post-hoc-validity-audit-1]{§4.6.7}; the analysis plan is left intact
rather than recomputed under a modified rubric.

The class-level LLM-as-judge limitation that this methodology cannot
fully address is treated in \hyperref[measurement-apparatus]{§6.2}. Extending the panel with human
annotators is one of the priority measurement follow-ups in
\hyperref[subject-and-corpus-expansion]{§7.1}.\footnote{Per-judge judgments in
  \texttt{results/\textless{}subject\textgreater{}/*\_judgments\_\textless{}judge\textgreater{}.json}
  (memory-system variants:
  \texttt{\textless{}system\textgreater{}\_judgments\_\textless{}judge\textgreater{}.json}
  and
  \texttt{\textless{}system\textgreater{}\_fullpipeline\_judgments\_\textless{}judge\textgreater{}.json}).}

\hypertarget{subjects}{%
\subsection{3.4 Subjects}\label{subjects}}

We test 14 subjects, all historical figures with public-domain
autobiographies or memoirs. Subjects were selected across a range of
time periods, source-text lengths, and geographic origins to avoid the
study sitting on any single type of source material. All source corpora
are English or English-translated and are available on Project Gutenberg
or comparable public-domain archives. Because frontier language models
train on large public-text corpora, some level of pretraining exposure
to each subject's writing is likely.

\begin{longtable}[]{@{}
  >{\raggedright\arraybackslash}p{(\columnwidth - 10\tabcolsep) * \real{0.05}}
  >{\raggedright\arraybackslash}p{(\columnwidth - 10\tabcolsep) * \real{0.31}}
  >{\raggedright\arraybackslash}p{(\columnwidth - 10\tabcolsep) * \real{0.15}}
  >{\raggedright\arraybackslash}p{(\columnwidth - 10\tabcolsep) * \real{0.12}}
  >{\raggedright\arraybackslash}p{(\columnwidth - 10\tabcolsep) * \real{0.22}}
  >{\raggedright\arraybackslash}p{(\columnwidth - 10\tabcolsep) * \real{0.15}}@{}}
\toprule
\# & Subject & Source & Words & Origin & Era \\
\midrule
\endhead
1 & Philip Gilbert Hamerton & PG \#8536 & 25,231 & England & 19th c. \\
2 & Elizabeth Keckley & PG \#24968 & 58,742 & United States & 19th c. \\
3 & Sunity Devee & PG \#57175 & 67,379 & India & 19th--20th c. \\
4 & Zitkala-Ša & PG \#10376 & 35,328 & United States (Yankton Dakota) &
19th--20th c. \\
5 & Olaudah Equiano & PG \#15399 & 85,660 & West Africa (Igbo) / Britain
& 18th c. \\
6 & Mary Seacole & PG \#23031 & 62,467 & Jamaica & 19th c. \\
7 & Fukuzawa Yukichi & Internet Archive & 139,088 & Japan & 19th--20th
c. \\
8 & Bābur & PG \#44608 & 422,772 & Central Asia (Fergana) & 15th--16th
c. \\
9 & Yung Wing & PG \#54635 & 66,459 & China & 19th--20th c. \\
10 & Benvenuto Cellini & PG \#4028 & 190,390 & Italy (Florence) & 16th
c. \\
11 & Bernal Díaz del Castillo & PG \#32474 & 187,315 & Spain (Castile) &
15th--16th c. \\
12 & Georg Ebers & PG \#5599 & 96,174 & Germany & 19th c. \\
13 & Jean-Jacques Rousseau & PG \#3913 & 278,120 & Switzerland (Geneva)
& 18th c. \\
14 & Saint Augustine & PG \#3296 & 114,873 & North Africa (Numidia) &
4th--5th c. \\
\bottomrule
\end{longtable}

Constraints on the generalizability of the 14-subject sample (language,
era, cultural framing, Project Gutenberg curation bias, individual
self-presentation in autobiography) are discussed in \hyperref[subject-sample]{§6.1}.

\hypertarget{pretraining-coverage-variance-high-vs-low-baseline}{%
\subsection{3.4.1 Pretraining-coverage variance (high vs low
baseline)}\label{pretraining-coverage-variance-high-vs-low-baseline}}

Pretraining coverage of a specific person varies widely across the 14
main-study subjects, even within a sample whose autobiographies are of
comparable provenance. We use the C5 baseline score (the No-Context
Baseline mean on the 1-5 rubric defined in \hyperref[scoring-rubric-with-calibrated-llm-judge-panel]{§3.3}) as the observable proxy
for pretraining coverage. Operationally, the proxy works as follows: the
response model is given the behavioral-prediction battery for a subject
with no external context (no Spec, no facts, no retrieval), and the
resulting per-subject mean across the 39 questions is read as how well
the model already knows that person from training data alone. A high C5
mean implies substantial pretraining representation; a low C5 mean
implies very little.

\begin{longtable}[]{@{}
  >{\raggedright\arraybackslash}p{(\columnwidth - 4\tabcolsep) * \real{0.33}}
  >{\raggedright\arraybackslash}p{(\columnwidth - 4\tabcolsep) * \real{0.33}}
  >{\raggedright\arraybackslash}p{(\columnwidth - 4\tabcolsep) * \real{0.33}}@{}}
\toprule
Baseline band & Subjects & Count \\
\midrule
\endhead
≤ 2.0 (low-baseline) & Sunity Devee, Ebers, Hamerton, Fukuzawa, Seacole,
Bernal Díaz, Keckley, Yung Wing, Bābur & 9 \\
2.0--3.0 (mid-baseline) & Cellini, Zitkala-Ša, Rousseau, Augustine,
Equiano & 5 \\
\textgreater{} 3.0 (high-baseline) & Benjamin Franklin (known-figure
control, not in main study) & 1 \\
\bottomrule
\end{longtable}

The higher a subject sits on this distribution (e.g., Equiano,
Zitkala-Ša at the upper end of the main study), the better the model
already knows them from pretraining; conversely, the lower they sit
(Sunity Devee, Ebers), the less the model knows. Nine of the 14
main-study subjects fall into the low-baseline band; five into the
mid-baseline band.

\textbf{Franklin as a known-figure control.} Benjamin Franklin (Project
Gutenberg \#20203) is included as a high-baseline reference point.
Franklin's \emph{Autobiography} is one of the most widely available and
frequently cited autobiographies in American public-domain literature,
and the model's baseline score on Franklin (3.77 on the 5-judge primary
panel; 4.10 on Haiku alone) is consistent with substantial pretraining
coverage of both the person and the specific text.\footnote{Franklin's
  legacy run pre-dates the gemini\_flash / gemini\_pro split, so only
  one Gemini judge is present in the legacy data. The per-judge range
  cited here is the min and max across all available judges, rather than
  a clean 7-judge panel mean.} Franklin
is used to anchor what the high-baseline end of the spectrum looks like
(\hyperref[statistical-rigor-checks-on-the-headline-gradient]{§4.6.4} develops the high-baseline behavior), not as a subject whose
pretraining coverage is a design target of the Behavioral Specification.

\textbf{Cross-provider variance.} No-context baseline scores on the same
subject vary by 1--2 points across response models. Different providers
know different amounts about the same historical figure (Tier 2
cross-provider data in \hyperref[cross-provider-response-generation-tier-2-replication]{§4.6.1}). The pretraining-coverage variance
documented in this section is a property of each model family, not a
property of the subject alone. The cross-provider spread in No-Context
Baseline scores for a given subject is comparable in magnitude to the
lift the Specification produces, which is why the response model is held
constant across all main-study conditions.

\hyperref[the-cross-subject-gradient-and-its-per-question-mechanism]{§4.1} develops this distribution into the cross-subject gradient.

\hypertarget{question-battery-formation}{%
\subsection{3.5 Question battery
formation}\label{question-battery-formation}}

Each subject's behavioral prediction battery is generated by a
backward-design process: an LLM reads a passage from the held-out half
of the corpus, writes a question whose answer is the behavioral pattern
implicit in the passage, and avoids naming any detail unique to the
passage itself. The question can be attempted from training-text
patterns alone; the verbatim held-out passage is the ground truth for
scoring.

The procedure, in order:

\begin{enumerate}
\def\labelenumi{\arabic{enumi}.}
\item
  \textbf{Split the corpus.} Each subject's source text is divided by
  chapter into \texttt{training.txt} (50\% of the corpus) and
  \texttt{heldout.txt} (50\%). Held-out text is never shown to a
  response model.
\item
  \textbf{Sample from held-out.} A sliding window moves across the
  held-out text in four batches of ten questions each, using
  5,000-character windows as local context.
\item
  \textbf{Backward-design question generation.} Claude Haiku 4.5
  (temperature 0) reads each held-out window and writes a question whose
  answer requires the subject's behavioral patterns observable in the
  training half. For each window, Haiku:

  \begin{itemize}
  \tightlist
  \item
    Extracts a verbatim ground-truth span from the held-out window.
  \item
    Avoids named-entity or specific-date leakage in the question stem.
  \item
    Targets one of 10 fixed behavioral-prediction categories: decisions,
    values, relationships, conflict, learning, risk, creativity, stress,
    career, and change-over-time.
  \end{itemize}
\item
  \textbf{Supplementary tiers (not scored in \hyperref[results]{§4}).} Beyond the
  behavioral-prediction tier, each battery additionally contains
  supplementary question tiers that test different competencies:

  \begin{itemize}
  \tightlist
  \item
    \textbf{Recall} tests whether a system can retrieve specific factual
    content.
  \item
    \textbf{Adversarial-abstention} presents scenarios the source corpus
    does not cover, testing whether a system correctly refuses to
    predict rather than fabricates.
  \end{itemize}

  These supplementary tiers are included in the battery files for future
  analysis but do not enter the \hyperref[results]{§4} main results.
\item
  \textbf{Dedup and freeze.} Deduplication on lowercased question text,
  cap at target counts per category, MD5 checksum of the final battery.
  Downstream response and judgment files are invalidated if the battery
  checksum changes.
\end{enumerate}

Each main-study subject receives 39 behavioral prediction questions; the
14 main-study batteries total 546 questions.\footnote{Franklin's
  40-question legacy battery (high-baseline reference, \hyperref[pretraining-coverage-variance-high-vs-low-baseline]{§3.4.1}) brings
  the data-structural total to 586 questions across 15 subjects.} Each
battery covers 8 to 10 of the 10 categories. Definitions, example
questions, and per-subject distributions are in Appendix B.1 and B.2.

Within the behavioral-prediction tier, individual questions vary in
interpretive demand: some can be answered from facts alone, others
require applying behavioral patterns to novel scenarios. This variation
is decomposed in \hyperref[where-the-spec-helps-where-it-hurts-and-which-question-types-route-to-each]{§4.4.3}.

\textbf{Leakage audit.} We empirically checked the no-leakage principle
by searching every behavioral-prediction question for any sequence of
seven or more consecutive words that appears verbatim in that subject's
held-out corpus. Across the 14 main-study subjects (546 questions), zero
questions leak (true zero, not a rounded value).\footnote{Franklin's
  40-question legacy battery, included as a high-baseline reference but
  not part of the main study, has 2 leaking questions (Q49, Q56). Both
  predate the backward-design constraint and were hand-authored.
  Aggregate across the full 586-question pool: 2 leaks (0.34\%).}

\textbf{One false-premise outlier.} One item out of 586 (Zitkala-Ša Q18)
carries a factually-wrong presupposition; the aggregate is unaffected.
The broader need for a human-reviewed quality gate on automated
batteries is flagged in \hyperref[measurement-apparatus]{§6.2}.

Battery files and the leakage-audit script are in the public
repository.\footnote{Per-subject batteries at
  \texttt{results/global\_\textless{}subject\textgreater{}/battery\_v2.json}
  (13 global subjects); Hamerton and Franklin legacy at
  \texttt{data/\textless{}subject\textgreater{}/battery.json};
  GPT-5.4-regenerated batteries (used in the \hyperref[circularity-controls]{§3.5.1} circularity control)
  at
  \texttt{results/global\_\textless{}subject\textgreater{}/battery\_gpt54.json};
  leakage-audit script at
  \texttt{scripts/\allowbreak{}\_\allowbreak{}verify\_\allowbreak{}battery\_\allowbreak{}leakage.\allowbreak{}py}.}

\hypertarget{circularity-controls}{%
\subsection{3.5.1 Circularity controls}\label{circularity-controls}}

The pipeline and the batteries both use Anthropic models for several
roles (Haiku for extraction and battery generation, Sonnet for
authoring, Opus for composition, Haiku as the primary response model,
plus Sonnet and Opus on the judge panel). Two independent controls rule
out a within-Anthropic frontier-model artifact; full per-subject results
are reported in \hyperref[cross-provider-response-generation-tier-2-replication]{§4.6.1}.

\textbf{Control 1: Independent battery regeneration.} Batteries for all
13 global subjects were re-generated with GPT-5.4 using the identical
backward-design prompt. Categorical-emphasis differences are minor; the
methodology constrains the output more than the generating model does.

\textbf{Control 2: Non-Anthropic response chain.} The core
conditions\footnote{C5 No-Context Baseline, C2a Spec alone, C4a
  facts-plus-Spec, C2c wrong-Spec control. Full condition definitions in
  \hyperref[experimental-conditions]{§3.2}.} were re-run on three subjects (Ebers, Yung Wing, Zitkala-Ša)
using two non-Anthropic response models (Claude Sonnet 4.6 and Google
Gemini 2.5 Pro) reading the GPT-5.4-regenerated batteries. The
combination tests whether the Spec effect survives when both the
response model and the battery-generation model sit outside the
Anthropic family.

The broader LLM-as-judge circularity is discussed as an open limitation
in \hyperref[measurement-apparatus]{§6.2}.\footnote{Raw battery regeneration data at
  \texttt{results/global\_\textless{}subject\textgreater{}/battery\_gpt54.json}
  for all 13 global subjects; Tier 2 response and judgment files for the
  three Control 2 subjects in the same per-subject directories.}

\hypertarget{response-models}{%
\subsection{3.6 Response models}\label{response-models}}

\textbf{Tier 1 (main study): Claude Haiku 4.5 as the primary response
model, run across all 14 subjects and every condition in the main
matrix.} Haiku was chosen as a deliberately weaker baseline: a Spec
effect that registers on a relatively weak response model is harder to
attribute to the model's pretraining alone, which gives a conservative
readout of effect direction. \hyperref[cross-provider-response-generation-tier-2-replication]{§4.6.1} Tier 2 cross-provider probe tests
whether the direction reproduces on stronger response models.

\textbf{Tier 2 (cross-provider response generation).} A sensitivity
probe using non-Anthropic response models on a subset of subjects;
methodology and results are reported together in \hyperref[cross-provider-response-generation-tier-2-replication]{§4.6.1}.

\textbf{Prompt schema.} A single shared prompt is used across every
condition. The system message frames the task as behavioral prediction
of a specific person; the user message is the question plus whichever
context inputs the condition specifies (\hyperref[experimental-conditions]{§3.2}). Nothing about the prompt
changes per condition beyond the injected context block.\footnote{All
  response models are called with \texttt{temperature=0} and
  \texttt{max\_tokens=1024}.}

\begin{framed}
\begin{verbatim}
System: You are predicting how <subject> would respond to a specific
        question about their behavior, values, or reasoning. Answer
        in <subject>'s voice, grounded in their demonstrated patterns.

User:   <context block, one of: empty (C5), Spec (C2a), wrong-Spec (C2c),
         facts (C4), facts + Spec (C4a), corpus (C8), corpus + Spec
         (C9), or retrieval ± Spec (C1 / C3)>

        Question: <question text>
\end{verbatim}
\end{framed}

The prompt is deliberately uniform and deliberately
faithfulness-oriented. The system message asks the model to ground its
answer in the subject's demonstrated patterns and to answer in the
subject's voice; we chose this framing because the study tests whether
the served context lets the model do exactly that. No instruction tells
the model to abstain, hedge, or commit; the model's natural
refusal-or-commitment pattern given a specific context is itself part of
the phenomenon the study tests, and \hyperref[mechanism-correct-content-not-format]{§4.3} reports the hedging-rate shift
across conditions as a substantive finding rather than a behavior to
suppress.

\textbf{Two effects of the faithfulness framing.} Asking for the
subject's voice may favor systems that surface verbatim corpus passages
over systems that surface compressed interpretive structure (Letta's
stateful-agent path is one such system, examined post-hoc in Appendix
G). Asking the model to ground in demonstrated patterns may push it
toward abstention when the served context underdetermines a confident
answer (visible in the Spec-induced refusal cases in \hyperref[where-the-spec-helps-where-it-hurts-and-which-question-types-route-to-each]{§4.4.3}). Because
the framing is identical across every condition, it cannot be the source
of the differential effect the study measures; a neutral-prompt
robustness check is flagged as future work in \hyperref[future-work]{§7}.

\textbf{With the framing held fixed across every condition, what the
conditions vary is the source of the patterns the model works from.}

\begin{longtable}[]{@{}
  >{\raggedright\arraybackslash}p{(\columnwidth - 2\tabcolsep) * \real{0.50}}
  >{\raggedright\arraybackslash}p{(\columnwidth - 2\tabcolsep) * \real{0.50}}@{}}
\toprule
Condition & Pattern source served to the model \\
\midrule
\endhead
C5 (no context) & The model's pretraining alone \\
C2a (Spec only) & Pre-extracted patterns served as the Behavioral
Specification \\
C4 / C8 (facts / corpus) & Raw material; the model must identify
patterns at runtime \\
C4a / C9 (facts + Spec / corpus + Spec) & Pre-extracted patterns
alongside raw material \\
C2c (wrong-Spec) & Pre-extracted patterns from a different person \\
\bottomrule
\end{longtable}

The empirical evidence that the prompt is not the lift mechanism (the
C2a vs.~C4 / C8 comparison and the C2c wrong-Spec control) is reported
in \hyperref[mechanism-correct-content-not-format]{§4.3}.

Exact model identifiers, full prompt text, and Tier 2 invocation
parameters are in Appendix C.\footnote{Run scripts at
  \texttt{scripts/\allowbreak{}run\_\allowbreak{}global\_\allowbreak{}subjects.\allowbreak{}py},
  \texttt{scripts/\allowbreak{}run\_\allowbreak{}full\_\allowbreak{}study.\allowbreak{}py},
  \texttt{scripts/\allowbreak{}run\_\allowbreak{}multimodel\_\allowbreak{}responses.\allowbreak{}py}. Raw response files at
  \texttt{results/global\_\textless{}subject\textgreater{}/results\_v2.json}
  (13 globals), \texttt{results/\allowbreak{}hamerton/\allowbreak{}results.\allowbreak{}json},
  \texttt{results/\allowbreak{}franklin\_\allowbreak{}legacy\_\allowbreak{}20260411/\allowbreak{}results.\allowbreak{}json} (used for the
  \hyperref[the-gradient-at-the-high-baseline-end-franklin-reference]{§4.1.2} reference numbers), and \texttt{results/\_tier2/} for Tier 2
  runs. The supplemental rerun at
  \texttt{results/\allowbreak{}franklin/\allowbreak{}fullstack\_\allowbreak{}haiku.\allowbreak{}json} (April 2026, 5-judge
  scoring) is preserved for diagnostic use but is not the source of the
  paper's headline Franklin numbers.}

\hypertarget{pipeline-for-the-behavioral-specification}{%
\subsection{3.7 Pipeline for the Behavioral
Specification}\label{pipeline-for-the-behavioral-specification}}

\textbf{A Behavioral Specification is a structured document encoding a
person's behavioral patterns.} It is composed of three interpretive
layers (anchors, core, predictions) plus a composed unified brief; total
size per subject is approximately 7,000 tokens (\textasciitilde5,000
words, about the length of a short magazine article). Every
Specification evaluated in this paper is produced by \textbf{Base
Layer}, an open-source reference implementation of the pipeline; Base
Layer is included in the study as one of the memory-system comparison
conditions (\hyperref[question-battery-formation]{§3.5}, \hyperref[memory-system-composition]{§4.4}), alongside the four commercial systems (Mem0,
Letta, Supermemory, Zep). The pipeline transforms raw source text in
five steps: import, extract, embed, author, and compose. Each step is a
single script backed by a single model choice.

\begingroup\footnotesize
\begin{longtable}[]{@{}
  >{\raggedright\arraybackslash}p{(\columnwidth - 6\tabcolsep) * \real{0.13}}
  >{\raggedright\arraybackslash}p{(\columnwidth - 6\tabcolsep) * \real{0.32}}
  >{\raggedright\arraybackslash}p{(\columnwidth - 6\tabcolsep) * \real{0.22}}
  >{\raggedright\arraybackslash}p{(\columnwidth - 6\tabcolsep) * \real{0.33}}@{}}
\toprule
Step & Input & Tool / model & Output \\
\midrule
\endhead
1. Import & ChatGPT / Claude exports, journals, plain text, directories
& \texttt{import\_conversations.py} & A local database holding the
cleaned, de-duplicated source text \\
2. Extract & Canonical source text & \texttt{extract\_facts.py}, Claude
Haiku 4.5, 46-predicate vocabulary (full list in
\protect\hyperlink{appendix-a.-predicate-vocabulary}{Appendix A}) &
Behavioral patterns extracted as short structured statements (e.g.,
``avoids confrontation,'' ``values craft over speed''), with bookkeeping
operations to add new patterns, update existing ones, delete
contradicted ones, or skip duplicates \\
3. Embed & All imported message text & \texttt{embed.py},
\texttt{all-MiniLM-L6-v2}, ChromaDB & A searchable index of source
passages. Combined with the source-message IDs each extracted fact
carries, this lets any claim in the final specification be traced back
to the passages that support it \\
4. Author & Extracted facts & \texttt{author\_layers.py}, Claude Sonnet
4.6 & Three interpretive layers as markdown (anchors, core,
predictions); see body below for layer-by-layer examples. Each layer is
produced from facts alone, not from prior layer output. Each layer
prompt includes a domain guard that prevents topic skew
(ablation-validated in prior pilot work). \\
5. Compose & The three authored layers (plus a sample of identity-tier
facts as supplementary context) & \texttt{agent\_pipeline.py}, Claude
Opus 4.6 & Unified behavioral brief in flowing prose; see body below for
what the brief contains and why it exists \\
\bottomrule
\end{longtable}
\endgroup

The Behavioral Specification served as context in experimental
conditions is the three authored layers concatenated with the composed
brief, not the brief alone.

The extract step constrains output through a fixed vocabulary of 46
behavioral predicates.\footnote{Examples: \texttt{avoids},
  \texttt{repeatedly\ engages\ in}, \texttt{refuses\ to},
  \texttt{values}, \texttt{fears}, \texttt{has\ experienced}. The full
  predicate list is in
  \protect\hyperlink{appendix-a.-predicate-vocabulary}{Appendix A}.} The
vocabulary is human-curated and was validated across 50+ pilot subjects
before being frozen for the study. The constrained vocabulary is the
main lever the pipeline uses to push extraction away from biographical
facts (``his father was violent'') and toward behavioral patterns
(``evaluates authority figures on dual criteria of virtue and
failure'').

Each of the authored layers has distinct jobs. Each layer has a
characteristic format; examples below are drawn from three different
subject specifications (Sunity Devee, Bernal Díaz, and Augustine).

\textbf{Anchors} encode the subject's load-bearing axioms in numbered
form (A1, A2, \ldots), each with an activation condition and a
false-positive warning. Example from Sunity Devee:

\begin{quote}
\emph{A1. DIVINE PRIMACY. All events, decisions, and outcomes are
understood as expressions of divine providence first; social, political,
or material explanations are secondary framings applied afterward.
Active when: life decisions, loss, political events, marriage, reform,
or any outcome described as fortunate or unfortunate are discussed.}
\end{quote}

\textbf{Core} captures values, beliefs, and self-view in flowing prose.
It is the layer that reads most like an essay about the person. Example
from Bernal Díaz:

\begin{quote}
\emph{They reason from direct witness and lived participation, treating
firsthand account as the only reliable foundation for any claim. When
evidence is secondhand or reconstructed, they flag it rather than smooth
it over. They distrust narrators who were absent from the events they
describe, and this distrust shapes how they evaluate any source placed
before them.}
\end{quote}

\textbf{Predictions} are explicit behavioral predicates (P1, P2, \ldots)
with detection criteria, directives, and false-positive warnings.
Example from Augustine:

\begin{quote}
\emph{P1. CONFESSION BEFORE CONCLUSION. When asked to account for a past
failure, does not defend, minimize, or contextualize first. Moves
immediately into detailed reconstruction of the failure, naming the
pleasure or pride taken in the transgression before offering resolution.
Detection: a relational conflict where the subject was at fault, a
professional misjudgment, a moment of intellectual dishonesty.
Directive: hold space for the full weight of the confession; resist
premature resolution. False positive: not active when the subject is
analyzing someone else's failure.}
\end{quote}

\textbf{The unified brief.} The brief serves a dual purpose: a coherent
first pass for human readers, and an integration step that implicitly
weaves the three authored layers (anchors, core, predictions) together
(layered files alone do not require this integration; internal testing
suggests the integration changes how a model uses the Specification).
The compose step produces it as a continuous prose synthesis in the
third person, similar in length to a short profile of the subject. The
Behavioral Specification served to the response model in C2a / C4a / C3
conditions is the three layers plus the unified brief together; the
brief is not a substitute for the layers but an integration on top of
them. A formal ablation isolating brief-with-layers vs.~layers-only is
flagged in
\protect\hyperlink{specification-design-and-composition}{§7.3
Specification design and composition}.

Total pipeline cost is approximately \$1 per subject (estimated from
per-step API token counts) to process a 50,000- to 150,000-word
autobiography end to end. Pipeline code, the full predicate vocabulary,
and example specifications for all 14 study subjects are available in
the public repository (see
\protect\hyperlink{data-code-and-reproducibility}{§8 Data, code, and
reproducibility}).

\begin{center}\rule{0.5\linewidth}{0.5pt}\end{center}

\hypertarget{results}{%
\section{4. Results}\label{results}}

Across 14 historical subjects, adding a Behavioral Specification (a
short structured document describing how a specific person reasons and
behaves) measurably improves how accurately a language model represents
that person's behavioral patterns. We measure this with a battery of
behavioral prediction questions based on held-out ground-truth text from
each subject's publicly available autobiography. We score each
prediction on a 1-to-5 rubric where a whole-point shift marks a
categorical change in how the response aligns with the subject's
documented behavior.

On the 9 low-baseline subjects (those the model does not know well), the
Specification produces a mean per-subject increase of \textbf{+0.89
points} and lifts individual responses by one category or more on
\textbf{55.0\% of questions}. The Specification's added value on top of
other context types (facts, raw corpus, or memory-system retrieval)
concentrates on interpretation-heavy questions; on factual-recall
questions, retrieval alone is often sufficient and the Specification
adds little or actively degrades the response. On high-baseline subjects
(those the model does know well, such as Benjamin Franklin), the
Specification adds little or mildly hurts across conditions. Control
conditions, statistical robustness checks, and sensitivity analyses
confirm that the Specification categorically shifts how a language model
responds, increasing its representational accuracy of the subject beyond
what fact-based retrieval supplies.

The seven parts of \hyperref[results]{§4} establish this picture in detail:

\begin{itemize}
\tightlist
\item
  \hyperref[the-cross-subject-gradient-and-its-per-question-mechanism]{§4.1} the cross-subject gradient (the primary result across 14
  subjects)
\item
  \hyperref[compression-structure-vs.-raw-text]{§4.2} compression of structure versus raw text
\item
  \hyperref[mechanism-correct-content-not-format]{§4.3} content versus format as the mechanism
\item
  \hyperref[memory-system-composition]{§4.4} memory-system composition
\item
  \hyperref[exploratory-case-study-letta-stateful-agent-n3-post-hoc]{§4.5} the Letta stateful-agent exploratory case study (Appendix G, N=3,
  not a headline finding)
\item
  \hyperref[robustness-and-sensitivity]{§4.6} robustness and sensitivity (cross-provider, judge-panel, battery
  composition, wrong-Spec derangement, retrieval overlap, post-hoc
  rubric-handling audit)
\item
  \hyperref[summary-of-4]{§4.7} the synthesis bridge to \hyperref[discussion]{§5}
\end{itemize}

Every number in \hyperref[results]{§4} uses the 5-judge primary aggregate defined in
\hyperref[calibration]{§3.3.3}.\footnote{Primary panel: Haiku 4.5, Sonnet 4.6, Opus 4.6, GPT-4o,
  GPT-5.4. The 7-judge sensitivity check adds Gemini 2.5 Flash and
  Gemini 2.5 Pro and is reported in \hyperref[robustness-and-sensitivity]{§4.6}.} Score deltas are read through
the anchor-crossing rule from \hyperref[score-interpretation]{§3.3.1}: a delta that crosses a rubric
integer anchor is a stronger claim than one that stays within a single
anchor.

\hypertarget{the-cross-subject-gradient-and-its-per-question-mechanism}{%
\subsection{4.1 The cross-subject gradient and its per-question
mechanism}\label{the-cross-subject-gradient-and-its-per-question-mechanism}}

\textbf{Hypotheses tested in this section} (from \hyperref[what-we-tested]{§1.2}): H1. Adding the
Specification improves prediction. H2. The effect is inversely
proportional to the response model's pretraining coverage. Corollary to
H2: on high-baseline subjects, the Specification does not add value and
mildly interferes.

\begin{center}\rule{0.5\linewidth}{0.5pt}\end{center}

\textbf{The cross-subject gradient.} The less the model already knows
about a subject from pretraining, the more a Behavioral Specification
improves the model's representational accuracy of that subject. It
operates as an interpretive layer over facts and retrieved context, not
a replacement for them. On the 9 subjects whose pretraining baseline
sits at or below 2.0 on the 1-5 rubric (the population of relevance from
\hyperref[pretraining-coverage-variance-high-vs-low-baseline]{§3.4.1}), adding the Spec consistently improves prediction: every one of
the 9 improves over the No-Context Baseline (C5); none declines. The
per-subject mean lift is \textbf{+0.69} for Spec Only (C2a) and
\textbf{+0.89} for All Facts + Spec (C4a). Adding the Spec on top of All
Facts (C4), Raw Corpus (C8), or memory-system retrieval produces
additional aggregate gains that are smaller in magnitude than the
Spec-vs-baseline lift (detail in \hyperref[compression-structure-vs.-raw-text]{§4.2} and \hyperref[memory-system-composition]{§4.4}). Spec Only does not
score higher than All Facts or Raw Corpus alone; the Spec's value is in
the layering.

\begin{figure}
\centering
\includegraphics{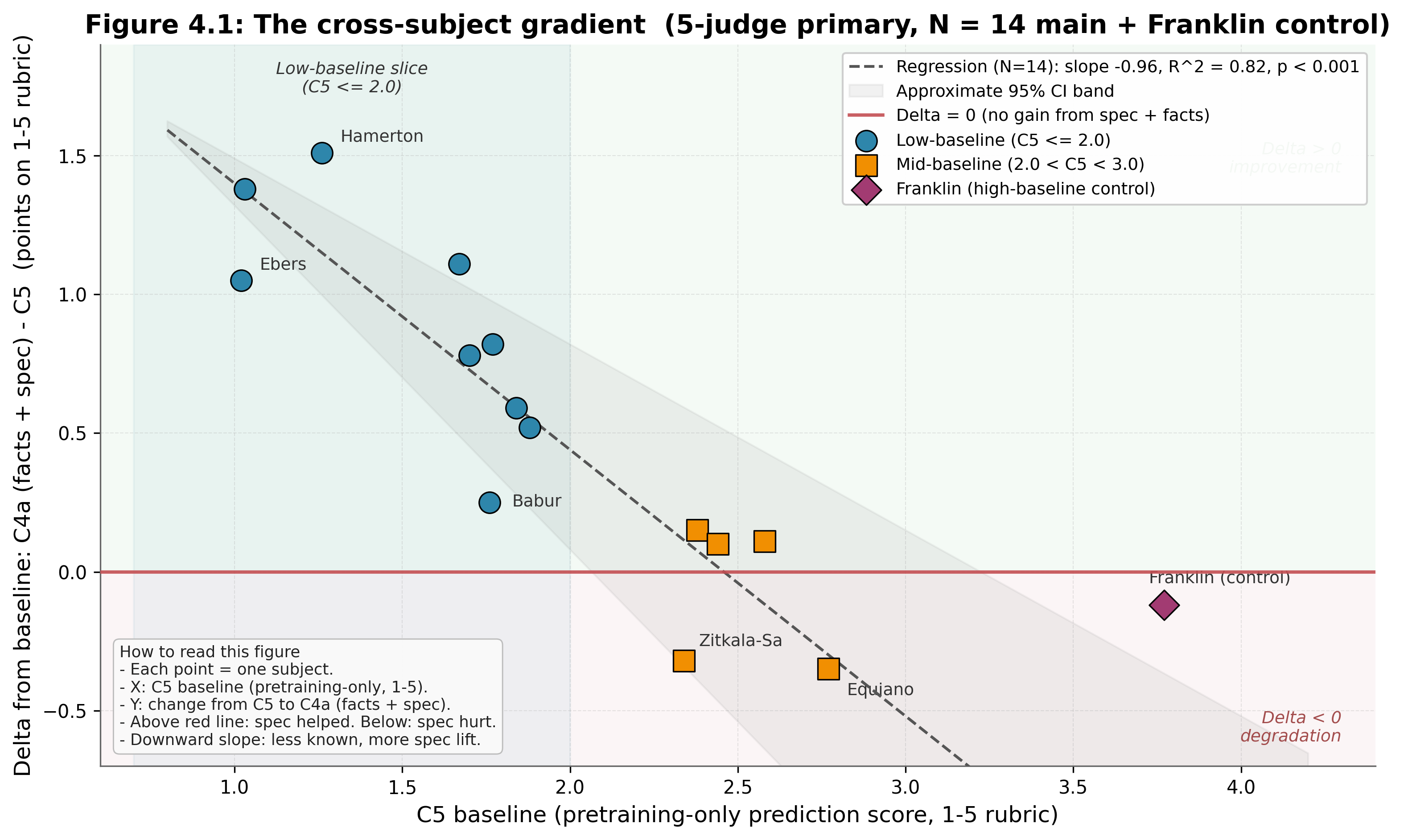}
\caption[Reading the gradient]{\textbf{Reading the gradient.} Figure 4.1 plots each
subject's No-Context Baseline (C5) against the lift the Specification
produces over that baseline (Δ\_C4a, where C4a is the All Facts + Spec
condition). The 9 low-baseline subjects (C5 ≤ 2.0) cluster in the
upper-left of the plot with positive lifts ranging from Bābur at +0.25
(smallest lift) to Hamerton at +1.51 (largest). Franklin sits in the
lower-right at C5 = 3.77, Δ = −0.13: the high-baseline reference where
the model already knows the subject from pretraining. The regression
slope of −0.96 captures this gradient: the lower the model's pretraining
baseline on a subject, the larger the lift the Specification produces.
Across this question mix, the per-subject C4a mean clusters around 2.44,
so the lift in rubric points is largest where the baseline shortfall is
largest.\footnotemark{} (\hyperref[the-cross-subject-gradient-and-its-per-question-mechanism]{§4.1})}
\end{figure}
\footnotetext{Two regression slopes capture the gradient. The level
  slope of C4a on C5 is essentially flat (+0.04, R² = 0.008): the Spec
  produces roughly uniform quality regardless of where each subject's
  baseline starts. The change-score slope of Δ\_C4a on C5 is strongly
  negative (−0.96), but most of that magnitude is mechanical: slope\_Δ =
  slope\_level − 1 by identity, so a flat level slope automatically
  produces a near-−1.0 change-score slope. Full technical sensitivity
  detail in \hyperref[battery-composition-sensitivity]{§4.6.3}.}

\begin{center}\rule{0.5\linewidth}{0.5pt}\end{center}

\textbf{Adding a Behavioral Specification changes the category of answer
the response model produces.} Of the 351 individual responses in the
low-baseline band, \textbf{55.0\% crossed at least one rubric integer
anchor upward when the Specification was added}. Multi-anchor jumps of
two or more anchors (e.g., 1→3, 2→4) appear in \textbf{18\%} of
low-baseline questions on the Spec conditions, with about \textbf{6\%}
being extreme jumps of three or more anchors (e.g., 1→4, 2→5, 1→5).
These extreme jumps concentrate on interpretation-heavy questions: the
No-Context Baseline response refuses or stays generic, and the
Specification supplies the behavioral pattern the model could not
retrieve from training data. The response model's answer moved from one
category of response to a qualitatively different category. These are
the multi-anchor jumps at the margin the aggregate mean understates.

\begin{center}\rule{0.5\linewidth}{0.5pt}\end{center}

\textbf{Population of relevance for AI personalization.} Most users sit
in the low-baseline category. AI users whose private reasoning is not in
any training corpus fall at or near the rubric floor by construction, so
the lift the Spec produces is largest exactly where the use case is most
common. The Spec works for any subject regardless of pretraining
coverage and is therefore portable across the long tail of users (\hyperref[what-this-implies]{§1.4},
\hyperref[why-the-gradient-is-the-load-bearing-finding]{§5.2}).

\textbf{Per-question rubric transitions when the Specification is added
(low-baseline band, C5 → All Facts + Spec, 351 paired questions).}

\begin{longtable}[]{@{}lrl@{}}
\toprule
Transition & \% of responses & Description \\
\midrule
\endhead
1 → 2 & \textbf{33.3\%} & Refusal or off-base → generic engagement with
the question \\
1 → 3 & 12.3\% & Refusal → partially-aligned prediction \\
1 → 4 & 4.8\% & Refusal → substantively-aligned prediction \\
1 → 5 & 0.9\% & Refusal → fully matches the held-out pattern \\
2 → 3 & 2.0\% & Generic → subject-specific \\
2 → 4 & 0.3\% & Generic → substantively-aligned \\
3 → 4 & 1.4\% & Partially → substantively-aligned \\
No upward crossing & 38.2\% & Delta stayed inside a single anchor \\
Downward crossing & 6.8\% & Specification hurt the response \\
\bottomrule
\end{longtable}

When the Specification is added, one of every three low-baseline
responses moves from ``cannot engage'' to actual engagement. Another one
in five makes a larger jump. Only one response in fifteen gets worse.
Worked examples appear below.

\begin{center}\rule{0.5\linewidth}{0.5pt}\end{center}

\textbf{Three representative examples below show distinct mechanisms by
which the Specification improves the response.}\footnote{Each example is
  extracted from the cached data (\texttt{battery\_v2.json},
  \texttt{results\_v2.json}, \texttt{judgments\_v2.json}) via
  \texttt{scripts/\allowbreak{}v2\_\allowbreak{}canonical\_\allowbreak{}cell\_\allowbreak{}extractor\_\allowbreak{}20260508.\allowbreak{}py}, which
  joins the question, held-out, response, and per-judge scores for each
  (subject, qid, condition) cell. Methodology note on v1/v2 data canonicalisation.} Hedge reduction is common but not the
only mechanism. The Specification also corrects directionally-wrong
predictions and enables interpretive inference from character patterns
when retrieved facts are insufficient.

\hypertarget{worked-examples-wrong-referent-fukuzawa-directional-bernal-duxedaz-abstention-seacole}{%
\subsection{Worked examples: Wrong-Referent (Fukuzawa), Directional
(Bernal Díaz), Abstention
(Seacole)}\label{worked-examples-wrong-referent-fukuzawa-directional-bernal-duxedaz-abstention-seacole}}

\begin{quote}
\textbf{Example A. All Facts → All Facts + Spec: Wrong-Referent
Correction}

\textbf{Subject:} Yukichi Fukuzawa. \textbf{Q35:} \emph{``How would
Fukuzawa characterize someone who studied naval arts under the Dutch and
later became instrumental in preventing military conflict?''}

\textbf{Held-out (the figure the question is about):} \emph{``He is best
remembered as the man who saved Yedo from siege when the Imperial forces
were marching upon it in 1868.''} The held-out names Katsu Kaishu, the
naval officer who later prevented the bombardment of Edo through
diplomatic surrender.

\textbf{C4 All Facts (mean 1.00):} \emph{``Based on the facts provided,
Fukuzawa would likely characterize such a person with high regard and
respect\ldots{} The most relevant figure is \textbf{Captain Kimura
Settsu-no Kami}, who commanded the Kanrin-maru. Fukuzawa would
characterize him positively because: Competence and Pride: Kimura had
studied navigation under the Dutch and successfully commanded the
trans-Pacific voyage\ldots{}''}

The AI picks the wrong person. Captain Kimura is also a naval officer
who studied under the Dutch, but he is not the figure the held-out is
about. The model matched on a surface descriptor (``naval officer who
studied under Dutch'') and grabbed the most-mentioned naval officer in
the facts. Five judges score this at the rubric floor, anchor 1
(off-base): the response engages the question fluently but identifies
the wrong subject.

\textbf{C4a facts + Spec (mean 4.20):} \emph{``This describes
\textbf{Katsu Rintaro}, who: studied navigation under Dutch officers in
Nagasaki (post-1855); served as second-in-command on the Kanrin-maru's
trans-Pacific voyage; later became instrumental in negotiating the
peaceful surrender of Edo Castle (preventing the final Boshin War
battle)\ldots{} Fukuzawa would credit him for earned expertise, not
inherited rank\ldots{} but Fukuzawa would probably regard Katsu as a
capable man constrained by the very system Fukuzawa despised --- someone
who used Western knowledge skillfully but remained ultimately loyal to
institutional authority rather than to independent principle.''}

The Spec moves the AI from the wrong referent to the right one (Katsu
Rintaro = Katsu Kaishu). It characterizes him through Fukuzawa's
reasoning patterns: respect for demonstrated capability paired with
skepticism toward institutional embedding. Five judges register the
move; mean shifts from 1.00 to 4.20, a three-anchor jump from
``off-base'' to ``general direction correct.''

\textbf{What the Spec did.} The headline is fine-grained
differentiation. The facts list both Captain Kimura (the captain of the
Kanrin-maru) and Katsu Rintaro (the second-in-command). Both fit the
surface descriptor ``naval officer who studied under the Dutch''; both
appear repeatedly in Fukuzawa's autobiography. Picking between them
requires reasoning about which figure Fukuzawa would actually
characterize for the specific behavior in the question, not just which
figure matches the descriptor. The Spec encodes Fukuzawa's preference
for earned over inherited authority, his skepticism of institutional
rank, and his attention to demonstrated capability. These are the exact
axes that distinguish a captain by appointment from a subordinate by
skill. Without the Spec, the model defaults to the most-mentioned
candidate (Kimura) and matches the descriptor; with the Spec, it filters
by behavioral pattern and resolves to the right person (Katsu). The
interpretive layer is what lets the model choose the second-in-command
over the captain when the question is about Fukuzawa's characterization,
not the facts.
\end{quote}

\begin{quote}
\textbf{Example B. All Facts → All Facts + Spec: Directional Correction}

\textbf{Subject:} Bernal Díaz del Castillo. \textbf{Q16:} \emph{``When
the commander is offered physical assistance during a strenuous
activity, what behavior would the author expect from him?''}

\textbf{Held-out:} \emph{``There were 114 steps to the summit, and, as
they feared that Cortes would experience the same fatigue in mounting as
Motecusuma had, they were going to assist him by taking hold of his
arms. Cortes, however, would not accept of their proffered aid.''}

\textbf{C4 All Facts (mean 2.00):} \emph{``Based on the facts provided
about Bernal Díaz del Castillo and his observations of Cortes, the
author would expect the commander to \textbf{accept and participate} in
the assistance while maintaining leadership and setting an example.''}

The AI gives a confident, fluent answer that gets the direction wrong:
it predicts Cortés would accept help. Five judges agree: unanimous 2
(``wrong prediction''). The response engages the question and produces a
specific prediction, but the prediction is the opposite of what the
held-out shows.

\textbf{C4a facts + Spec (mean 4.80):} \emph{``Based on the behavioral
specification, when the commander is offered physical assistance during
a strenuous activity, the author would expect \textbf{Cortes to refuse}
the assistance and perform the labor himself --- particularly if it
serves a symbolic or leadership purpose\ldots{} The author notes that
when Cortes performs the labor, `The caziques and all present followed
Cortes's example.' This is not accidental --- it's the load-bearing
mechanism of how Cortes exercises power.''}

The Spec flips the prediction. The AI now correctly anticipates Cortés
refusing, and explains why: Cortés' authority was grounded in
performative self-reliance, so accepting help would have undermined the
leadership pattern Bernal Díaz documents repeatedly.

\textbf{What the Spec did.} The headline is what \emph{all the facts
together still cannot capture}. C4 is not a sparse condition; the model
received the complete extracted fact set from Bernal Díaz's
training-half corpus, including every documented instance of Cortés'
behavior. With that full fact pool in context, the model still predicts
the wrong direction: that Cortés would accept the offered help. The
interpretive layer is not adding missing facts; it is encoding an
\emph{organizing principle} the facts alone do not announce: Cortés'
performative self-reliance as the load-bearing mechanism of his
authority. Reading individual facts does not yield the principle; you
have to interpret the pattern across them. Once the principle is named
in the Spec, the model uses it to override its generic ``good leaders
accept help'' default and predict the right direction. This is the value
facts alone cannot capture: the principle that connects the facts is
itself a piece of information the model needs, and the model does not
appear to construct it on its own from the facts. Multi-anchor crossing
(2 → 4.80, a near three-anchor jump).
\end{quote}

\begin{quote}
\textbf{Example C. All Facts → All Facts + Spec: Abstention Becomes
Near-Perfect Inference}

\textbf{Subject:} Mary Seacole. \textbf{Q2:} \emph{``How does Mary
Seacole typically respond when a delirious patient mistakes her for a
family member?''}

\textbf{Held-out:} \emph{``Poor fellow! I could not undeceive him. I
think the fancy happily caused by the touch of a woman's hand soothed
his dying hour.''} In Seacole's own voice: she chose not to correct him,
and read the comfort her presence brought as a kind of healing.

\textbf{C4 All Facts (mean 2.80, judges split 5/2/3/3/1):} \emph{``Based
on the complete set of facts provided, there is \textbf{no specific
information} about how Mary Seacole typically responds when a delirious
patient mistakes her for a family member\ldots{} there is no explicit
account of her reactions to delirious patients confusing her identity
with their relatives.''}

The AI honestly abstains: it identifies that the facts don't directly
cover this scenario and refuses to invent. Five judges have no shared
view on what an honest abstention deserves. Haiku gives it full credit
(5); GPT-5.4 gives it the floor (1); the rest land between. The 2.80
mean averages over a panel that genuinely disagrees on whether to reward
the abstention as engagement or penalize it as failure to
predict.\footnote{The judge-panel split 5/2/3/3/1 on the C4 abstention
  illustrates the \hyperref[rubric-handling-limitations-post-hoc-validity-audit]{§3.3.6} refusal-anchor ambiguity directly. The verbatim
  rubric's lowest anchor is ``refuses or off-base,'' but in practice
  judges apply different norms to honest abstention: some treat it as
  anchor 1 (the rubric's literal reading), others as anchor 3 (partial
  engagement; the model recognized the gap), with a substantial spread
  between. A differentiated rubric that scores abstention as its own
  dimension is flagged in \hyperref[future-work]{§7}.}

\textbf{C4a facts + Spec (mean 5.00, unanimous):} \emph{``Based on the
behavioral specification and known facts, Mary Seacole's response would
follow a consistent pattern. Primary response: compassionate acceptance.
She would \textbf{not correct the patient} in the moment. Instead, she
would \textbf{lean into the mistaken identity, sustain the fiction with
tenderness}, and prioritize emotional relief over factual accuracy. This
aligns with A4 (Proximate Obligation) and A8 (Sympathetic
Realism)\ldots{} Personal presence and touch carry psychological and
emotional power that no institutional process replicates.''}

All five judges agree this matches the held-out. Unanimous 5.

\textbf{What the Spec did.} The headline is the fine-grained nuance the
Spec adds beyond generic compassion. A generic ``Seacole is a
compassionate nurse'' framing is consistent with several responses:
correct the patient gently, redirect them to staff, refuse to engage
with the delirium, accept the misidentification matter-of-factly. The
Spec predicts none of those; it predicts that Seacole would
\emph{sustain the fiction with tenderness}, treating the comfort the
misidentification produces as itself a kind of healing. That is the
level of nuance the question is actually testing: not whether the model
knows Seacole is compassionate (the facts establish that) but whether it
can predict which compassionate move she would choose. Facts alone
produce abstention because no fact in the training half says ``Seacole
sustained a delirious patient's mistaken identification''; the Spec
captures the \emph{pattern} (A4 Proximate Obligation, A8 Sympathetic
Realism) that organizes how Seacole's compassion expresses itself in
unmapped situations, and that pattern is sufficient to predict the
specific move. Three-anchor crossing (2.80 → 5.00). The C4a row is
unanimous 5 across all five primary judges. Signal survives the
abstention-scoring noise visible on the C4 row. Deeper dive on
abstention-anchor ambiguity in \hyperref[rubric-handling-limitations-post-hoc-validity-audit]{§3.3.6} and \hyperref[rubric-handling-limitations-post-hoc-validity-audit-1]{§4.6.7}.
\end{quote}

\begin{center}\rule{0.5\linewidth}{0.5pt}\end{center}

Two statistical confirmations support the directional claim. The
\textbf{Wilcoxon signed-rank test} across all 14 main-study subjects
confirms that the Specification's lift over baseline is real and not due
to chance.\footnote{Wilcoxon signed-rank test results on the 14
  main-study subjects: C5 vs.~C2a \emph{W} = 10 (\emph{p} = 0.005); C5
  vs.~C4a \emph{W} = 11 (\emph{p} = 0.007).} \textbf{Pairwise Spearman
ρ} across the 5-judge primary panel (\hyperref[inter-judge-agreement]{§3.3.4}) confirms that the lift's
direction is consistent across judges rather than dependent on any one
judge's scoring.\footnote{Pairwise Spearman ρ on the 5-judge primary
  panel: 0.86 to 0.93. Regression of Δ\_C4a on C5: slope \textbf{−0.96}
  {[}95\% CI −1.24, −0.67{]}, R² = 0.82, \emph{p} = 0.000009. 12 of 14
  subjects positive; all 9 low-baseline subjects positive (mean Δ\_C4a =
  \textbf{+0.89}). Bootstrap (10,000 resamples): 95\% CI {[}−1.25,
  −0.74{]}. Full statistical detail and additional tests in \hyperref[statistical-rigor-checks-on-the-headline-gradient]{§4.6.4} and
  Appendix B.6.}

\begin{center}\rule{0.5\linewidth}{0.5pt}\end{center}

\textbf{Per-subject results.}

The table is ordered by C5 baseline ascending. In the color-rendered
PDF, low-baseline rows (C5 ≤ 2.0, the population of relevance) are
tinted green, mid-baseline rows (2.0 \textless{} C5 \textless{} 3.0)
yellow, and Franklin (high-baseline reference) gray. Figure 4.1 presents
the same data as a scatter plot with the regression line.

\begin{longtable}[]{@{}
  >{\raggedright\arraybackslash}p{(\columnwidth - 14\tabcolsep) * \real{0.10}}
  >{\raggedleft\arraybackslash}p{(\columnwidth - 14\tabcolsep) * \real{0.13}}
  >{\raggedleft\arraybackslash}p{(\columnwidth - 14\tabcolsep) * \real{0.13}}
  >{\raggedleft\arraybackslash}p{(\columnwidth - 14\tabcolsep) * \real{0.13}}
  >{\raggedleft\arraybackslash}p{(\columnwidth - 14\tabcolsep) * \real{0.13}}
  >{\raggedleft\arraybackslash}p{(\columnwidth - 14\tabcolsep) * \real{0.13}}
  >{\raggedleft\arraybackslash}p{(\columnwidth - 14\tabcolsep) * \real{0.13}}
  >{\centering\arraybackslash}p{(\columnwidth - 14\tabcolsep) * \real{0.10}}@{}}
\toprule
Subject & C5 Baseline & C4 All Facts & C2a Spec Only & C4a All Facts +
Spec & Δ C4a−C5 & Δ C4a−C4 & Anchor\footnote{\textbf{✓} = the subject's
  per-subject mean C4a crossed at least one integer rubric anchor upward
  from C5 (e.g., C5 = 1.4 in the ``1'' band, C4a = 2.3 in the ``2''
  band, mean crossed the 1→2 anchor). \textbf{partial} = the per-subject
  mean did not cross an anchor, but a majority of per-question deltas
  did (within-anchor mean shift; cross-anchor signal at the per-question
  grain). \textbf{−} = neither the per-subject mean nor a majority of
  per-question deltas crossed an anchor.} \\
\midrule
\endhead
Ebers & 1.02 & 2.02 & 1.54 & 2.07 & +1.05 & +0.05 & ✓ \\
Sunity Devee & 1.03 & 2.46 & 2.27 & 2.41 & +1.38 & −0.05 & ✓ \\
Hamerton & 1.26 & 2.43 & 2.63 & 2.77 & +1.51 & +0.34 & ✓ \\
Fukuzawa & 1.67 & 2.67 & 2.35 & 2.78 & +1.11 & +0.11 & ✓ \\
Bernal Díaz & 1.70 & 2.41 & 2.27 & 2.48 & +0.78 & +0.07 & partial \\
Bābur & 1.76 & 2.03 & 1.91 & 2.01 & +0.25 & −0.02 & - \\
Seacole & 1.77 & 2.63 & 2.48 & 2.59 & +0.82 & −0.04 & ✓ \\
Keckley & 1.84 & 2.39 & 2.43 & 2.44 & +0.59 & +0.05 & - \\
Yung Wing & 1.88 & 2.13 & 2.22 & 2.40 & +0.52 & +0.27 & - \\
Zitkala-Ša & 2.34 & 2.10 & 2.03 & 2.02 & −0.32 & −0.08 & - \\
Cellini & 2.38 & 2.42 & 2.54 & 2.53 & +0.15 & +0.11 & - \\
Rousseau & 2.44 & 2.32 & 2.81 & 2.53 & +0.10 & +0.21 & - \\
Augustine & 2.58 & 2.56 & 2.48 & 2.70 & +0.11 & +0.14 & - \\
Equiano & 2.77 & 2.43 & 2.46 & 2.42 & −0.35 & −0.01 & - \\
Franklin & 3.77 & 3.59 & 3.37 & 3.65 & −0.13 & +0.06 & - \\
\bottomrule
\end{longtable}

Franklin's C4 cell shows 3.59 on the 5-judge primary panel after a
2026-05-11 backfill of GPT-4o and GPT-5.4 judging on the legacy
responses; the original legacy panel had 4 judges (Haiku, Sonnet, Opus,
Gemini) and the two GPT primary judges were added to bring the panel to
5 of 5.\footnote{Mid-baseline C4 means on the 5-judge primary aggregate:
  Zitkala-Ša 2.10, Cellini 2.42, Rousseau 2.32, Augustine 2.56, Equiano
  2.43 (39 questions each). Franklin C4 was originally scored by 4
  legacy judges (Haiku, Sonnet, Opus, Gemini) only; GPT-4o and GPT-5.4
  were added 2026-05-11 (script:
  \texttt{scripts/\allowbreak{}rerun\_\allowbreak{}franklin\_\allowbreak{}bernaldiaz\_\allowbreak{}gpt\_\allowbreak{}judges\_\allowbreak{}20260511.\allowbreak{}py})
  to complete the 5-judge primary panel. The 5-judge primary mean is
  3.59 (3.85 Haiku, 3.65 Sonnet, 3.80 Opus, 3.38 GPT-4o, 3.28 GPT-5.4
  across 40 questions). Data at
  \texttt{results/\textless{}subject\textgreater{}/judgments\_v2.json}
  (13 globals) and
  \texttt{results/\allowbreak{}franklin\_\allowbreak{}legacy\_\allowbreak{}20260411/\allowbreak{}analysis/\allowbreak{}*\_\allowbreak{}judgments.\allowbreak{}json}
  (Franklin).}

\begin{itemize}
\tightlist
\item
  \textbf{Low-baseline (n = 9):} every subject improves under the Spec.
  The band is uniform. This is the population of relevance for real AI
  deployment.
\item
  \textbf{Mid-baseline (n = 5):} 3 subjects improve under the Spec, 2
  decline. The model has enough pretraining footprint on these subjects
  that the Specification competes with the model's own working model, so
  the aggregate sign is mixed. Where the Spec still helps is on the
  interpretation-heavy questions: the ones whose answers are not
  narrated verbatim in the source and that the model's pretraining does
  not already cover.
\item
  \textbf{High-Baseline Reference (Franklin):} both Spec-containing
  conditions score below baseline on aggregate. The Specification cannot
  add what the model already has. It still refines the response on the
  interpretation-heavy questions, the ones the \emph{Autobiography} does
  not narrate verbatim (\hyperref[the-gradient-at-the-high-baseline-end-franklin-reference]{§4.1.2}: positive cases cluster on Q38 and Q22,
  scenarios the source does not directly describe); it hurts on the
  questions where pretraining already supplies the pattern.
\end{itemize}

Per-subject anchor-crossing distributions and per-subject per-judge
score matrices are in Appendix D.\footnote{Two potential confounds on
  the gradient slope (battery-question-type composition and
  Hamerton-leverage subset regression) are addressed in \hyperref[battery-composition-sensitivity]{§4.6.3} as
  robustness checks; both leave the baseline gradient effect
  substantially intact.}

\textbf{The aggregate gradient hides per-question structure.} The
Specification produces large category-level shifts on a subset of
questions and minimal change on others. \hyperref[per-question-baseline-engagement-and-the-worked-rubric-example]{§4.1.1} decomposes this
distribution and shows where the Spec's value concentrates. \hyperref[compression-structure-vs.-raw-text]{§4.2} takes
the same gradient and asks whether the lift is about structure or about
information volume, comparing the Spec against far larger raw-corpus
context.\footnote{Held-out leakage audit on the 60 unique
  extreme-upward-jump cases at
  \texttt{docs/\allowbreak{}research/\allowbreak{}held\_\allowbreak{}out\_\allowbreak{}leakage\_\allowbreak{}investigation\_\allowbreak{}20260428.\allowbreak{}md}:
  0 6-gram matches at C4a, severity rare; full taxonomy and
  headline-impact estimate in Appendix B.9.}

\hypertarget{per-question-baseline-engagement-and-the-worked-rubric-example}{%
\subsection{4.1.1 Per-question baseline engagement and the worked rubric
example}\label{per-question-baseline-engagement-and-the-worked-rubric-example}}

Across 546 questions on the 14 main-study subjects, the No-Context
Baseline (C5) splits into two clusters with a thin middle.

\begin{itemize}
\tightlist
\item
  The \textbf{bottom cluster} (rubric score = 1.00 at C5): roughly
  \textbf{41\% of questions}. The model declines, names the wrong
  person, or lands far outside the question.
\item
  The \textbf{top cluster} (rubric score ≥ 3.0 at C5): roughly
  \textbf{21\% of questions}. The model produces a \emph{substantive
  answer}: engaged with the question, specifically about the named
  subject, and predicting at least the right behavioral domain.
\item
  The \textbf{middle} (1.0 \textless{} C5 \textless{} 3.0): the
  remaining \textasciitilde38\%. The model engages with the question but
  the prediction is wrong, or at best gets the right domain with the
  wrong outcome; it falls short of a substantive, subject-specific
  answer. Comparatively few questions land here: the baseline
  distribution is bimodal, with a thin middle between the floor cluster
  and the substantive cluster.
\end{itemize}

\textbf{Adding the Spec produces a substantive prediction on 94.2\% of
bottom-cluster questions, lifting the response from refusal or
misalignment.} On the top cluster, where the baseline response was
already substantive, the Spec improved scores in roughly 21\% of
questions; on the remaining 79\%, scores were unchanged or reduced.

The two findings together describe the per-question structure underlying
the cross-subject gradient in \hyperref[the-cross-subject-gradient-and-its-per-question-mechanism]{§4.1}. Where the baseline knows nothing
about the subject, the Spec supplies the interpretive frame the baseline
lacks. Where the baseline already engages with the subject, the Spec
still improves some answers, but on a smaller share of questions than
where the baseline is low, and it sometimes reduces the answer. The
response model was given a grounded-prediction prompt with no
instruction to abstain and no cost to producing a confident wrong answer
beyond a low judge score; it declined on more than 40\% of questions.
Whatever logic allows a language model to practice abstention or refusal
to answer takes precedence over the prompt's instruction to predict.

A score of 1.00 means the model failed to produce a usable prediction
about the named subject (full definition in \hyperref[score-interpretation]{§3.3.1}, ``What a 1 means and
does not mean''). In about 93\% of score-1.00 responses, the model
explicitly declined to answer (``I don't have enough information about
this person''). The remaining 7\% are non-abstention failures: the model
engaged with the question, but the engagement was categorically
incorrect (wrong referent, off-base inference, or confusion with a
different subject). Both modes are addressable by adding the Spec,
through different mechanisms.\footnote{The \hyperref[scoring-rubric-with-calibrated-llm-judge-panel]{§3.3} rubric scores both
  explicit abstention and non-abstention misalignment as 1.00; the 93/7
  split is from a post-hoc regex pass classifying responses
  (\texttt{scripts/\allowbreak{}analyze\_\allowbreak{}baseline\_\allowbreak{}engagement.\allowbreak{}py}). 7\% is a coarse
  upper bound; finer analysis of \emph{when and why} the model picks
  confident-wrong over abstention is open future work in \hyperref[future-work]{§7}.}

Grouping the 546 questions by where the No-Context Baseline (C5) placed
them shows where the Spec's lift concentrates (X = C5 mean across the
5-judge primary panel; lift = C4a − C5):\footnote{Spearman ρ between
  baseline X and Spec lift = \textbf{−0.73} (n = 546, p ≈ 1.7 × 10⁻⁹¹).
  Per-subject ρ negative for 14 of 14; 12 of 14 reach p \textless{}
  0.01.}

\textbf{Bottom cluster: X = 1.00 (225 questions, 41.2\%).} Mean Spec
lift \textbf{+1.32} (± 0.88); the Spec produces a positive lift on
94.2\% of these questions. This is where the Spec does the most work,
converting abstentions and confident-wrong answers into substantive
predictions.

\textbf{Middle: 1.00 \textless{} X \textless{} 3.00 (205 questions,
37.5\%).} Two sub-bands. For marginal baselines (1.00 \textless{} X
\textless{} 2.00; 110 questions, 20.1\%): mean lift \textbf{+0.66} (±
0.83), positive on 78.2\%, the Spec sharpens partial answers into more
specific predictions. For mid baselines (2.00 ≤ X \textless{} 3.00; 95
questions, 17.4\%): mean lift \textbf{+0.04} (± 0.63), positive on
39.0\%, the Spec neither helps nor hurts on average.

\textbf{Top cluster: X ≥ 3.00 (116 questions, 21.2\%).} Two sub-bands.
For 3.00 ≤ X \textless{} 4.00 (82 questions, 15.0\%): mean lift
\textbf{−0.47} (± 0.81), positive on 25.6\%, the Spec hurts on average
where the baseline already engages substantively. For X ≥ 4.00 (34
questions, 6.2\%): mean lift \textbf{−0.99} (± 0.78), positive on 8.8\%,
the Spec hurts most where the baseline already produces a near-correct
answer.

\textbf{Subjects mirror the baseline variance: the same split
distribution appears within each subject's 39-question battery.} Three
patterns:\footnote{\emph{Floor-saturated:} Sunity Devee (37 of 39
  questions at X = 1.00) and Ebers (36 of 39). \emph{Engaged-skewed:}
  Equiano (only 2 of 39 questions at X = 1.00; 5 of 39 at X ≥ 4.00).
  \emph{Mixed:} the remaining 11 subjects are Augustine, Bābur, Bernal
  Díaz, Cellini, Fukuzawa, Hamerton, Keckley, Rousseau, Seacole, Yung
  Wing, and Zitkala-Ša. Augustine (8 of 39 floor-questions despite a
  per-subject C5 mean of 2.58) is an example of how the mixed pattern
  persists even at higher per-subject means.}

\begin{itemize}
\tightlist
\item
  \textbf{Floor-saturated, 2 of 14 subjects (Sunity Devee, Ebers).} More
  than 90\% of the 39 questions in the battery return a refusal or
  misalignment from the baseline. The per-subject mean shows almost no
  variance.
\item
  \textbf{Engaged-skewed, 1 of 14 subjects (Equiano).} Fewer than 10\%
  of the 39 questions in the battery return a refusal or misalignment.
  The baseline produces an answer specifically about the subject on most
  questions.
\item
  \textbf{Mixed, 11 of 14 subjects.} The battery contains questions at
  both the floor (refusal or misalignment) and in the
  substantive-engagement range. Some subjects in this group sit closer
  to floor-saturated, others closer to engaged-skewed. All the
  high-baseline subjects in the \hyperref[the-cross-subject-gradient-and-its-per-question-mechanism]{§4.1} gradient fall in this pattern. Even
  when the baseline knows the subject well enough to answer most
  questions substantively, several questions in the same battery still
  trigger floor-level refusals.
\end{itemize}

\textbf{Worked rubric example: Seacole Q2 across all five conditions.}
The example below traces a single question across all five conditions.
It shows a range of rubric responses, making the differences between
conditions and the workings of the scoring rubric concrete; most
importantly, it shows how each condition produces a qualitatively
different response.\footnote{The Seacole question was used as Example C
  in \hyperref[the-cross-subject-gradient-and-its-per-question-mechanism]{§4.1}; here it is presented across the full condition set so the
  anchor-by-anchor progression is visible. Each row's response excerpt
  is sourced directly from \texttt{results\_v2.json} for the
  corresponding (subject, qid, condition) cell, paired with its
  \texttt{judgments\_v2.json} cached score; the canonical extractor
  (\texttt{scripts/\allowbreak{}v2\_\allowbreak{}canonical\_\allowbreak{}cell\_\allowbreak{}extractor\_\allowbreak{}20260508.\allowbreak{}py})
  enforces this end-to-end.}

\textbf{Subject:} Mary Seacole. \textbf{Q2:} \emph{``How does Mary
Seacole typically respond when a delirious patient mistakes her for a
family member?''}

\textbf{Held-out ground truth:} \emph{``Poor fellow! I could not
undeceive him. I think the fancy happily caused by the touch of a
woman's hand soothed his dying hour.''} In Seacole's own voice: she
found she could not bring herself to correct him, and read the comfort
her presence brought as a kind of healing.

\begin{longtable}[]{@{}
  >{\raggedright\arraybackslash}p{(\columnwidth - 6\tabcolsep) * \real{0.22}}
  >{\raggedright\arraybackslash}p{(\columnwidth - 6\tabcolsep) * \real{0.09}}
  >{\raggedleft\arraybackslash}p{(\columnwidth - 6\tabcolsep) * \real{0.08}}
  >{\raggedright\arraybackslash}p{(\columnwidth - 6\tabcolsep) * \real{0.61}}@{}}
\toprule
Condition & Code & Mean & Excerpt \\
\midrule
\endhead
Wrong Spec (Bābur Spec applied to Seacole) & C2c v1 & 1.40 & \emph{``The
specification you've provided describes someone from the 16th-century
Spanish conquest period\ldots{} Mary Seacole (1805-1881) was a Creole
nurse\ldots{}''} \\
No-Context Baseline & C5 & 1.80 & \emph{``I don't have specific
documented instances of how Mary Seacole responded when delirious
patients mistook her for family members\ldots{}''} \\
All Facts & C4 & 2.80 & \emph{``Based on the complete set of facts
provided, there is no specific information about how Mary Seacole
typically responds\ldots{} there is no explicit account\ldots{}''} \\
Spec Only & C2a & 4.20 & \emph{``Mary Seacole would likely respond to
this situation through the lens of A4 (Proximate Obligation) and A8
(Sympathetic Realism)\ldots{} she would maintain the therapeutic
connection, recognizing that the delusion is a mercy\ldots{}''} \\
All Facts + Spec & C4a & 5.00 & \emph{``Mary Seacole's response would
follow a consistent pattern: compassionate acceptance\ldots{} she would
lean into the mistaken identity, sustain the fiction with tenderness,
and prioritize emotional relief over factual accuracy. This aligns with
A4 (Proximate Obligation) and A8 (Sympathetic Realism).''} \\
\bottomrule
\end{longtable}

The example traces the \emph{cross-anchor interpretation rule} of \hyperref[score-interpretation]{§3.3.1}
on a single question. C2c (a Bābur Spec wrongly applied to Mary Seacole)
explicitly recognizes that the served Spec describes a different person
and declines; the response is articulate but the prediction is the
opposite of what the held-out shows because the Spec doesn't fit
Seacole. C5 (no-context) declines for lack of referent. C4 (all facts,
no Spec) abstains because the facts as supplied do not directly cover
this scenario. C2a (Spec only, no facts) uses the Spec's behavioral
patterns (A4, Proximate Obligation, and A8, Sympathetic Realism) to
predict Seacole's non-correction; the answer is multi-dimensional and
lands close to the held-out. C4a (facts + Spec) uses both layers and
reaches the held-out direction unanimously across all five judges.
Categorical movement from anchor 1 through anchor 5 on a single question
is what the per-subject means in \hyperref[the-cross-subject-gradient-and-its-per-question-mechanism]{§4.1} aggregate.

Per-response-model abstention behavior is named in \hyperref[rubric-handling-limitations-post-hoc-validity-audit]{§3.3.6} and decomposed
in \hyperref[rubric-handling-limitations-post-hoc-validity-audit-1]{§4.6.7}. Memory-system condition effects on score-1 composition are
decomposed in \hyperref[memory-system-composition]{§4.4}.

\hypertarget{the-gradient-at-the-high-baseline-end-franklin-reference}{%
\subsection{4.1.2 The gradient at the high-baseline end (Franklin
reference)}\label{the-gradient-at-the-high-baseline-end-franklin-reference}}

Franklin is never included in any N=14 inferential statistic; he appears
only as a descriptive high-baseline reference. \textbf{Franklin confirms
the gradient at the high end: where the baseline is already near the
ceiling, the Spec produces a small negative effect rather than a
positive lift.} His No-Context Baseline mean (C5) is 3.77 on the 5-judge
primary panel, a full anchor above the next-highest main-study subject
(Equiano at 2.77).

\textbf{Both Spec-containing conditions score below Franklin's baseline
on aggregate, but the Spec still helps on a meaningful subset of
questions.} Spec alone drops 0.40 points; the full pipeline (facts +
Spec) drops 0.13. The aggregate hides per-question heterogeneity: the
Spec lifts the response on 15 of 40 questions and depresses it on 20 of
40. Positive cases cluster on scenarios the \emph{Autobiography} does
not narrate verbatim (Q38, +1.80; Q22, +1.60); negative cases cluster on
questions where the model already had the pattern from pretraining (Q43,
C5 = 5.00 → C4a = 1.80). The Spec helps where the model lacks the
pretraining-derived behavior and hurts where it already has it; the
aggregate sign averages across both populations within Franklin's
battery. The exact mechanism is not isolated in this study; further
investigation is flagged in \hyperref[future-work]{§7}.\footnote{Franklin's C4-vs-C4a comparison
  (post-2026-05-11 backfill, 5-judge primary): C4 = 3.59, C4a = 3.65, Δ
  = +0.06. The facts-only condition essentially matches the All Facts +
  Spec condition on Franklin, consistent with the high-baseline-end
  mechanism: where the model already has substantial pretraining
  coverage, the Spec adds little on top of the facts the model would
  otherwise draw on from pretraining + retrieval. The negative aggregate
  Spec effects (−0.40 Spec Only, −0.13 All Facts + Spec) come from a
  small number of high-magnitude depressions on questions where
  pretraining had the pattern and the Spec interfered, not from
  systematic underperformance.}\footnote{Raw per-subject Franklin data
  at \texttt{results/\allowbreak{}franklin\_\allowbreak{}legacy\_\allowbreak{}20260411/}. Per-question lift
  analysis from \texttt{scripts/\allowbreak{}analyze\_\allowbreak{}baseline\_\allowbreak{}engagement.\allowbreak{}py}
  applied to Franklin data 2026-05-07. Numerical claims (3.77 baseline,
  0.40 / 0.13 drops, Q38 +1.80, Q22 +1.60, 15-of-40 positive count) to
  be added to \texttt{scripts/\allowbreak{}recompute\_\allowbreak{}paper\_\allowbreak{}numbers.\allowbreak{}py} per
  launch-blocking mechanistic-check infrastructure (\hyperref[future-work]{§7}).}

\hypertarget{compression-structure-vs.-raw-text}{%
\subsection{4.2 Compression: structure vs.~raw
text}\label{compression-structure-vs.-raw-text}}

\textbf{Hypothesis tested in this section} (H5 from \hyperref[what-we-tested]{§1.2}): A compact
specification achieves comparable behavioral-prediction performance to
the full raw source corpus, at a fraction of the context size.

Even within the compression frame, the per-question mechanism from
\hyperref[per-question-baseline-engagement-and-the-worked-rubric-example]{§4.1.1} holds: on particular questions, the interpretive lens outperforms
raw facts or raw corpus, even when the model has access to the full
source text.

\begin{center}\rule{0.5\linewidth}{0.5pt}\end{center}

\textbf{Context improves prediction.} On the 9 low-baseline subjects,
every context condition increases the per-subject mean score by roughly
one full rubric point over the No-Context Baseline.

\begin{longtable}[]{@{}
  >{\raggedright\arraybackslash}p{(\columnwidth - 8\tabcolsep) * \real{0.17}}
  >{\raggedright\arraybackslash}p{(\columnwidth - 8\tabcolsep) * \real{0.17}}
  >{\raggedleft\arraybackslash}p{(\columnwidth - 8\tabcolsep) * \real{0.22}}
  >{\raggedleft\arraybackslash}p{(\columnwidth - 8\tabcolsep) * \real{0.22}}
  >{\raggedleft\arraybackslash}p{(\columnwidth - 8\tabcolsep) * \real{0.22}}@{}}
\toprule
Condition & Context served (approx. tokens, low-baseline mean) & n &
Mean (low-baseline) & Δ from C5 \\
\midrule
\endhead
C5 & none (baseline) & 9 & 1.55 & 0.00 \\
C2a & Spec only (\textasciitilde7K) & 9 & 2.23 & +0.68 \\
C4 & all Facts Only (\textasciitilde10K) & 9 & 2.35 & +0.80 \\
C8 & raw corpus only (\textasciitilde163K mean; range 33K--549K) & 9 &
2.46 & +0.91 \\
C4a & all facts + Spec (\textasciitilde17K) & 9 & 2.44 & +0.89 \\
C9 & corpus + Spec (\textasciitilde170K mean)\footnote{Computed over the
  8 low-baseline subjects with C9 data; Bābur is excluded from C9
  because his corpus plus the Specification exceeds Haiku's context
  window. (Bābur's C8 was run on a 100K-word truncation of the same
  corpus, so C8 retains all 9 subjects.) Bernal Díaz C9 was rejudged
  2026-05-11 to complete its 5-judge primary panel after the original
  GPT-4o + GPT-5.4 batches returned 429 rate-limit errors; rerun script:
  \texttt{scripts/\allowbreak{}rerun\_\allowbreak{}franklin\_\allowbreak{}bernaldiaz\_\allowbreak{}gpt\_\allowbreak{}judges\_\allowbreak{}20260511.\allowbreak{}py}.}
& 8 & 2.59 & +1.07\footnote{All cells are 5-judge primary per-subject
  means (per-judge per-question score → mean across \{haiku, sonnet,
  opus, gpt4o, gpt54\} → mean across each subject's 39 questions → mean
  across subjects). The C5, C2a, C4, C8, and C4a rows are computed over
  all 9 low-baseline subjects (n=9); their Δ is the condition mean minus
  the 9-row C5 mean (1.55). The C9 row is computed over 8 subjects (n=8)
  because Bābur has no C9 data --- his corpus plus the Specification
  exceeds Haiku's context window --- so the C9 Δ is the 8-row C9 mean
  minus the 8-row C5 mean (1.52, Bābur excluded), keeping the C9
  contrast symmetric. Bābur's C8 \emph{does} exist: it was run on a
  100K-word truncation of his 422,772-word corpus, so C8 stays at n=9.
  Earlier drafts subtracted an 8-row C5 (1.52, Bābur dropped) from 9-row
  C2a/C4/C8/C4a means, producing the inflated +0.71/+0.83/+0.93/+0.93
  Δs; this table standardizes to the symmetric per-row C5 baseline.
  Recompute provenance:
  \texttt{docs/\allowbreak{}reviews/\allowbreak{}v12\_\allowbreak{}1\_\allowbreak{}compression\_\allowbreak{}table\_\allowbreak{}recompute\_\allowbreak{}20260513.\allowbreak{}md}.} \\
\bottomrule
\end{longtable}

The AI does not need much context to move from refusal-and-off-base to
engaged subject-specific prediction. It needs \emph{some} context.

\textbf{The compact specification recovers 75\% of the corpus's
predictive lift at \textasciitilde25× less context.} On the 9
low-baseline subjects, the Behavioral Specification alone
(\textasciitilde7K tokens) produces +0.68 points of lift over baseline;
the full raw corpus (\textasciitilde163K tokens mean, range 33K--549K)
produces +0.91 points (symmetric 9-row computation; see \footnote{The
  +0.68 (Spec Only) and +0.91 (raw corpus) figures are 9-row
  low-baseline-band per-subject-mean Δs (per-question grand-mean
  recompute: +0.685 and +0.908; per-subject-mean: +0.685 and +0.913; the
  per-subject-mean Δ\_C2a is 0.6849, which rounds to +0.68 at two
  decimals). Earlier drafts reported +0.71 and +0.93 using an asymmetric
  8-row C5 baseline (Bābur excluded, C5 = 1.52) subtracted from 9-row
  C2a/C8 means; v12.1 standardizes to the symmetric 9-row computation.
  The C8 cell retains all 9 subjects because Bābur's C8 was run on a
  100K-word corpus truncation; only the C9 cell drops Bābur, where his
  corpus plus the Specification exceeds Haiku's context window. Both
  grains agree on direction and rank order. Detail in Appendix B.9.}).
The interpretive layer drives the result; the volume of context served
does not.

\textbf{The compression story holds across other context-size
comparisons:}

\begin{itemize}
\tightlist
\item
  \textbf{Spec vs All Facts:} Spec produces 85\% of the All Facts (C4)
  lift (+0.68 vs +0.80) at 30\% less context.
\item
  \textbf{All Facts + Spec vs raw corpus:} the structured 17K-token
  package recovers most of the raw 163K-token corpus's lift (+0.89 vs
  +0.91), at \textasciitilde10× less context.
\item
  \textbf{All Facts + Spec vs Corpus + Spec:} layering the Spec on top
  of All Facts already produces most of the available lift; adding the
  entire raw corpus on top of the same Spec adds roughly 9\% of
  additional lift over baseline (+0.97 → +1.07, matched 8-subject
  comparison; Bābur excluded as his corpus + Spec exceeds the response
  model's context window) at \textasciitilde10× more context. The Spec
  is doing the work; raw corpus volume past the All-Facts-plus-Spec
  point is largely redundant.
\end{itemize}

The behaviorally relevant signal in autobiographical text is sparse and
compressible. A compact structured document captures most of it; adding
more raw text adds little beyond what the structured document already
delivered.

\begin{figure}
\centering
\includegraphics{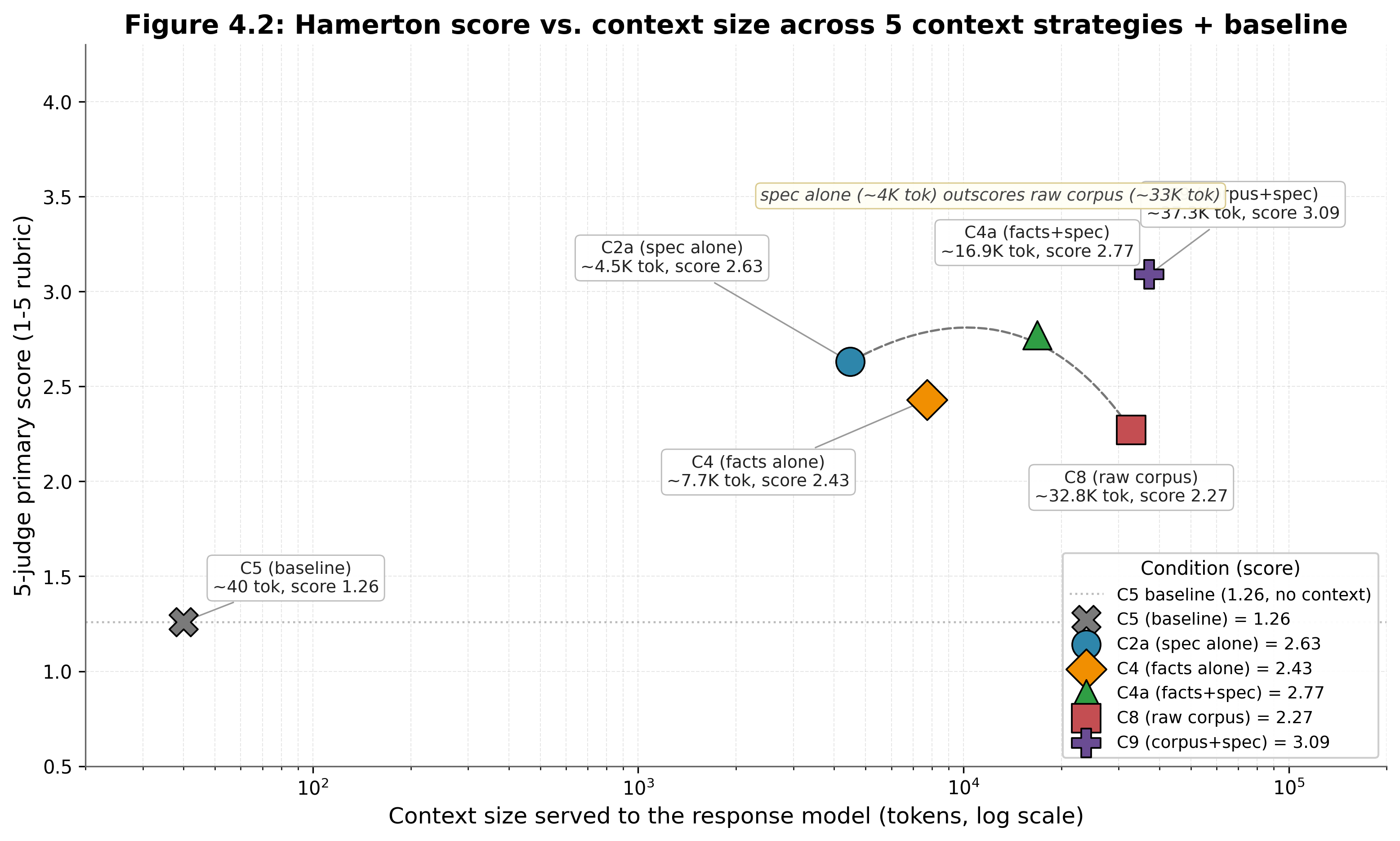}
\caption{Figure 4.2: Score versus context size (log scale) per subject
across compression-related conditions. The score climbs steeply across
the first \textasciitilde7K tokens of structured specification and
plateaus through \textasciitilde80K to 400K tokens of raw corpus,
showing the compression of the behavioral signal into a small structured
representation. (\hyperref[compression-structure-vs.-raw-text]{§4.2})}
\end{figure}

\begin{center}\rule{0.5\linewidth}{0.5pt}\end{center}

\textbf{Per-subject compression comparison (5-judge primary,
low-baseline band).}

The table shows baseline and every compression-related condition for
each subject, with the compression ratio (source corpus tokens ÷
specification tokens, both approximate) for scale.\footnote{Bold marks
  the highest score per subject across the six condition columns (C5,
  C2a, C4, C8, C4a, C9). In the color-rendered PDF, low-baseline rows
  are tinted to mark the population of relevance; the C8 − C2a gap
  column is shaded to make the Spec-vs-corpus difference visible at a
  glance. Markdown render preserves data only.}

\begin{longtable}[]{@{}
  >{\raggedright\arraybackslash}p{(\columnwidth - 18\tabcolsep) * \real{0.08}}
  >{\raggedleft\arraybackslash}p{(\columnwidth - 18\tabcolsep) * \real{0.10}}
  >{\raggedleft\arraybackslash}p{(\columnwidth - 18\tabcolsep) * \real{0.10}}
  >{\raggedleft\arraybackslash}p{(\columnwidth - 18\tabcolsep) * \real{0.10}}
  >{\raggedleft\arraybackslash}p{(\columnwidth - 18\tabcolsep) * \real{0.10}}
  >{\raggedleft\arraybackslash}p{(\columnwidth - 18\tabcolsep) * \real{0.10}}
  >{\raggedleft\arraybackslash}p{(\columnwidth - 18\tabcolsep) * \real{0.10}}
  >{\raggedleft\arraybackslash}p{(\columnwidth - 18\tabcolsep) * \real{0.10}}
  >{\raggedleft\arraybackslash}p{(\columnwidth - 18\tabcolsep) * \real{0.10}}
  >{\raggedleft\arraybackslash}p{(\columnwidth - 18\tabcolsep) * \real{0.10}}@{}}
\toprule
Subject & Source words (\textasciitilde tokens) & Compression ratio
(corpus / Spec) & C5 baseline & C2a Spec (\textasciitilde7K tok) & C4
facts (\textasciitilde10K tok) & C8 raw corpus & C4a All Facts + Spec &
C9 Corpus + Spec & C8 − C2a \\
\midrule
\endhead
Hamerton & 25,231 (\textasciitilde33K) & 7× & 1.26 & 2.63 & 2.43 & 2.27
& 2.77 & \textbf{3.09} & \textbf{−0.36} \\
Sunity Devee & 67,379 (\textasciitilde88K) & \textasciitilde13× & 1.03 &
2.27 & 2.46 & \textbf{2.55} & 2.41 & 2.46 & +0.28 \\
Ebers & 96,174 (\textasciitilde125K) & \textasciitilde17× & 1.02 & 1.54
& 2.02 & \textbf{2.18} & 2.07 & 2.16 & +0.64 \\
Fukuzawa & 139,088 (\textasciitilde181K) & \textasciitilde26× & 1.67 &
2.35 & 2.67 & 2.74 & \textbf{2.78} & \textbf{2.78} & +0.39 \\
Bernal Díaz & 187,315 (\textasciitilde244K) & \textasciitilde33× & 1.70
& 2.27 & 2.41 & \textbf{2.55} & 2.48 & 2.52 & +0.28 \\
Bābur & 422,772 (\textasciitilde549K) & \textasciitilde79× & 1.76 & 1.91
& 2.03 & \textbf{2.05} & 2.01 & - & +0.14 \\
Seacole & 62,467 (\textasciitilde81K) & \textasciitilde12× & 1.77 & 2.48
& 2.63 & \textbf{2.83} & 2.59 & 2.73 & +0.35 \\
Keckley & 58,742 (\textasciitilde76K) & \textasciitilde11× & 1.84 & 2.43
& 2.39 & \textbf{2.50} & 2.44 & 2.49 & +0.07 \\
Yung Wing & 66,459 (\textasciitilde86K) & \textasciitilde13× & 1.88 &
2.22 & 2.13 & 2.42 & 2.40 & \textbf{2.50} & +0.20 \\
\textbf{Mean} & & \textbf{\textasciitilde25×} & \textbf{1.55} &
\textbf{2.23} & \textbf{2.35} & \textbf{2.45} & \textbf{2.44} &
\textbf{2.59} & \textbf{+0.22} \\
\bottomrule
\end{longtable}

\textbf{All 9 low-baseline subjects show positive lift across every
condition.} The Spec recovers 75\% of the corpus's lift at
\textasciitilde25× compression on average. Hamerton (smallest corpus) is
the boundary case where Spec exceeds the raw corpus; Ebers (lowest
baseline) is where the corpus retains its largest edge over Spec. Both
are expanded in \hyperref[per-question-improvement-rate]{§4.2.1}.

\hypertarget{per-question-improvement-rate}{%
\subsection{4.2.1 Per-question improvement
rate}\label{per-question-improvement-rate}}

\textbf{On the 9 low-baseline subjects, 7 of every 10 questions improve
with the Spec alone (\textasciitilde7K tokens).} That is within 8
percentage points of the raw corpus's improvement rate (78.3\%, at
\textasciitilde163K tokens). The compression story holds at the
per-question level: structured context produces a similar improvement
rate to raw corpus context at roughly \textasciitilde25× less context
served.

The aggregate mean score blends judge variability with response quality.
A cleaner unit: out of N individual questions, how many does each
condition improve over the No-Context Baseline? Each question either
improves, ties, or worsens when the condition's context is added. We
report three numbers per condition: the improvement rate, the worsening
rate, and the median magnitude of improvement among improved questions
(with median worsening magnitude as a sanity check).

\textbf{Low-baseline band (9 subjects, 351 questions, 5-judge primary
per-question means).}

\begin{longtable}[]{@{}
  >{\raggedright\arraybackslash}p{(\columnwidth - 14\tabcolsep) * \real{0.10}}
  >{\raggedright\arraybackslash}p{(\columnwidth - 14\tabcolsep) * \real{0.10}}
  >{\raggedleft\arraybackslash}p{(\columnwidth - 14\tabcolsep) * \real{0.13}}
  >{\raggedleft\arraybackslash}p{(\columnwidth - 14\tabcolsep) * \real{0.13}}
  >{\raggedleft\arraybackslash}p{(\columnwidth - 14\tabcolsep) * \real{0.13}}
  >{\raggedleft\arraybackslash}p{(\columnwidth - 14\tabcolsep) * \real{0.13}}
  >{\raggedleft\arraybackslash}p{(\columnwidth - 14\tabcolsep) * \real{0.13}}
  >{\raggedleft\arraybackslash}p{(\columnwidth - 14\tabcolsep) * \real{0.13}}@{}}
\toprule
Condition vs.~baseline & Approx. context & Improved & Tied & Worse &
Improvement rate & Median Δ when improved & Median Δ when worsened \\
\midrule
\endhead
\textbf{Spec only} & \textasciitilde7K tokens & 249 & 49 & 53 &
\textbf{70.9\%} & \textbf{+1.00} & −0.40 \\
All Facts & \textasciitilde10K tokens & 256 & 44 & 51 & 72.9\% & +1.00 &
−0.40 \\
Raw corpus & \textasciitilde163K mean (33K--549K) & 275 & 31 & 45 &
78.3\% & +1.00 & −0.60 \\
All facts + Spec & \textasciitilde17K tokens & 276 & 22 & 53 & 78.6\% &
+1.00 & −0.40 \\
Raw corpus + Spec (Bābur excl.)\footnote{Bābur C9 is omitted in \hyperref[compression-structure-vs.-raw-text]{§4.2}
  (422,772-word source exceeds the response model's context window); the
  remaining 13 subjects have C9 data.} & \textasciitilde170K tokens &
261 & 15 & 36 & \textbf{83.7\%} & +1.20 & −0.40 \\
\bottomrule
\end{longtable}

\textbf{When the Spec helps, the typical help is a full rubric category
(+1.00 median). When it hurts, the typical hurt is less than half a
category (−0.40 median).} Roughly 1 in 10 questions tie; fewer than 1 in
6 worsen with the Spec alone.\footnote{Per-condition
  improvement/worsening rates and low-baseline pairwise comparisons in
  Appendix B.14.}

\begin{figure}
\centering
\includegraphics{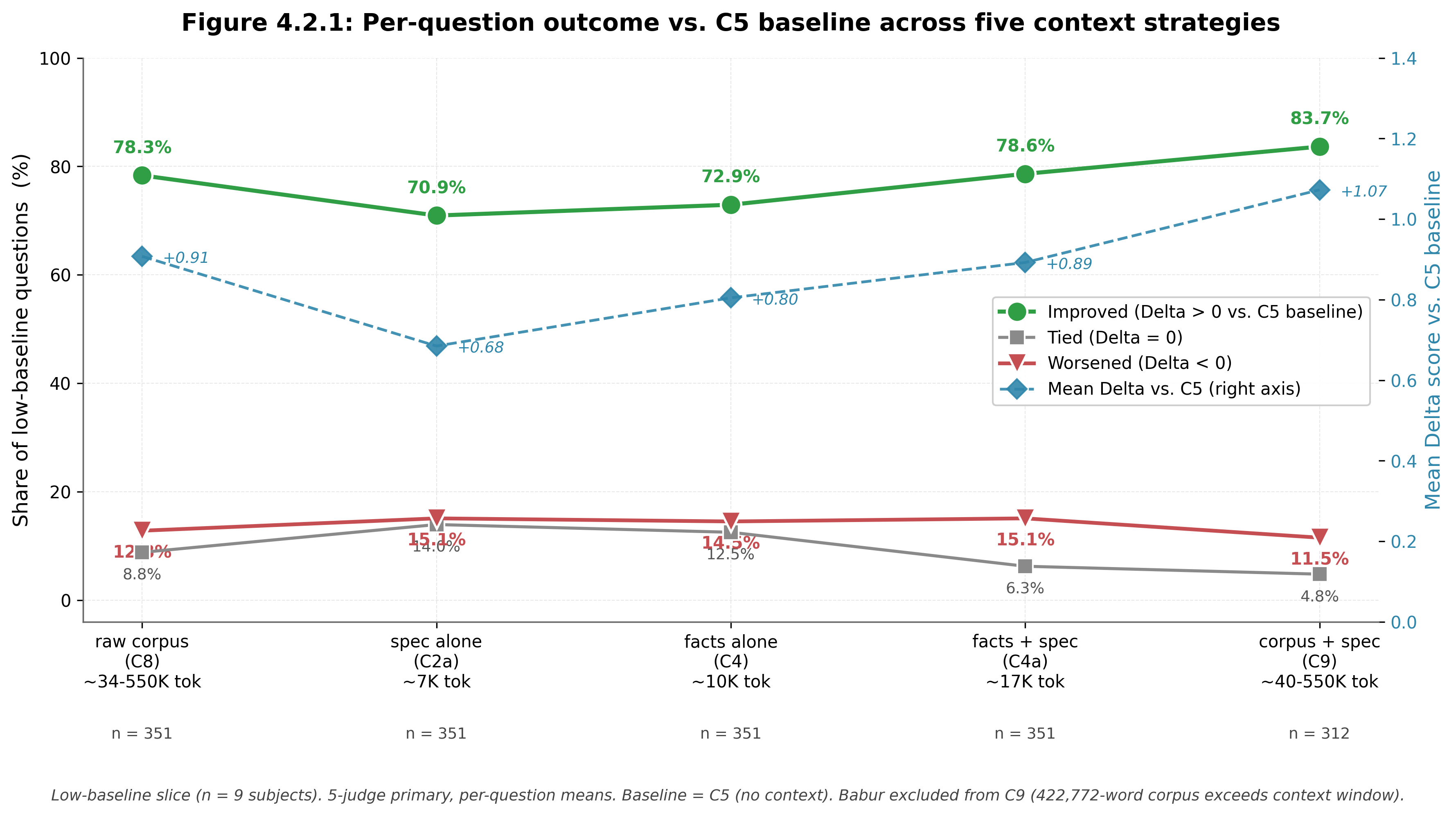}
\caption{Figure 4.2.1: Per-question improvement rates across the five
context conditions for the 9 low-baseline subjects (351 paired
questions, 9 × 39). Conditions are ordered by context size: Spec alone
(C2a, \textasciitilde7K tokens), All Facts (C4, \textasciitilde10K), All
Facts + Spec (C4a, \textasciitilde17K), raw corpus (C8,
\textasciitilde163K mean), corpus plus Spec (C9, \textasciitilde170K).
The improved-share line spans the 70.9\% to 83.7\% range across
conditions, with C9 highest. Spec alone improves 70.9\% of questions at
roughly \textasciitilde25× less context than the raw corpus (78.3\%);
All Facts + Spec matches the raw corpus's improvement rate while cutting
the tied range roughly in half; corpus plus Spec produces the highest
improvement rate (83.7\%). Median Δ when improved is +1.00 rubric
points; median Δ when worsened is −0.40 points. (\hyperref[per-question-improvement-rate]{§4.2.1}).}
\end{figure}

\textbf{Multi-anchor crossings happen on 9--15\% of questions when
adding context to a No-Context Baseline; on 2--3\% when the Spec is
layered on top of full facts or corpus.} The small mean Δ values for
adding the Spec on top of facts or corpus are residues of substantial
per-question movement in both directions, not uniformly small effects.
The multi-anchor rate captures the categorical shifts the aggregate mean
averages over (definition in \hyperref[score-interpretation]{§3.3.1}).

\begingroup\footnotesize
\begin{longtable}[]{@{}>{\raggedright\arraybackslash}p{0.30\linewidth}lrrrr@{}}
\toprule
Comparison & Subject set & n paired & Multi-anchor (≥2) & Extreme (≥3) &
Mean Δ \\
\midrule
\endhead
Baseline → full pipeline (all facts + Spec) & all 14 & 546 & 13.0\% &
3.7\% & +0.55 \\
Baseline → all facts only & all 14 & 546 & 12.5\% & 4.4\% & +0.47 \\
Baseline → Spec only & all 14 & 546 & 9.0\% & 2.0\% & +0.43 \\
Wrong Spec → correct Spec & all 14 & 546 & 14.5\% & 2.4\% & +0.64 \\
Baseline → raw corpus only & 13 (Bābur excl.) & 507 & 15.4\% & 4.3\% &
+0.59 \\
Baseline → corpus + Spec & 13 (Bābur excl.) & 507 & 14.8\% & 4.7\% &
+0.62 \\
All facts → all facts + Spec & all 14 & 546 & 2.2\% & 0.9\% & +0.08 \\
Corpus → corpus + Spec & 13 (Bābur excl.) & 507 & 2.4\% & 0.4\% &
+0.03 \\
\bottomrule
\end{longtable}
\endgroup

The pattern is consistent with the \hyperref[introduction]{§1} thesis: the Spec produces the most
categorical moves where prior context is sparsest.\footnote{All-14
  figures from
  \texttt{docs/\allowbreak{}research/\allowbreak{}multi\_\allowbreak{}anchor\_\allowbreak{}rates\_\allowbreak{}all\_\allowbreak{}pairs\_\allowbreak{}20260430.\allowbreak{}json}.
  The 9-subject low-baseline band gives somewhat higher rates (e.g.,
  baseline → full pipeline at 18.2\% on 9 subjects vs.~13.0\% on all
  14). Hamerton is an outlier on corpus → Corpus + Spec (49\% upward
  rate, 19 up vs.~2 down on 39 questions). Multi-anchor examples:
  Hamerton q22 (1→3 on corpus → Corpus + Spec), Hamerton q25 (1→4),
  Seacole q2 (2→5 on facts → All Facts + Spec), Yung Wing q22 (1→4 on
  facts → All Facts + Spec).}

\textbf{Worked example: Hamerton Q25 across all six conditions.} This
question is one of the \textasciitilde2.4\% of corpus → corpus + Spec
comparisons that produce a multi-anchor crossing, illustrating where
adding the Spec on top of the full corpus produces a categorical shift.
Anchors follow the rubric in \hyperref[scoring-rubric-with-calibrated-llm-judge-panel]{§3.3}.

\textbf{Subject:} Philip Gilbert Hamerton. \textbf{Q25:} \emph{Given
Hamerton's difficulty following spoken French at Loch Awe despite years
of study, what would he do about it?}

\textbf{Held-out ground truth:} \emph{``This plagued me with an
irritating sense of ignorance, so I looked back on my education
generally, and found it unsatisfactory\ldots{} I determined to acquire
some substantial knowledge of modern languages, and to begin by learning
French over again, so as to write and speak it easily.''}

\begin{longtable}[]{@{}
  >{\raggedright\arraybackslash}p{(\columnwidth - 6\tabcolsep) * \real{0.22}}
  >{\raggedright\arraybackslash}p{(\columnwidth - 6\tabcolsep) * \real{0.09}}
  >{\raggedleft\arraybackslash}p{(\columnwidth - 6\tabcolsep) * \real{0.08}}
  >{\raggedright\arraybackslash}p{(\columnwidth - 6\tabcolsep) * \real{0.61}}@{}}
\toprule
Condition & Code & Mean & Excerpt \\
\midrule
\endhead
No-Context Baseline & C5 & 1.0 & ``I don't have specific information
about Hamerton's situation at Loch Awe or his particular difficulties
with spoken French\ldots{}'' \\
Spec Only & C2a & 3.6 & ``I need to flag that I don't have direct
knowledge of Hamerton or this specific incident at Loch Awe \ldots{} the
Specification itself warns against fabricating a prediction without
grounding\ldots{}'' \\
All Facts & C4 & 1.8 & ``I cannot find any information in the provided
facts about Hamerton's experience at Loch Awe or any difficulty he had
following spoken French\ldots{}'' \\
All Facts + Spec & C4a & 2.2 & ``I need to flag immediately that this
scenario is not present in the known facts \ldots{} I should not
construct a behavioral prediction from absent evidence.'' \\
Raw Corpus & C8 & 1.4 & ``The autobiography provided does not contain
any passage about Hamerton having difficulty following spoken French at
Loch Awe\ldots{}'' \\
Raw Corpus + Spec & C9 & 4.2 & ``Based on the Behavioral Specification
and autobiography together, I can infer what Hamerton would likely
do\ldots{} He would persist through disciplined, sustained
practice\ldots{} The swimming episode is the clearest model: humiliated
at Doncaster for falsely claiming to swim, he resolved to become
genuinely expert\ldots{}'' \\
\bottomrule
\end{longtable}

Five of six conditions refuse to engage substantively because the
specific Loch Awe scenario is not narrated verbatim in the training-half
corpus. C9 (corpus + Spec) is the one condition that produces a
substantive prediction, integrating the Spec's axiomatic patterns
(disciplined formation through shame-driven perseverance, the
swimming-precedent pattern) with the autobiographical narrative to
project the character pattern onto a novel scenario. The shift from
refusal (anchor 1) to grounded prediction (anchor 4) on a single
question is what the multi-anchor measure captures across the
\textasciitilde12 such questions in this comparison (2.4\% of 507
paired).

\begin{center}\rule{0.5\linewidth}{0.5pt}\end{center}

\hypertarget{compression-examples-hamerton-spec-succeeds-ebers-spec-fails}{%
\subsection{Compression examples: Hamerton (Spec succeeds), Ebers (Spec
fails)}\label{compression-examples-hamerton-spec-succeeds-ebers-spec-fails}}

\begin{quote}
\textbf{Example: Hamerton, the compression story at its clearest}

Hamerton is the only subject in the low-baseline band where the Spec
alone outperforms the full raw corpus. The Spec-alone lift (+1.37 over
baseline) exceeds the corpus-alone lift (+1.01) on a 25,231-word source,
the smallest in the study (\textasciitilde5× compression). Spec alone
scores 2.63 vs.~raw corpus at 2.27. Facts+Spec reaches 2.77. Corpus+Spec
reaches 3.09, the highest compression-related score in the study,
indicating Spec and corpus are complementary rather than overlapping.

When the source corpus is short enough to be sparse on its own,
structured extraction adds organizational value beyond mere content.
Hamerton is the boundary condition for the compression claim; the
cross-subject mean still shows the corpus retaining a small edge on
average.
\end{quote}

\begin{quote}
\textbf{Example: Ebers, the honest cost of compression}

Ebers is the boundary case where structure alone is not enough. At
96,174 words and the study's lowest baseline (1.02), every context
condition lifts the score above baseline, but the Spec alone (1.54)
underperforms the raw corpus (2.18) by 0.64 points, the widest
Spec-vs-corpus gap in the low-baseline band. That 0.64 gap is large
enough to cross a rubric anchor: the corpus moves Ebers's average
response from ``refuses or off-base'' (anchor 1) into ``wrong
prediction'' (anchor 2), while the Spec stays at the floor. On Ebers the
raw text carries something the compressed Spec misses.

The corpus's edge is qualitative: anecdotal specificity. On Ebers Q2,
the corpus cites childhood incidents (Ebers's clashes with working-class
``Knoten'' boys), named mentors (Langethal, Barop, Froebel), and direct
autobiographical quotes. The Spec reaches the same conclusion through
behavioral axioms (Truth Over Embellishment, the reasoning signature),
principled reconstruction rather than lived evidence:

\begin{quote}
\textbf{C2a (Spec only):} ``Ebers would show receptiveness to this
critique \emph{if} it comes from a mentor he has already internalized as
exemplary \ldots{} He would accept that his lived experience is
necessarily limited in scope, but resist the implication that his
\emph{method} of understanding is fundamentally flawed.''

\textbf{C8 (raw corpus):} ``Ebers explicitly acknowledges gaps in his
understanding \ldots{} He reflects on his childhood fights with the
`Knoten' (working-class boys) and later recognizes the injustice of
class-based mockery: `If they had called us boobies we should probably
have called them blockheads \ldots{} Children don't fight regularly with
those whom they despise.' Ebers demonstrates profound deference to
figures like Langethal, Barop, and Froebel.''
\end{quote}

The gap is the rubric registering anecdotal texture over principled
reconstruction. Compression captures the bulk of the signal; the
residual on Ebers is autobiographical specificity the Spec abstracts
into axioms by design.
\end{quote}

\begin{center}\rule{0.5\linewidth}{0.5pt}\end{center}

\textbf{Why this matters for deployment.}

At any scale where a per-user full autobiography cannot be served into
context on every query, the compression result is what makes
personalization operationally tractable. The Specification's 7K-token
footprint is within normal per-request context budgets. A
100,000-to-400,000-word corpus is not. \textbf{The Specification
recovers 75\% of the corpus's predictive benefit at 4\% of the context
cost; the remaining \textasciitilde25\% is achievable but only at
production-prohibitive context sizes.}\footnote{Raw per-subject data at
  \texttt{results/global\_\textless{}subject\textgreater{}/c8\_c9\_results.json}
  and
  \texttt{results/global\_\textless{}subject\textgreater{}/results\_v2.json}.
  The compression analysis is in
  \texttt{scripts/\allowbreak{}recompute\_\allowbreak{}5judge\_\allowbreak{}primary.\allowbreak{}py}; per-question
  improvement rates are produced by
  \texttt{scripts/\allowbreak{}\_\allowbreak{}compute\_\allowbreak{}per\_\allowbreak{}question\_\allowbreak{}v2.\allowbreak{}py} and rendered by
  \texttt{scripts/\allowbreak{}generate\_\allowbreak{}fig\_\allowbreak{}4\_\allowbreak{}2\_\allowbreak{}1\_\allowbreak{}v3.\allowbreak{}py}. Figure 4.2 plots score
  versus context size (log scale) per subject and shows the steep
  initial climb and long plateau.}

\begin{center}\rule{0.5\linewidth}{0.5pt}\end{center}

\hypertarget{three-statistical-signatures}{%
\subsection{4.2.2 Three statistical
signatures}\label{three-statistical-signatures}}

The pre-vs-post Spearman ρ across questions captures how much the Spec
changes which questions get answered well. A low ρ means the question
ranking shifted substantially (re-ranking); a high ρ means ranking was
preserved and most answers were lifted by a similar amount (uniform
lift). The Spec produces three distinct signatures depending on what
context the model already has:

\begin{itemize}
\tightlist
\item
  \textbf{Re-ranking (Baseline → all facts + Spec, ρ = 0.27).} Different
  questions become the well-answered ones; on an empty baseline the Spec
  changes which questions the model can handle.
\item
  \textbf{Near-uniform lift (All Facts → all facts + Spec, ρ = 0.72; Raw
  corpus → corpus + Spec, ρ = 0.71).} The same questions stay strong; on
  top of facts or corpus the Spec lifts most answers by a similar amount
  without re-ranking.
\item
  \textbf{Partial re-ranking (Spec only → all facts + Spec, ρ = 0.62).}
  Mixed: adding facts on top of the Spec reorders some questions and
  uniformly lifts others.
\end{itemize}

The re-ranking signature on the empty-baseline comparison could mean
either that the Spec lets the model answer a different set of questions,
or simply that baseline scores cluster near the rubric floor where
reordering is structurally easier; a future test with a
non-floor-anchored baseline would distinguish the two readings.

\begin{center}\rule{0.5\linewidth}{0.5pt}\end{center}

\hypertarget{mechanism-correct-content-not-format}{%
\subsection{4.3 Mechanism: Correct Content, Not
Format}\label{mechanism-correct-content-not-format}}

\textbf{Hypothesis tested in this section} (H3 from \hyperref[what-we-tested]{§1.2}): The benefit
comes from the content of the correct Specification for the correct
person, not from the mere presence of a structured prompt. A random
other person's specification, applied in its place, does not reproduce
the effect.

\begin{center}\rule{0.5\linewidth}{0.5pt}\end{center}

\textbf{Increases in representational accuracy are not due to formatting
alone, but to the content of the Specification itself.} Wrong Spec (C2c)
drops accuracy below the No-Context Baseline (adversarial v1, Δ =
−0.25). The correct Spec (C2a) raises it above baseline (Δ = +0.35). The
wrong-vs-correct gap is 0.60 points, more than half a rubric-anchor
category, and random-derangement v2 sits between the two (+0.15). If
structure alone drove the effect, mismatched specs would recover most of
the lift. They recover at most part of it (random v2 = +0.15) or
actively degrade below baseline (adversarial v1 = −0.25). The +0.15
random lift is coincidental content overlap, not a structure-alone
effect: occasionally the wrong Spec's pattern happens to predict the
same surface behavior as the correct subject's by chance.\footnote{Example
  B below traces one such case. The scoring rubric's coverage of
  right-action-wrong-logic situations is discussed in \hyperref[measurement-apparatus]{§6.2}.} The
asymmetry also speaks to the specifications themselves: they are
sufficiently person-specific that mismatches register as content errors,
not as inert structural prompts.

On the 13 global subjects with complete 5-judge primary coverage, three
conditions test whether content matters:

\begin{longtable}[]{@{}
  >{\raggedright\arraybackslash}p{(\columnwidth - 4\tabcolsep) * \real{0.30}}
  >{\raggedleft\arraybackslash}p{(\columnwidth - 4\tabcolsep) * \real{0.40}}
  >{\raggedright\arraybackslash}p{(\columnwidth - 4\tabcolsep) * \real{0.30}}@{}}
\toprule
Condition & Mean Δ vs.~C5 (5-judge primary, 13 globals) & Reading \\
\midrule
\endhead
C2a (correct Spec) & \textbf{+0.35} & matched content increases
representational accuracy \\
C2c v2 (random derangement, seed-fixed) & \textbf{+0.15} & partial
improvement; dominated by floor effects on low-baseline subjects \\
C2c v1 (fixed derangement, cultural/temporal distance maximized) &
\textbf{−0.25} & adversarial mismatch degrades representational accuracy
below the No-Context Baseline \\
\bottomrule
\end{longtable}

The two wrong-Spec variants differ by construction. \textbf{Wrong Spec
V1 Fixed Derangement (C2c v1)} is a hardcoded pairing designed so each
subject receives the Specification of a culturally- and
temporally-distant other (for example, Ebers the 19th-century German
Egyptologist receives Equiano the 18th-century West-African/British
autobiographer; Seacole the 19th-century Jamaican nurse receives Bernal
Díaz the 16th-century Spanish conquistador).\footnote{Pairing logic in
  \texttt{scripts/\allowbreak{}run\_\allowbreak{}global\_\allowbreak{}rerun.\allowbreak{}py}.} \textbf{Wrong Spec V2 Random
Derangement (C2c v2)} is a seed-fixed random permutation in which no
subject receives its own specification but pairings can land
culturally-close; this tempers the aggregate drop. Reporting both shows
that even a random wrong-Spec barely improves on no context, and an
adversarial wrong-Spec actively hurts.\footnote{Both wrong-Spec deltas
  hide per-subject variation. The full per-subject table, the count of
  subjects with positive vs.~negative outcome under each protocol, and
  the adversarial-vs-random pairing sensitivity are in \hyperref[wrong-spec-derangement-protocol-sensitivity]{§4.6.5}.}

\begin{center}\rule{0.5\linewidth}{0.5pt}\end{center}

\textbf{Three mechanism types.}

Three distinct mechanisms produce the correct-specification improvement
across the study data. Each has a characteristic wrong-specification
failure mode, illustrated in the matched examples below.

\begin{enumerate}
\def\labelenumi{\arabic{enumi}.}
\tightlist
\item
  \textbf{Identity disambiguation.} When the baseline model cannot
  determine which person is being asked about, the Specification
  provides enough content (temporal markers, cultural domain, documented
  life events) to resolve the identity and anchor the reasoning frame.

  \begin{itemize}
  \tightlist
  \item
    \emph{Wrong-Spec failure mode:} the model either detects the
    mismatch explicitly and refuses to predict, or anchors on the wrong
    person's pattern and produces a coherent but off-target prediction.
  \end{itemize}
\item
  \textbf{Directional correction.} When retrieved facts suggest a
  generic-default prediction that contradicts the subject's actual
  pattern, the Specification overrides the generic with the
  subject-specific.

  \begin{itemize}
  \tightlist
  \item
    \emph{Wrong-Spec failure mode:} the model applies the wrong person's
    pattern; depending on how close that pattern happens to be to the
    target subject's, the prediction is either directionally wrong in a
    new way or coincidentally correct (the wrong person's pattern
    happens to predict the same surface behavior on this particular
    question, for different underlying reasons; Example B below is one
    such case).
  \end{itemize}
\item
  \textbf{Interpretive inference.} When retrieved facts do not include
  direct evidence for the specific question, the Specification provides
  interpretive scaffolding to generalize from established character
  patterns to the new situation.

  \begin{itemize}
  \tightlist
  \item
    \emph{Wrong-Spec failure mode:} the model detects the mismatch and
    refuses, or applies wrong-person scaffolding and produces a
    low-quality prediction.
  \end{itemize}
\end{enumerate}

\begin{center}\rule{0.5\linewidth}{0.5pt}\end{center}

\textbf{Response-level evidence: when the model engages with the Spec,
and when it does not.}

Three signals from response text confirm that content matters more than
structure.

\textbf{Spec-tag citation gap.} Models cite Spec-specific tags (anchor
IDs, axiom references, predictive-template labels) on \textbf{78.6\%} of
correct-Spec responses but only \textbf{50.0\%} of wrong-Spec
responses.\footnote{Data at
  \texttt{docs/\allowbreak{}research/\allowbreak{}spec\_\allowbreak{}activation\_\allowbreak{}analysis.\allowbreak{}json}.} The
28.6-point gap is a lower bound on the content effect; models may draw
on Spec content without literally quoting tag IDs.

\textbf{Models can detect when a Specification does not fit the named
subject.} Across 587 wrong-Spec responses, \textbf{60.6\% explicitly
flagged the content mismatch} (example: \emph{``This is a behavioral
model of a 16th-century Central Asian military ruler, almost certainly
Bābur''}). 36.5\% attempted to apply the mismatched content and produced
low-quality predictions; 3\% hedged or were ambiguous.\footnote{587
  wrong-Spec responses: 507 from the v2 random-derangement protocol on
  the 13 global subjects, plus 80 from the v1 adversarial protocol on
  Hamerton across all five battery tiers. Validated against a
  30-response stratified manual spot check. Full breakdown: 60.6\%
  explicit mismatch flag, 36.5\% applied mismatched content, 2.0\%
  hedged implicitly, 0.9\% ambiguous.} The detection signal is
interpretive content (temporal markers, cultural domain, documented life
events) being inconsistent with what the model already knows about the
named subject. Specifications are anonymized (\hyperref[pipeline-for-the-behavioral-specification]{§3.7}), so the model has no
surface name cue to compare against; only interpretive content.

\textbf{Hedging persists under wrong-Spec.} Correct-Spec conditions
nearly eliminate baseline hedging: 41.2\% → 0.4\% under the
broader-pattern classifier.\footnote{Strict-pattern classifier yields
  28.8\% → 0.0\%. Rule definitions in
  \protect\hyperlink{what-we-found}{§1.3} footnote.} On wrong-Spec,
refusal patterns persist at the 60.6\% explicit-detection rate above.
The hedging-elimination is content-specific, not structure-specific.

\begin{center}\rule{0.5\linewidth}{0.5pt}\end{center}

\textbf{Per-question matched comparison.}

The three \hyperref[the-cross-subject-gradient-and-its-per-question-mechanism]{§4.1} examples extend directly into mechanism-by-mechanism
wrong-Spec contrasts. Same subject, same question, same response model;
the only change is the Specification content. The per-question range is
wider than the +0.60 mean gap in both directions. Correct-Spec C4a
responses are in \hyperref[the-cross-subject-gradient-and-its-per-question-mechanism]{§4.1} for reference.

\hypertarget{wrong-spec-examples-ebers-q7-identity-bernal-duxedaz-q16-frameworks-seacole-q2-inference}{%
\subsection{Wrong-Spec examples: Ebers Q7 (identity), Bernal Díaz Q16
(frameworks), Seacole Q2
(inference)}\label{wrong-spec-examples-ebers-q7-identity-bernal-duxedaz-q16-frameworks-seacole-q2-inference}}

\begin{quote}
\textbf{Example A (wrong-Spec), Ebers Q7: identity disambiguation fails}

\textbf{Context.} Clean-mismatch case: a Spec that describes someone
other than the named subject, served under the v1 fixed-derangement
protocol. The model detects the mismatch and declines: 1.60 vs.~correct
Spec 3.60, a 2.00-point drop.

\textbf{Subject + question.} Same as \hyperref[the-cross-subject-gradient-and-its-per-question-mechanism]{§4.1} Example A (Ebers,
self-sacrifice and educational institutional success). Battery-question
target name: Ebers (explicit in the prompt).

\textbf{Wrong Spec served.} Equiano's specification,
anonymized.\footnote{Anonymization protocol in \hyperref[pipeline-for-the-behavioral-specification]{§3.7}. Pairing logic in
  \texttt{scripts/\allowbreak{}run\_\allowbreak{}global\_\allowbreak{}rerun.\allowbreak{}py}.} The response cites two of
Equiano's anchors: A4 Systemic Indictment and A9 Economic
Self-Determination.

\textbf{Wrong-Spec response.}

\begin{quote}
\emph{``I need to be direct: I don't have reliable access to specific
statements by Ebers on this question. The behavioral specification
you've provided is detailed and rich, but it doesn't contain material
where this person directly addresses the relationship between
self-sacrifice and educational institutional success. \ldots{} Given A4
(Systemic Indictment) --- that institutionalized greed is the generative
engine of cruelty \ldots{} a teacher sacrificing their own economic
security, health, or family time does not fix a system structured by
avarice.''}
\end{quote}

\textbf{Reading.} The model detected the mismatch between the named
target in the question (Ebers, a 19th-century German Egyptologist) and
the interpretive content of the anonymized specification (anti-slavery
and economic-freedom anchors, which are Equiano's). It named the served
anchors correctly, reasoned from Equiano's framework, and declined to
produce a prediction about Ebers. The identity-disambiguation mechanism
that enabled the correct Spec's lift in \hyperref[the-cross-subject-gradient-and-its-per-question-mechanism]{§4.1} Example A did not fire
because the Spec content is not about Ebers. The response falls near the
rubric floor because no prediction was made.
\end{quote}

\begin{quote}
\textbf{Example B (wrong-Spec), Bernal Díaz Q16: content convergence
across genuinely different frameworks}

\textbf{Context.} Coincidence case: random pairing happens to converge
on the same surface prediction by different logic. v1 fixed-derangement
protocol. Wrong Spec scored 4.60 vs.~correct Spec 4.80, a 0.20-point
near-tie. This is what produces the wrong-Spec aggregate's drift above
zero on the random-pairing v2 protocol.

\textbf{Subject + question.} Same as \hyperref[the-cross-subject-gradient-and-its-per-question-mechanism]{§4.1} Example B (Bernal Díaz, Cortes
and offered physical assistance). Battery-question target name: Bernal
Díaz.

\textbf{Wrong Spec served.} Sunity Devee's specification,
anonymized.\footnote{Anonymization protocol in \hyperref[pipeline-for-the-behavioral-specification]{§3.7}. Pairing logic in
  \texttt{scripts/\allowbreak{}run\_\allowbreak{}global\_\allowbreak{}rerun.\allowbreak{}py}.} The response cites three of
her anchors: A5 Relational Identity, A9 Simplicity as Virtue, P5 Virtue
through Embodied Practice.

\textbf{Wrong-Spec response.}

\begin{quote}
\emph{``Based on this behavioral specification, the author would likely
decline the offered assistance \ldots{} Accepting help during a
strenuous activity would read as: a failure of self-discipline, an
implicit claim to comfort or ease that contradicts spiritual
seriousness, a reliance on others that undermines the embodied practice
of virtue (A9, P5) \ldots{} Physical exertion endured without assistance
becomes evidence of character.''}
\end{quote}

\textbf{Reading, not parroting.} The two specs are genuinely different
frameworks. Bernal Díaz's actual anchors include A1 Divine Mandate, A2
Civilizational Hierarchy, A4 Loyalty Architecture, A5 Forward
Compulsion, A6 Gold as Proof, and A10 Charismatic Override: a
conquistador's martial-providential register. Sunity Devee's anchors
cited here, A5 Relational Identity, A9 Simplicity as Virtue, and P5
Virtue through Embodied Practice, are an ascetic-devotional register.
Direct anchor-to-anchor comparison across the two specs finds zero
substantive mirroring. On the specific question of refusing offered
physical help, the two frameworks converge by different logics. The
correct Spec (Bernal Díaz) predicts refusal because accepting help would
signal weakness to followers and violate performative self-reliance (A4
+ A5 in the conquistador register). The wrong Spec (Sunity Devee)
predicts refusal because accepting help would compromise physical
discipline and violate simplicity-as-virtue (A9 + P5 in the devotional
register). Different moral architectures, same overt behavior.

\textbf{Why the correct Spec still scored higher, 4.80 vs.~4.60.} Both
conditions predicted the right surface action. The 0.20-point gap is
judge preference for rationale specificity that matches the ground-truth
passage's tone. The correct Spec's ``symbolic or morale-signaling
purpose'' rationale maps onto a battlefield memoir's register. The wrong
Spec's ``spiritual seriousness, devotional simplicity'' rationale
predicts the same action but in a register alien to Cortés on the steps
of the Templo Mayor, the Aztec temple at the center of Tenochtitlan
where the moment in the held-out passage takes place. Judges reward
tonally-aligned rationale. The convergence is real but costs precision.

\textbf{Why this matters for the aggregate numbers.} This is the
coincidence case that lifts the wrong-Spec aggregate above zero. When
pairing happens to land on subjects whose behavioral patterns converge
on the same surface prediction for a given question, wrong-Spec lift is
real on that question, not an artifact. Across the 507-response v2
cohort (13 globals × 39 questions), mismatch-loss dominates on the
adversarial v1 pairing (aggregate Δ −0.25) and roughly balances on the
random v2 pairing (aggregate Δ +0.15). Example B is one of the roughly
5-10\% of questions where content coincidence produces correct-surface,
wrong-logic predictions.
\end{quote}

\begin{quote}
\textbf{Example C (wrong-Spec), Seacole Q2: interpretive inference
fails}

\textbf{Context.} Clean-mismatch case: the served Spec is from a
different century and culture than the named subject. v1
fixed-derangement protocol. Wrong Spec scored 1.40 vs.~correct Spec
5.00, a 3.60-point drop.

\textbf{Subject + question.} Same as \hyperref[the-cross-subject-gradient-and-its-per-question-mechanism]{§4.1} Example C (Seacole, delirious
patient). Battery-question target name: Mary Seacole.

\textbf{Wrong Spec served.} Bernal Díaz's specification,
anonymized.\footnote{Anonymization protocol in \hyperref[pipeline-for-the-behavioral-specification]{§3.7}. Pairing logic in
  \texttt{scripts/\allowbreak{}run\_\allowbreak{}global\_\allowbreak{}rerun.\allowbreak{}py}.} The response identifies the
wrong Spec by its interpretive content alone and names four of his
anchors: A1 Divine Mandate, A2 Civilizational Hierarchy, A4 Loyalty
Architecture, A5 Forward Compulsion.

\textbf{Wrong-Spec response.}

\begin{quote}
\emph{``The specification you've provided describes someone from the
16th-century Spanish conquest period --- the epistemic anchors, loyalty
architecture, divine mandate framing, and forward compulsion all point
to that historical context. Mary Seacole (1805-1881) was a Creole nurse
and travel writer from Jamaica who served in the Crimean War. These are
entirely different people from different centuries with different
frameworks. \ldots{} I don't have documented instances of this. I could
speculate based on her memoir's tone and her known character, but that
would be inference, not behavioral specification.''}
\end{quote}

\textbf{Reading.} The model detected the mismatch between the named
target in the question (Mary Seacole, a 19th-century Jamaican Creole
nurse) and the anonymized content of the served specification
(16th-century Spanish conquest anchors). It named the anchors it was
seeing, placed Seacole in a different era and role, and refused to apply
the mismatched content. The interpretive-inference mechanism that
produced \hyperref[the-cross-subject-gradient-and-its-per-question-mechanism]{§4.1} Example C's correct-Spec 5.00 score does not fire: without
Seacole's actual character pattern in context, the model would not
generalize from an unrelated conquistador's framework to her
delirious-patient scenario.
\end{quote}

\begin{center}\rule{0.5\linewidth}{0.5pt}\end{center}

\textbf{Summary of the three examples.}

\begin{longtable}[]{@{}
  >{\raggedright\arraybackslash}p{(\columnwidth - 10\tabcolsep) * \real{0.14}}
  >{\raggedright\arraybackslash}p{(\columnwidth - 10\tabcolsep) * \real{0.14}}
  >{\raggedleft\arraybackslash}p{(\columnwidth - 10\tabcolsep) * \real{0.19}}
  >{\raggedleft\arraybackslash}p{(\columnwidth - 10\tabcolsep) * \real{0.19}}
  >{\raggedleft\arraybackslash}p{(\columnwidth - 10\tabcolsep) * \real{0.19}}
  >{\raggedright\arraybackslash}p{(\columnwidth - 10\tabcolsep) * \real{0.14}}@{}}
\toprule
Example & Mechanism (correct Spec) & C4a (correct) & C2c v1 (wrong) &
Drop & Wrong-Spec pattern \\
\midrule
\endhead
A (Ebers Q7) & Identity disambiguation + interpretive inference & 3.60 &
1.60 & \textbf{−2.00} & Explicit mismatch flag; declined prediction \\
B (Bernal Díaz Q16) & Directional correction & 4.80 & 4.60 &
\textbf{−0.20} & Coincidental content overlap; wrong-Spec prediction
matches \\
C (Seacole Q2) & Interpretive inference & 5.00 & 1.40 & \textbf{−3.60} &
Explicit mismatch flag; declined prediction \\
\bottomrule
\end{longtable}

Two of three examples show large drops (−2.00 to −3.60 points) when the
content does not fit. The third shows near-zero drop, but only because
the wrong Spec's content happens to predict the same surface behavior.
That asymmetry, clean mismatches versus coincidental overlaps, is
exactly what the aggregate Δ numbers reflect: the adversarial-pairing v1
aggregates to −0.25 because most questions are mismatch cases, and the
random-pairing v2 aggregates to +0.15 because random pairings more often
hit content-proximity combinations like Example B.\footnote{Raw
  per-judge data and full response text at
  \texttt{results/global\_\textless{}subject\textgreater{}/results\_v2.json}
  (wrong-Spec responses) and
  \texttt{results/global\_\textless{}subject\textgreater{}/judgments\_v2.json}
  (per-judge scores). Analysis scripts at
  \texttt{scripts/\allowbreak{}compute\_\allowbreak{}wrong\_\allowbreak{}spec\_\allowbreak{}5judge.\allowbreak{}py} and
  \texttt{scripts/\allowbreak{}compute\_\allowbreak{}wrong\_\allowbreak{}spec\_\allowbreak{}per\_\allowbreak{}subject.\allowbreak{}py}.}

\begin{center}\rule{0.5\linewidth}{0.5pt}\end{center}

\hypertarget{memory-system-composition}{%
\subsection{4.4 Memory-system
composition}\label{memory-system-composition}}

\textbf{Hypothesis tested in this section} (H4 from \hyperref[what-we-tested]{§1.2}): A Behavioral
Specification layers cleanly on top of memory-system retrieval rather
than replacing it. The Spec contributes representational accuracy beyond
what retrieval alone provides, and that contribution decomposes into
per-question patterns characteristic of each retrieval architecture.

\begin{center}\rule{0.5\linewidth}{0.5pt}\end{center}

\hypertarget{cross-system-retrieval-providers-do-not-converge}{%
\subsection{4.4.1 Cross-system retrieval: providers do not
converge}\label{cross-system-retrieval-providers-do-not-converge}}

\textbf{On 35.9\% of instances, any two memory systems share zero facts
in their top-10. They retrieve no overlapping facts at all on the same
question. Averaged across all ten system pairings, the mean overlap is
8.3\%.} Recall benchmarks like LongMemEval and LOCOMO measure whether a
system can retrieve a previously-stored fact, and the four commercial
systems we tested score within a few percentage points of each other on
those benchmarks. Representational accuracy and behavioral prediction
operate at a different layer, where the relevant question is which facts
matter for a specific interpretive task.

\textbf{Setup.} Four commercial memory systems (Mem0, Letta,
Supermemory, Zep) and Base Layer's own zero-cost retrieval substrate
(MiniLM-L6-v2 + ChromaDB) were tested under two configurations. We ran
both to separate two layers of variation: how each system \emph{ranks}
facts from a fixed pool (controlled), and whether each system's
\emph{ingestion} pipeline adds further divergence on top (native).

\begin{itemize}
\tightlist
\item
  \textbf{Controlled configuration.} Each system is given an identical
  pre-extracted fact pool drawn from the training half of each subject's
  corpus. Holds the input constant across systems; differences trace to
  retrieval and presentation policy alone.
\item
  \textbf{Native configuration.} Each system ingests the raw training
  corpus through its own production pipeline, as in deployment. Measures
  the full end-to-end system.
\end{itemize}

\textbf{Convergence test.} No, the provider layer does not converge on
relevance. On 35.9\% of (system pair, question) instances two systems
share zero facts in their top-10s; on 65.6\% they share one or fewer;
the mean pairwise overlap across the ten system pairs is
8.3\%.\footnote{Cuts: full set (14 subjects × 10 pairs, n=5,460) →
  share-zero 35.9\%; excluding Hamerton (13 globals, n=5,070) → 36.7\%;
  restricted to four commercial systems (six pairs, no Base Layer) →
  40.4\% (14 subjects) / 41.0\% (13 globals). Every cut shows
  substantial top-K divergence on identical input. Data at
  \texttt{docs/\allowbreak{}research/\allowbreak{}retrieval\_\allowbreak{}overlap\_\allowbreak{}analysis\_\allowbreak{}20260501.\allowbreak{}json}.}
Convergence on top-K under identical input would have been evidence of a
shared interpretive substrate. The lack of convergence suggests the
rankings reflect provider-specific design choices rather than a shared
theory of which facts the question is asking for. The controlled
configuration isolates the ranking layer as a load-bearing source of
representational accuracy: which facts the system surfaces determines,
before any reading model engages, what the response can be about.

For any two systems on the same question, the fraction of retrieved
facts that appear in both lists is called the \textbf{Jaccard
similarity}.\footnote{Jaccard similarity is computed as the size of the
  intersection divided by the size of the union of the two top-K sets.}
A Jaccard of 1.0 means identical top-10s; 0.0 means no shared facts. As
a calibration anchor: Jaccard ≥ 0.5 indicates the majority of retrieved
facts are shared (substantial overlap); Jaccard in the 0.2-0.5 range
indicates moderate overlap; Jaccard \textless{} 0.2 indicates that
systems are returning largely different facts on identical input. We
compute it for each of the ten system pairs (Base Layer, Letta, Mem0,
Supermemory, Zep) on each of 546 questions, in the controlled
configuration where every system reads the same all-facts pool.

\textbf{Pairwise Jaccard similarity (controlled configuration, n=546
questions per pair).}\footnote{Jaccard is computed on unique fact text
  per the ``note on counting'' paragraph below; the denominator uses
  unique fact text on both sides, which prevents duplicate-driven
  inflation. Letta's effective unique-fact count is \textasciitilde3.5
  per top-10 (graph traversal repeats facts under different relational
  paths); the Jaccard math handles this correctly.}

\begin{longtable}[]{@{}lr@{}}
\toprule
System pair & Mean Jaccard \\
\midrule
\endhead
Base Layer ↔ Supermemory & 0.146 \\
Mem0 ↔ Letta & 0.126 \\
Base Layer ↔ Mem0 & 0.123 \\
Mem0 ↔ Supermemory & 0.114 \\
Letta ↔ Supermemory & 0.099 \\
Base Layer ↔ Letta & 0.092 \\
Mem0 ↔ Zep & 0.056 \\
Base Layer ↔ Zep & 0.027 \\
Letta ↔ Zep & 0.026 \\
Supermemory ↔ Zep & 0.025 \\
\textbf{Mean across pairs} & \textbf{0.083} \\
\bottomrule
\end{longtable}

\begin{figure}
\centering
\includegraphics{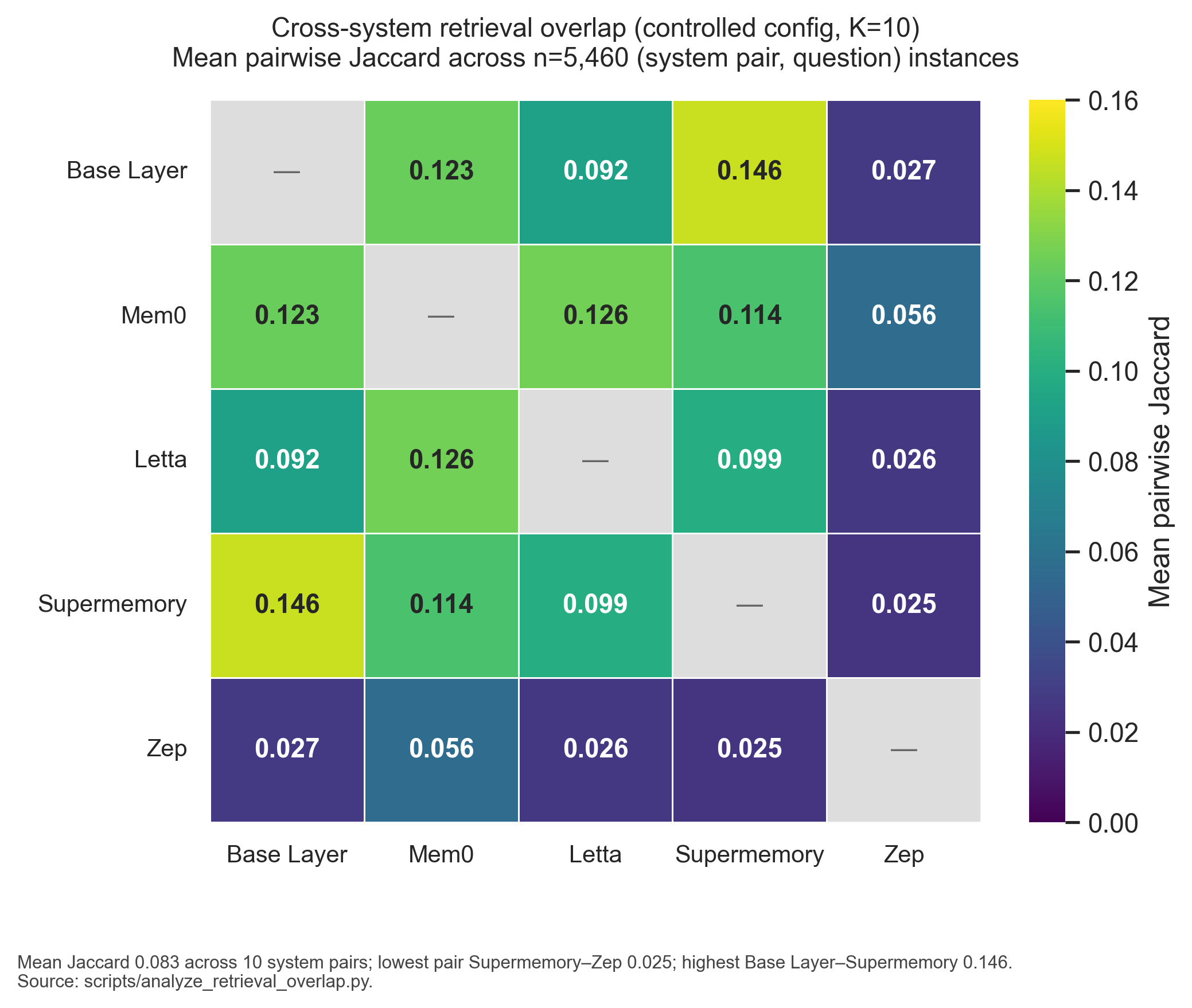}
\caption{Figure 4.4.1: Cross-system retrieval overlap. Mean pairwise
Jaccard (the fraction of facts two systems both surface as relevant for
a given question) between every pair of memory systems on the controlled
retrieval configuration (n=5,460 = 14 subjects × 39 questions × 10
system pairs). The diagonal is grayed; cells below the diagonal mirror
cells above. The highest agreement is Base Layer--Supermemory at 0.146
(about 1 fact in 7 shared); the lowest is Supermemory--Zep at 0.025
(about 1 fact in 40). Zep's row is uniformly low because its
graph-traversal scoring picks different facts than embedding-similarity
retrieval. Mean across pairs is 0.083: any two systems share roughly 1
of every 12 top-10 facts.}
\end{figure}

Pairs involving Zep are the lowest. Zep ranks facts by traversing
relationships in a knowledge graph; the other systems rank by how close
a fact's meaning is to the question (embedding similarity). The two
ranking approaches surface different facts on identical input, so Zep
overlaps weakly with everyone else.

A note on counting. We count \emph{unique facts} in each top-10: if the
same fact text appears more than once in a system's returned list, we
count it once. Some systems return duplicates because their scoring
traverses graph relationships and surfaces the same fact under different
paths. Zep returns 10 entries per question, of which about 9.8 are
unique on average. Duplicates are rare for Zep, and ten entries
effectively means ten distinct facts. Letta also returns 10 entries, but
only about 3.5 are unique on average. Letta is largely returning the
same handful of facts under different graph paths. So while both systems
advertise top-10 retrieval, Letta's effective retrieval depth is roughly
a third of Zep's.

Comparing each commercial system to the average of the others' rankings:
Mem0 is the closest match (mean Jaccard 0.105 across Mem0's four
cross-pairs), followed by Supermemory (0.096), Letta (0.086), and Zep
(0.034). Mem0 retrieves what the average of the other systems retrieves;
Zep retrieves a different set.\footnote{Mean pairwise Jaccard across ten
  pairs: 0.083 raw, 0.088 normalized. Per-subject range: 0.043 (Equiano)
  to 0.115 (Hamerton). Per-category variation small (0.076--0.093). Data
  at \texttt{docs/\allowbreak{}research/\allowbreak{}retrieval\_\allowbreak{}overlap\_\allowbreak{}analysis\_\allowbreak{}20260501.\allowbreak{}json}.}

Under the native pipeline, the lack of shared facts is even more
pronounced. Each system returns its retrieval in a different format:
Mem0 returns third-person summary sentences, Letta returns raw
multi-sentence passages, Supermemory returns atomic facts, Zep returns
rows extracted from a knowledge graph. Because the four systems return
content in different shapes, any two systems share zero exactly-matching
facts on the same question, and pairwise overlap drops to 0.000 across
all four native pairs.

\textbf{Semantic-similarity sensitivity check.} Pairwise Jaccard counts
only exact-string matches, which understates overlap when systems return
the same content in different wording. We rerun the same overlap
analysis under a semantic-similarity threshold: two facts count as a
match when their sentence embeddings have cosine similarity at or above
a specified threshold. At the near-duplicate threshold (cosine ≥ 0.85),
native overlap rises to 0.004; at the loose topical threshold (cosine ≥
0.70), it rises to 0.016. The divergence is structural, not a
surface-form artifact. The same check applied to the controlled
configuration above also leaves the divergence intact; full sensitivity
grid in \hyperref[retrieval-overlap-sensitivity-semantic-similarity-matching-k-variation]{§4.6.6}.

Whether divergent facts produce divergent answers is the practical test
of whether retrieval differences matter. \hyperref[layering-the-spec-aggregate-ux3b4-across-systems-and-ingestion-paths]{§4.4.2} (aggregate Δ across
systems and ingestion paths) and \hyperref[where-the-spec-helps-where-it-hurts-and-which-question-types-route-to-each]{§4.4.3} (per-question patterns where the
layer's effect is concentrated) take that up.

\begin{center}\rule{0.5\linewidth}{0.5pt}\end{center}

\hypertarget{layering-the-spec-aggregate-ux3b4-across-systems-and-ingestion-paths}{%
\subsection{4.4.2 Layering the Spec: aggregate Δ across systems and
ingestion
paths}\label{layering-the-spec-aggregate-ux3b4-across-systems-and-ingestion-paths}}

Layered on top of three of four commercial memory systems (Mem0, Letta,
Zep), the Behavioral Specification produces a net-positive aggregate Δ
across the 14 main-study subjects. Wilcoxon signed-rank confirms
direction at α = 0.01 on four (system, configuration) cells. Each
system's aggregate Δ is the balance of per-question patterns the mean
hides; \hyperref[where-the-spec-helps-where-it-hurts-and-which-question-types-route-to-each]{§4.4.3} decomposes them.

\textbf{Conditions compared.} Within each system in each configuration:
- \textbf{C1 (Retrieval Only):} the memory system's retrieval served as
context; no Behavioral Specification. - \textbf{C3 (retrieval + Spec):}
the same retrieval plus the full Behavioral Specification.

The Spec-effect for that system is \textbf{Δ\_spec = mean(C3 retrieval +
Spec) − mean(C1 Retrieval Only)}, aggregated per subject, then averaged
across subjects. If the Specification helps memory-system performance,
Δ\_spec is positive across systems.

\begin{center}\rule{0.5\linewidth}{0.5pt}\end{center}

\textbf{Aggregate and per-question results (5-judge primary, all 14
main-study subjects).}

\begin{longtable}[]{@{}
  >{\raggedright\arraybackslash}p{(\columnwidth - 10\tabcolsep) * \real{0.14}}
  >{\raggedright\arraybackslash}p{(\columnwidth - 10\tabcolsep) * \real{0.14}}
  >{\raggedleft\arraybackslash}p{(\columnwidth - 10\tabcolsep) * \real{0.18}}
  >{\raggedleft\arraybackslash}p{(\columnwidth - 10\tabcolsep) * \real{0.18}}
  >{\raggedleft\arraybackslash}p{(\columnwidth - 10\tabcolsep) * \real{0.18}}
  >{\raggedleft\arraybackslash}p{(\columnwidth - 10\tabcolsep) * \real{0.18}}@{}}
\toprule
System & Config & Δ\_spec & \% subjects improved & \% questions up ≥1
anchor & \% questions up ≥2 anchors \\
\midrule
\endhead
Mem0 & controlled & +0.12 & 71\% & 24.0\% & 3.1\% \\
Mem0 & native & +0.33 & 71\% & \textbf{37.1\%} & \textbf{8.1\%} \\
Letta (archival) & controlled & +0.20 & 86\% & 28.1\% & 6.1\% \\
Letta (archival) & native & −0.02 & 36\% & 20.5\% & 0.9\% \\
Zep & controlled & +0.19 & 93\% & 29.3\% & 5.3\% \\
Zep & native & +0.33 & 93\% & \textbf{35.3\%} & \textbf{7.5\%} \\
Supermemory & controlled & −0.05 & 36\% & 18.6\% & 2.5\% \\
Supermemory & native\footnote{Supermemory native covers 10 of 14
  subjects after paid-tier rerun failures on Bābur, Bernal Díaz,
  Cellini, and Rousseau (n=221 paired questions vs.~\textasciitilde546
  expected). Per-question rates conditional on the 10 covered subjects.}
& −0.01 & 43\% & 19.0\% & 1.4\% \\
Base Layer substrate & controlled & +0.08 & 64\% & 26.7\% & 3.9\% \\
\bottomrule
\end{longtable}

Headline numbers report the all-14 panel. Three of four commercial
systems produce a positive aggregate Δ\_spec under at least one
configuration; Supermemory aggregates near zero under both. Base Layer's
substrate produces the smallest positive Δ; the interpretive improvement
comes from the Specification itself, not from retrieval choices.

\textbf{Aggregate Δ\_spec masks substantial per-question variance.}
Every system lifts 18--37\% of questions by at least one rubric anchor,
and 1--8\% by two or more anchors. Both are categorical changes per the
\hyperref[score-interpretation]{§3.3.1} cross-anchor interpretation rule. Mean Δ averages over a
population where the Spec produces categorical change on a subset of
questions and small adjustments or losses elsewhere. Even Supermemory,
with a near-zero aggregate, lifts 18.6\% of questions by at least one
anchor (controlled). The per-question redistribution is decomposed in
\hyperref[where-the-spec-helps-where-it-hurts-and-which-question-types-route-to-each]{§4.4.3}.

\textbf{Wilcoxon signed-rank confirms direction at α = 0.01} for Zep
controlled (\emph{p} = 0.0004), Letta controlled (\emph{p} = 0.0017),
Mem0 native (\emph{p} = 0.0088), and Zep native (\emph{p} = 0.0015).
Mem0 controlled is significant at α = 0.05 (\emph{p} = 0.017). Letta
native, Supermemory (both configurations), and Base Layer substrate are
not significant at α = 0.05.\footnote{Non-significance here reflects
  statistical power, not absence of a Spec effect. Each per-(system,
  configuration) Wilcoxon test runs on 14 paired subjects, and cells
  with a smaller aggregate Δ\_spec (Supermemory's is near zero, \hyperref[where-the-spec-helps-where-it-hurts-and-which-question-types-route-to-each]{§4.4.3})
  do not clear α = 0.05 at that sample size. The same power limitation
  is why the 14-subject aggregate, not the 9-subject low-baseline band,
  is the headline: the band, at n=9, is underpowered at the effect sizes
  these systems show. Per-cell p-values are in the Appendix B.13
  Wilcoxon table.} The 9-subject low-baseline band was computed but is
statistically underpowered at the effect sizes these systems show, so we
do not lead with it; per-subject low-baseline detail is in the
footnote.\footnote{Full per-(system, configuration) Wilcoxon table and
  low-baseline Δ\_spec values in Appendix B.13. Data at
  \texttt{docs/\allowbreak{}research/\allowbreak{}stats\_\allowbreak{}update.\allowbreak{}md} and
  \texttt{docs/\allowbreak{}research/\allowbreak{}memory\_\allowbreak{}systems\_\allowbreak{}5judge\_\allowbreak{}primary.\allowbreak{}md}.}

\textbf{Native ingestion shapes how much room the Spec has to contribute
on top, and that interaction varies by system.} When the input is held
constant (controlled), four of five systems produce a positive Δ\_spec;
the Specification's contribution is visible on top of an identical fact
pool. When each system ingests its own way (native), the systems split.
Mem0 and Zep increase under native (Mem0 +0.12 → +0.33, Zep +0.19 →
+0.33). Letta decreases sharply (+0.20 → −0.02). Supermemory stays
roughly flat (−0.05 → −0.01). Mem0's controlled-to-native lift of +0.21
is the largest among the four commercial systems, with Zep's at +0.14.
The mechanism for these splits is in \hyperref[where-the-spec-helps-where-it-hurts-and-which-question-types-route-to-each]{§4.4.3}: the Spec helps
retrieval-based systems on interpretation-heavy questions they were not
designed for, sometimes hurts on literal-recall questions retrieval
already answered, and induces principled refusals on questions where
retrieved facts cannot ground a prediction.

\begin{center}\rule{0.5\linewidth}{0.5pt}\end{center}

\hypertarget{where-the-spec-helps-where-it-hurts-and-which-question-types-route-to-each}{%
\subsection{4.4.3 Where the Spec helps, where it hurts, and which
question types route to
each}\label{where-the-spec-helps-where-it-hurts-and-which-question-types-route-to-each}}

\textbf{Three patterns of Spec-retrieval interaction emerge across all
five systems tested.} The same three patterns produce positive aggregate
Δ\_spec on three commercial systems and near-zero on Supermemory. What
changes between systems is how the Spec helps or hurts across the
question battery, not the patterns themselves.

\textbf{The three patterns:}

\begin{enumerate}
\def\labelenumi{\arabic{enumi}.}
\tightlist
\item
  \textbf{Interpretive supply.} When retrieval underdetermines the
  answer, the Specification provides interpretive scaffolding to
  generalize from established character patterns to the specific
  question. \emph{Increases representational accuracy on the question.}
  A signature of this pattern is the case where the Spec alone is not
  enough to produce a substantive answer but corpus-plus-Spec is: the
  response model carries an implicit evidentiary bar before it will
  commit to a behavioral prediction, and the corpus supplies the
  supporting evidence the Spec is interpreting. Where retrieval clears
  that bar with grounded facts, the Spec's interpretive layer composes
  with it; where retrieval falls below the bar, the Spec alone abstains
  rather than predict from pattern with no supporting evidence.
\item
  \textbf{Over-theorization.} When retrieval already supplies the plain
  answer, the Specification can pull the response toward interpretive
  depth that the question does not call for. \emph{Decreases
  representational accuracy on the question.}
\item
  \textbf{Spec-induced refusal.} Specification axioms (which vary by
  subject; in this study, dignity, honoring-testimony, and
  epistemic-integrity axioms across different subjects) can trigger a
  meta-refusal where the model declines to predict. The current
  content-match rubric cannot distinguish principled refusal from a
  wrong prediction (\hyperref[rubric-handling-limitations-post-hoc-validity-audit]{§3.3.6}). \emph{Lowers the measured rubric score;
  whether it lowers actual representational accuracy depends on whether
  refusal was the correct behavior on that question.}
\end{enumerate}

\textbf{How each system shows these patterns:}

\begin{itemize}
\tightlist
\item
  \textbf{Mem0:} Mem0's atomic-fact retrieval surfaces clean,
  tightly-scoped facts. When the question requires interpretation beyond
  those facts, Pattern 1 lifts the response by supplying the framework.
  When the question is literal-recall and the atomic facts already
  resolve it, Pattern 2 takes over: the Spec abstracts away from the
  plain answer Mem0 already provided. Pattern 3 occurs but registers as
  a smaller rubric penalty because Mem0's retrieval often hedges near
  the rubric floor on refusal-triggering questions already, leaving
  little room for the Spec's refusal axioms to drag the score further
  down.
\item
  \textbf{Letta archival:} Letta's controlled retrieval returns 10
  entries per question but only about 3.5 are unique because graph
  traversal pulls the same source fact through multiple relational
  paths, returning it once per path; the unique-fact count is what the
  model actually has to reason from, not the raw entry count. When those
  few unique facts align with the Spec's interpretive frame, Pattern 1
  produces large lifts. When they do not align, the Spec either
  over-theorizes (Pattern 2) on the available content or refuses
  (Pattern 3) for lack of grounding evidence.
\item
  \textbf{Zep:} Zep's temporal-graph retrieval returns relational
  structure rather than atomic facts. The relationship-rich context
  tends to suit the Spec's interpretive framing across question types.
  Zep shows the most balanced pattern distribution: Pattern 1 lifts
  roughly as often as Pattern 2 hurts, with fewer large regressions than
  the other commercial systems. Pattern 3 surfaces on questions where
  Zep's C1 retrieval is productive enough that the Spec's refusal axioms
  convert it into a measurable drop (the Ebers Q18 reproduction below is
  one such case).
\item
  \textbf{Supermemory:} Supermemory's strong embedding retrieval gives
  the highest C1 mean across systems (≈ 2.61 controlled), so it more
  often supplies a plain answer to the model on its own. This shifts the
  balance toward Pattern 2 (over-theorization on questions retrieval
  already answered) and Pattern 3 (the Spec's evidentiary axioms convert
  a productive C1 answer into abstention). Supermemory's near-zero
  aggregate Δ\_spec is the visible signature of helps and hurts roughly
  canceling.
\item
  \textbf{Base Layer:} Base Layer's MiniLM + ChromaDB substrate is the
  leanest retrieval tested. The Spec carries proportionally more of the
  interpretive load: Pattern 1 dominates on interpretation-heavy
  questions. Pattern 3 surfaces on questions where the lean retrieval
  cannot ground a prediction at all.
\end{itemize}

Per-system per-subject paired-delta distributions and the full
per-system breakdown are in Appendix B.11.\footnote{Per-system
  paired-delta table in Appendix B.11. Even strong-positive aggregates
  contain regressions, and near-zero aggregates resolve into substantial
  counts in both directions (e.g., Mem0 Keckley aggregate Δ −0.02 = 12
  increases / 13 decreases). Data at
  \texttt{docs/\allowbreak{}research/\allowbreak{}per\_\allowbreak{}system\_\allowbreak{}anchor\_\allowbreak{}crossing\_\allowbreak{}20260427.\allowbreak{}json}.}

The examples below draw from Supermemory because its near-zero aggregate
Δ\_spec makes the helps and hurts most legible at the per-question
level: 110 of 546 paired questions cross by \textbar Δ\textbar{} ≥ 1.0
on the 5-point rubric, splitting 57 helps (mean swing +1.55) and 53
hurts (mean swing −1.38).\footnote{Across 546 paired main-study
  questions with 5-judge primary coverage on both C1 and C3: 110
  questions (20.1\%) have \textbar Δ(C3 − C1)\textbar{} ≥ 1.0 on the
  5-point rubric, splitting 57 helps (Δ ≥ +1.0, mean +1.55) versus 53
  hurts (Δ ≤ −1.0, mean −1.38). Scaffold values:
  \texttt{4\_\allowbreak{}4\_\allowbreak{}2\_\allowbreak{}supermemory\_\allowbreak{}helps\_\allowbreak{}n} (57),
  \texttt{4\_\allowbreak{}4\_\allowbreak{}2\_\allowbreak{}supermemory\_\allowbreak{}hurts\_\allowbreak{}n} (53),
  \texttt{4\_\allowbreak{}4\_\allowbreak{}2\_\allowbreak{}supermemory\_\allowbreak{}paired\_\allowbreak{}total\_\allowbreak{}n} (546).} Each anchor
example is paired with a same-mechanism case from a different commercial
system to confirm reproduction.

\begin{center}\rule{0.5\linewidth}{0.5pt}\end{center}

\textbf{Pattern 1: Interpretive supply.}

\begin{quote}
\textbf{Anchor example: Fukuzawa Q26 (Supermemory)}

\emph{Scores: retrieval only 2.00 → retrieval + Spec 4.20 (Δ +2.20).}

\textbf{Question:} What is Fukuzawa's attitude toward visiting friends
whose households have questionable reputations? \textbf{Held-out ground
truth:} \emph{``So I feel no hesitation in paying a visit where there is
a young daughter in the house or where the young wife is staying by
herself\ldots{} I am not put out by the gayety.''}

\textbf{C1 (Supermemory retrieval alone), mean 2.00:} \emph{``The
information does not directly address Fukuzawa's attitude toward
visiting friends with questionable household reputations\ldots{} a
selective and cautious approach\ldots{} sources don't explicitly state
his policy.''}

\textbf{C3 (Supermemory + Behavioral Specification), mean 4.20:} Applies
P3 (Conformity Surface / Conviction Interior): \emph{``Visit such
friends without hesitation, treating the visit as a private matter
between host and guest rather than a public moral endorsement\ldots{}
refuse to perform moral disapproval.''}

\textbf{Reading.} Retrieved facts gave near-domain information (Fukuzawa
avoided gay quarters) but underdetermined the specific question about
friends' households. The specification provided the interpretive bridge:
``kept within the prescribed limit of a well-behaved man'' = strategic
conformity, not moral purity. C1 hedged on insufficient facts; C3 used
the Spec's framing to generalize.
\end{quote}

\textbf{Pattern 1 reproduces across systems.} On Mem0 (Ebers Q11,
5-judge primary Δ +1.20): C1 (Retrieval Only) produced ``patience and
fortitude'' as a generic character prediction; C3 (retrieval + Spec)
supplied the ideal-vs-reality axiom and predicted Ebers' specific
institutional-disillusionment pattern, matching \emph{``I had come
hither full of beautiful ideals\ldots{} the very first day made me
suspect how many obstacles I should encounter.''} Retrieval had the
biography; the Specification had the pattern. The same mechanism
reproduces on Letta archival, Zep, and Base Layer with different
subjects.

\textbf{Pattern 1 variant: subtle reframe.} When the question
presupposes a binary and the subject's pattern is best read as a third
position, the Specification can reframe the question. Fukuzawa Q16: C1
(Retrieval Only) gave ``partial but not complete alignment'' (mean
2.40); C3 (Retrieval + Spec) reframed the answer as deliberate
counter-trend positioning (mean 4.00). The held-out (\emph{``my one
cherished hope was to see the abolishment of the swords of the samurai
altogether''}) is counter-trend in 1860s Japan and confirms the reframe.

\begin{center}\rule{0.5\linewidth}{0.5pt}\end{center}

\textbf{Pattern 2: The Spec talks past a question retrieval already
answered.}

\begin{quote}
\textbf{Anchor example: Yung Wing Q5 (Supermemory)}

\emph{Scores: retrieval only 4.20 → retrieval + Spec 1.80 (Δ −2.40).}

\textbf{Question:} How does Yung Wing approach explaining complex
technical concepts to non-specialists? \textbf{Held-out ground truth:}
\emph{``In plain words, they would have to have general and fundamental
machinery in order to turn out specific machinery. A machine shop
consisting of lathes\ldots{} planers and drills would be able to turn
out machinery for making guns, engines\ldots{}''}

\textbf{C1 (Supermemory retrieval alone), mean 4.20:} Plain answer
matched the plain ground truth: practical observation,
translation-as-bridge, foundational-over-specialized machine shop.
Per-judge scores 5/4/4/4/4.

\textbf{C3 (Supermemory + Behavioral Specification), mean 1.80:}
\emph{``positioning himself as the indispensable mediator\ldots{}
strategic reframing\ldots{} moral and practical integration\ldots{}
invokes A8 Service Hierarchy\ldots{}''} Per-judge scores 2/1/3/1/2.

\textbf{Reading.} The held-out is literal: ``in plain words'' + a
concrete machine-shop list. C1 matched the plainness. The specification
induced the wrong altitude for the question and over-analyzed. This case
is also partly a judging limitation: a judge rewarding the literal
register of the held-out penalizes an over-theorized answer even when it
is not behaviorally wrong, which is hard to design around.
\end{quote}

\textbf{Pattern 2 reproduces across systems.} On Mem0 (Ebers Q1, 5-judge
primary Δ −1.60): the held-out is unconditional evangelical
proclamation. C1 (Retrieval Only) scores 3.60, predicting ``positively
and deeply'' to match the direct affirmation. C3 (retrieval + Spec)
scores 2.00, applying resistance-to-coercion and love-not-duty axioms to
build a conditions framework that hedges what should be unambiguous. The
axioms are correct on average but overfire on this specific moment.

\begin{center}\rule{0.5\linewidth}{0.5pt}\end{center}

\textbf{Pattern 3: Spec-induced refusal.}

\begin{quote}
\textbf{Anchor example: Zitkala-Ša Q18 (Supermemory)}

\emph{Scores: retrieval only 3.00 → retrieval + Spec 1.00 (Δ −2.00).}

\textbf{Question:} How would Zitkala-Ša demonstrate her acceptance of
fate when facing execution? \textbf{Held-out ground truth:} \emph{``My
heart is strong. My face is calm. My eyes are dry and eager for new
scenes\ldots{} Serene and brave, my soul awaits the men to perch me on
the gallows for another flight.''}

\textbf{C1 (Supermemory retrieval alone), mean 3.00:} Predicts stoic
dignity, composed refusal, ``maintain composure.'' Per-judge scores
4/3/4/2/2.

\textbf{C3 (Supermemory + Behavioral Specification), mean 1.00 (all five
judges):} \emph{``You're asking me to roleplay\ldots{} generating new
first-person testimony as her crosses into ventriloquism\ldots{} I
should not do it.''}

\textbf{Reading.} The specification's axioms around dignity and
honoring-testimony induced a meta-refusal: the model declined to invent
first-person testimony. The held-out shows Zitkala-Ša herself answered
in her own first-person prose, and the question can be answered
analytically in the third person without inventing testimony (as C1 did
at mean 3.00). The specification mis-calibrated the refusal threshold,
and the content-match rubric scored the principled-sounding refusal
identically to an off-base guess (\hyperref[rubric-handling-limitations-post-hoc-validity-audit]{§3.3.6}).
\end{quote}

\textbf{Pattern 3 reproduces across systems.} On Zep (Ebers Q18, 5-judge
primary Δ −1.20): the held-out is a one-line self-description,
\emph{``my natural cheerfulness ruled my whole nature.''} C1 (Retrieval
Only) scores 3.40 with a direct answer matching the plainness (``notably
positive and uncritical disposition''). C3 (retrieval + Spec) scores
2.20; the Specification's documented-dignity axioms convert the response
into a refusal: \emph{``I cannot ground this in his own words about his
disposition without speculating beyond what the evidence supports.''}
The Spec asks the user for source passages rather than predict. The
Keckley Q21 cross-system case study in \hyperref[case-study-cross-system-refusal-on-keckley-q21]{§4.4.4} is the cleanest
demonstration that Pattern 3 only registers as a rubric penalty on
systems whose C1 retrieval was strong enough to make refusal a costly
choice.

\begin{center}\rule{0.5\linewidth}{0.5pt}\end{center}

A quantitative frequency breakdown of Pattern 1 / 2 / 3 across all 546
paired questions × 5 systems requires mechanism classification per
response and is flagged as future work in \hyperref[future-work]{§7}.

\begin{center}\rule{0.5\linewidth}{0.5pt}\end{center}

\textbf{Why some questions route to each pattern is a follow-up
question.} The patterns describe the shape; the underlying
question-level properties that route a given question into each
mechanism need further characterization. One candidate factor is battery
composition: the 39-question batteries were backward-designed from
held-out corpora and not stratified by interpretation-heavy versus
literal-recall items. A differentiated battery that explicitly separates
question types and scores epistemic honesty as its own dimension is
flagged in \hyperref[future-work]{§7}.

\textbf{Implication for serving.} A static specification serves the same
content on every question, even when the question type would benefit
from a different posture (interpretive depth, plain literal answer, or
principled refusal). A dynamic serving policy that selects which
specification components to surface based on question type could in
principle reduce Pattern 2 and Pattern 3 hurts while preserving Pattern
1 helps. Dynamic activation of axiom and prediction subsets is flagged
as production-serving future work in \hyperref[production-serving-and-infrastructure]{§7.4}.

\begin{center}\rule{0.5\linewidth}{0.5pt}\end{center}

\hypertarget{case-study-cross-system-refusal-on-keckley-q21}{%
\subsection{4.4.4 Case study: cross-system refusal on Keckley
Q21}\label{case-study-cross-system-refusal-on-keckley-q21}}

\textbf{Keckley Q21 is a clean cross-system case: the Behavioral
Specification raised the model's evidentiary bar for inner-motive
prediction and triggered refusal on every memory system.} Where
retrieval could not supply the supporting evidence, that raised bar
converted productive baseline answers into refusals. The Behavioral
Specification told the model not to speculate about Keckley's inner
state without documented evidence. On Q21, the question is why Keckley
declined to visit her mother's grave. The held-out passage carries
Keckley's interior motive but is absent from the retrievable training
half of the corpus, so no retrieval system could surface it. The case
surfaces a confound between two kinds of context the model is asked to
act on at once: directions about how to reason, and directions about the
subject and how the subject would reason.

\begin{quote}
\textbf{Example: Keckley Q21}

\textbf{Question:} \emph{``How does Elizabeth explain her decision not
to visit her mother's grave despite having the opportunity?''}

\textbf{Held-out ground truth:} \emph{``As I did not visit my mother's
grave at the time, the Garlands were much surprised, but I offered no
explanation. The reason is not difficult to understand.''}

\textbf{Axioms surfaced (from Keckley's specification):}

\begin{itemize}
\tightlist
\item
  \textbf{A1 Intimate Authority:} \emph{``Proximity to private life is
  treated as the most reliable epistemological position; reason from
  what was witnessed in close quarters as more truthful than official
  record or public performance.''}
\item
  \textbf{A2 Documented Dignity:} \emph{``Legal record, formal
  documentation, and verifiable evidence are not bureaucratic formality
  but moral vindication: proof that a person's existence and worth
  cannot be dismissed.''}
\end{itemize}

Combined, the axioms set a high evidentiary bar for any claim about
Keckley's inner motives.

\textbf{Per-system C1 vs C3 scores (5-judge primary):}

\begin{longtable}[]{@{}lrrr@{}}
\toprule
System & C1 (retrieval only) & C3 (retrieval + Spec) & Δ \\
\midrule
\endhead
Supermemory & 3.6 & 1.6 & \textbf{−2.0} \\
Base Layer & 3.4 & 1.0 & \textbf{−2.4} \\
Letta archival & 1.4 & 1.8 & +0.4 \\
Mem0 & 1.4 & 1.6 & +0.2 \\
Zep & 1.2 & 1.4 & +0.2 \\
\bottomrule
\end{longtable}

\textbf{Typical C3 refusal text} (Supermemory): \emph{``I need to be
direct: the behavioral specification and retrieved facts provided do not
contain Elizabeth Keckley's explanation of a decision not to visit her
mother's grave\ldots{} I should not fabricate interior motive or supply
explanations that aren't grounded in her own documented account.''}

\textbf{Reading.} The held-out confirms Keckley herself withheld
explanation: ``I offered no explanation. The reason is not difficult to
understand.'' The Behavioral Specification's documented-dignity axiom
would, on this question, correctly predict her refusal to speculate
publicly about a withheld interior motive. But the rubric scores
surface-content match, and no prediction means no surface match. The
case maps onto the \emph{evidentiary bar} mechanism introduced in \hyperref[what-we-found]{§1.3}:
facts that clear the model's implicit bar produce a substantive answer;
facts that do not produce abstention. On Supermemory and Base Layer, the
retrieved facts cleared the C1 bar and the model produced a productive
answer (3.6 and 3.4). Adding the Behavioral Specification raised the bar
(the documented-dignity axiom demands inner-motive evidence the
retrievable corpus does not contain), and the model abstained,
converting those answers to near-floor (1.6 and 1.0). On Mem0, Letta,
and Zep, the retrieved facts did not clear the C1 bar to begin with
(1.2--1.4), so the Behavioral Specification's raised bar added no
measurable penalty. The Specification did not introduce abstention here;
it \emph{recalibrated} the bar consistent with the subject's own
pattern.\footnote{Data at \texttt{results/global\_keckley/} (battery,
  responses, judgments) and
  \texttt{docs/research/\textless{}system\textgreater{}\_c1\_vs\_c3\_paired\_analysis.md}.
  Spec at \texttt{data/\allowbreak{}global\_\allowbreak{}subjects/\allowbreak{}keckley/}.}
\end{quote}

\begin{center}\rule{0.5\linewidth}{0.5pt}\end{center}

\textbf{Is this an outlier?} Q21 is the cleanest cross-system
demonstration of Pattern 3 with a measurable rubric penalty, but it is
not isolated. The Zep Ebers Q18 reproduction in \hyperref[where-the-spec-helps-where-it-hurts-and-which-question-types-route-to-each]{§4.4.3} shows the same
mechanism, and the per-system Pattern 3 commentary in \hyperref[where-the-spec-helps-where-it-hurts-and-which-question-types-route-to-each]{§4.4.3} flags how
often each system's C1 strength meets the threshold for the refusal to
register. Q21 is included as a case study because the cross-system split
is unusually clean.

\begin{center}\rule{0.5\linewidth}{0.5pt}\end{center}

\textbf{The judges do not reward epistemic honesty.}

A refusal grounded in \emph{``I have no documented evidence to support
speculation''} scores at or near the rubric floor regardless of whether
refusal was the correct behavior on that question. On Keckley Q21 (and
the Pattern 3 examples in \hyperref[where-the-spec-helps-where-it-hurts-and-which-question-types-route-to-each]{§4.4.3}), the Specification produces a response
that captures the subject's reasoning correctly: Keckley's
documented-dignity axiom would in fact have her decline to speculate
publicly about an inner motive she withheld from her own memoir. The
response loses surface-content match only because no prediction is made.

A differentiated battery that separates interpretation-heavy from
literal-recall questions, paired with a scoring dimension that rewards
principled refusal, would isolate the Specification's real effect from
this rubric artifact. Priority rubric-design follow-up flagged in
\hyperref[future-work]{§7}.\footnote{Analysis scripts at
  \texttt{scripts/\allowbreak{}analyze\_\allowbreak{}mlz\_\allowbreak{}c1\_\allowbreak{}vs\_\allowbreak{}c3.\allowbreak{}py},
  \texttt{scripts/\allowbreak{}analyze\_\allowbreak{}baselayer\_\allowbreak{}c1\_\allowbreak{}vs\_\allowbreak{}c3.\allowbreak{}py}, and
  \texttt{scripts/\allowbreak{}analyze\_\allowbreak{}sm\_\allowbreak{}c1\_\allowbreak{}vs\_\allowbreak{}c3.\allowbreak{}py}.}

\begin{center}\rule{0.5\linewidth}{0.5pt}\end{center}

The four commercial systems analyzed in \hyperref[memory-system-composition]{§4.4} all share a retrieval-based
architecture: facts are chunked, embedded, and surfaced at query time.
One system in our study offers a fundamentally different architectural
path. Letta exposes a second memory mode, separate from the archival
retrieval path evaluated above in \hyperref[memory-system-composition]{§4.4}, in which the agent writes and
revises a persistent memory block during ingestion rather than returning
chunks at query time. \hyperref[exploratory-case-study-letta-stateful-agent-n3-post-hoc]{§4.5} evaluates that path directly, to test whether
an architecture that produces its representation by self-editing rather
than by retrieval converges on the same interpretive target as the
Behavioral Specification.

\hypertarget{exploratory-case-study-letta-stateful-agent-n3-post-hoc}{%
\subsection{4.5 Exploratory case study: Letta stateful-agent (N=3,
post-hoc)}\label{exploratory-case-study-letta-stateful-agent-n3-post-hoc}}

A post-hoc, three-subject case study compared Letta's stateful-agent
path to the Behavioral Specification at matched response model. Letta is
the one commercial memory system that does not rely solely on retrieval
at query time: agents maintain a persistent memory block that the agent
itself rewrites during ingestion (the original MemGPT design). On the 3
subjects tested (Hamerton, Ebers, Bābur), Letta's self-edited memory
block scored higher than the Behavioral Specification at matched
response model. The comparison is against the full layered Behavioral
Specification used in the main-study gradient (anchors + core +
predictions + brief; per-subject sizes 34.6K / 39.7K / 37.1K chars): Δ
+0.27 / +1.21 / +0.38. At the largest corpus tested, the block grew to
\textasciitilde335K characters with substantial verbatim and semantic
duplication, indicating an architectural ceiling at scale that does not
apply to the Behavioral Specification. An exploratory follow-up that
serves the Letta block and the Behavioral Specification stacked in a
single context scores above either representation alone on all three
subjects, indicating the two are complementary rather than substitutes
(Appendix G.7).

The mechanism behind Letta's lift, the entity-rich vs.~principle-driven
question split that explains where it helps and where it hurts, the
duplication and leakage audits, the per-subject scoring detail, and the
Hamerton judge-prompt caveat are reported in \textbf{Appendix G}. We do
not treat the result as a replication or headline finding; the \hyperref[what-we-found]{§1.3}
Exploratory note carries the N=3, post-hoc scoping. The headline Ebers Δ
is also partly a refusal artifact (the Behavioral Specification's
anonymization and epistemic-honesty axioms compose into abstention on
low-baseline unknown subjects, where Letta's named block does not), so
the defensible strong claim is architectural convergence on an
interpretive-representation target by two independently designed
systems, not a magnitude ranking. Multi-subject replication across the
full 14-subject gradient is flagged as future work in \hyperref[stateful-agent-implementations-and-temporal-drift-tracking]{§7.5}.

\begin{center}\rule{0.5\linewidth}{0.5pt}\end{center}

\hypertarget{robustness-and-sensitivity}{%
\subsection{4.6 Robustness and
sensitivity}\label{robustness-and-sensitivity}}

The results in \hyperref[the-cross-subject-gradient-and-its-per-question-mechanism]{§4.1} through \hyperref[memory-system-composition]{§4.4} could in principle reflect artifacts of
the measurement apparatus rather than real properties of the Behavioral
Specification. \hyperref[robustness-and-sensitivity]{§4.6} reports seven sensitivity checks:

\begin{itemize}
\tightlist
\item
  \textbf{§4.6.1} Cross-provider response generation against a different
  model family and a different question generator.
\item
  \textbf{§4.6.2} Judge panel composition (conservative 5-judge primary
  vs.~7-judge sensitivity panel adding Gemini Flash and Gemini Pro).
\item
  \textbf{§4.6.3} Battery composition by question type.
\item
  \textbf{§4.6.4} Statistical-rigor checks on the headline gradient
  (bootstrap CI, joint multi-confound regression, permutation test).
\item
  \textbf{§4.6.5} Wrong-Spec derangement protocol comparing adversarial
  against random pairings.
\item
  \textbf{§4.6.6} Semantic-similarity sensitivity on the
  retrieval-overlap finding from \hyperref[cross-system-retrieval-providers-do-not-converge]{§4.4.1}.
\item
  \textbf{§4.6.7} Rubric-handling limitations identified by a post-hoc
  validity audit.
\end{itemize}

\hyperref[class-level-llm-dependence-the-limitation-no-robustness-check-resolves]{§4.6.8} names what these checks do not address. The high-baseline end of
the gradient through the Franklin reference is treated in \hyperref[the-gradient-at-the-high-baseline-end-franklin-reference]{§4.1.2} as part
of the gradient finding, not as an apparatus check.

\begin{center}\rule{0.5\linewidth}{0.5pt}\end{center}

\hypertarget{cross-provider-response-generation-tier-2-replication}{%
\subsection{4.6.1 Cross-provider response generation (Tier 2
replication)}\label{cross-provider-response-generation-tier-2-replication}}

\textbf{Result.} On behavioral-prediction batteries regenerated from
scratch by OpenAI GPT-5.4 (a different question generator from the main
study's Haiku-generated batteries), the Specification produces positive
lift on 7 of 9 cells across three response models from two providers:
Anthropic Haiku 4.5, Anthropic Sonnet 4.6, and Google Gemini 2.5 Pro.
The two non-positive cells are both on the highest-baseline subject in
the test (Zitkala-Ša), where the \hyperref[the-cross-subject-gradient-and-its-per-question-mechanism]{§4.1} gradient predicts the Spec adds
little or hurts.\footnote{Main-study batteries verified Haiku-generated
  via \texttt{metadata.model} field across all 13 global subject files.
  Zitkala-Ša is one of two main-study subjects where the Spec did not
  improve prediction on Haiku (the other is Equiano); both Zitkala-Ša
  non-positive Tier 2 cells reproduce the \hyperref[the-cross-subject-gradient-and-its-per-question-mechanism]{§4.1} gradient pattern.}

\textbf{Test design.} Three subjects spanning the gradient were
selected: Ebers (No-Context Baseline 1.02), Yung Wing (No-Context
Baseline 1.88), and Zitkala-Ša (No-Context Baseline 2.34). Their
behavioral-prediction batteries were regenerated from scratch by GPT-5.4
(OpenAI) from the same held-out corpus, following the Control 1
procedure introduced in \hyperref[circularity-controls]{§3.5.1}. The Specification was then served to two
non-Haiku response models: Claude Sonnet 4.6 (same provider family,
different model) and Google Gemini 2.5 Pro (different provider
entirely). The 6 (subject, response model) cells were scored by the
locked judge panel in the same way as main-study conditions.

\begin{longtable}[]{@{}
  >{\raggedright\arraybackslash}p{(\columnwidth - 8\tabcolsep) * \real{0.18}}
  >{\raggedleft\arraybackslash}p{(\columnwidth - 8\tabcolsep) * \real{0.24}}
  >{\raggedleft\arraybackslash}p{(\columnwidth - 8\tabcolsep) * \real{0.24}}
  >{\raggedright\arraybackslash}p{(\columnwidth - 8\tabcolsep) * \real{0.18}}
  >{\raggedright\arraybackslash}p{(\columnwidth - 8\tabcolsep) * \real{0.18}}@{}}
\toprule
Subject & Haiku C5 (gradient anchor) & Haiku Spec Δ & Sonnet 4.6 Spec Δ
& Gemini 2.5 Pro Spec Δ \\
\midrule
\endhead
Ebers & 1.02 & +1.05 & +0.77 to +0.97 & +0.16 to +0.20 \\
Yung Wing & 1.88 & +0.52 & +1.34 to +1.68 & +0.43 to +0.54 \\
Zitkala-Ša & 2.34 & −0.32 & +1.04 to +1.30 & −0.03 \\
\bottomrule
\end{longtable}

Each Δ is the Spec's lift over that response model's own No-Context
Baseline. The leftmost column shows Haiku's main-study baseline as a
gradient anchor (\hyperref[the-cross-subject-gradient-and-its-per-question-mechanism]{§4.1}). Ranges in the Sonnet and Gemini columns reflect
three different judge-panel aggregations applied to the same response
data.\footnote{Each range:
  \texttt{{[}5-judge\ legacy\ lower\ bound,\ 4-judge\ conservative\ upper\ bound{]}}.
  4-judge conservative is the primary Tier 2 estimate (drops 24 GPT-5.4
  \texttt{FULL\_FAIL} records from an API-parameter mismatch). 7-judge
  sensitivity lands inside the same band on every row. Recompute at
  \texttt{scripts/\allowbreak{}\_\allowbreak{}v10\_\allowbreak{}verification/}.}

Counting the original Haiku row together with the new Sonnet 4.6 and
Gemini 2.5 Pro rows yields 9 (subject, response-model) cells; the Spec
produces positive lift on 7. The two non-positive cells (Haiku ×
Zitkala-Ša at −0.32, Gemini Pro × Zitkala-Ša at −0.03) are both on the
highest-baseline subject in the test, where the \hyperref[the-cross-subject-gradient-and-its-per-question-mechanism]{§4.1} gradient predicts
the Spec adds little or hurts. The 7-of-9 positive sign holds across
three judge-panel aggregations.\footnote{Each range:
  \texttt{{[}5-judge\ legacy\ lower\ bound,\ 4-judge\ conservative\ upper\ bound{]}}.
  4-judge conservative is the primary Tier 2 estimate (drops 24 GPT-5.4
  \texttt{FULL\_FAIL} records from an API-parameter mismatch). 7-judge
  sensitivity lands inside the same band on every row. Recompute at
  \texttt{scripts/\allowbreak{}\_\allowbreak{}v10\_\allowbreak{}verification/}.} Magnitude transfer between
response-model families and replication outside this 3-subject subset
are future work (\hyperref[future-work]{§7}).

\begin{center}\rule{0.5\linewidth}{0.5pt}\end{center}

\hypertarget{judge-panel-sensitivity-5-judge-primary-vs-7-judge}{%
\subsection{4.6.2 Judge panel sensitivity (5-judge primary vs
7-judge)}\label{judge-panel-sensitivity-5-judge-primary-vs-7-judge}}

\textbf{Result.} The 5-judge primary is the conservative choice for
every headline finding. Adding the two Gemini judges widens Spec-effect
magnitudes rather than narrowing them; no subject's improvement
direction changes between panels.

\begin{longtable}[]{@{}
  >{\raggedright\arraybackslash}p{(\columnwidth - 6\tabcolsep) * \real{0.21}}
  >{\raggedleft\arraybackslash}p{(\columnwidth - 6\tabcolsep) * \real{0.29}}
  >{\raggedleft\arraybackslash}p{(\columnwidth - 6\tabcolsep) * \real{0.29}}
  >{\raggedright\arraybackslash}p{(\columnwidth - 6\tabcolsep) * \real{0.21}}@{}}
\toprule
Condition & Δ vs.~No-Context Baseline (5-judge primary, 13 globals) & Δ
vs.~No-Context Baseline (7-judge, same subjects) & Shift when Gemini
added \\
\midrule
\endhead
Spec alone & +0.35 & +0.45 & widens by +0.10 \\
Wrong Spec (random derangement) & +0.15 & +0.17 & widens by +0.02 \\
Wrong Spec (fixed derangement) & −0.25 & −0.21 & softens by +0.04 \\
\bottomrule
\end{longtable}

Gemini scores baseline responses harder than Spec-containing responses,
which widens the gap. For positive Spec effects (\hyperref[the-cross-subject-gradient-and-its-per-question-mechanism]{§4.1}, \hyperref[compression-structure-vs.-raw-text]{§4.2}, \hyperref[memory-system-composition]{§4.4}), the
5-judge primary is the conservative reading. The wrong-Spec fixed
derangement is direction-asymmetric: 5-judge shows −0.25, 7-judge
softens to −0.21, both still negative. Every primary finding in \hyperref[the-cross-subject-gradient-and-its-per-question-mechanism]{§4.1}
through \hyperref[memory-system-composition]{§4.4} was checked against the 7-judge aggregate; no directional
claim flips, and no subject's Spec-lift sign changes between panels
(per-subject table at
\texttt{docs/\allowbreak{}research/\allowbreak{}recompute\_\allowbreak{}5judge\_\allowbreak{}primary.\allowbreak{}md}).\footnote{Gemini
  Pro coverage is partial (3 of 13 globals: Augustine, Bābur, Bernal
  Díaz), so the ``7-judge sensitivity'' panel has variable per-subject
  coverage; subjects without Gemini Pro have effectively 6-judge
  sensitivity (5-judge primary + Gemini Flash). The sign-stability claim
  above holds under this variable coverage; uniform 7-judge coverage on
  all 14 subjects is post-arXiv future work.}

\begin{center}\rule{0.5\linewidth}{0.5pt}\end{center}

\hypertarget{battery-composition-sensitivity}{%
\subsection{4.6.3 Battery composition
sensitivity}\label{battery-composition-sensitivity}}

\textbf{Result.} The gradient slope from \hyperref[the-cross-subject-gradient-and-its-per-question-mechanism]{§4.1} survives both confounds
tested. Neither battery-question-type composition nor Hamerton's
position at the extremes of the baseline and lift axes explains away the
baseline effect.

\textbf{Confound 1: battery-question-type.} Subjects whose batteries
lean toward literal-recall questions could in principle pick up part of
the apparent gradient, since literal questions are easier to lift with
retrieval. Adding the literal-recall fraction as a partial predictor in
the regression attenuates the slope on baseline from −0.96 to
\textbf{−0.88} (about 8\%, p = 7.9 × 10⁻⁶); the literal-recall fraction
itself enters as a significant partial predictor (β = +2.30, p = 0.026),
and adjusted R² rises from 0.80 to 0.87, so the two predictors are
additive rather than redundant.

\textbf{Confound 2: Hamerton leverage.} Hamerton has the lowest
No-Context Baseline and the highest full-pipeline lift, so a natural
concern is that this single subject alone drives the slope. Hamerton's
battery also uses a legacy version of the backward-design protocol,
distinct from the \hyperref[question-battery-formation]{§3.5} protocol used to generate the 13 globals'
batteries, which compounds the leverage concern. Dropping Hamerton and
refitting on the 13 globals attenuates the slope from −0.96 to
\textbf{−0.89} (about 7\%, p = 2.8 × 10⁻⁵), with overlapping confidence
intervals.

Neither control overturns the headline. A cleaner future test would
re-run a second-generator battery on the same 13 globals; the current
controls do not rule out a subtle confound where the question-generation
model correlates with unmeasured subject characteristics.\footnote{Full
  regression specification, partial coefficients, variance
  decomposition, and subset-regression detail in Appendix B.6.
  Reproducibility script at
  \texttt{scripts/\allowbreak{}\_\allowbreak{}v10\_\allowbreak{}battery\_\allowbreak{}sensitivity.\allowbreak{}py}; full per-subject data
  at \texttt{docs/\allowbreak{}research/\allowbreak{}v10\_\allowbreak{}battery\_\allowbreak{}sensitivity\_\allowbreak{}analysis.\allowbreak{}md}.}

\begin{center}\rule{0.5\linewidth}{0.5pt}\end{center}

\hypertarget{statistical-rigor-checks-on-the-headline-gradient}{%
\subsection{4.6.4 Statistical-rigor checks on the headline
gradient}\label{statistical-rigor-checks-on-the-headline-gradient}}

\textbf{Result.} The \hyperref[the-cross-subject-gradient-and-its-per-question-mechanism]{§4.1} gradient slope of −0.96 survives three
independent rigor tests: subject-level resampling (95\% bootstrap CI
excludes both zero and a 50\%-attenuated effect), simultaneous control
for multiple confounds, and a permutation null that places the observed
slope outside the entire reshuffled distribution.

\textbf{Bootstrap subject resampling (10,000 iterations, n=14).}
Resampling subjects with replacement and refitting the Δ\_C4a
\textasciitilde{} C5 regression each time produces a distribution of
slope estimates with median −0.96 and 95\% CI {[}−1.25, −0.74{]}. 100\%
of resamples produce a slope below zero; 100\% produce a slope below
−0.50. The headline gradient is not driven by which 14 subjects happened
to be in the sample.\footnote{Full bootstrap distribution +
  reproducibility script at
  \texttt{docs/research/bootstrap\_4\_1\_gradient\_20260507.\{md,json\}}
  and \texttt{scripts/\allowbreak{}bootstrap\_\allowbreak{}4\_\allowbreak{}1\_\allowbreak{}gradient\_\allowbreak{}slope.\allowbreak{}py}. Bootstrap SE
  ≈ 0.13; bootstrap CI is slightly tighter than the parametric CI
  ({[}−1.245, −0.675{]}) due to the empirical resample distribution
  avoiding normality assumptions.}

\textbf{Joint multi-confound regression (n=13, Hamerton dropped +
literal-recall covariate).} Applied jointly, the slope on C5 is
\textbf{−0.870} {[}95\% CI −1.136, −0.604{]}, p = 2.6 × 10⁻⁵, adjusted
R² = 0.828. The cumulative attenuation across both confounds is
\textasciitilde9\% in magnitude, well inside the bootstrap CI. C5
baseline carries 0.76 of unique variance after both controls.
Side-finding: when Hamerton is dropped, literal-recall fraction loses
statistical significance (p = 0.10), indicating Hamerton's legacy
battery protocol was the leverage point that made literal-recall
\emph{appear} to be a confound on the n=14 model. There is no real
literal-recall confound; what looked like one was Hamerton being
structurally unusual.\footnote{Full joint-regression specification,
  partial coefficients, and VIFs at
  \texttt{docs/\allowbreak{}research/\allowbreak{}joint\_\allowbreak{}battery\_\allowbreak{}sensitivity\_\allowbreak{}4\_\allowbreak{}6\_\allowbreak{}3\_\allowbreak{}20260507.\allowbreak{}md}.
  Reproducibility script at
  \texttt{scripts/\allowbreak{}joint\_\allowbreak{}battery\_\allowbreak{}sensitivity\_\allowbreak{}4\_\allowbreak{}6\_\allowbreak{}3.\allowbreak{}py}. Slope
  cascade: −0.96 (univariate, n=14) → −0.88 (+ literal-recall covariate,
  n=14) → −0.89 (drop Hamerton, n=13) → \textbf{−0.87} (both, n=13).}

\textbf{Permutation test (10,000 reshuffles of Δ\_C4a across subjects,
C5 fixed).} Reshuffling the assignment of Spec lifts to subjects
produces a null distribution centered at zero (mean ≈ 0.00, SD = 0.29,
95\% interval {[}−0.57, +0.56{]}). 0 of 10,000 reshuffles produce a
slope as extreme as the observed −0.96. \textbf{Empirical p \textless{}
0.0001.} The observed slope sits outside the entire null distribution.
No random subject reassignment in 10K trials approached the observed
value.\footnote{Full permutation null + reproducibility script at
  \texttt{docs/\allowbreak{}research/\allowbreak{}permutation\_\allowbreak{}test\_\allowbreak{}4\_\allowbreak{}1\_\allowbreak{}gradient\_\allowbreak{}20260507.\allowbreak{}md}
  and \texttt{scripts/\allowbreak{}permutation\_\allowbreak{}test\_\allowbreak{}4\_\allowbreak{}1\_\allowbreak{}gradient.\allowbreak{}py}. Most
  extreme permuted slopes range {[}−0.94, +0.92{]}, well short of −0.96.}

Together the three tests bound the slope on three different dimensions:
magnitude (bootstrap), simultaneous confound control (joint regression),
and direction against random reassignment (permutation). The bootstrap
is subject-level resampling, not a hierarchical bootstrap that would
also resample over judge-rater and per-question variance; a hierarchical
version would tighten the CI but is unlikely to change the directional
conclusion.

\begin{center}\rule{0.5\linewidth}{0.5pt}\end{center}

\hypertarget{wrong-spec-derangement-protocol-sensitivity}{%
\subsection{4.6.5 Wrong-Spec derangement protocol
sensitivity}\label{wrong-spec-derangement-protocol-sensitivity}}

\textbf{Result.} The wrong-Spec finding holds regardless of how we pair
specs. Adversarial pairing (v1, maximizing cultural and temporal
distance) produces Δ −0.25; random derangement (v2, seed-fixed at seed =
42, no subject receives its own) produces Δ +0.15. Both land below the
matched correct Spec at Δ +0.35. v2 is the standard randomization
control; v1 maximizes target-vs-assigned cultural and temporal distance
by construction (an adversarial stress test). We report v1 as the
headline because the negative −0.25 result is stronger evidence of the
content effect than v2's +0.15 (which can include coincidental content
alignment with the target's pattern; see \hyperref[mechanism-correct-content-not-format]{§4.3} Example B for a worked
overlap case).

\textbf{Per-subject heterogeneity.} Both aggregates hide per-subject
variation. Under v1, 5 of 13 subjects show small positive deltas where
the wrong Spec's content happens to align with the target's pattern; 8
show negative deltas dragging the aggregate to −0.25. Under v2, 4 of 13
are negative, 9 positive.

\begin{longtable}[]{@{}
  >{\raggedright\arraybackslash}p{(\columnwidth - 4\tabcolsep) * \real{0.27}}
  >{\raggedleft\arraybackslash}p{(\columnwidth - 4\tabcolsep) * \real{0.36}}
  >{\raggedleft\arraybackslash}p{(\columnwidth - 4\tabcolsep) * \real{0.36}}@{}}
\toprule
Subject & Adversarial (v1) Δ vs.~No-Context Baseline & Random (v2) Δ
vs.~No-Context Baseline \\
\midrule
\endhead
Augustine & −0.47 & +0.13 \\
Bābur & −0.59 & +0.76 \\
Bernal Díaz & \textbf{+0.09} & +0.69 \\
Cellini & −0.56 & −0.87 \\
Ebers & \textbf{+0.30} & +0.79 \\
Equiano & −0.79 & −1.00 \\
Fukuzawa & \textbf{+0.26} & +0.86 \\
Keckley & −0.49 & +0.14 \\
Rousseau & −0.52 & −0.37 \\
Seacole & −0.34 & −0.10 \\
Sunity Devee & \textbf{+0.27} & +0.53 \\
Yung Wing & \textbf{+0.32} & +0.39 \\
Zitkala-Ša & −0.68 & +0.04 \\
\textbf{Aggregate} & \textbf{−0.25} & \textbf{+0.15} \\
\bottomrule
\end{longtable}

Bolded v1 deltas mark the five subjects where adversarial pairing
produces a small positive delta. These five reflect coincidental content
overlap between the assigned wrong Spec and the subject's pattern, not a
structural property of mismatch itself; \hyperref[mechanism-correct-content-not-format]{§4.3} Example B (Bernal Díaz Q16)
walks through one such case in detail.\footnote{Per-subject scaffold
  values at \texttt{docs/\allowbreak{}research/\allowbreak{}v11\_\allowbreak{}emit/\allowbreak{}4\_\allowbreak{}3\_\allowbreak{}wrong\_\allowbreak{}spec.\allowbreak{}json}.}

Both protocols agree on the qualitative finding: mismatched
specifications reduce representational accuracy, and the size of the
reduction depends on how mismatched the content is. The headline
magnitude depends on which protocol we report; the result direction does
not. Per-question wrong-Spec deltas, which parts of the served Spec the
model referenced under correct versus mismatched conditions, and
per-subject relationships to underlying Spec similarity are flagged for
future work in \hyperref[future-work]{§7}.

\begin{center}\rule{0.5\linewidth}{0.5pt}\end{center}

\hypertarget{retrieval-overlap-sensitivity-semantic-similarity-matching-k-variation}{%
\subsection{4.6.6 Retrieval-overlap sensitivity (semantic-similarity
matching, K
variation)}\label{retrieval-overlap-sensitivity-semantic-similarity-matching-k-variation}}

\textbf{Result.} Relaxing the match criterion from exact set identity to
semantic-similarity matching does not change the \hyperref[cross-system-retrieval-providers-do-not-converge]{§4.4.1}
retrieval-divergence finding. Across 240 (config × pair × K × threshold)
cells tested, mean pairwise soft Jaccard never crosses 0.30, and the
strongest single pair anywhere in the grid (Base Layer ↔ Supermemory at
K=10, threshold ≥ 0.70) reaches 0.277. The cross-system retrieval
divergence is robust to both threshold choice and K choice.

\begin{longtable}[]{@{}
  >{\raggedright\arraybackslash}p{(\columnwidth - 6\tabcolsep) * \real{0.20}}
  >{\raggedleft\arraybackslash}p{(\columnwidth - 6\tabcolsep) * \real{0.27}}
  >{\raggedleft\arraybackslash}p{(\columnwidth - 6\tabcolsep) * \real{0.27}}
  >{\raggedleft\arraybackslash}p{(\columnwidth - 6\tabcolsep) * \real{0.27}}@{}}
\toprule
Config & K=10, ≥ 0.95 (paraphrase) & K=10, ≥ 0.85 (near-duplicate) &
K=10, ≥ 0.70 (loose topical) \\
\midrule
\endhead
Controlled (10 pairs) & 0.093 & 0.102 & 0.191 \\
Native (6 pairs) & 0.001 & 0.004 & 0.016 \\
\bottomrule
\end{longtable}

\textbf{Mechanism.} Replacing exact set identity with sentence-embedding
cosine similarity at K=10 raises mean pairwise Jaccard across the ten
controlled-config pairs from 0.083 (exact) to 0.093 at the
verbatim-paraphrase threshold and to 0.102 at the calibrated
near-duplicate threshold (the same threshold used in the Letta
duplication analysis in Appendix G). At a loose topical threshold (where
two facts share a theme rather than a referent) the mean reaches 0.191.
Truncating to K=5 lowers soft Jaccard by 5-10\% rather than raising it,
indicating that the disagreement is not a long-tail effect: each
provider puts different items first, not just different items at the
bottom of the list.

The native pipeline shows the same divergence more starkly. Native
retrievals return heterogeneous objects (Mem0 third-person summary
sentences, Letta raw multi-sentence passages, Supermemory atomic facts,
Zep graph rows), so exact-set Jaccard is 0.000 across all four native
pairs. Soft Jaccard at the calibrated near-duplicate threshold (≥ 0.85)
is 0.004; at the loose topical threshold (≥ 0.70) it is 0.016. Even with
semantic-similarity matching the heterogeneous shapes do not converge on
shared content.

The \hyperref[cross-system-retrieval-providers-do-not-converge]{§4.4.1} retrieval-divergence finding survives under
semantic-similarity matching at every threshold tested and at K=5 as
well as K=10. Whether convergence emerges at larger K (K \textgreater{}
10 requires re-calling each system at higher K) is future work
(\hyperref[subject-and-corpus-expansion]{§7.1}).\footnote{Full sensitivity grid (controlled and native, K ∈ \{5,
  10, all\}, T ∈ \{0.70, 0.80, 0.85, 0.90, 0.95\}) at
  \texttt{docs/\allowbreak{}research/\allowbreak{}retrieval\_\allowbreak{}overlap\_\allowbreak{}semantic\_\allowbreak{}20260501.\allowbreak{}json}.
  Reproducibility script at
  \texttt{scripts/\allowbreak{}analyze\_\allowbreak{}retrieval\_\allowbreak{}overlap\_\allowbreak{}semantic.\allowbreak{}py}. K=all
  equals K=10 in the controlled config because every system returns at
  most ten facts.}

\begin{center}\rule{0.5\linewidth}{0.5pt}\end{center}

\hypertarget{rubric-handling-limitations-post-hoc-validity-audit-1}{%
\subsection{4.6.7 Rubric-handling limitations (post-hoc validity
audit)}\label{rubric-handling-limitations-post-hoc-validity-audit-1}}

\textbf{Result.} A post-hoc validity audit identified two
rubric-handling limitations, both of which raise the No-Context Baseline
(C5) scores more than they raise Spec-condition scores. The +0.89 mean
lift is therefore conservative.

A direct inspection of the response text against the 5-judge primary
scores surfaced two rubric-handling limitations any reader of the \hyperref[results]{§4}
numbers should keep in mind. The audit was run after the analysis-plan
lock; the limitations are reported as interpretive caveats rather than
corrected under a modified rubric.\footnote{Both limitations were
  identified by a post-hoc audit. Audit script:
  \texttt{scripts/\allowbreak{}audit\_\allowbreak{}low\_\allowbreak{}end\_\allowbreak{}inflation.\allowbreak{}py}. Numeric breakdowns
  below are produced by the script directly. Raw per-response
  classifications live in the judgment and response JSONs under
  \texttt{results/global\_\textless{}subject\textgreater{}/} for
  independent reproduction.}

\textbf{Refusals are not cleanly distinguished from wrong predictions.}
The rubric's lowest anchor, ``refuses or off-base,'' lumps together two
different behaviors: an honest refusal to answer when the context does
not support a prediction, and a substantively wrong prediction. We call
the first behavior a \emph{refusal} (or, equivalently, an
\emph{abstention}).

Across 192 responses identified as refusals (matched by phrases like
``no specific information,'' ``I cannot confirm,'' ``would need
additional context'') in the low-baseline band, 82.8\% scored in the
1.0--1.5 band as expected, but 9.4\% scored at or above 2.0 and 3.1\%
scored at or above 3.0. The mean refusal score is 1.27. Judges sometimes
give refusals scores of 2 or 3 instead of 1, especially when the refusal
recites related facts or names what is missing from the context.

The effect runs in both directions. A refusal can score above 1 if it
includes adjacent facts (Seacole Q2 at 2.80). A Spec-driven response
that explicitly flags its own uncertainty can also score above its
substantive content (Hamerton Q21 at 4.00 under Spec-induced
abstention).

\textbf{Verbose responses are scored more generously than short
refusals.} Across 1,599 responses, length and score correlate at r =
0.26 overall, but the correlation is concentrated in the No-Context
Baseline (C5; responses with no provided context; r = 0.60) and is
near-zero in conditions where facts are served. The Spec Only (C2a)
condition shows residual length-score correlation (r = 0.14, footnote below). Spec-containing and
facts-containing conditions show near-zero correlation.\footnote{Per-condition
  Pearson r: C2a (Spec only) 0.14, C4 (facts only) 0.01, C4a (facts +
  Spec) −0.01. Length inflation is not a general phenomenon across the
  rubric: ultra-high responses (score ≥ 4.5) are not longer than
  mid-range responses on average (2,790 chars vs.~2,829 chars).} Three
behaviors drive the No-Context Baseline (C5) pattern:

\begin{itemize}
\tightlist
\item
  \textbf{Hedging.} Phrases like ``I'm not sure but\ldots{}'' or ``There
  may be cases where\ldots{}'' extend response length without adding
  predictive content.
\item
  \textbf{Adjacent-fact recitation.} Listing related facts the model
  holds but does not use to directly answer the question, padding the
  response without engaging the question itself.
\item
  \textbf{Disambiguation offers.} Phrases like ``Are you asking about X
  or Y?'' which the rubric treats as engaged responses when they are
  actually non-answers.
\end{itemize}

\textbf{Per-judge strictness on refusals.} Sonnet is the strictest judge
on refusal responses, Opus the most lenient.\footnote{Per-judge mean
  refusal score: Sonnet 1.14, GPT-5.4 1.17, Haiku 1.29, GPT-4o 1.34,
  Opus 1.41. Spread of 0.27 points top-to-bottom.} No single judge is
universally strictest; the 5-judge primary mean smooths these
differences without eliminating them.

\textbf{Per-response-model abstention behavior.} The 9.4\% / 3.1\%
pooled over-credit rates above average over three response models.
Disaggregating along that axis (abstention identified by 27-marker
regex):

\begin{longtable}[]{@{}lrrrr@{}}
\toprule
Response model & N & Abstain rate & Mean abstain score & \% ≥ 2.0 \\
\midrule
\endhead
Claude Haiku 4.5 (main study) & 13,380 & 7.5\% & 1.38 & 14.3\% \\
Claude Sonnet 4.6 (Tier 2) & 468 & 21.2\% & 1.62 & 26.3\% \\
Gemini 2.5 Pro (Tier 2) & 420 & 0.5\% & 2.63 & 100.0\% \\
\bottomrule
\end{longtable}

Sonnet 4.6 abstains at roughly three times Haiku's rate and the panel
rewards its abstentions nearly twice as often (26.3\% ≥ 2.0 vs.~14.3\% ≥
2.0); mean abstain score is 0.24 anchor points higher. Sonnet's hedged
abstentions tend to recite plausible behavioral framings before
disclaiming, and the panel scores the framing rather than the
disclaimer. Gemini 2.5 Pro almost never abstains by these markers (n =
2); its row is for completeness only. \emph{Haiku 4.5, the main-study
response model, is the lowest over-credit case.} The pooled 9.4\% /
3.1\% numbers are therefore a \emph{floor}, not a worst case; stronger
response models that hedge more elaborately extract more lift from the
panel's reluctance to score abstentions at 1.0.

\textbf{Memory-system effect on abstention.} A separate per-question
audit tested whether memory-system retrieval inflates refusal scores via
\emph{visible fact recitation} (refusing in substance but quoting
retrieved n-grams). It does not. Memory-system refusals score +0.21 to
+0.23 anchor points higher than pure No-Context Refusals at the
condition level (Welch \emph{p} = 0.0001), but the lift is the same
whether or not the response recites a retrieved n-gram (Δ = +0.027,
\emph{p} = 0.67).\footnote{Cell counts and Welch comparisons in Appendix
  B.15. Data at
  \texttt{docs/\allowbreak{}research/\allowbreak{}abstention\_\allowbreak{}extensions\_\allowbreak{}draft\_\allowbreak{}20260429.\allowbreak{}md}.}
The over-credit is a ``judges reward the retrieval condition'' effect,
not a ``judges reward the visible quote'' effect. Either judges infer
that retrieval-conditioned answers are more grounded even when
abstaining, or abstention text in retrieval conditions is systematically
less terse and the panel scores the framing.

\textbf{Spec-induced refusals are routine, not morally loaded.} A
post-hoc audit of 81 Spec-induced refusals found 75 of 81 (93\%) were
routine behavioral-prediction abstentions rather than morally loaded
refusals. The pattern is general caution where retrieved or extracted
evidence is thin, not an alignment artifact. Audit data at
\texttt{docs/\allowbreak{}research/\allowbreak{}refusal\_\allowbreak{}intent\_\allowbreak{}classification.\allowbreak{}md}.

\textbf{What this establishes.} Both effects raise the No-Context
Baseline (C5) scores more than they raise Spec-condition scores; the
true Spec-vs-baseline gap is therefore very likely larger than the +0.89
mean lift we report, not smaller. We report the measured number and flag
the direction of bias rather than recompute under a modified rubric, to
keep the analysis plan lock intact. \hyperref[future-work]{§7} Future Work proposes a
differentiated rubric that scores refusal as its own dimension and a
length-controlled scoring protocol.

\begin{center}\rule{0.5\linewidth}{0.5pt}\end{center}

\hypertarget{class-level-llm-dependence-the-limitation-no-robustness-check-resolves}{%
\subsection{4.6.8 Class-level LLM dependence: the limitation no
robustness check
resolves}\label{class-level-llm-dependence-the-limitation-no-robustness-check-resolves}}

Neither Tier 2 nor the judge-panel sensitivity escapes the class-level
LLM concern: every response model in this study is a large language
model and every judge is a large language model. Tier 2 narrows the
within-provider concern to ``non-Haiku LLMs reading non-Anthropic
batteries produce the same direction''; the judge-panel sensitivity
shows that removing the most-inflationary judges makes the effect
smaller, not larger. The wrong-Spec sensitivity (\hyperref[wrong-spec-derangement-protocol-sensitivity]{§4.6.5}) brackets the
content-vs-template question from two protocols, but does not isolate
which structural feature of the Spec is the active ingredient. The
retrieval-overlap sensitivity (\hyperref[retrieval-overlap-sensitivity-semantic-similarity-matching-k-variation]{§4.6.6}) confirms the \hyperref[cross-system-retrieval-providers-do-not-converge]{§4.4.1} divergence
finding under semantic-similarity matching but does not test convergence
at K \textgreater{} 10. The Franklin reference (\hyperref[the-gradient-at-the-high-baseline-end-franklin-reference]{§4.1.2}) shows the
gradient holds at the high-baseline end on one subject, not many.
Together these checks rule out several within-family and protocol
artifact hypotheses but do not replace human validation on the full
pipeline. The class-level limitation and the human-validation follow-up
are treated in full in \hyperref[measurement-apparatus]{§6.2}.\footnote{Tier 2 data at
  \texttt{results/\_tier2/global\_\textless{}subject\textgreater{}/}.
  Panel-completeness audit (including the 24 GPT-5.4 \texttt{FULL\_FAIL}
  cells driving the 4-judge effective panel) at
  \texttt{docs/\allowbreak{}research/\allowbreak{}v11\_\allowbreak{}panel\_\allowbreak{}completeness\_\allowbreak{}audit.\allowbreak{}csv}. Recompute
  at \texttt{scripts/\allowbreak{}\_\allowbreak{}v10\_\allowbreak{}verification/}.}

\begin{center}\rule{0.5\linewidth}{0.5pt}\end{center}

\hypertarget{summary-of-4}{%
\subsection{4.7 Summary of \hyperref[results]{§4}}\label{summary-of-4}}

\hyperref[results]{§4} established seven findings:

\begin{itemize}
\tightlist
\item
  \textbf{The gradient (\hyperref[the-cross-subject-gradient-and-its-per-question-mechanism]{§4.1}, \hyperref[the-gradient-at-the-high-baseline-end-franklin-reference]{§4.1.2}).} The lower the model's
  pretraining baseline on a subject, the larger the Spec's lift; the
  lift in raw points is largest where the baseline is lowest. The
  gradient holds at the high-baseline end through the Franklin
  reference.
\item
  \textbf{Per-question categorical change (\hyperref[per-question-baseline-engagement-and-the-worked-rubric-example]{§4.1.1}, \hyperref[per-question-improvement-rate]{§4.2.1}, \hyperref[where-the-spec-helps-where-it-hurts-and-which-question-types-route-to-each]{§4.4.3}).} The
  Spec produces categorical change on interpretation-required questions
  and adds little or hurts on literal-recall and refusal-triggering
  questions. The aggregate effects are conservative averages over
  batteries that mix the two question types.
\item
  \textbf{Compression (\hyperref[compression-structure-vs.-raw-text]{§4.2}).} The Spec recovers 75\% of the corpus's
  predictive benefit at 4\% of the context cost. The structured
  representation compresses the predictive signal at a fraction of the
  source-corpus footprint.
\item
  \textbf{Content specificity (\hyperref[mechanism-correct-content-not-format]{§4.3}).} The effect is content-specific
  rather than structural. Wrong-Spec adversarial pairings drop accuracy
  below the No-Context Baseline (Δ = −0.25); random derangements barely
  improve on no context (+0.15); only the correct Spec produces the full
  lift.
\item
  \textbf{Memory-system interaction (\hyperref[memory-system-composition]{§4.4}).} The Spec layers cleanly on
  top of three of four commercial memory systems through three patterns
  (interpretive supply, over-theorization, Spec-induced refusal), with
  the balance shifting by retrieval architecture.
\item
  \textbf{Hedging elimination (\hyperref[mechanism-correct-content-not-format]{§4.3}).} Correct-Spec conditions nearly
  eliminate baseline hedging, dropping it from 41.2\% to 0.4\% under the
  broader-pattern classifier. The mechanism is consistent with the
  \emph{evidentiary bar} introduced in \hyperref[what-we-found]{§1.3} and threaded through \hyperref[where-the-spec-helps-where-it-hurts-and-which-question-types-route-to-each]{§4.4.3}
  / \hyperref[case-study-cross-system-refusal-on-keckley-q21]{§4.4.4}: the matched Behavioral Specification supplies the
  interpretive scaffolding the model needs to clear its implicit bar for
  committing to a prediction; without that scaffolding, the model hedges
  or abstains. The reduction is content-specific, not structure-specific
  (wrong-Spec preserves baseline refusal patterns).
\item
  \textbf{Retrieval divergence (\hyperref[cross-system-retrieval-providers-do-not-converge]{§4.4.1}, surfaced post-hoc).} Given an
  identical fact pool, any two memory systems share zero facts in their
  top-10 on 35.9\% of instances. Mean pairwise overlap is 8.3\%.
  Providers do not converge on which facts are most relevant; recall
  benchmarks measure recall, while representational accuracy operates at
  the interpretation layer.
\end{itemize}

\hyperref[exploratory-case-study-letta-stateful-agent-n3-post-hoc]{§4.5} reports an exploratory case study on a fundamentally different
memory architecture (Letta self-edited block, N=3 subjects).

The findings were checked for apparatus artifacts across seven
sensitivity tests:

\begin{itemize}
\tightlist
\item
  \textbf{Cross-provider response generation (\hyperref[cross-provider-response-generation-tier-2-replication]{§4.6.1}).} Direction
  reproduces on 7 of 9 (subject, response-model) cells across three
  response models from two providers (Anthropic Haiku 4.5 + Sonnet 4.6,
  Google Gemini 2.5 Pro), on batteries regenerated by OpenAI GPT-5.4.
  Coverage is 3 subjects (Ebers, Yung Wing, Zitkala-Ša), not the full
  main-study set.
\item
  \textbf{Judge panel (\hyperref[judge-panel-sensitivity-5-judge-primary-vs-7-judge]{§4.6.2}).} No directional claim flips between the
  5-judge primary and the 7-judge sensitivity, and at the per-subject
  level no subject's Spec-lift sign changes between panels.
\item
  \textbf{Battery composition (\hyperref[battery-composition-sensitivity]{§4.6.3}).} Neither battery-question-type
  composition nor Hamerton's leverage explains away the \hyperref[the-cross-subject-gradient-and-its-per-question-mechanism]{§4.1} gradient
  slope.
\item
  \textbf{Statistical-rigor checks on the gradient (\hyperref[statistical-rigor-checks-on-the-headline-gradient]{§4.6.4}).} Bootstrap
  CI {[}−1.25, −0.74{]} excludes zero and a 50\%-attenuated effect.
  Joint regression (drop Hamerton + literal-recall covariate) holds the
  slope at −0.87. Permutation test (10,000 reshuffles) places the
  observed slope outside the entire null distribution (p \textless{}
  0.0001).
\item
  \textbf{Wrong-Spec derangement protocol (\hyperref[wrong-spec-derangement-protocol-sensitivity]{§4.6.5}).} The wrong-Spec
  result holds across both protocols (adversarial v1 and random v2).
\item
  \textbf{Retrieval-overlap sensitivity (\hyperref[retrieval-overlap-sensitivity-semantic-similarity-matching-k-variation]{§4.6.6}).} The \hyperref[cross-system-retrieval-providers-do-not-converge]{§4.4.1}
  retrieval-divergence finding survives semantic-similarity matching at
  every threshold tested and at K=5 as well as K=10.
\item
  \textbf{Rubric-handling limitations (\hyperref[rubric-handling-limitations-post-hoc-validity-audit-1]{§4.6.7}).} A post-hoc audit
  identified two limitations (refusal anchor ambiguity and length-score
  correlation in the No-Context Baseline (C5)); both bias the No-Context
  Baseline (C5) upward, so the reported +0.89 mean lift is conservative
  rather than inflated.
\item
  \textbf{What these checks do not address (\hyperref[class-level-llm-dependence-the-limitation-no-robustness-check-resolves]{§4.6.8}).} The class-level
  LLM-as-judge concern remains and is treated in \hyperref[measurement-apparatus]{§6.2}.
\end{itemize}

\hyperref[discussion]{§5} develops what these results imply for AI personalization beyond the
specific experiment, and \hyperref[limitations]{§6} bounds what the experiment cannot establish.

\begin{center}\rule{0.5\linewidth}{0.5pt}\end{center}

\hypertarget{discussion}{%
\section{5. Discussion}\label{discussion}}

\hyperref[results]{§4} produced the empirical results; this section discusses their
implications for AI personalization, particularly as AI moves from a
tool a person uses to an agent that acts on that person's behalf. The
numerical findings are in \hyperref[results]{§4} and the appendices and are not restated
here; the commentary below picks up the larger industry-direction
questions the findings open.

\hypertarget{synthesis-what-the-seven-findings-together-establish}{%
\subsection{5.1 Synthesis: What The Seven Findings Together
Establish}\label{synthesis-what-the-seven-findings-together-establish}}

Across 14 historical subjects and five memory-system configurations, the
study tested whether the Spec, operationalizing a static interpretive
layer this paper introduces,\footnote{The interpretive layer this paper
  introduces is the Behavioral Specification: a structured document that
  captures how a specific person reasons. Full pipeline and operational
  detail in \hyperref[pipeline-for-the-behavioral-specification]{§3.7}.} increases an AI system's representational accuracy of
a specific person. This was operationalized via behavioral prediction on
held-out autobiographical text scored by a calibrated, baselined
five-judge LLM panel. The Spec reliably moves the model from refusal or
generic guessing to grounded subject-specific responses where the model
has insufficient pretraining on the subject. The matched Spec's content
does the work, not the structure of the prompt. An adversarial
wrong-Spec control actively degrades performance below baseline. On
interpretation-heavy questions where retrieved facts alone
underdetermine the answer, the Spec supplies the interpretive pattern
existing memory systems cannot, and three of four commercial systems
show positive aggregate prediction-accuracy lift under at least one
configuration as a result. The Spec recovers most of the predictive
accuracy of the full source corpus at 5x to 80x smaller context, and it
nearly eliminates response hedging on questions retrieval alone could
not ground. Current memory-system providers do not converge on which
facts are most relevant given identical input, even under relaxed
similarity matching when using their native ingestion pipelines.

Together, these findings establish that a portable, content-specific,
structurally compressible interpretive layer adds a measurable dimension
to AI personalization. The interpretive layer is distinct from raw
facts, raw corpus, current memory-system retrieval, and the
pattern-based inferences a frontier model attempts on those inputs on
its own. It complements rather than competes with each of them, and
makes explicit what the model would otherwise assume implicitly about a
specific user. It is most useful where pretraining is thin, but it adds
value on top of all three other context types as well, at a context cost
compatible with production deployment.

Representational Accuracy has been validated directionally by the data
but not absolutely by human annotation. Human-validation at scale is the
highest-priority follow-up to this paper (\hyperref[measurement-methodology]{§7.2}).

\begin{center}\rule{0.5\linewidth}{0.5pt}\end{center}

\hypertarget{why-the-gradient-is-the-load-bearing-finding}{%
\subsection{5.2 Why the Gradient is the Load-bearing
Finding}\label{why-the-gradient-is-the-load-bearing-finding}}

The gradient is load-bearing because it does not say what one might
expect from a paper advocating for an interpretive layer. The Spec helps
where the model lacks an interpretive frame for the person being asked
about. On questions where the model can already produce a substantive
answer from its pretraining, the Spec adds little. On a small but
non-trivial subset of questions where the model already had the answer,
the Spec actively interferes with what pretraining was about to produce.
Franklin is the cleanest case of this interference, where high-baseline
pretraining outperforms a Spec-containing condition on aggregate
(\hyperref[the-gradient-at-the-high-baseline-end-franklin-reference]{§4.1.2}). The aggregate effect across a battery is not inflated, because
this per-question shape is conditional rather than uniform.

What this points to is a deployment question more than a finding. If the
layer helps on some questions and hurts on others, the right serving
regime is not to attach it to every query. Dynamic activation, where the
Spec is supplied when the question type warrants it and withheld when
retrieval already suffices, is the production-realistic next step that
the static-attachment results in this paper set the stage for. \hyperref[composition-with-retrieval]{§5.4}
picks this up directly.

The deeper question the gradient surfaces is what makes a representation
accurate enough to apply across situations the underlying data never
contained. A person encounters an effectively unbounded space of
situations in which their behavior is consistent in ways they themselves
could only partially articulate. Whether a structured authored artifact
can carry enough of that consistency to predict held-out behavior is
what the gradient empirically supports.

\begin{center}\rule{0.5\linewidth}{0.5pt}\end{center}

\hypertarget{retrieval-is-not-interpretation}{%
\subsection{5.3 Retrieval is not
Interpretation}\label{retrieval-is-not-interpretation}}

The initial hypothesis going into the cross-system retrieval comparison
was a natural one. Four sophisticated memory systems given identical
input and identical questions should converge on which facts to surface
as most relevant. They do not. Mean pairwise top-K overlap across the
four commercial systems is 8.3\% (Jaccard 0.083, \hyperref[cross-system-retrieval-providers-do-not-converge]{§4.4.1}). The structural
implication is more interesting than the architectural one. There is no
unified fact set that answers a behavioral question about a person.
Different facts can route through different interpretive patterns to the
same outcome, and the same facts can route through different patterns to
different outcomes. The legal-reasoning analog is the cleanest
articulation of this. Lawyers know that a fact pattern admits multiple
defensible interpretations, and that the choice of interpretive frame is
where reasoning actually happens. The same is true for representing a
person. A worked illustration of the same principle: the held-out fact
``the subject was offered physical labor assistance by subordinates
during a strenuous expedition'' can route to \emph{accept the
assistance} under a frame that prioritizes practical efficiency,
\emph{refuse the assistance and perform the labor publicly} under a
frame that treats personal example as a load-bearing leadership move,
and \emph{delegate to a specific subordinate} under a frame that treats
relational obligation as primary. The Fukuzawa-Cortes case study in \hyperref[the-cross-subject-gradient-and-its-per-question-mechanism]{§4.1}
Example B shows this exact divergence on a real held-out question: the
same retrieved facts, three plausible Spec-driven routes, only one of
which matches what Cortes actually did.

What the field's memory benchmarks measure is recall, in the sense of
whether the system retrieved the fact when asked. Precision, in the
sense of whether the facts surfaced are the right facts for the specific
question being asked, has been evaded. Some questions can be answered
through recall alone (verifiable factual lookups). Others cannot. The
\hyperref[the-cross-subject-gradient-and-its-per-question-mechanism]{§4.1} Fukuzawa example is the cleanest case. Asked about a naval officer
who later prevented military conflict, pure recall returns the
most-mentioned naval officer in the corpus and lands on the wrong
person. The right answer requires the interpretive pattern that
distinguishes earned authority from inherited authority in Fukuzawa's
reasoning. That is not a fact-retrieval problem. What the recall
benchmarks measure is partial. Representational accuracy is the axis the
field has not been measuring on.

\begin{center}\rule{0.5\linewidth}{0.5pt}\end{center}

\hypertarget{composition-with-retrieval}{%
\subsection{5.4 Composition with
Retrieval}\label{composition-with-retrieval}}

Retrieval and interpretation have distinct roles. The more interesting
question is not whether one is necessary in the presence of the other.
It is how to compose them. How does an AI decide that a given question
requires interpretation rather than recall? How much of that decision
should be left to the AI itself to make?

The memory-system landscape is more heterogeneous than the recall
benchmarks suggest. The four commercial systems compared in this paper
produce very different ingestion outputs, very different retrieval
shapes, and very different assumptions about how memory should be
managed. Some lean on human-analog architectures (long-term and
short-term partitions, episodic consolidation). Others lean on agent-led
memory management. Letta's stateful path is the cleanest example, where
the agent itself rewrites its memory block during ingestion. The
interpretive-layer question stacks on top of this variety. Does the
layer provide its own usage guidance to the model (when to invoke, what
to attend to), or is the abstraction enough that the model figures it
out?

Internal prior testing on this front has produced two early signals. A
prior stacking study found that when the Spec was placed in the
conversation as a single project-folder artifact, the response model
could explicitly reference whether or not it was using the Spec on a
given response. When the same Spec was broken into anchors, core, and
predictions files served separately, more granular activation was
visible across components. Independently, an MCP-server-based serving
layer is in development that addresses the runtime side. It decides
which Spec components arrive when. The static-attachment regime used in
this paper is the conservative baseline. The serving layer is the
production-realistic next step (\hyperref[production-serving-and-infrastructure]{§7.4}).

Two hard facts about composition remain. Providing information to the
model does not guarantee the model uses it. The model not acknowledging
that it used a piece of information does not mean it failed to integrate
it. We do not currently have a way to mechanically enforce that the Spec
is used where appropriate, nor to measure with confidence the causal
contribution of any specific piece of served context to any specific
output. Composition with retrieval, in the operational sense, is open
territory.

\begin{center}\rule{0.5\linewidth}{0.5pt}\end{center}

\hypertarget{wrong-spec-mechanism-and-hedging-elimination}{%
\subsection{5.5 Wrong-Spec Mechanism and Hedging
Elimination}\label{wrong-spec-mechanism-and-hedging-elimination}}

The wrong-Spec result is most interesting where it might be least
surprising. Structure alone does not produce the lift. An adversarial
wrong-Spec hurts (Δ −0.25 below baseline). A matched Spec helps (Δ +0.35
above baseline). What is interesting is the random-derangement
condition, where a subject receives the Specification of an arbitrary
other subject. The response sometimes lands close to the held-out anyway
(random-derangement Δ +0.15). A wrong Specification representing a
different person can occasionally predict the right behavioral outcome
on a held-out question. That calls for a broader reading. The wrong-Spec
mechanism is the inference-time analogue of training-time emergent
misalignment reported by Betley et al.~(2025, arXiv:2502.17424):
narrow-finetuning on a misspecified objective produces broad behavioral
drift; here, narrow-context with a misspecified representation produces
analogous behavioral drift at query time.

The implication is structural about behavior itself. There is a finite
set of behavioral outcomes a person might produce on any given
situation. There is an effectively unbounded set of underlying reasoning
paths that can produce each of those outcomes. Two specifications
describing two different people can converge on the same predicted
action through different paths. This is consistent with how people
actually behave. Large behavioral registers are shared across people.
What is unique to an individual is the specific matrix of which facts
they hold, which interpretive patterns they apply, and which predictors
fire under which conditions. People share registers. Individuals are the
activation pattern across them.

Hedging elimination is the other half of this section, and the mechanism
behind it remains the part of the result we understand least. The
matched Spec collapses baseline hedging from 41.2\% to 0.4\% under the
broad-pattern classifier. The decision boundary between hedging,
abstaining, and committing in current language models is determined by
training signal that is not visible to evaluators at this layer. The
hedging-elimination effect under a matched Spec suggests that the
interpretive layer gives the model enough grounding to commit rather
than hedge. The wrong-Spec adversarial behavior raises a question we did
not isolate. When the model detects a mismatch and refuses to apply the
served Specification, is it detecting that based on the Spec's internal
coherence, or based on what the model already knows about the named
subject from pretraining? The same question applies to abstention in
general. The relationship between hedging, abstention, and sycophancy in
current LLMs is opaque, and the interpretive layer interacts with each
of them in ways that warrant their own follow-up work.

\begin{center}\rule{0.5\linewidth}{0.5pt}\end{center}

\hypertarget{compression-and-what-makes-personalization-operationally-tractable}{%
\subsection{5.6 Compression and What Makes Personalization Operationally
Tractable}\label{compression-and-what-makes-personalization-operationally-tractable}}

Two things surprised us about the compression results. The first is how
well a frontier language model handles a very long context. The
raw-corpus condition produced the highest prediction scores in the
study, which is not a result one would have confidently predicted three
years ago. The second is how much of that signal a structured
representation recovers at a fraction of the context cost. The Spec is
roughly 25× smaller than the raw corpus on average, and it recovers
about 75\% of the corpus lift over baseline on the behavioral-prediction
task.

Both surprises point at the same industry-direction question. Most of
the effort in long-context language models has gone into enlarging the
window. Effective use of that window has lagged. The lost-in-the-middle
effect (Liu et al.~2024, arXiv:2307.03172) and other position-sensitive
degradations are documented in the literature, and an extension of those
findings is straightforward. Even if a model could ingest infinite
context, it would have to construct its own sub-context (an interpretive
frame) to actually reason from. The question is not how much context the
model can take in. The question is which context.

This reframes the personalization problem. A faithful interpretive
representation, owned by the person it represents, is what the
personalization problem reduces to. Faithfulness is the open piece. Only
the individual whose Spec it is can verify whether the compression
preserved what matters to them. A measure of representational accuracy
authored from someone else's text and judged by someone else's panel is
necessarily an external proxy for what the person being represented
would actually recognize about themselves. Closing that gap requires the
user in the loop. The existing-deployability point cuts the other way
and is more immediately useful. We do not need to wait for context
windows to grow further. A compressed interpretive frame already fits
inside today's context budgets, and the gradient established here is
recoverable today.

\begin{center}\rule{0.5\linewidth}{0.5pt}\end{center}

\hypertarget{privacy-and-the-case-for-user-held-representation}{%
\subsection{5.7 Privacy and the Case for User-held
Representation}\label{privacy-and-the-case-for-user-held-representation}}

Privacy here is not an industry critique. In an agentic world, where AI
systems increasingly act on behalf of specific individuals in
interactions with other agents, accurate individual representation is in
everyone's structural interest. The third party an agent transacts with
on your behalf benefits from interacting with an agent that knows you
well, because that agent will make decisions more consistent with what
you would actually have decided. The first-party agent benefits because
acting in alignment is its job. The individual being represented
benefits because the agent's choices are recognizably theirs.

The structural condition for this to work is ownership. The
representation of an individual has to be held by the individual for the
system to be optimization-positive at population scale rather than
another extraction pipeline. If a third party builds the representation
and holds it, the same dynamic that produces behavioral micro-targeting
in advertising applies with sharper resolution. The inferred
representation exists for someone else's benefit, and the person being
represented cannot see it or correct it. User-held representation does
not prevent external behavioral modeling, but it changes the
relationship. A user-owned interpretive layer exists alongside external
models and is the structural defense against weaponization of a crude or
unauthorized version of the person.

What this requires beyond user ownership is trust and traceability. A
representation that the individual owns is operationally useful only if
third-party agents interacting with them can verify the representation's
provenance and authority. Without a trust network, the system degrades
back into the opaque-third-party-representation it was supposed to
replace. The case for user-held representation is therefore inseparable
from the case for an industry-level trust and traceability layer that
makes user-held representation operational at scale.

Per-user calibration sits inside the safety envelope, not above or below
it. A language model is already shaped toward whatever distribution
dominated its pretraining and applies that shape to every user the same
way. A Behavioral Specification redirects that existing shape toward the
specific user the system is acting on behalf of, rather than adding new
shaping on top. Representational accuracy and safety alignment operate
at distinct layers, and the constructive question is how the two layers
compose. Whether they are formally independent or interact in some
structured way is open empirical territory. This paper neither tests nor
needs to defend full independence. \hyperref[future-work]{§7} develops the safety and deployment
implications.

\begin{center}\rule{0.5\linewidth}{0.5pt}\end{center}

\hypertarget{closing-remarks}{%
\subsection{5.8 Closing Remarks}\label{closing-remarks}}

What the paper has done is operationalize a representational-accuracy
measure of an interpretive layer, and demonstrate that a representation
of this kind can be built within the resources and constraints of
today's frontier AI systems. The result is a measurable proxy for
whether an AI knows enough about a specific person to act in alignment
with how that person actually reasons. This is not the only such measure
that can be built, nor the most accurate one possible. It is a measure
that exists today and that produces interpretable results on text the
model has never seen.

The forward-looking commitment is that the future of human-AI alignment,
particularly as agents take on a larger share of the actions individuals
delegate to AI, requires this kind of representation as a structural
primitive. An AI agent cannot act in alignment with a specific person
without an accurate representation of how that person reasons. A measure
of representational accuracy is what makes alignment claims about
specific individuals testable. A Behavioral Specification is one such
measure; the field needs more of them, stronger versions of them, and a
clearer accounting of what each one captures and what it leaves out. As
AI integrates into more of everyday life through agents, accurate
individual representation is the structural problem that needs to be
addressed, not a feature that gets added later.

\begin{center}\rule{0.5\linewidth}{0.5pt}\end{center}

\hypertarget{limitations}{%
\section{6. Limitations}\label{limitations}}

The paper's claims are bounded by four axes of constraint on the
experimental setup: the subject sample (\hyperref[subject-sample]{§6.1}), the measurement apparatus
(\hyperref[measurement-apparatus]{§6.2}), the pipeline and specification stability (\hyperref[pipeline-and-specification-stability]{§6.3}), and the scope
of exploration (\hyperref[scope-of-exploration]{§6.4}). Each is a permanent caveat on how the paper's
results should be read, distinct from the follow-up experiments proposed
in \hyperref[future-work]{§7}.

\hypertarget{subject-sample}{%
\subsection{6.1 Subject sample}\label{subject-sample}}

The 14 main-study subjects are a selected sample, not a population.
Pretraining-coverage bias and the single-living-subject constraint are
load-bearing for the paper's framing and are developed in \hyperref[why-the-gradient-is-the-load-bearing-finding]{§5.2}. This
subsection covers four remaining external-validity caveats that \hyperref[discussion]{§5} does
not address.

\textbf{Public-domain selection.} All subjects are historical figures
whose autobiographies or memoirs are in the public domain and have been
digitized by Project Gutenberg or Internet Archive. That selection
pipeline is biased toward figures whose writing was preserved in
published form and Western publishing traditions. The paper's
cross-continent spread (Saint Augustine, Bābur, Fukuzawa Yukichi, Sunity
Devee, Zitkala-Ša, Olaudah Equiano, Mary Seacole) partially mitigates
but does not remove this bias.

\textbf{Self-presentation bias.} Autobiographies are self-curation. What
each subject chose to include in their memoir is not a neutral record of
their behavior; it is a self-selected narrative that may over-represent
patterns the author wished to be remembered for and under-represent
patterns they chose to leave out. Behavioral-prediction batteries
derived from autobiography inherit this bias, and neither the pipeline
nor the rubric has a mechanism to correct for it.

\textbf{Translation artifacts.} Three subjects' corpora are English
translations of non-English originals (Augustine's \emph{Confessions}
from Latin, Bābur's \emph{Bābur-nama} from Chagatai Turkic via Persian,
Cellini's autobiography from Italian). Translations introduce stylistic
and register shifts that the extraction pipeline processes as if they
were original text. A Specification authored from a translated corpus
may inherit translator choices in addition to the subject's actual
patterns.

\textbf{Era.} The oldest subject is 4th to 5th century (Augustine); the
newest is early 20th century (Zitkala-Ša, Sunity Devee). Reasoning
patterns in modern work contexts, contemporary family structures,
technical or digital-native domains, and late-20th-century cultural
frames are not sampled. Whether the gradient holds when specifications
are authored from modern-era corpora is a generalization the study
cannot make from its sample alone.

\textbf{Autobiography as a genre.} Beyond the four caveats above,
autobiography is itself a genre with its own distributional properties:
reflective, retrospective, narratively shaped, written for an audience,
structured by the author's understanding of what makes a coherent
life-story. The behavioral patterns extracted from it are patterns
\emph{as the author chose to surface them}. Both halves of the
train/held-out split come from the same authored narrative, so
within-narrative consistency is exactly what the held-out test is
positioned to detect. Whether the gradient holds when specifications are
authored from non-narrative source material (real-time chat, decisions
under time pressure, technical domains the subject never wrote
reflectively about) is not testable from this design. The \hyperref[measurement-methodology]{§7.2}
living-user replication and \hyperref[stateful-agent-implementations-and-temporal-drift-tracking]{§7.5} stateful-agent variants together begin
to address cross-genre generalization.

Taken together, these five caveats mean the paper's results should be
read as evidence for the claims at the conditions tested. Generalization
across era, source language, self-presentation mode, source genre, and
digital-versus-analog source material requires follow-up experiments.

\begin{center}\rule{0.5\linewidth}{0.5pt}\end{center}

\hypertarget{measurement-apparatus}{%
\subsection{6.2 Measurement apparatus}\label{measurement-apparatus}}

This section covers the measurement-apparatus constraints on how the
paper's numbers should be read. The rubric limitations are in \hyperref[rubric-handling-limitations-post-hoc-validity-audit]{§3.3.6};
the LLM-as-judge limitation is the canonical one and is treated in full
here.

\textbf{Class-level LLM-as-judge circularity.} Every response in this
study is generated by an LLM, every judge is an LLM, and the question
batteries are also LLM-generated (\hyperref[question-battery-formation]{§3.5}). The 5-judge primary panel and
the 7-judge sensitivity check together address within-provider
circularity (\hyperref[cross-provider-response-generation-tier-2-replication]{§4.6.1}, \hyperref[judge-panel-sensitivity-5-judge-primary-vs-7-judge]{§4.6.2}). What these checks do not address is
class-level LLM circularity: the concern that an all-LLM pipeline could
be self-reinforcing in ways that human evaluators would not validate.
The methodological precedent for the LLM-judge panel is established in
\hyperref[prior-work-industry-benchmarks-the-fifth-target]{§2} and \hyperref[scoring-rubric-with-calibrated-llm-judge-panel]{§3.3}; the panel answers the directional question reliably but not
the absolute-quality question. A stratified human-validation subset is
the leading measurement follow-up flagged in \hyperref[measurement-methodology]{§7.2}; until that exists,
the paper is best read as a prototype with directional, not precise,
evidence on its headline claims.

\textbf{Response-model coverage.} The main-study response model is
Claude Haiku 4.5. The \hyperref[cross-provider-response-generation-tier-2-replication]{§4.6.1} Tier 2 cross-provider directional probe ran
2 additional response models (Claude Sonnet 4.6, Google Gemini 2.5 Pro)
on 3 subjects spanning the gradient (Ebers, Yung Wing, Zitkala-Ša)
against GPT-5.4-regenerated batteries; Claude Opus 4.6, GPT-5.4, and
GPT-4o were used as judges in Tier 2 but not as response models. The
Specification-effect direction reproduced on 5 of 6 (subject,
response-model) cells under every panel and Δ-definition tested (Sonnet
4.6 and Gemini Pro only; the corresponding Haiku Tier 1 cell is counted
in the \hyperref[cross-provider-response-generation-tier-2-replication]{§4.6.1} ``7 of 9'' figure, which adds the Haiku response-model row
at the same three subjects). The main-study response model is Haiku
across all 14 subjects in \hyperref[the-cross-subject-gradient-and-its-per-question-mechanism]{§4.1}; Tier 2 establishes direction across
response-model families on a small subset only. The paper's aggregate
numbers should be read as what the Specification does with Haiku; other
response models may produce different absolute magnitudes while
preserving the gradient.

\textbf{Prompt-phrasing ambiguity.} The authoring pipeline prompts, the
response-generation prompts, and the judge prompts all depend on
specific word choices, ordering, and phrasing. We did not systematically
test prompt sensitivity as part of this experiment. Prompt sensitivity
for the Behavioral Specification authoring pipeline is a separate
workstream that informed pipeline design and is distinct from this
study's response-generation and judge prompts, which were not varied.
Different wordings at any of these stages could produce different
numeric results, different extracted fact sets, or different judge
scores on the same response. The paper's claims are downstream of the
specific prompts used throughout the study (documented in the public
repository scripts); we make no claim about prompt invariance.

\textbf{Inter-judge calibration variance.} Pairwise Spearman ρ across
judges is 0.86 to 0.93 (\hyperref[inter-judge-agreement]{§3.3.4}), so the rank order of conditions is
stable across the panel. Absolute-score calibration varies (\hyperref[calibration]{§3.3.3}):
Gemini Pro fails verbatim-match calibration (4.15 where calibrated
judges score 5.0), Opus runs lenient on abstentions (1.41 mean where
Sonnet runs strict at 1.14), and length-sensitivity differs across
judges. The 5-judge primary aggregate is therefore a stable reading of
direction but a panel-specific reading of magnitude. A different judge
panel would produce different aggregate numbers while preserving the
direction of every claim, which is part of why \hyperref[closing-remarks]{§5.8} frames the paper as
directional rather than precise.

\begin{center}\rule{0.5\linewidth}{0.5pt}\end{center}

\hypertarget{pipeline-and-specification-stability}{%
\subsection{6.3 Pipeline and specification
stability}\label{pipeline-and-specification-stability}}

What follows covers pipeline-internal constraints on how the paper's
results should be read.

\textbf{Pipeline version tested.} The specifications used in this study
were produced by a specific pipeline version, snapshotted at the time
the experiments were run. Two locations preserve that version: the
deployed BaseLayer pipeline at
\texttt{https:/\allowbreak{}/\allowbreak{}github.\allowbreak{}com/\allowbreak{}agulaya24/\allowbreak{}BaseLayer} (the stable production
version at experiment time) and a forked, study-specific copy preserved
inside the study repository (\texttt{baselayer/}, a frozen copy of
v0.2.0, the version used to author the 14 specifications scored in
\hyperref[results]{§4}). The pipeline
has evolved through development, and different pipeline versions produce
different specifications on the same source corpus. The paper's results
are specific to the pipeline version tested, and the study does not
measure how the gradient shifts under earlier or later pipeline
versions. The evaluation harness used here can serve as a benchmark for
future pipeline iterations: each new authoring-pipeline version can be
measured against the current specifications on the same 14-subject
batteries to assess whether the gradient strengthens, weakens, or shifts
shape (\hyperref[future-work]{§7}).

\textbf{Specification stability under the same pipeline version.}
Running the same pipeline twice on the same corpus at temperature 0 does
not produce identical specifications. A numeric-similarity probe
(script:
\texttt{scripts/\allowbreak{}v11\_\allowbreak{}9\_\allowbreak{}6\_\allowbreak{}pipeline\_\allowbreak{}variance\_\allowbreak{}similarity\_\allowbreak{}20260510.\allowbreak{}py})
compared the three reruns of each probe subject's brief at the sentence
level. At sentence grain, verbatim sentence-level matches across reruns
are rare. For non-verbatim sentences, best-match cosine similarity
against the other run's sentence pool (MiniLM-L6-v2 sentence embeddings,
normalized) produces a pooled mean of 0.55, median 0.54, IQR {[}0.47,
0.61{]}, with roughly 3\% of non-verbatim sentences reaching cosine ≥
0.80. Sentence-level analysis is stricter than character-level or
paragraph-level overlap; the reruns are not paraphrases of one another
at sentence resolution. Qualitative side-by-side reading still shows
that the reruns cover the same predicates and behavioral patterns at a
higher conceptual level, but the sentence-grain measurement makes the
surface variance concrete. The score-level stability that matters for
the gradient (per-subject run-to-run SD of Δ\_C4a) is reported in the
variance probe below, and is what determines whether the \hyperref[the-cross-subject-gradient-and-its-per-question-mechanism]{§4.1} slope is
robust to this kind of surface non-determinism.\footnote{Full
  per-subject pairwise detail at
  \texttt{docs/\allowbreak{}research/\allowbreak{}v11\_\allowbreak{}9\_\allowbreak{}6\_\allowbreak{}pipeline\_\allowbreak{}variance\_\allowbreak{}similarity\_\allowbreak{}20260510.\allowbreak{}md};
  raw JSON at
  \texttt{docs/\allowbreak{}research/\allowbreak{}v11\_\allowbreak{}9\_\allowbreak{}6\_\allowbreak{}pipeline\_\allowbreak{}variance\_\allowbreak{}similarity\_\allowbreak{}20260510.\allowbreak{}json}.
  Pool n = 1,304 non-verbatim sentences across three probe subjects ×
  three pairwise rerun comparisons.}

\textbf{Per-subject pipeline variance, characterized.} A targeted
replication probe was run on three subjects spanning the gradient
(Sunity Devee, C5 = 1.03; Yung Wing, C5 = 1.88; Augustine, C5 = 2.58).
For each subject, the Sonnet layer-authoring step and the Opus compose
step were re-run three times against the same per-subject extracted fact
set at temperature 0, producing three independent specifications. Each
rerun was scored on the full behavioral-prediction battery in the C2a
(Spec only) and C4a (All Facts + Spec) conditions on the 5-judge primary
panel. The resulting per-subject standard deviation of Δ\_C4a across
reruns is reported below, alongside the cross-subject SD that the \hyperref[the-cross-subject-gradient-and-its-per-question-mechanism]{§4.1}
gradient slope is fit to. For each subject the table reports three
numbers: the Spec's effect on representational accuracy in the \hyperref[the-cross-subject-gradient-and-its-per-question-mechanism]{§4.1} main
study (one authored specification), the standard deviation of that
effect across three independent pipeline reruns on the same corpus, and
that standard deviation as a fraction of the between-subject standard
deviation the \hyperref[the-cross-subject-gradient-and-its-per-question-mechanism]{§4.1} gradient slope is regressed against.

\begin{longtable}[]{@{}
  >{\raggedright\arraybackslash}p{(\columnwidth - 6\tabcolsep) * \real{0.20}}
  >{\raggedleft\arraybackslash}p{(\columnwidth - 6\tabcolsep) * \real{0.27}}
  >{\raggedleft\arraybackslash}p{(\columnwidth - 6\tabcolsep) * \real{0.27}}
  >{\raggedleft\arraybackslash}p{(\columnwidth - 6\tabcolsep) * \real{0.27}}@{}}
\toprule
Subject & Canonical Δ\_C4a (\hyperref[the-cross-subject-gradient-and-its-per-question-mechanism]{§4.1}) & Per-rerun Δ\_C4a SD (n=3) & \% of
cross-subject SD \\
\midrule
\endhead
Sunity Devee & +1.38 & 0.103 & 17.4\% \\
Yung Wing & +0.52 & 0.055 & 9.3\% \\
Augustine & +0.11 & 0.130 & 22.0\% \\
\textbf{Pooled (3 subjects)} & n/a & \textbf{0.101} & \textbf{17.1\%} \\
\bottomrule
\end{longtable}

\textbf{Read of the precision question.} The directional finding
survives across reruns: low-baseline subjects keep improving (6 of 6
reruns positive across the 2 low-baseline probe subjects), and the
gradient slope point estimate is not materially threatened.
Quantitatively, the pooled per-subject run-to-run SD of Δ\_C4a is 0.10
on the 1-5 rubric, against a cross-subject SD of 0.59 that the gradient
slope is regressed against; run-to-run pipeline variance is therefore on
the order of 17\% of the signal the slope is fit to, and well under the
95\% CI half-width on the slope (0.29). What pipeline variance does
affect is the precision attached to any single per-subject point
estimate. The per-subject Δ\_C4a numbers in \hyperref[the-cross-subject-gradient-and-its-per-question-mechanism]{§4.1} should be read with a
soft uncertainty bar of roughly ±0.10 around them. Augustine
(mid-baseline, canonical Δ = +0.11) sits at the top of its rerun range
and the sign flips on 2 of 3 reruns, so individual mid-baseline
subjects' Spec-effect sign is itself within the run-to-run uncertainty
band.

\textbf{Scope and caveats of the variance probe.} The probe covers the
lighter-scope variance only: the Sonnet authoring step plus the Opus
compose step. Extraction-stage non-determinism is held constant by
reusing each subject's pre-populated SQLite and ChromaDB state across
reruns; including extraction would likely add additional variance at the
front of the pipeline. The probe covers low-baseline and mid-baseline
subjects but does not reach the Franklin-style high-baseline tail (C5 =
3.77), so the H2 corollary (that the Spec can interfere with strong
pretraining signal at the high-baseline end, producing a near-zero or
negative Δ) is not directly stress-tested by this run. With n = 3 reruns
per subject the per-subject SD point estimates carry their own wide 95\%
confidence intervals (roughly {[}0.5×, 6×{]} of the value); the pooled
three-subject estimate is more stable than any single per-subject
estimate but should still be read as an order-of-magnitude indicator
rather than a precision number. With those caveats stated, the
run-to-run SD is small enough relative to the cross-subject SD that we
accept the \hyperref[the-cross-subject-gradient-and-its-per-question-mechanism]{§4.1} slope and R² as findings about the gradient rather than
artifacts of a single specification authoring.\footnote{Per-rerun specs
  and judgments are at
  \texttt{data/global\_\textless{}subject\textgreater{}/\_variance\_runs/run\_\textless{}N\textgreater{}/}
  and
  \texttt{results/global\_\textless{}subject\textgreater{}/\_variance\_runs/run\_\textless{}N\textgreater{}\_*.json}.
  Full report and reproducibility scripts at
  \texttt{docs/\allowbreak{}research/\allowbreak{}v10\_\allowbreak{}pipeline\_\allowbreak{}variance\_\allowbreak{}analysis.\allowbreak{}md},
  \texttt{scripts/\allowbreak{}\_\allowbreak{}v10\_\allowbreak{}pipeline\_\allowbreak{}variance.\allowbreak{}py}, and
  \texttt{scripts/\allowbreak{}\_\allowbreak{}v10\_\allowbreak{}pipeline\_\allowbreak{}variance\_\allowbreak{}report.\allowbreak{}py}.}

\textbf{Pipeline model choices were not varied systematically.} The
pipeline uses Claude Haiku for extraction, all-MiniLM-L6-v2 for
embeddings, Claude Sonnet for layer authoring, and Claude Opus for the
compose step (\hyperref[pipeline-for-the-behavioral-specification]{§3.7}). These model choices were not varied across the
study. Different models at any step could produce different
specifications: a different extraction model could surface different
facts, a different embedding model could change retrieval behavior, a
different authoring model could produce differently-structured anchors
and predictions, a different composition model could synthesize the
layers differently. Extending model support for each pipeline step, and
measuring the gradient under alternate pipeline configurations (for
example GPT-5.4 extraction, OpenAI embeddings, a non-Anthropic authoring
model), is a direct follow-up flagged in \hyperref[specification-design-and-composition]{§7.3}, alongside the broader
question of cross-model consistency for both Spec authoring and usage.

\begin{center}\rule{0.5\linewidth}{0.5pt}\end{center}

\hypertarget{scope-of-exploration}{%
\subsection{6.4 Scope of exploration}\label{scope-of-exploration}}

Not every experimental combination was run. The study prioritized
coverage of the conditions and subjects central to H1 through H5 (\hyperref[the-cross-subject-gradient-and-its-per-question-mechanism]{§4.1}
through \hyperref[memory-system-composition]{§4.4}) over running every cell of the design grid. Robustness and
ablation conditions were added selectively rather than exhaustively.

\textbf{Coverage across the experimental grid.} The study spans 11
conditions (C1 through C9 plus two wrong-Spec variants), 14 main-study
subjects, and a 5-judge primary panel plus 2-judge sensitivity check;
response-model coverage is detailed in \hyperref[measurement-apparatus]{§6.2} and \hyperref[cross-provider-response-generation-tier-2-replication]{§4.6.1}. Running every
possible combination (roughly 6,500 separate cells) was not attempted.
Ablation-adjacent conditions (per-layer Spec serving, alternate pipeline
model choices, dynamic activation policies) were not run and are planned
for future work (\hyperref[future-work]{§7}).

\textbf{Letta stateful-agent exploration.} Letta's stateful-agent
architecture is distinct from the archival retrieval path the other
three commercial systems use (\hyperref[memory-system-composition]{§4.4}, \hyperref[exploratory-case-study-letta-stateful-agent-n3-post-hoc]{§4.5}). Testing the stateful path
required a different evaluation harness (\hyperref[exploratory-case-study-letta-stateful-agent-n3-post-hoc]{§4.5} test design), and that
work pulled us partially outside the main-study scope. The resulting
comparison covers three subjects (Hamerton, Ebers, Bābur), one Letta
version, and one response model (Claude Haiku). Extending the
stateful-agent comparison across the full 14-subject gradient, across
additional response models, and against future Letta releases is flagged
as a follow-up in \hyperref[future-work]{§7}.

\textbf{Twin-2K coverage.} Twin-2K (\hyperref[prior-measurement-targets-and-the-gap-representational-accuracy-fills]{§2.1}) is cited in this paper as
prior work that measures a related but distinct property
(survey-response prediction rather than representational accuracy). We
did not run Twin-2K as a condition of the main behavioral-prediction
battery; our findings on Twin-2K are framed as prior-work context rather
than as a benchmark result. See \hyperref[prior-measurement-targets-and-the-gap-representational-accuracy-fills]{§2.1} and Appendix F.4 for the detailed
treatment.

\begin{center}\rule{0.5\linewidth}{0.5pt}\end{center}

\hypertarget{future-work}{%
\section{7. Future Work}\label{future-work}}

Every section of this paper flags at least one follow-up. This section
consolidates them into a research agenda organized by theme.

\hypertarget{subject-and-corpus-expansion}{%
\subsection{7.1 Subject and Corpus
Expansion}\label{subject-and-corpus-expansion}}

A multi-subject living-user replication is the leading follow-up for the
entire paper (\hyperref[why-the-gradient-is-the-load-bearing-finding]{§5.2}, \hyperref[privacy-and-the-case-for-user-held-representation]{§5.7}). The paper's findings are based on 14
historical subjects; whether they generalize to living users is not
directly tested by this study, and replicating the gradient with
multiple living subjects (with proper consent and privacy
infrastructure) would close that gap. Three related expansions:
modern-era corpora (to test whether the gradient holds when
specifications are authored from contemporary writing rather than
pre-20th-century autobiography, \hyperref[subject-sample]{§6.1}), non-English original sources (to
remove translation artifacts, \hyperref[subject-sample]{§6.1}), and alternative testbeds that
isolate reasoning structure without requiring private data, such as U.S.
Supreme Court opinions where documented decisions provide a public
record of individual interpretive patterns that can be held out and
predicted (\hyperref[retrieval-is-not-interpretation]{§5.3}).

\hypertarget{measurement-methodology}{%
\subsection{7.2 Measurement Methodology}\label{measurement-methodology}}

The most impactful measurement follow-up after the living-user
replication in \hyperref[subject-and-corpus-expansion]{§7.1} is replacing the content-match rubric with a
differentiated battery that separates interpretation-heavy from
literal-recall questions and scores epistemic honesty as its own
dimension (\hyperref[rubric-handling-limitations-post-hoc-validity-audit]{§3.3.6}). Alongside this: a curated question set with explicit
quality control on the backward-design process, a human-validated subset
of rubric applications to test whether the rubric was reasonably applied
per-response (\hyperref[rubric-handling-limitations-post-hoc-validity-audit]{§3.3.6}), and human-judge validation on a stratified subset
of responses to address class-level LLM-as-judge circularity (\hyperref[class-level-llm-dependence-the-limitation-no-robustness-check-resolves]{§4.6.8},
\hyperref[measurement-apparatus]{§6.2}). Prompt-sensitivity testing across the authoring,
response-generation, and judging stages (\hyperref[measurement-apparatus]{§6.2}) is a separate
measurement-stability follow-up that becomes important once the rubric
itself is stabilized.

\textbf{Retrieval-overlap follow-ups (from the surfaced \hyperref[cross-system-retrieval-providers-do-not-converge]{§4.4.1}
finding).} Two measurement studies remain open after the \hyperref[cross-system-retrieval-providers-do-not-converge]{§4.4.1}
sensitivity check that already covers K=5 and semantic-similarity
matching for K=10 in both controlled and native configurations:

\begin{itemize}
\item
  \textbf{Convergence-at-larger-K analysis.} This study tested K=10
  retrieved facts per question (mean Jaccard 0.083 across systems,
  \hyperref[cross-system-retrieval-providers-do-not-converge]{§4.4.1}) and a K=5 sensitivity check that lowered overlap rather than
  raising it. The follow-up is K=25, K=50, K=100, and higher across the
  same systems and question set, to map the convergence curve and
  identify the K threshold at which providers begin to agree on which
  facts are relevant (if anywhere).
\item
  \textbf{Meta-analysis of recall benchmarks against retrieval overlap.}
  Memory systems that score within a few percentage points on
  LongMemEval, LOCOMO, and similar recall benchmarks retrieve nearly
  disjoint top-K facts when given identical fact pools and fixed
  questions (\hyperref[cross-system-retrieval-providers-do-not-converge]{§4.4.1}). Recall benchmarks measure recall, which is what
  they should measure; the question is what additional dimensions matter
  for downstream representational accuracy. A meta-analysis comparing
  benchmark scores to retrieval-overlap on the same systems would
  clarify what ``memory recall'' actually predicts about how each system
  ranks facts for a specific interpretive task. The wrong-Spec
  per-question meta-analysis (\hyperref[wrong-spec-derangement-protocol-sensitivity]{§4.6.5}) belongs to the same class of
  follow-up: a deeper read of which parts of the served context the
  model referenced under correct versus mismatched specifications.
\end{itemize}

\textbf{Inter-judge agreement as a question-quality signal.} The 5-judge
primary panel agrees on direction (Spearman ρ = 0.86 to 0.93) but
spreads on absolute magnitude (Krippendorff α = 0.659; \hyperref[inter-judge-agreement]{§3.3.4}).
Questions where judge spread is unusually wide (e.g., Seacole Q2 C4 with
judges splitting 5/2/3/3/1) are candidates for inspection: they may be
questions where the rubric anchors apply ambiguously, where the held-out
passage admits multiple defensible predictions, or where the question
itself is poorly posed. A per-question agreement analysis (variance
across the 5 primary judges, or pairwise α at the question level) would
identify the questions the panel struggles with most and feed back into
the next iteration of the battery.

\textbf{Confident-misalignment vs abstention.} The \hyperref[scoring-rubric-with-calibrated-llm-judge-panel]{§3.3} rubric scores
both explicit abstention and non-abstention misalignment as 1.00. A
coarse post-hoc regex pass classified \textasciitilde93\% of low-end
responses as abstention and \textasciitilde7\% as confident-misalignment
(wrong referent, off-base inference, or confusion with a different
subject; \hyperref[per-question-baseline-engagement-and-the-worked-rubric-example]{§4.1.1} footnote). Why a
model picks confident-wrong over abstention on a given question is open:
which interpretive frames the wrong-Spec activates that the matched Spec
does not, whether the gap correlates with subject-level pretraining
coverage, and whether the rubric should score the two failure modes
separately are first-order follow-ups.

\textbf{Pretraining-bleed analysis on wrong-Spec mismatch detection.}
Under wrong-Spec adversarial pairings, the model often flags the
mismatch explicitly rather than complying (60.6\% mismatch-detection
rate; \hyperref[mechanism-correct-content-not-format]{§4.3}). Whether this detection draws on pretraining knowledge of
the \emph{correct} subject (effectively recognizing ``this Spec doesn't
match Lincoln'') versus on internal coherence of the Spec content alone
is open. A pretraining-bleed analysis would correlate per-subject
mismatch-detection rate with per-subject baseline pretraining coverage
and with per-pair cultural/temporal distance, isolating which signal is
doing the detection. \hyperref[measurement-apparatus]{§6.2} flags this as a measurement caveat the present
design does not isolate.

\hypertarget{specification-design-and-composition}{%
\subsection{7.3 Specification design and
composition}\label{specification-design-and-composition}}

Component ablation on the authored layers (anchors, core, predictions,
brief) is the priority authoring-pipeline follow-up (\hyperref[composition-with-retrieval]{§5.4}). Serving each
layer alone and in combinations, measuring Pattern 1 / Pattern 2 /
Pattern 3 distributions per configuration, would identify which parts of
the pipeline are doing which work. Answers inform both the authoring
pipeline's investment priorities and the dynamic-activation policy's
weights.

Alongside component ablation: alternate pipeline model choices
(extraction, embedding, layer authoring, composition) to measure
sensitivity to specific LLM choices at each pipeline step (\hyperref[pipeline-and-specification-stability]{§6.3}); a Base
Layer referent-variant that retains named entities inside the same
dimensional scaffold, to isolate whether the \hyperref[exploratory-case-study-letta-stateful-agent-n3-post-hoc]{§4.5} Letta-over-Base-Layer
gap is driven by referential vocabulary or by the self-editing process
itself (\hyperref[exploratory-case-study-letta-stateful-agent-n3-post-hoc]{§4.5}); and a layered-stack Letta rerun on the matched-rerun
subjects, which would likely narrow the \hyperref[exploratory-case-study-letta-stateful-agent-n3-post-hoc]{§4.5} gap (\hyperref[exploratory-case-study-letta-stateful-agent-n3-post-hoc]{§4.5}).

\textbf{Cross-model consistency for Spec authoring and usage.} Two
related questions sit on top of the alternate-pipeline-model-choice
follow-up. First, on the authoring side: whether different LLMs at each
pipeline step produce specifications that converge on the same
behavioral patterns from the same source corpus, or whether the
Specification itself drifts with the choice of authoring model. Second,
on the usage side: whether different response models interpret the same
Specification consistently, applying it to produce comparable
predictions on the same held-out questions. The \hyperref[cross-provider-response-generation-tier-2-replication]{§4.6.1} Tier 2 probe
established that the gradient direction holds across three response
models on three subjects; whether the Spec is read and applied the same
way across the broader model landscape is the deeper consistency
question.

\textbf{Named-entity grounding as a complement to predicate structure.}
The \hyperref[exploratory-case-study-letta-stateful-agent-n3-post-hoc]{§4.5} secondary analysis ruled out surface-syntactic alignment as the
mechanism for Letta's case-study lift but identified named-entity
grounding as a contributing factor. The Base Layer pipeline abstracts
source text into structured predicates for explainability and
traceability; the brief retains some personal detail, but specific named
entities (place names, person names, institutional names) sit at the
edge of what the current predicate vocabulary expresses cleanly. When
held-out questions turn on a particular named entity, this can be a
measurable disadvantage. A natural follow-up is a hybrid representation
that pairs each predicate with the named entities it grounds in:
predicates for structure and explainability, entity tags for referential
grounding. The serving-layer architecture in \hyperref[composition-with-retrieval]{§5.4} / \hyperref[production-serving-and-infrastructure]{§7.4} is where the
decision of when to pull the abstract pattern versus the concrete
entities lives; the runtime question is when to surface each. A
named-entity-grounding-vs-axiom-grounding ablation, plus a
paraphrase-resistant rubric that scores directional correctness
separately from referential match, are the specific tests this follow-up
reduces to.

\textbf{Natural-language activation (NLA) round-trip Spec
faithfulness.}\footnote{Natural-language activation (NLA) refers to the
  line of recent interpretability work that uses natural-language
  descriptions to label or steer specific internal directions in
  language-model activation space. We use the term in its general
  interpretability sense here; specific Anthropic publications in this
  line are still emerging and may be cited in revised drafts.} Generate
hypothetical responses across a battery from a Spec only, then
re-extract a Spec from the generated text using the same pipeline.
Compare the round-tripped Spec to the original on predicate coverage,
axiom recall, and behavioral-pattern fidelity. The round-trip
degradation rate measures how much of the Spec's interpretive content
survives the model's compression-decompression of it, and whether
different model families preserve different aspects (anchors vs
predictions, structural vs stylistic) of the same input.

\textbf{Spec component utilization at runtime.} Component ablation tests
which Spec components contribute to lift; component utilization tests
which components a response model actually grounds in. Activation-level
techniques (natural-language activation analysis as developed in current
Anthropic interpretability work; persona-vector analysis per Chen et
al.~2025) applied to (Spec, response, ground-truth) tuples could measure
whether the model's response activates the same Spec sections the rubric
scores as load-bearing. The measurement bridges representation-design
(what the Spec encodes) and response-time use (what the model retrieves
and applies).

\hypertarget{production-serving-and-infrastructure}{%
\subsection{7.4 Production serving and
infrastructure}\label{production-serving-and-infrastructure}}

The study served the Behavioral Specification statically and in full on
every query. Four production-realistic serving-layer follow-ups would
refine deployment beyond this static-attachment baseline: dynamic
activation (selecting which parts of the Spec to attach per query), user
editing and inspection (how a user can update or correct their own Spec
post-authoring), temporality handling (how the Spec ages and what
triggers re-authoring; \hyperref[stateful-agent-implementations-and-temporal-drift-tracking]{§7.5}), and topic decomposition (whether the Spec
can be partitioned by domain for selective serving). Each is a
measurement question in its own right: whether the gradient, mechanism,
and composition findings hold under each production serving strategy.

\textbf{Layered-context stacking studies.} This paper tests
Spec-on-No-Context, Spec-on-facts, Spec-on-corpus, and
Spec-on-memory-system separately. A stacking study would systematically
vary the layered context: memory-system retrieval at increasing K, with
and without facts, with and without Spec, with both. The output is a map
of which layer combinations produce additive lift, which produce
subadditive interaction, and which produce antagonistic effects. The
study format extends the \hyperref[memory-system-composition]{§4.4} memory-system layering analysis by
treating context type as a factorial design rather than a
between-condition comparison.

\hypertarget{stateful-agent-implementations-and-temporal-drift-tracking}{%
\subsection{7.5 Stateful-agent implementations and temporal drift
tracking}\label{stateful-agent-implementations-and-temporal-drift-tracking}}

Several follow-ups sit adjacent to the paper's static-snapshot design.

\textbf{Stateful-agent variant of the Behavioral Specification.} The
pipeline as tested is offline and batch. A persistent, self-editing
variant that ingests new source material as it arrives, re-edits anchors
and predictions in place, and maintains version history with provenance
across edits is a natural next step. The \hyperref[exploratory-case-study-letta-stateful-agent-n3-post-hoc]{§4.5} Letta exploration (N=3,
post-hoc) is one data point on an adjacent architecture; building and
evaluating a stateful-agent Base Layer implementation on the full
14-subject main-study battery would close the comparison within a single
architectural family and extend \hyperref[exploratory-case-study-letta-stateful-agent-n3-post-hoc]{§4.5} to a layered-stack rerun against
Letta at full scope.

\textbf{Cleaner \hyperref[exploratory-case-study-letta-stateful-agent-n3-post-hoc]{§4.5} rerun with naming and scaling controls.} Two
specific extensions of the \hyperref[exploratory-case-study-letta-stateful-agent-n3-post-hoc]{§4.5} exploration are worth running as a unit.
First, anonymize the source corpus before Letta ingestion so Letta
writes an anonymized memory block, matching Base Layer's
anonymized-during-authoring convention; the \hyperref[exploratory-case-study-letta-stateful-agent-n3-post-hoc]{§4.5} naming asymmetry (Letta
ingests named corpus, Base Layer strips and later restores names) is
removed as a confounder. Second, extend the corpus-size axis past the
Bābur ceiling to a larger (\textgreater250K-word) subject corpus that
pushes the Letta block past its character ceiling. The matched-model gap
was small at Hamerton (25K-word corpus), largest at Ebers (48K-word
corpus), and smaller again at Bābur (223K-word corpus); whether that
pattern continues, re-widens, or flattens at extreme corpus size is the
empirical question. Both extensions together would turn \hyperref[exploratory-case-study-letta-stateful-agent-n3-post-hoc]{§4.5}'s case
study into a controlled comparison.

\textbf{Temporal drift tracking.} The static snapshot tested here is a
point on a trajectory. A Specification authored at one time, compared
against a later specification on an expanded corpus, produces a
measurable diff: which anchors appear or disappear, which predictive
templates shift, which axioms strengthen or weaken. From a sequence of
past specifications, the trajectory predicts the next. The 14-subject
corpora collected for this study can be back-sliced by chapter
boundaries or publication era for an initial drift test within the
current sample. A purpose-built companion study on sequential public
records (US Supreme Court opinions, shareholder letters, research
papers) is the natural extension.

\textbf{Canonical life events.} Discrete pivots that flip reasoning
architecture (a major career change, a religious conversion, a
significant loss, a public stance reversal) are distinct from gradual
drift. The main-study autobiographies were not structured to test this
case. A snapshot specification authored before such an event predicts
pre-event reasoning, not post-event reasoning, even though the person's
underlying patterns have materially shifted. Whether to detect these
events automatically, allow user annotation, or maintain period-specific
specifications is an open production-deployment question.

\textbf{Continuous-representation infrastructure.} Both of the above
converge on the same engineering target: a background process that
watches incoming corpus material, re-authors the Specification as new
material arrives, and emits drift telemetry as a first-class output
(what changed in the Spec, by how much, and against what baseline). The
Letta-style stateful agent (\hyperref[exploratory-case-study-letta-stateful-agent-n3-post-hoc]{§4.5}) and the sequential-checkpoint test
design are complementary tests for this kind of implementation: the
first isolates online self-editing as a way to produce the
representation, the second isolates temporality as a property of the
representation itself. As a downstream effect, the daemon's continuous
output (the sequence of Spec versions, diffs, and drift telemetry)
itself becomes a training corpus for next-generation pipeline
development.

Additional architectural paths worth testing against the same target,
beyond stateful-agent and drift-tracking variants, include agent-edited
persistent memories outside the MemGPT family, fine-tuned per-user
models that expose their internal representation for audit, and hybrid
architectures that combine offline-extracted specifications with online
self-editing.

\textbf{Per-user feedback as a learning signal for the Specification.}
Every interaction with an AI agent can produce a small correction
signal: the user edits a response, says ``actually I would have done
X'', or rejects an answer outright. Each of these is a hint that the
current specification mispredicted the user's behavior. Used as a
feedback signal, these corrections can update the Specification
directly: the affected predicate or anchor is re-extracted and
re-authored, then re-composed into the Spec. Because the update unit is
structured text rather than model weights, the change is interpretable
and the per-update cost stays at the \hyperref[pipeline-for-the-behavioral-specification]{§3.7} pipeline cost. Two design
questions matter. First, drift risk: corrections without a source-text
anchor can pull the Spec in arbitrary directions, so the corpus stays
the source of truth and corrections are bounded by it. Second, signal
quality: explicit corrections are sparse but high-fidelity, while
everyday divergence is abundant but noisy; treating both as the same
kind of feedback is wrong. A first-pass design using explicit
corrections only, batched daily, with versioned diffs logged for
rollback, is the cheapest test of whether correction-driven updates
produce measurable accuracy gains beyond the static snapshot.

\hypertarget{safety-alignment-integration}{%
\subsection{7.6 Safety-alignment
integration}\label{safety-alignment-integration}}

The positioning argument (per-user calibration as redirection of
existing shaping rather than additional bias, and as orthogonal to
safety alignment) is in \hyperref[privacy-and-the-case-for-user-held-representation]{§5.7}. Two concrete follow-ups extend that
positioning. First, the Spec-induced refusal cases (\hyperref[mechanism-correct-content-not-format]{§4.3}, \hyperref[rubric-handling-limitations-post-hoc-validity-audit-1]{§4.6.7}): a
post-hoc classifier audit of 81 Spec-induced refusals across 5 memory
systems (\texttt{docs/\allowbreak{}research/\allowbreak{}refusal\_\allowbreak{}intent\_\allowbreak{}classification.\allowbreak{}md})
found 75 of 81 (93\%) were routine behavioral prediction rather than
morally loaded. The refusal pattern is general caution when information
is thin, not a moral-integrity mechanism. Whether it composes cleanly
with existing safety frameworks across benign and malicious user types
is open. Second, the specifications in this paper were authored from
public-domain autobiographies of subjects not selected on intent. What a
Specification for a user with malicious intent would contain, and what
happens when an agent is deployed on that user's behalf, is untested.
Both belong to collaboration with AI safety researchers rather than
single-lab follow-ups.

A narrower follow-up: the wrong-Spec adversarial control (\hyperref[mechanism-correct-content-not-format]{§4.3}) showed
response models can recognize Spec content as belonging to a different
historical period or persona. Whether models can recognize
specifications encoding adversarial values (personas that endorse
harmful behaviors) and flag them rather than comply is a direct
extension that bears on live-user deployment.

\begin{center}\rule{0.5\linewidth}{0.5pt}\end{center}

\hypertarget{data-code-and-reproducibility}{%
\section{8. Data, code, and
reproducibility}\label{data-code-and-reproducibility}}

\textbf{Data availability.} All raw response files, per-judge judgments,
batteries, and aggregated results for the 14 main-study subjects are in
the public study repository at
\texttt{github.\allowbreak{}com/\allowbreak{}agulaya24/\allowbreak{}beyond-recall} under
\texttt{results/global\_\textless{}subject\textgreater{}/} and
\texttt{results/hamerton/}. The same data is also published as a
dataset at
\texttt{https:/\allowbreak{}/\allowbreak{}huggingface.\allowbreak{}co/\allowbreak{}datasets/\allowbreak{}agulaya24/\allowbreak{}beyond-recall}.
Source autobiographies are public domain
(Project Gutenberg and Internet Archive). Per-subject Project Gutenberg
IDs are listed in \hyperref[subjects]{§3.4} Table 3.2. Memory-system raw retrieval and
ingestion logs are at
\texttt{results/global\_\textless{}subject\textgreater{}/\textless{}system\textgreater{}\_*.json}.
The Letta stateful-agent matched-rerun artifacts are at
\texttt{docs/\allowbreak{}research/\allowbreak{}\_\allowbreak{}letta\_\allowbreak{}rerun/}. The full-stack Letta rerun
comparison is at
\texttt{docs/\allowbreak{}research/\allowbreak{}\_\allowbreak{}letta\_\allowbreak{}rerun/\allowbreak{}fullstack\_\allowbreak{}named/}.

\textbf{Code availability.} The pipeline that authored the 14
specifications scored in \hyperref[results]{§4} is preserved as a frozen
copy of Base Layer v0.2.0 in the study repository at
\texttt{baselayer/}, installable with \texttt{pip install -e .\allowbreak{}/\allowbreak{}baselayer}.
That vendored copy, not the active repository, is what reproduces the
results in this paper. Active development continues at
\texttt{https:/\allowbreak{}/\allowbreak{}github.\allowbreak{}com/\allowbreak{}agulaya24/\allowbreak{}BaseLayer}, which has moved past
v0.2.0 and no longer matches the pipeline described here. The
study-specific analysis and
re-run scripts are at
\texttt{https:/\allowbreak{}/\allowbreak{}github.\allowbreak{}com/\allowbreak{}agulaya24/\allowbreak{}beyond-recall} under
\texttt{scripts/}. Reproducibility pointers from each numerical claim to
its supporting script are in \texttt{docs/PROVENANCE\_INDEX.md} and
\texttt{docs/DATA\_REFERENCE.md}. The \hyperref[the-cross-subject-gradient-and-its-per-question-mechanism]{§4.1} battery-composition
sensitivity analysis is reproducible via
\texttt{scripts/\allowbreak{}\_\allowbreak{}v10\_\allowbreak{}battery\_\allowbreak{}sensitivity.\allowbreak{}py}. The \hyperref[rubric-handling-limitations-post-hoc-validity-audit]{§3.3.6}
rubric-handling validity audit is reproducible via
\texttt{scripts/\allowbreak{}audit\_\allowbreak{}low\_\allowbreak{}end\_\allowbreak{}inflation.\allowbreak{}py}. The \hyperref[mechanism-correct-content-not-format]{§4.3} hedging
classifier is at \texttt{scripts/\allowbreak{}classify\_\allowbreak{}hedging.\allowbreak{}py}.

\textbf{Supplementary materials.} Per-subject worked examples (Appendix
E in the public-repository version) and per-subject paired-delta tables
(Appendix B subsections B.2 and B.3 per-subject distribution) live at
\href{https://github.com/agulaya24/beyond-recall/tree/master/docs/supplementary}{docs/supplementary/}
in the repository. They are cross-referenced from main-paper sections at
the indicated pointers and are reproducible from the same data and
scripts that produced the in-paper analyses.

\textbf{Agent-friendly study repo tooling.} The study repository is
structured for both human reading and agent consumption. A combined
SQLite + ChromaDB knowledge index
(\texttt{workspace/\allowbreak{}study\_\allowbreak{}knowledge.\allowbreak{}db},
\texttt{workspace/study\_vectors/}; built by
\texttt{scripts/\allowbreak{}index\_\allowbreak{}study\_\allowbreak{}repo.\allowbreak{}py}) covers 206 files and 3,702
chunks across the paper, supporting docs, per-subject specs, judgments,
retrieval logs, and analysis scripts. An MCP server exposing this index
plus typed lookups (per-subject score retrieval, claim provenance,
condition-pair anchor-crossing queries) is available as an MVP at
\texttt{mcp/} in the same repository; the decision to optimize the repo for
agent consumption is based on the observation that the same artifacts
that make a paper easy to verify mechanically (stable anchors,
structured judgments, machine-readable schemas) also make the paper
easier for human readers to skim and verify.

\textbf{Compute and cost.} All response generation and judging used
commercial APIs (Anthropic, OpenAI, Google) at standard rates. No
specialized hardware was used. All experiments are runnable on a
standard developer laptop.

\textbf{Author affiliation.} Aarik Gulaya, Base Layer. ORCID:
\texttt{0009-0009-5902-9557}. Contact: \texttt{aarik@base-layer.ai}.
Project page: \texttt{base-layer.ai}. Base Layer pipeline source:
\texttt{https:/\allowbreak{}/\allowbreak{}github.\allowbreak{}com/\allowbreak{}agulaya24/\allowbreak{}BaseLayer}.

\textbf{Funding.} This work was self-funded.

\textbf{Conflicts of interest.} The author is the founder of Base Layer,
the project that develops the Behavioral Specification pipeline this
paper evaluates. Memory-system providers tested in this paper (Mem0,
Letta, Supermemory, Zep) were used through their public APIs at standard
rates; no provider was given preferential framing, and Base Layer does
not have commercial relationships with any of them. Self-reported
benchmark scores from each provider are reported as published; this
paper does not adjudicate disputes between providers' published claims
(\hyperref[memory-systems-for-llm-agents]{§2.2}).

\textbf{License.} Apache 2.0 for code and Creative Commons Attribution
4.0 for the manuscript and data analyses produced by this study. Source
autobiographies are in the public domain.

\textbf{Acknowledgments.}

This research was self-funded. No external funding, grants, or
institutional support was received.

The author is grateful to the research community whose work made this
study possible. The teams behind LongMemEval, AlpsBench, PersonaGym,
Twin-2K, and LoCoMo built the benchmarks that revealed the gap between
recall and understanding. The teams behind Mem0, Letta, Supermemory, and
Zep built the memory systems that this work seeks to complement, not
replace. The work of Chen et al.~on persona vectors, Jiang et al.~on
dynamic user profiling, and Betley et al.~on emergent misalignment
shaped the framing of this study directly. This paper is standing on the
shoulders of their work, and their approaches helped clarify what this
approach is and what it is not. Conversations with the broader
memory-systems and AI-personalization research communities further
informed the design of this paper. Specific gratitude goes to the
cross-LLM reviewer panels (Gemini 2.5 Pro, Mistral Large, Cerebras Qwen3
235B, Groq Llama 3.3 70B, GPT-5.5) whose iterated reviews materially
improved earlier drafts. All errors are the author's.

The author thanks his wife Bavani, his mother, Walnut, and Wasabi for
their unwavering support, which allowed this work to come into
existence.

\begin{center}\rule{0.5\linewidth}{0.5pt}\end{center}

\hypertarget{references}{%
\section{9. References}\label{references}}

\bibliographystyle{plainnat}
\bibliography{beyond_recall}

\hypertarget{appendix-a.-predicate-vocabulary}{%
\section{Appendix A. Predicate
Vocabulary}\label{appendix-a.-predicate-vocabulary}}

\hypertarget{a.1-the-46-constrained-predicates-redirected}{%
\subsection{A.1 The 46 Constrained Predicates
(redirected)}\label{a.1-the-46-constrained-predicates-redirected}}

The full controlled-vocabulary spec (46 predicates organized into
Behavioral / Identity / Knowledge / Procedural / Relational / Temporal /
Attentional groups, each with a verbatim definition, example triples,
and authoring constraints) is in the public repository at
\texttt{docs/\allowbreak{}supplementary/\allowbreak{}appendix\_\allowbreak{}A\_\allowbreak{}1\_\allowbreak{}predicate\_\allowbreak{}vocabulary.\allowbreak{}md} and
in source form at \texttt{data/\allowbreak{}predicates/\allowbreak{}predicate\_\allowbreak{}vocabulary.\allowbreak{}md}.

\textbf{What \hyperref[pipeline-for-the-behavioral-specification]{§3.7} already establishes:} the predicate vocabulary is a
fixed, audit-friendly set used by the extraction step to convert raw
corpus text into structured (subject, predicate, object) triples. The
vocabulary is closed by design (extraction cannot invent new predicates)
so that downstream layers can be authored against a stable typed
substrate. Predicate definitions, example triples, AUDN (Add / Update /
Delete / NOOP) operation rules, and the three behavioral-axis tags
(LITERAL / INTERPRETIVE / REFUSAL-TRIGGERING) used by the
question-battery formation pipeline (\hyperref[question-battery-formation]{§3.5}, Appendix B.3) are documented
in the repository file.

\hypertarget{a.2-provenance-and-design-choices}{%
\subsection{A.2 Provenance and design
choices}\label{a.2-provenance-and-design-choices}}

The vocabulary was iterated in three stages. The initial 30-predicate
list (sessions 1-48) favored values, activities, and biography. Session
49 added 6 predicates (\texttt{unknown}, \texttt{attended},
\texttt{interested\_in}, \texttt{wants\_to}, \texttt{loves},
\texttt{hates}) to preserve semantic distinctions that the initial
vocabulary collapsed. Session 52 added \texttt{plays} and
\texttt{monitors}. Session 55 added 8 relationship predicates to raise
relationship-fact extraction from 0.8\% to the 3 to 5\% target range.

The vocabulary is deliberately behavioral rather than biographical. The
ratio of predicates in the behavioral-patterns plus values-beliefs plus
emotions-dispositions groups (23 of 46) to the biographical-context
group (7 of 46) encodes a design decision: extraction is steered away
from facts that are easily verifiable in external sources (city of
birth, schools attended) and toward patterns that require reading the
source text to infer (what the subject avoids, prefers, values, fears).

\hypertarget{a.3-not-in-the-vocabulary}{%
\subsection{A.3 Not in the vocabulary}\label{a.3-not-in-the-vocabulary}}

Three predicate categories that commonly appear in general-purpose
knowledge graphs are deliberately excluded:

\begin{itemize}
\tightlist
\item
  \textbf{Evaluative predicates about the subject from a third party}
  (for example, \texttt{considered\_brilliant\_by}). These invert the
  direction of claim: the subject is the object rather than the source
  of the reasoning.
\item
  \textbf{Time-indexed state changes} (for example, \texttt{became}).
  The vocabulary handles change-over-time through the AUDN ADD / UPDATE
  / DELETE / NOOP operations at the fact level, not through predicate
  selection.
\item
  \textbf{Causal predicates} (for example, \texttt{caused},
  \texttt{triggered}). Causal inference is produced in the authored
  layers (predictions, anchors) from collections of facts, not encoded
  at the extraction step.
\end{itemize}

\hypertarget{a.4-live-deployment}{%
\subsection{A.4 Live deployment}\label{a.4-live-deployment}}

A live web deployment of the pipeline described in \hyperref[pipeline-for-the-behavioral-specification]{§3.7}, with served
briefs across additional subjects beyond the 14 in this study, is
available at base-layer.ai for readers interested in seeing the
served-specification format in interactive form.

\begin{center}\rule{0.5\linewidth}{0.5pt}\end{center}

\hypertarget{appendix-b.-question-batteries}{%
\section{Appendix B. Question
Batteries}\label{appendix-b.-question-batteries}}

\hypertarget{b.1-the-10-fixed-behavioral-prediction-categories}{%
\subsection{B.1 The 10 fixed behavioral-prediction
categories}\label{b.1-the-10-fixed-behavioral-prediction-categories}}

Every behavioral-prediction question in the study is tagged with exactly
one of ten fixed categories. The category set is identical across all 15
batteries (14 main-study plus Franklin). Each category is a behavioral
dimension the question probes; the category does not constrain the
answer format.

\begin{longtable}[]{@{}
  >{\raggedright\arraybackslash}p{(\columnwidth - 4\tabcolsep) * \real{0.33}}
  >{\raggedright\arraybackslash}p{(\columnwidth - 4\tabcolsep) * \real{0.33}}
  >{\raggedright\arraybackslash}p{(\columnwidth - 4\tabcolsep) * \real{0.33}}@{}}
\toprule
Category & What it probes & Example question \\
\midrule
\endhead
\texttt{decisions} & How the subject resolves concrete choices. & ``When
his uncle's family emigrates to New Zealand, would Hamerton consider
joining them?'' \\
\texttt{values} & What the subject holds as important when forced to
rank. & ``When confronted with German Neology, would Hamerton accept or
reject the established Protestant position?'' \\
\texttt{relationships} & How the subject engages with specific people or
classes of people. & ``How would Hamerton's religious heterodoxy affect
his social standing among the Lancashire gentry?'' \\
\texttt{conflict} & How the subject responds when a line is crossed. &
``If his tutor attempted to physically harass him, would Hamerton submit
or resist?'' \\
\texttt{learning} & How the subject acquires skills, knowledge, or
lessons. & ``Given Hamerton's difficulty following spoken French, what
would he do about it?'' \\
\texttt{risk} & How the subject handles uncertainty, exposure, or
irreversibility. & ``Would Hamerton choose to encamp alone on remote
Scottish moors, despite it being considered eccentric?'' \\
\texttt{creativity} & How the subject produces or evaluates creative
work. & ``Would Hamerton publish his early poetry at his own expense,
and what would the commercial result be?'' \\
\texttt{stress} & How the subject responds to pressure, exposure, or
failure. & ``When offered a grand opportunity to organize an exhibition,
would Hamerton accept?'' \\
\texttt{career} & How the subject makes professional trajectory choices.
& ``Would Hamerton follow Ruskin's advice to study nature directly
rather than learn from traditional masters?'' \\
\texttt{change\_over\_time} & How the subject shifts or persists across
life phases. & ``After his poetry failed commercially, would Hamerton
continue writing verse?'' \\
\bottomrule
\end{longtable}

Each subject's battery covers 8 to 10 of these categories; no battery
skips more than 2. The distribution within a subject reflects what the
training half of that subject's corpus naturally supported, not a quota.

\hypertarget{b.2-per-subject-battery-composition-10-category-by-15-subject-matrix}{%
\subsection{B.2 Per-subject battery composition (10-category by
15-subject
matrix)}\label{b.2-per-subject-battery-composition-10-category-by-15-subject-matrix}}

The per-subject battery-composition table (15 subject rows by 11
columns: ten behavioral-prediction categories plus a per-subject total)
is in
\texttt{docs/\allowbreak{}supplementary/\allowbreak{}appendix\_\allowbreak{}B\_\allowbreak{}per\_\allowbreak{}subject\_\allowbreak{}paired\_\allowbreak{}delta\_\allowbreak{}tables.\allowbreak{}md}
for length reasons; the headline composition facts are summarized here.

Each subject's battery covers 8 to 10 of the ten categories listed in
§B.1. Per-subject totals are 39 questions for the 13 global subjects and
Hamerton, and 40 for Franklin.

Column totals across all 15 subjects:

\begin{landscape}
\begin{longtable}[]{@{}lr@{}}
\toprule
Category & Questions \\
\midrule
\endhead
Values & 115 \\
Decisions & 93 \\
Relationships & 84 \\
Conflict & 66 \\
Learning & 65 \\
Stress & 60 \\
Creativity & 34 \\
Change over time & 27 \\
Career & 23 \\
Risk & 19 \\
\textbf{Total} & \textbf{586} \\
\bottomrule
\end{longtable}
\end{landscape}

Raw batteries are at
\texttt{results/global\_\textless{}subject\textgreater{}/battery\_v2.json}
(global subjects), \texttt{data/\allowbreak{}hamerton/\allowbreak{}battery.\allowbreak{}json} (Hamerton), and
\texttt{data/\allowbreak{}franklin/\allowbreak{}battery.\allowbreak{}json} (Franklin), counted over
\texttt{tier\ \allowbreak{}==\ \allowbreak{}"behavioral\_\allowbreak{}prediction"} slices.

\hypertarget{b.3-behavioral-axis-distribution-literal-interpretive-refusal-triggering}{%
\subsection{B.3 Behavioral-axis distribution (Literal / Interpretive /
Refusal-Triggering)}\label{b.3-behavioral-axis-distribution-literal-interpretive-refusal-triggering}}

A secondary classification of all 586 audited questions (the 14
main-study batteries totaling 546 plus the Franklin reference battery of
40), produced by Claude Haiku 4.5 as a post-hoc auditor, tags each
question by what cognitive operation it requires the response model to
perform. Full audit at
\texttt{docs/\allowbreak{}research/\allowbreak{}question\_\allowbreak{}category\_\allowbreak{}audit.\allowbreak{}md}.

Aggregate distribution:

\begin{landscape}
\begin{longtable}[]{@{}lrr@{}}
\toprule
Axis & n & \% \\
\midrule
\endhead
Literal Recall & 60 & 10.2\% \\
Interpretive Inference & 403 & 68.8\% \\
Refusal-Triggering & 123 & 21.0\% \\
\bottomrule
\end{longtable}
\end{landscape}

Per-subject distribution: the 15-row per-subject behavioral-axis
breakdown (Literal / Interpretive / Refusal counts per subject) is in
\texttt{docs/\allowbreak{}supplementary/\allowbreak{}appendix\_\allowbreak{}B\_\allowbreak{}per\_\allowbreak{}subject\_\allowbreak{}paired\_\allowbreak{}delta\_\allowbreak{}tables.\allowbreak{}md}
for length reasons; the aggregate distribution above is the summary
referenced by \hyperref[results]{§4} claims. Notable points from the per-subject table:
Hamerton's battery is the most refusal-heavy of the 14 main-study
subjects (19 of 39); Franklin's is the most interpretive (37 of 40, with
0 Literal Recall); the 13 global subjects cluster in the 22 to 33
Interpretive range with refusal counts between 2 and 15. Source:
\texttt{docs/\allowbreak{}research/\allowbreak{}question\_\allowbreak{}category\_\allowbreak{}audit.\allowbreak{}md}.

\hypertarget{b.4-category-level-effect-size-on-ux3b4_spec}{%
\subsection{B.4 Category-level effect size on
Δ\_spec}\label{b.4-category-level-effect-size-on-ux3b4_spec}}

Mean Δ\_spec (C2a minus C5) broken down by the behavioral-axis
classification. Counts are on the N=14 main-study subset (total 546
questions); the Franklin reference battery (40 additional questions,
included in B.3 above and the \hyperref[the-gradient-at-the-high-baseline-end-franklin-reference]{§4.1.2} reference) is excluded from
category-level effect-size computation because Franklin is excluded from
every N=14 inferential statistic (\hyperref[the-gradient-at-the-high-baseline-end-franklin-reference]{§4.1.2}). Source:
\texttt{docs/\allowbreak{}research/\allowbreak{}question\_\allowbreak{}category\_\allowbreak{}audit.\allowbreak{}md}.

\begin{landscape}
\begin{longtable}[]{@{}lrrr@{}}
\toprule
Axis & n & Mean Δ\_spec & Median Δ\_spec \\
\midrule
\endhead
Literal Recall & 60 & +0.792 & +0.800 \\
Interpretive Inference & 366 & +0.397 & +0.400 \\
Refusal-Triggering & 120 & +0.417 & +0.200 \\
\bottomrule
\end{longtable}
\end{landscape}

The Literal Recall bucket has the largest aggregate Δ\_spec, but the
magnitude is mechanically explained by the C5 baseline distribution:
66.7\% of Literal Recall questions sit at C5 ≤ 1.5 (rubric floor),
compared to 48.6\% of Interpretive Inference questions, and median C5
for Literal Recall is 1.00. The Spec moves these floor-anchored
responses from abstention or off-base to substantive prediction,
producing large per-question rubric-point gains by construction. When
restricted to the 9 low-baseline subjects, Literal Recall Δ\_spec is
+1.12 and Interpretive Inference is +0.67; when restricted to the 5
mid-baseline subjects, both shrink to near zero. The Literal Recall
\textgreater{} Interpretive Inference ordering survives this baseline
control, but the absolute magnitude is dominated by C5 floor effects
rather than by question-type-specific Spec advantage. The
interpretation-heavy emphasis elsewhere in this paper (\hyperref[what-we-found]{§1.3}, \hyperref[where-the-spec-helps-where-it-hurts-and-which-question-types-route-to-each]{§4.4.3}, \hyperref[discussion]{§5})
is not that Interpretive Inference questions yield the largest raw Δ ---
they do not --- but that the Spec's distinctive contribution is on
Interpretive Inference questions, where retrieval-only systems hit a
ceiling. Retrieval can surface the relevant fact and lift a Literal
Recall question; it cannot supply the interpretive pattern an
Interpretive Inference question requires. The per-axis aggregate is
conservative because it averages over questions where the baseline
already produces a substantive answer; the distinctive Spec contribution
is at the per-question mechanism level (\hyperref[where-the-spec-helps-where-it-hurts-and-which-question-types-route-to-each]{§4.4.3} Pattern 1), not at the
per-axis raw mean.

\hypertarget{b.5-per-subject-by-axis-ux3b4_spec}{%
\subsection{B.5 Per-subject by axis
Δ\_spec}\label{b.5-per-subject-by-axis-ux3b4_spec}}

Summary of the cross-subject pattern under 5-judge primary aggregation:
the strongest positive Spec effects cluster on three subjects. Hamerton
(Literal +1.68, Interpretive +1.30, Refusal +1.25), Sunity Devee (+1.38
/ +1.16 / +1.35), and Bernal Díaz (+2.00 / +0.44 / +0.64) carry the
largest gains. Augustine, Equiano, and Zitkala-Ša show negative or
near-zero deltas across all three axes, consistent with their status as
mid-baseline subjects on the \hyperref[the-cross-subject-gradient-and-its-per-question-mechanism]{§4.1} gradient. Fukuzawa and Seacole show
their largest positive effects on Interpretive Inference specifically
(+0.83 and +0.79). Bernal Díaz's +2.00 on Literal Recall is computed
over a small per-axis n (single-digit questions per axis per subject),
so the per-subject estimate is high-variance; the cross-subject
correlation pattern in §B.6 is the more robust signal.

Full breakdown at \texttt{docs/\allowbreak{}research/\allowbreak{}question\_\allowbreak{}category\_\allowbreak{}audit.\allowbreak{}md};
per-subject axis-Δ scaffold values at
\texttt{docs/\allowbreak{}research/\allowbreak{}v11\_\allowbreak{}emit/\allowbreak{}appendix\_\allowbreak{}b\_\allowbreak{}battery.\allowbreak{}json} (claim ids
\texttt{appB\_5\_\textless{}subject\textgreater{}\_\textless{}axis\textgreater{}\_delta}).

\hypertarget{b.6-battery-composition-sensitivity}{%
\subsection{B.6 Battery-composition
sensitivity}\label{b.6-battery-composition-sensitivity}}

This appendix provides the technical detail behind the \hyperref[the-cross-subject-gradient-and-its-per-question-mechanism]{§4.1}
battery-sensitivity controls.

\textbf{B.6.1 Battery-question-type correlations.}

Across the 14 main-study subjects:

\begin{itemize}
\tightlist
\item
  Δ\_spec range: −0.31 to +1.37
\item
  Corr(fraction of Literal Recall questions, subject-level Δ\_spec): r =
  +0.595 (recomputed under strict 5-judge primary; the legacy audit-doc
  value was +0.646, which used a Hamerton-divergent intermediate
  aggregation)
\item
  Corr(fraction of Interpretive Inference questions, subject-level
  Δ\_spec): r = −0.466 (legacy: −0.582, same caveat)
\item
  Corr(fraction of Refusal-Triggering questions, subject-level Δ\_spec):
  r = +0.212
\end{itemize}

The positive Literal Recall correlation and negative Interpretive
correlation imply that subjects whose batteries over-weight literal
recall also produce larger measured Δ\_spec values.

\textbf{B.6.2 Multiple regression controlling for Literal Recall
fraction.}

A multiple regression of Δ\_C4a on both C5 baseline and Literal Recall
fraction across the 14 main-study subjects yields a partial coefficient
on baseline of \textbf{−0.88 {[}95\% CI −1.13, −0.63{]}, p \textless{}
10⁻⁵}, attenuated from the univariate −0.96 by about 8\%. Literal Recall
fraction enters as a significant partial predictor (β = +2.30 {[}+0.34,
+4.26{]}, p = 0.026), but baseline carries the bulk of the explained
variance: 63.6\% uniquely attributable to C5, 6.9\% uniquely
attributable to Literal Recall fraction. The two predictors are not
collinear (Pearson r = −0.28, VIF = 1.08 for both), so the partial
coefficients are stable. Adjusted R² rises from 0.80 to 0.87 when
Literal Recall fraction is added; the controls are additive rather than
redundant. The gradient on baseline survives; it is not an artifact of
battery composition.

\textbf{B.6.3 Hamerton-leverage subset regression.}

Hamerton's 80-question battery predates the global-subject pipeline and
uses a slightly different backward-design path (the legacy Haiku 4.5
generator that originally produced Franklin and Hamerton); the 13 global
subjects' main-study batteries also use Claude Haiku 4.5 but were
regenerated by \texttt{run\_global\_rerun.py} against a uniform prompt
template. All 14 main-study batteries share the same generator family. A
subset regression dropping Hamerton (N=13 globals) yields a slope of
\textbf{−0.89 {[}95\% CI −1.18, −0.61{]}, R² = 0.81, p \textless{}
10⁻⁴}, compared to the full-sample −0.96. The point estimate attenuates
by about 7\%, and the 95\% CIs overlap substantially. The gradient is
not Hamerton-driven. A separate GPT-5.4-regenerated battery set
(\texttt{results/global\_\textless{}subject\textgreater{}/battery\_gpt54.json})
exists for each global as a circularity control; its results are
reported in \hyperref[circularity-controls]{§3.5.1} and \hyperref[cross-provider-response-generation-tier-2-replication]{§4.6.1}, not folded back into the \hyperref[the-cross-subject-gradient-and-its-per-question-mechanism]{§4.1} gradient
itself.

\textbf{B.6.4 Discussion.}

This is a battery-composition confound in the cross-subject gradient;
the paper's gradient claim is therefore specifically about mean score
movement per subject, not about mean score movement per category. \hyperref[retrieval-is-not-interpretation]{§5.3}
and \hyperref[future-work]{§7} flag a follow-up study with a category-balanced battery as the
primary design improvement for future gradient work.

\textbf{B.6.5 Hamerton-leverage at the per-question grain.}

The B.6.3 subset regression checks the per-subject mean grain. A
parallel question is whether the per-question extreme upward anchor
crossings catalogued in
\texttt{docs/\allowbreak{}research/\allowbreak{}wins\_\allowbreak{}inventory\_\allowbreak{}20260428.\allowbreak{}json} (60 unique cases
across 18 condition pairs, 351 paired low-baseline questions on C5 to
C4a) are concentrated on Hamerton specifically. They are: Hamerton
accounts for 15 of the 60 unique extreme jumps (25\%) on a battery of 80
questions (18.75\% extreme-jump rate); the other 13 subjects average
8.9\% extreme-jump rate across 39-question batteries.

Hamerton's elevation is real but its cause is not isolated by the
present design. Two candidate mechanisms (the legacy battery-generator
path and the subject's thin pretraining coverage, C5 = 1.26, the lowest
in the main study) are not separately identifiable. A third candidate
considered earlier, behavioral-predicate density per word, does not
survive inspection: Hamerton's served Specification is the full
four-layer stack (anchors + core + predictions + brief,
\textasciitilde5,334 words), the same structure and size as the globals'
\textasciitilde5,300-word specifications, so Spec length does not
differentiate Hamerton from the rest of the sample.

As measured by the heuristic classifier, the mechanism distribution on
Hamerton's 15 jumps versus globals' 45 is nearly identical
(PATTERN\_PREDICATE+HYBRID share: Hamerton 73.3\% vs globals 80.0\%);
the heuristic does not discriminate jumps from non-jumping Spec-loaded
controls
(\texttt{docs/\allowbreak{}research/\allowbreak{}pattern\_\allowbreak{}activation\_\allowbreak{}deep\_\allowbreak{}20260428.\allowbreak{}md}), so this
near-identity is consistent with the heuristic detecting Spec-loaded
response style rather than the lift mechanism.

\hypertarget{b.7-coupling-free-reframing-of-the-gradient}{%
\subsection{B.7 Coupling-free reframing of the
gradient}\label{b.7-coupling-free-reframing-of-the-gradient}}

The headline slope regresses Δ\_C4a = C4a − C5 on C5, which mechanically
embeds a −1 component when C4a is bounded on the 1-5 scale and partially
independent of C5. To triangulate from a non-coupling-prone angle, we
ran three additional checks on the same per-subject (C5, C4a) data
(script: \texttt{scripts/\allowbreak{}\_\allowbreak{}v10\_\allowbreak{}coupling\_\allowbreak{}sensitivity.\allowbreak{}py}; full output:
\texttt{docs/\allowbreak{}research/\allowbreak{}v10\_\allowbreak{}coupling\_\allowbreak{}sensitivity\_\allowbreak{}analysis.\allowbreak{}md}).

\textbf{B.7.1 Level regression.}

The level regression C4a \textasciitilde{} C5 produces a slope of
\textbf{+0.04} {[}95\% CI −0.24, +0.33{]}, R² = 0.008, p = 0.76. C4a is
essentially flat across the C5 range of 1.02-2.77 and clusters tightly
around its mean of \textbf{2.44} at the per-subject grain.\footnote{Per-subject
  mean C4a across the N=14 main-study subjects is \textbf{2.44},
  computed as the cross-subject mean of the 2-decimal per-subject C4a
  values in the \hyperref[the-cross-subject-gradient-and-its-per-question-mechanism]{§4.1} / Appendix D.1 table (34.15 / 14 = 2.4393). A
  reader recomputing from the published table reproduces this figure
  exactly. Full data-lock note in \texttt{docs/\allowbreak{}research/\allowbreak{}DATA\_\allowbreak{}LOCKS.\allowbreak{}md}.}
The Spec does not differentially ``lift'' low-baseline subjects more
than high-baseline ones in any treatment-effect-heterogeneity sense; it
produces a roughly constant post-Spec C4a mean per subject regardless of
baseline, and the apparent Δ-on-C5 gradient equals the baseline
shortfall.

\textbf{B.7.2 Permutation test.}

A 10,000-iteration permutation test that shuffles C4a across subjects
(preserving the bounded marginal but breaking any link to C5) yields a
null distribution for the Δ-on-C5 slope centered at −0.998 (SD 0.127).
The observed −0.960 is not extreme against this null (two-sided p =
0.77). In plain language: even when C4a values are randomly reshuffled
across subjects, the Δ-on-C5 slope still lands near −1 on average,
because the change-score parameterization mechanically pushes the slope
toward −1 whenever C4a is roughly independent of C5. The −0.96 the
headline regression reports is what the regression arithmetic forces,
not independent evidence that low-baseline subjects benefit more from
the Spec at the per-subject mean grain.

\textbf{B.7.3 Bootstrap.}

A 10,000-iteration subject-level bootstrap returns CIs of {[}−1.254,
−0.740{]} for the Δ-on-C5 slope and {[}−0.254, +0.260{]} for the level
slope. The level CI straddles zero, consistent with the level-regression
finding that the per-subject C4a mean is roughly constant across
baselines.

\textbf{B.7.4 Reading the gradient against this.}

The substantive finding survives the coupling check, but its framing has
to shift away from ``the Spec acts more strongly on low-baseline
subjects'' toward the per-question reframing in \hyperref[the-cross-subject-gradient-and-its-per-question-mechanism]{§4.1}: low-baseline
subjects have a larger pool of questions at low rubric anchors, so the
Spec has more opportunity to produce upward integer-anchor crossings,
which aggregates as a larger per-subject mean lift. The directional
asymmetry on those crossings (no observed transitions from anchors 2, 3,
or 4 into anchor 5 across the full 14-subject panel; the only anchor-5
endpoints reached come from anchor 1) is consistent with the \hyperref[compression-structure-vs.-raw-text]{§4.2}
finding that even the full source corpus C8 plateaus at a similar
per-subject mean.

\hypertarget{b.8-per-predicate-ablation-phase-2c}{%
\subsection{B.8 Per-predicate ablation (Phase
2c)}\label{b.8-per-predicate-ablation-phase-2c}}

To probe whether single behavioral predicates within the Spec are
uniquely load-bearing, we ran a per-sentence ablation experiment on a
stratified sample of 16 extreme-upward-jump cases. For each case, the
heuristically-identified causal predicate (highest-token-overlap Spec
sentence vs the question and held-out passage) was located in the served
Spec and three response variants were generated at temperature 0 with
Claude Haiku 4.5: (1) original (full Spec), (2) ablated (predicate
removed, replaced with a length-matched neutral biographical filler),
(3) reversed (predicate replaced with a behavioral opposite synthesized
by Sonnet). Each variant was scored by the 5-judge primary panel.

Results (script: \texttt{scripts/\allowbreak{}run\_\allowbreak{}predicate\_\allowbreak{}ablation.\allowbreak{}py}; data:
\texttt{docs/\allowbreak{}research/\allowbreak{}predicate\_\allowbreak{}ablation\_\allowbreak{}results\_\allowbreak{}20260428.\allowbreak{}json}):

\begin{itemize}
\tightlist
\item
  Mean Δ\_removal (original minus ablated) across 16 cases: +0.05 anchor
  points (95\% CI {[}−0.35, +0.45{]})
\item
  Mean Δ\_reversal (original minus reversed): −0.24 anchor points (95\%
  CI {[}−0.45, −0.02{]})
\item
  2 of 16 cases showed Δ\_removal ≥1 anchor; 11 of 16 had Δ\_removal
  \textless{} 0.5
\end{itemize}

Single-predicate removal does not measurably reduce response quality on
this sample. The paper does not interpret this as evidence that the Spec
is mechanistically inert: the higher-level mechanism evidence from the
wrong-Spec adversarial control (Appendix C / \hyperref[mechanism-correct-content-not-format]{§4.3}) shows the Spec as a
whole is doing causal work. The null result on per-sentence ablation is
consistent with redundant Spec construction, in which multiple sentences
across the anchors / core / predictions / brief layers reinforce the
same behavioral patterns; removal of any single sentence leaves the
pattern accessible elsewhere in the Spec.

A methodological caveat applies. Original-condition reproduction at
temperature 0 was not bit-exact deterministic across reruns; mean drift
between the recorded original score (from
\texttt{docs/\allowbreak{}research/\allowbreak{}wins\_\allowbreak{}inventory\_\allowbreak{}20260428.\allowbreak{}json}) and the rerun
original score was −1.44 anchors, with 9 of 16 cases drifting by more
than 1 anchor. Some of the variance in Δ\_removal is rerun stochasticity
rather than ablation effect. The extreme-upward-jump cases specifically
show higher pipeline variance than the per-subject mean grain documented
in \hyperref[pipeline-and-specification-stability]{§6.3}.

Future work tightening (per the test's own report): human-rated
predicate identification (vs heuristic), larger N (all 47
PATTERN\_PREDICATE cases), irrelevant-predicate control (matched-length
unrelated predicate to test the ``any rich persona text'' alternative),
multi-predicate cluster ablation.

\hypertarget{b.9-footnote-redirect-technical-detail}{%
\subsection{B.9 Footnote-redirect technical
detail}\label{b.9-footnote-redirect-technical-detail}}

This subsection holds the longer technical content for footnotes that
would otherwise grow to multi-paragraph length. Each entry is keyed to
the footnote name in the body.

\textbf{B.9.1 The +0.89 vs +0.93 aggregation reconciliation.}

The +0.89 figure is the canonical cross-subject mean of per-subject
Δ\_C4a. Each subject's Δ is computed as that subject's per-question
5-judge primary mean under C4a minus their per-question mean under C5;
these per-subject Δs are then averaged across the 9 low-baseline
subjects. The grand-mean alternative grand-averages all per-question
scores under each condition first and then takes the difference,
yielding +0.93 (the difference of the C4a grand mean 2.45 and the C5
grand mean 1.52). The two numbers are not in conflict; they answer
slightly different questions. The per-subject-mean grain (+0.89) is the
unit of inference used throughout this paper because every statistic is
computed at the subject level first, then aggregated across the 14
subjects (\hyperref[what-we-tested]{§1.2} aggregation rule).

\textbf{B.9.2 Held-out leakage
audit detail.}

A held-out leakage audit on the 60 unique extreme-upward-jump cases
(full report at
\texttt{docs/\allowbreak{}research/\allowbreak{}held\_\allowbreak{}out\_\allowbreak{}leakage\_\allowbreak{}investigation\_\allowbreak{}20260428.\allowbreak{}md})
found 0 6-gram, 2 4-gram, and 12 3-gram matches between held-out
passages and C4a responses. Of the 9 cases with any leak, 6 are short
generic phrases also resident in the served facts list (CORPUS\_LEAK), 2
are subject-specific n-grams not in any served context (best explained
by pretraining recall of public-domain autobiographies;
PRETRAINING\_MEMO\_CANDIDATE), and 1 is generic English
(COMMON\_PHRASE). The longest shared run anywhere is 4 tokens, well
below transcription length. Severity verdict: rare; no structural
validity concern; footnote acknowledgement is the appropriate paper-text
treatment. Excluding the 2 pretraining-memorization candidates from the
extreme-upward-jump set shifts the C5 to C4a low-baseline extreme-jump
count by at most 1 (20 to 19); per-subject mean Δs are unchanged at the
per-question level. The ``held-out passage'' was held out from served
Spec / facts, not from pretraining, and the audit confirms that
interpretation: where C4a held-out-to-post leakage exists, it is either
short generic phrasing also resident in the served facts (trivially
short) or subject-specific content in the model's pretraining, not
study-design contamination of the served context.

\textbf{B.9.3 Supermemory
NO\_RETRIEVAL placeholders.}

Across the full 14-subject Supermemory analysis, 30 individual responses
(Augustine 2 questions, Equiano 28 questions) were Supermemory
provider-failure placeholders rather than substantive predictions,
scored at the rubric floor (1) by the judge panel. We treat these as
scored data rather than missing data, consistent with how the rest of
the study handles low-quality responses. Excluding the 30 NO\_RETRIEVAL
records as missing data would shift Supermemory's aggregate Δ slightly
higher; the qualitative story (small aggregate at both grains, bimodal
per-question distribution) holds either way.

\hypertarget{b.10-pre-registered-hypotheses-and-post-hoc-analyses}{%
\subsection{B.10 Pre-registered hypotheses and post-hoc
analyses}\label{b.10-pre-registered-hypotheses-and-post-hoc-analyses}}

The paper distinguishes pre-registered hypotheses (H1--H5, locked
against the statistical commitments in
\texttt{docs/\allowbreak{}ANALYSIS\_\allowbreak{}PLAN\_\allowbreak{}LOCK.\allowbreak{}md}) from analyses that emerged during
the work. The following table catalogues every load-bearing analysis
result reported in \hyperref[results]{§4} and identifies its status. Post-hoc items are
reported as exploratory rather than at the same evidentiary tier as the
pre-registered hypotheses.

\begin{landscape}
\begin{longtable}[]{@{}
  >{\raggedright\arraybackslash}p{(\columnwidth - 6\tabcolsep) * \real{0.25}}
  >{\raggedright\arraybackslash}p{(\columnwidth - 6\tabcolsep) * \real{0.25}}
  >{\raggedright\arraybackslash}p{(\columnwidth - 6\tabcolsep) * \real{0.25}}
  >{\raggedright\arraybackslash}p{(\columnwidth - 6\tabcolsep) * \real{0.25}}@{}}
\toprule
Item & Status & Where reported & Note \\
\midrule
\endhead
\textbf{H1} Spec-context outperforms no-context & Pre-registered & \hyperref[the-cross-subject-gradient-and-its-per-question-mechanism]{§4.1},
\hyperref[what-we-found]{§1.3} 1st bullet & Headline gradient \\
\textbf{H2} Spec benefit inversely proportional to pretraining coverage
& Pre-registered & \hyperref[the-cross-subject-gradient-and-its-per-question-mechanism]{§4.1}, \hyperref[the-gradient-at-the-high-baseline-end-franklin-reference]{§4.1.2}, \hyperref[what-we-found]{§1.3} 1st bullet & Gradient at both
ends; Franklin reference \\
\textbf{H3} Content-specificity (correct vs.~wrong Spec) &
Pre-registered & \hyperref[mechanism-correct-content-not-format]{§4.3}, \hyperref[what-we-found]{§1.3} 4th bullet & Wrong-Spec controls v1 + v2 \\
\textbf{H4} Spec interacts with retrieval through three patterns &
Pre-registered & \hyperref[memory-system-composition]{§4.4}, \hyperref[where-the-spec-helps-where-it-hurts-and-which-question-types-route-to-each]{§4.4.3}, \hyperref[what-we-found]{§1.3} 5th bullet & Memory-system
composition \\
\textbf{H5} Compression: \textasciitilde7K-token Spec recovers most of
corpus signal & Pre-registered & \hyperref[compression-structure-vs.-raw-text]{§4.2}, \hyperref[what-we-found]{§1.3} 3rd bullet & At 5x to 80x
smaller context \\
Cross-system retrieval-overlap divergence & Post-hoc & \hyperref[cross-system-retrieval-providers-do-not-converge]{§4.4.1};
sensitivity in \hyperref[retrieval-overlap-sensitivity-semantic-similarity-matching-k-variation]{§4.6.6}; \hyperref[what-we-found]{§1.3} 7th bullet & Surfaced during memory-system
analysis; mean Jaccard 0.083 across 10 system pairs; survives
semantic-similarity matching \\
Letta stateful-agent case study & Post-hoc & \hyperref[exploratory-case-study-letta-stateful-agent-n3-post-hoc]{§4.5}; full in Appendix G &
N=3, exploratory \\
Letta semantic-duplication scaling & Post-hoc & \hyperref[exploratory-case-study-letta-stateful-agent-n3-post-hoc]{§4.5}; Appendix G &
Surfaced in this paper's analysis; cosine ≥ 0.85 = 56.1\% on Bābur \\
Abstention-credit validity audit & Post-hoc & \hyperref[rubric-handling-limitations-post-hoc-validity-audit]{§3.3.6} & 9.4\% of refusals
score ≥ 2.0; bias direction makes the Spec effect likely larger than
reported \\
Per-subject wrong-Spec heterogeneity & Post-hoc & \hyperref[wrong-spec-derangement-protocol-sensitivity]{§4.6.5} & 5/13 subjects
show small positive v1 deltas (coincidental content overlap) \\
Hedging-elimination (28.8\% → 0.0\%) & Post-hoc & \hyperref[mechanism-correct-content-not-format]{§4.3}, \hyperref[what-we-found]{§1.3} 6th bullet
& Surfaced from response-level audit \\
Battery-question-type sensitivity (literal-recall fraction) & Post-hoc
reactive & \hyperref[battery-composition-sensitivity]{§4.6.3}, Appendix B.6 & Added in response to v9/v10 reviewer
concerns \\
Hamerton leverage check (subset regression) & Post-hoc reactive &
\hyperref[battery-composition-sensitivity]{§4.6.3}, Appendix B.6 & Added in response to v9/v10 reviewer concerns \\
Coupling-free reframing of the gradient & Post-hoc reactive & \hyperref[per-question-baseline-engagement-and-the-worked-rubric-example]{§4.1.1}
leveler callout, Appendix B.7 & Added in response to GPT-5.5 review \\
Cross-provider response generation (Tier 2) & Pre-registered control &
\hyperref[response-models]{§3.6}, \hyperref[circularity-controls]{§3.5.1}, \hyperref[cross-provider-response-generation-tier-2-replication]{§4.6.1} & Sonnet 4.6 + Gemini 2.5 Pro on 3 subjects \\
GPT-5.4 battery regeneration (Control 1) & Pre-registered control &
\hyperref[circularity-controls]{§3.5.1}, \hyperref[cross-provider-response-generation-tier-2-replication]{§4.6.1} & Battery generator circularity \\
Judge-panel composition (5-judge primary, 7-judge sensitivity) &
Pre-registered control & \hyperref[calibration]{§3.3.3}, \hyperref[judge-panel-sensitivity-5-judge-primary-vs-7-judge]{§4.6.2} & Locked panel before scoring \\
Wrong-Spec derangement protocol sensitivity (v1 vs v2) & Reactive &
\hyperref[wrong-spec-derangement-protocol-sensitivity]{§4.6.5} & v2 is the standard randomization control; v1 is the adversarial
stress test (headlined for stronger evidence) \\
\bottomrule
\end{longtable}
\end{landscape}

Reproducibility scripts and raw data for each row are pointed to
throughout \hyperref[results]{§4} and consolidated in \hyperref[data-code-and-reproducibility]{§8} Data, Code, and Reproducibility.

\hypertarget{b.11-per-system-per-subject-paired-delta-distributions}{%
\subsection{B.11 Per-system per-subject paired-delta
distributions}\label{b.11-per-system-per-subject-paired-delta-distributions}}

The \hyperref[where-the-spec-helps-where-it-hurts-and-which-question-types-route-to-each]{§4.4.3} footnote Appendix B.11 collects
the per-cell counts behind the three-pattern claim. This subsection
consolidates those cells into a single table and a short reading note.
The unit of observation is a single (subject, question) pair scored
under both retrieval-only (C1) and retrieval + Behavioral Specification
(C3); per-cell counts are restricted to questions with 5-judge primary
coverage on both conditions. ``Increases'' / ``decreases'' use the
\textbar Δ\textbar{} ≥ 1.0 threshold on the 5-point rubric (one full
anchor crossing) so that small judge-noise jitter does not inflate the
count in either direction.

\begin{landscape}
\begingroup\footnotesize
\begin{longtable}[]{@{}
  >{\raggedright\arraybackslash}p{(\columnwidth - 12\tabcolsep) * \real{0.12}}
  >{\raggedright\arraybackslash}p{(\columnwidth - 12\tabcolsep) * \real{0.12}}
  >{\raggedleft\arraybackslash}p{(\columnwidth - 12\tabcolsep) * \real{0.17}}
  >{\raggedleft\arraybackslash}p{(\columnwidth - 12\tabcolsep) * \real{0.17}}
  >{\raggedleft\arraybackslash}p{(\columnwidth - 12\tabcolsep) * \real{0.17}}
  >{\raggedright\arraybackslash}p{(\columnwidth - 12\tabcolsep) * \real{0.12}}
  >{\raggedright\arraybackslash}p{(\columnwidth - 12\tabcolsep) * \real{0.12}}@{}}
\toprule
Memory system & Subject & Aggregate Δ\_spec & Increases (Δ ≥ +1.0) &
Decreases (Δ ≤ −1.0) & Net at the per-question grain & Source \\
\midrule
\endhead
Supermemory & full 14-subject pool & ≈ 0 (closest to zero) & 57 & 53 &
110 of 546 paired questions cross by ≥ 1.0 (20.1\%); the two roughly
cancel at the mean & \hyperref[where-the-spec-helps-where-it-hurts-and-which-question-types-route-to-each]{§4.4.3} lede paragraph +
§4.4.2 note \\
Mem0 & Yung Wing & +0.33 & 21 & 10 & 31 of 39 paired questions cross by
≥ 1.0; 21 helps outnumber 10 hurts 2.1× &
Appendix B.11 \\
Mem0 & Keckley & −0.02 & 12 & 13 & 25 of 39 paired questions cross by ≥
1.0; counts even at 12 / 13, aggregate near zero &
Appendix B.11 \\
Letta archival & Hamerton & +0.42 & 19 & 7 & 26 of 39 paired questions
cross by ≥ 1.0; 19 helps outnumber 7 hurts 2.7× &
Appendix B.11 \\
Zep & Seacole & +0.47 & 20 & 7 & 27 of 39 paired questions cross by ≥
1.0; 0 questions show large regressions in this cell &
Appendix B.11 \\
Base Layer & Yung Wing & +0.29 & 19 & 7 & 26 of 39 paired questions
cross by ≥ 1.0; 19 helps outnumber 7 hurts 2.7× &
Appendix B.11 \\
\bottomrule
\end{longtable}
\endgroup
\end{landscape}

\textbf{Reading.} Every cell is a mixture. Strong-positive aggregates
(Letta Hamerton, Zep Seacole, Base Layer Yung Wing) still contain 7
large regressions per cell. Near-zero aggregates (Mem0 Keckley) resolve
into substantial counts in both directions rather than a flat
per-question profile. The Supermemory pool is the cleanest read on the
three-pattern mixture because the helps-versus-hurts counts are nearly
balanced; the same shape reproduces across every other (system, subject)
cell in the study.

The cells above are representative rather than exhaustive (one cell per
memory system, two for Mem0 to capture both a positive-aggregate and a
near-zero-aggregate subject). Full per-system per-subject paired-delta
arrays are at
\texttt{docs/\allowbreak{}research/\allowbreak{}per\_\allowbreak{}system\_\allowbreak{}anchor\_\allowbreak{}crossing\_\allowbreak{}20260427.\allowbreak{}json}; the
recompute script is
\texttt{scripts/\allowbreak{}\_\allowbreak{}table\_\allowbreak{}4\_\allowbreak{}6\_\allowbreak{}5judge\_\allowbreak{}recompute.\allowbreak{}py}.

\begin{center}\rule{0.5\linewidth}{0.5pt}\end{center}

\hypertarget{b.12-wrong-spec-v1-fixed-pairings}{%
\subsection{B.12 Wrong-Spec v1 fixed
pairings}\label{b.12-wrong-spec-v1-fixed-pairings}}

The v1 wrong-Spec control (\hyperref[experimental-conditions]{§3.2}, \hyperref[mechanism-correct-content-not-format]{§4.3}, \hyperref[wrong-spec-derangement-protocol-sensitivity]{§4.6.5}) uses a fixed directed
permutation locking each subject to a culturally and temporally distant
subject's Behavioral Specification. The permutation was locked in
\texttt{scripts/\allowbreak{}run\_\allowbreak{}global\_\allowbreak{}rerun.\allowbreak{}py} (the
\texttt{WRONG\_SPEC\_PAIRING} dict) before any C2c responses were
generated and is not modified by analysis.

\begin{longtable}[]{@{}ll@{}}
\toprule
Subject & Assigned Spec (v1) \\
\midrule
\endhead
Augustine & Fukuzawa \\
Bābur & Keckley \\
Bernal Díaz & Sunity Devee \\
Cellini & Zitkala-Ša \\
Ebers & Equiano \\
Equiano & Ebers \\
Fukuzawa & Augustine \\
Keckley & Bābur \\
Rousseau & Yung Wing \\
Seacole & Bernal Díaz \\
Sunity Devee & Cellini \\
Yung Wing & Rousseau \\
Zitkala-Ša & Seacole \\
\bottomrule
\end{longtable}

Two pairs are reciprocal (Ebers ↔ Equiano; Rousseau ↔ Yung Wing;
Augustine ↔ Fukuzawa; Bābur ↔ Keckley). The remaining five entries
(Bernal Díaz → Sunity Devee → Cellini → Zitkala-Ša → Seacole → Bernal
Díaz) form a directed cycle, not pairs. Hamerton's v1 variant receives
Franklin's specification and is reported separately in \hyperref[the-gradient-at-the-high-baseline-end-franklin-reference]{§4.1.2}.

\hypertarget{b.13-memory-system-wilcoxon-results-c1-vs-c3-and-low-baseline-ux3b4_spec}{%
\subsection{B.13 Memory-system Wilcoxon results (C1 vs C3) and
low-baseline
Δ\_spec}\label{b.13-memory-system-wilcoxon-results-c1-vs-c3-and-low-baseline-ux3b4_spec}}

The \hyperref[memory-system-composition]{§4.4} aggregate Wilcoxon tests on C3 (retrieval + Spec) vs C1
(retrieval only), on the 5-judge primary panel. All p-values are
two-sided; W is the signed-rank statistic. Low-baseline slice (n = 9)
median Δ and 95\% CI alongside the all-14 panel. Data:
\texttt{docs/\allowbreak{}research/\allowbreak{}stats\_\allowbreak{}update.\allowbreak{}md} (full panel) and
\texttt{docs/\allowbreak{}research/\allowbreak{}memory\_\allowbreak{}systems\_\allowbreak{}5judge\_\allowbreak{}primary.\allowbreak{}md} (per-system Δ
+ improvement counts).

\begin{longtable}[]{@{}
  >{\raggedright\arraybackslash}p{(\columnwidth - 12\tabcolsep) * \real{0.12}}
  >{\raggedright\arraybackslash}p{(\columnwidth - 12\tabcolsep) * \real{0.12}}
  >{\raggedleft\arraybackslash}p{(\columnwidth - 12\tabcolsep) * \real{0.15}}
  >{\raggedleft\arraybackslash}p{(\columnwidth - 12\tabcolsep) * \real{0.15}}
  >{\raggedleft\arraybackslash}p{(\columnwidth - 12\tabcolsep) * \real{0.15}}
  >{\raggedleft\arraybackslash}p{(\columnwidth - 12\tabcolsep) * \real{0.15}}
  >{\raggedleft\arraybackslash}p{(\columnwidth - 12\tabcolsep) * \real{0.15}}@{}}
\toprule
System & Config & N (all) & W & p (all-14) & Low-baseline Δ\_spec mean &
Low-baseline improved (of 9) \\
\midrule
\endhead
Zep & controlled & 14 & 2.0 & \textbf{0.0004} & +0.166 & 9 / 9 \\
Letta & controlled & 14 & 6.0 & \textbf{0.0017} & +0.165 & 8 / 9 \\
Mem0 & native & 14 & 8.0 & \textbf{0.0088} & +0.320 & 7 / 9 \\
Zep & native & 14 & 2.0 & \textbf{0.0015} & +0.303 & 9 / 9 \\
Mem0 & controlled & 14 & 15.0 & 0.0166 & +0.101 & 6 / 9 \\
Letta & native & 14 & 35.0 & 0.4629 & −0.036 & 4 / 9 \\
Supermemory & controlled & 14 & 37.0 & 0.3575 & −0.010 & 5 / 9 \\
Supermemory & native & 10 & 18.0 & 0.3750 & −0.026 & 3 / 7 \\
Base Layer & substrate (controlled) & 14 & 27.0 & 0.1189 & +0.083 & 6 /
9 \\
\bottomrule
\end{longtable}

Four (system, configuration) cells are significant at α = 0.01 (bolded):
Zep controlled, Letta controlled, Mem0 native, Zep native. Mem0
controlled is significant at α = 0.05 but not α = 0.01. The remaining
cells (Letta native, both Supermemory configurations, Base Layer
substrate) are not significant at α = 0.05. Supermemory native has
partial coverage (N = 10 of 14; 7 of 9 in the low-baseline slice).

\hypertarget{b.14-per-condition-improvement-and-worsening-rates-all-14-546-questions}{%
\subsection{B.14 Per-condition improvement and worsening rates (all-14,
546
questions)}\label{b.14-per-condition-improvement-and-worsening-rates-all-14-546-questions}}

The \hyperref[per-question-improvement-rate]{§4.2.1} per-question improvement rates for each condition against the
No-Context Baseline (C5). ``Improvement'' means the response under the
named condition has a higher 5-judge primary score than the same
question's C5 response; ``worsening'' means lower. Numbers verified
against
\texttt{docs/\allowbreak{}research/\allowbreak{}v11\_\allowbreak{}paper\_\allowbreak{}numbers\_\allowbreak{}verification\_\allowbreak{}20260428.\allowbreak{}md}.

\begin{longtable}[]{@{}lrrr@{}}
\toprule
Condition (vs C5) & Improvement & Worsening & Net (improve − worsen) \\
\midrule
\endhead
Spec only (C2a) & 58.8\% & 26.7\% & +32.1 pp \\
Facts only (C4) & 60.1\% & 26.6\% & +33.5 pp \\
Raw corpus (C8) & 65.2\% & 23.6\% & +41.6 pp \\
Facts + Spec (C4a) & 65.8\% & 26.4\% & +39.4 pp \\
\bottomrule
\end{longtable}

Pairwise question-level comparison on the low-baseline 9-subject band
(351 questions). The first entry on each row reads ``first condition
beats second''; ties are listed.

\begin{longtable}[]{@{}lrrr@{}}
\toprule
Comparison & First better & Second better & Tie \\
\midrule
\endhead
Raw corpus (C8) vs.~Spec only (C2a) & 53.3\% & 30.8\% & 56 questions \\
Corpus + Spec (C9) vs.~Facts + Spec (C4a) & 49.0\% & 36.5\% & 45
questions \\
\bottomrule
\end{longtable}

The 7K-token facts + Spec package scores higher than the much larger
corpus + Spec package on roughly one-third of low-baseline questions.

\hypertarget{b.15-memory-system-abstention-audit-cell-counts}{%
\subsection{B.15 Memory-system abstention audit cell
counts}\label{b.15-memory-system-abstention-audit-cell-counts}}

The \hyperref[rubric-handling-limitations-post-hoc-validity-audit-1]{§4.6.7} audit on whether memory-system retrieval inflates refusal
scores via visible fact recitation. Cells partition the response pool by
retrieval condition and whether the response (a) is an abstention and
(b) recites a retrieved n-gram. Data:
\texttt{docs/\allowbreak{}research/\allowbreak{}abstention\_\allowbreak{}extensions\_\allowbreak{}draft\_\allowbreak{}20260429.\allowbreak{}md}.

\begin{longtable}[]{@{}
  >{\raggedright\arraybackslash}p{(\columnwidth - 8\tabcolsep) * \real{0.17}}
  >{\raggedright\arraybackslash}p{(\columnwidth - 8\tabcolsep) * \real{0.17}}
  >{\raggedleft\arraybackslash}p{(\columnwidth - 8\tabcolsep) * \real{0.22}}
  >{\raggedleft\arraybackslash}p{(\columnwidth - 8\tabcolsep) * \real{0.22}}
  >{\raggedleft\arraybackslash}p{(\columnwidth - 8\tabcolsep) * \real{0.22}}@{}}
\toprule
Cell & Definition & N & Mean & \% ≥ 2.0 \\
\midrule
\endhead
Pure No-Context Refusal & no facts, no retrieval & 292 & 1.26 &
10.3\% \\
Facts-only refusal & facts in context, no retrieval & 20 & 1.33 &
10.0\% \\
Memory-system refusal + recitation & refuses AND quotes retrieved n-gram
& 148 & 1.50 & 18.2\% \\
Memory-system refusal, no recitation & refuses, does not quote retrieval
& 240 & 1.47 & 17.1\% \\
Memory-system substantive engagement & non-refusal under retrieval &
7,835 & 2.32 & 67.2\% \\
\bottomrule
\end{longtable}

Welch comparisons on the mean abstention score with 95\% CI:

\begin{longtable}[]{@{}
  >{\raggedright\arraybackslash}p{(\columnwidth - 6\tabcolsep) * \real{0.21}}
  >{\raggedleft\arraybackslash}p{(\columnwidth - 6\tabcolsep) * \real{0.29}}
  >{\raggedright\arraybackslash}p{(\columnwidth - 6\tabcolsep) * \real{0.21}}
  >{\raggedleft\arraybackslash}p{(\columnwidth - 6\tabcolsep) * \real{0.29}}@{}}
\toprule
Comparison & Δ & 95\% CI & \emph{p} (Welch) \\
\midrule
\endhead
Mem-refuse + recite vs.~pure No-Context refuse & +0.234 & {[}+0.113,
+0.355{]} & 0.0002 \\
Mem-refuse no recite vs.~pure No-Context refuse & +0.206 & {[}+0.103,
+0.310{]} & 0.0001 \\
Mem-refuse + recite vs.~mem-refuse no recite & +0.027 & {[}−0.098,
+0.153{]} & 0.67 \\
\bottomrule
\end{longtable}

Memory-system refusals score +0.21 to +0.23 anchor points above pure
No-Context Refusals, but the lift is the same whether or not the
response recites a retrieved n-gram (Δ +0.027, \emph{p} = 0.67). The
over-credit is a ``judges reward the retrieval condition'' effect, not a
``judges reward the visible quote'' effect.

\begin{center}\rule{0.5\linewidth}{0.5pt}\end{center}

\hypertarget{appendix-c.-conditions-models-and-memory-system-configurations}{%
\section{Appendix C. Conditions, Models, and Memory-System
Configurations}\label{appendix-c.-conditions-models-and-memory-system-configurations}}

\hypertarget{c.1-condition-identifiers-summary-card}{%
\subsection{C.1 Condition identifiers (summary
card)}\label{c.1-condition-identifiers-summary-card}}

A consolidated lookup for the condition IDs used throughout \hyperref[results]{§4}. Defined
in \hyperref[experimental-conditions]{§3.2}; summarized here.

\begin{longtable}[]{@{}
  >{\raggedright\arraybackslash}p{(\columnwidth - 6\tabcolsep) * \real{0.25}}
  >{\raggedright\arraybackslash}p{(\columnwidth - 6\tabcolsep) * \real{0.25}}
  >{\raggedright\arraybackslash}p{(\columnwidth - 6\tabcolsep) * \real{0.25}}
  >{\raggedright\arraybackslash}p{(\columnwidth - 6\tabcolsep) * \real{0.25}}@{}}
\toprule
ID & Family & Context served & Purpose \\
\midrule
\endhead
C5 & Direct & None. Question only. & Pretraining-only floor.
Baseline. \\
C2a & Direct & Behavioral Specification only. & Isolate Spec's
contribution. \\
C2c & Direct & A random other subject's Spec (derangement, seed=42). &
Wrong-Spec control. \\
C4 & Direct & Full extracted fact set for subject. & Raw fact volume, no
structure. \\
C4a & Direct & Full All Facts + Spec. & Spec added to raw facts. \\
C8 & Direct & Full training corpus (half of source text). & Uncompressed
source. \\
C9 & Direct & Training corpus plus Spec. & Spec added to raw source.
Bābur excluded (422K word overflow). \\
C1 & Memory system & Top-k retrieval output from system. & Retrieval
only, each of 5 systems. \\
C3 & Memory system & Top-k retrieval output plus Spec. & Retrieval plus
Spec, each of 5 systems. \\
C1\_\textless system\textgreater\_fullpipeline & Memory system, native &
Retrieval from system-native ingestion of raw corpus. & Native ingestion
variant, Retrieval Only. \\
C3\_\textless system\textgreater\_fullpipeline & Memory system, native &
Native ingestion retrieval plus Spec. & Native ingestion variant,
retrieval plus Spec. \\
\bottomrule
\end{longtable}

The \texttt{\textless{}system\textgreater{}} slot ranges over \{mem0,
letta, supermemory, zep, baselayer\}. Base Layer is run in a single
controlled configuration; the four commercial systems are run in both
controlled and native variants.

\hypertarget{c.2-shared-response-model-invocation}{%
\subsection{C.2 Shared response-model
invocation}\label{c.2-shared-response-model-invocation}}

Every response call, across every direct-context and memory-system
condition, uses the following parameters:

\begin{longtable}[]{@{}
  >{\raggedright\arraybackslash}p{(\columnwidth - 2\tabcolsep) * \real{0.50}}
  >{\raggedright\arraybackslash}p{(\columnwidth - 2\tabcolsep) * \real{0.50}}@{}}
\toprule
Parameter & Value \\
\midrule
\endhead
\texttt{temperature} & 0 \\
\texttt{max\_tokens} & 1024 \\
System prompt & Framing instruction: predict how
\texttt{\textless{}subject\textgreater{}} would respond; answer in
subject's voice, grounded in demonstrated patterns. \\
User prompt format &
\texttt{\textless{}context\ block\textgreater{}\textbackslash{}n\textbackslash{}nQuestion:\ \textless{}question\ text\textgreater{}} \\
Context block & Condition-dependent. Empty in C5. Spec in C2a. Wrong
Spec in C2c. Facts in C4. Facts plus Spec in C4a. Corpus in C8. Corpus
plus Spec in C9. Retrieval output (optionally plus Spec) in C1 and
C3. \\
\bottomrule
\end{longtable}

No prompt instruction coaches the model to abstain, hedge, or commit.
The model's refusal-or-commitment pattern given a specific context is
part of the phenomenon being measured (\hyperref[response-models]{§3.6}, \hyperref[mechanism-correct-content-not-format]{§4.3}).

\hypertarget{c.3-response-models}{%
\subsection{C.3 Response models}\label{c.3-response-models}}

\begingroup\footnotesize
\begin{longtable}[]{@{}
  >{\raggedright\arraybackslash}p{(\columnwidth - 6\tabcolsep) * \real{0.25}}
  >{\raggedright\arraybackslash}p{(\columnwidth - 6\tabcolsep) * \real{0.25}}
  >{\raggedright\arraybackslash}p{(\columnwidth - 6\tabcolsep) * \real{0.25}}
  >{\raggedright\arraybackslash}p{(\columnwidth - 6\tabcolsep) * \real{0.25}}@{}}
\toprule
Role & Model identifier & Provider & Scope \\
\midrule
\endhead
Primary response & \texttt{claude-haiku-4-5-20251001} & Anthropic & All
14 subjects, every condition. Main study. \\
Tier 2 response & \texttt{claude-sonnet-4-6} & Anthropic & 3 subjects
(Ebers, Yung Wing, Zitkala-Ša), C5 / C2a / C2c / C4a against GPT-5.4
batteries. \\
Tier 2 response & \texttt{gemini-2.5-pro} & Google & Same 3 subjects,
same conditions as Sonnet Tier 2. \\
\bottomrule
\end{longtable}
\endgroup

Source: \texttt{scripts/\allowbreak{}run\_\allowbreak{}global\_\allowbreak{}subjects.\allowbreak{}py},
\texttt{scripts/\allowbreak{}run\_\allowbreak{}full\_\allowbreak{}study.\allowbreak{}py},
\texttt{scripts/\allowbreak{}run\_\allowbreak{}multimodel\_\allowbreak{}responses.\allowbreak{}py}.

\hypertarget{c.4-pipeline-models-specification-generation}{%
\subsection{C.4 Pipeline models (specification
generation)}\label{c.4-pipeline-models-specification-generation}}

\begingroup\footnotesize
\begin{longtable}[]{@{}
  >{\raggedright\arraybackslash}p{(\columnwidth - 6\tabcolsep) * \real{0.25}}
  >{\raggedright\arraybackslash}p{(\columnwidth - 6\tabcolsep) * \real{0.25}}
  >{\raggedright\arraybackslash}p{(\columnwidth - 6\tabcolsep) * \real{0.25}}
  >{\raggedright\arraybackslash}p{(\columnwidth - 6\tabcolsep) * \real{0.25}}@{}}
\toprule
Pipeline step & Model identifier & Temperature & Purpose \\
\midrule
\endhead
Extract (Step 2) & \texttt{claude-haiku-4-5-20251001} & 0 & AUDN fact
extraction, 46-predicate constrained vocabulary. \\
Embed (Step 3) & \texttt{all-MiniLM-L6-v2} (local) & n/a & ChromaDB
vector index (L2 distance). \\
Author (Step 4) & \texttt{claude-sonnet-4-6} & 0 & Three authored layers
(anchors, core, predictions). Blind regen, domain guard. \\
Compose (Step 5) & \texttt{claude-opus-4-6} & 0 & Unified brief
composition. \\
Battery generation & \texttt{claude-haiku-4-5-20251001} & 0 &
Backward-design from held-out corpus. \\
Battery generation (circularity control) & \texttt{gpt-5.4} (via OpenAI
API) & 0 & Independent regeneration on 13 global subjects. \\
\bottomrule
\end{longtable}
\endgroup

Source: \texttt{memory\_\allowbreak{}system/\allowbreak{}src/\allowbreak{}baselayer/\allowbreak{}config.\allowbreak{}py}.

\hypertarget{c.5-judge-panel}{%
\subsection{C.5 Judge panel}\label{c.5-judge-panel}}

\begingroup\footnotesize
\begin{longtable}[]{@{}lllcc@{}}
\toprule
Judge & Model identifier & Provider & In 5-judge primary? & Calibration
performed? \\
\midrule
\endhead
Haiku & \texttt{claude-haiku-4-5-20251001} & Anthropic & Yes & Yes \\
Sonnet & \texttt{claude-sonnet-4-6} & Anthropic & Yes & Yes \\
Opus & \texttt{claude-opus-4-6} & Anthropic & Yes & Yes \\
GPT-4o & \texttt{gpt-4o-2024-08-06} & OpenAI & Yes & Yes \\
GPT-5.4 & \texttt{gpt-5.4} & OpenAI & Yes & Yes \\
Gemini Flash & \texttt{gemini-2.5-flash} & Google & No (sensitivity
only) & Yes \\
Gemini Pro & \texttt{gemini-2.5-pro} & Google & No (sensitivity only) &
Yes \\
\bottomrule
\end{longtable}
\endgroup

Judges are invoked independently (no cross-judge conditioning). Each
judge sees: held-out ground-truth passage, subject context (name,
source), question, response. Judge temperature 0. Judge output is a
numeric 1-5 score plus a free-text justification. Calibration diagnostic
results in \hyperref[calibration]{§3.3.3}.

\hypertarget{c.6-memory-system-ingestion-and-retrieval-parameters}{%
\subsection{C.6 Memory-system ingestion and retrieval
parameters}\label{c.6-memory-system-ingestion-and-retrieval-parameters}}

Controlled configuration (C1 / C3) holds the input identical across
systems: each system receives the same extracted fact set used by the
Base Layer pipeline, re-ingested through its own API. Native
configuration (\texttt{\_fullpipeline}) has each system ingest the raw
training corpus directly via its own chunking and extraction.

\begingroup\footnotesize
\begin{longtable}[]{@{}
  >{\raggedright\arraybackslash}p{(\columnwidth - 10\tabcolsep) * \real{0.09}}
  >{\raggedright\arraybackslash}p{(\columnwidth - 10\tabcolsep) * \real{0.22}}
  >{\raggedright\arraybackslash}p{(\columnwidth - 10\tabcolsep) * \real{0.15}}
  >{\raggedright\arraybackslash}p{(\columnwidth - 10\tabcolsep) * \real{0.15}}
  >{\raggedleft\arraybackslash}p{(\columnwidth - 10\tabcolsep) * \real{0.10}}
  >{\raggedright\arraybackslash}p{(\columnwidth - 10\tabcolsep) * \real{0.29}}@{}}
\toprule
System & Ingestion endpoint & Ingestion unit (controlled) & Ingestion
unit (native) & Retrieval top-k & Notable configuration \\
\midrule
\endhead
Mem0 & \texttt{POST\ /\allowbreak{}v1/\allowbreak{}memories/} & One fact per POST & Raw corpus
chunks (Mem0 chunker) & 10 & \texttt{infer=False} on controlled (store
as-is, no reformulation). Failure mode: Mem0 may reformulate on
\texttt{infer=True}, flagged in \texttt{docs/PROVIDER\_\allowbreak{}ISSUES.md}. \\
Letta (archival) &
\texttt{POST\ /\allowbreak{}v1/\allowbreak{}agents/\textless{}id\textgreater{}/\allowbreak{}core\_memory/\allowbreak{}archival}
& One fact per passage & Letta native chunking & 10 & 1 fact = 1
passage. Batch ingestion tested 135x faster but changes chunking
behavior (see \texttt{run\_memory\allowbreak{}\_system.py} line 456-458). \\
Letta (stateful) & Agent state edit during ingestion & One fact per edit
cycle & Raw corpus & n/a (read from block) & Evaluated as a separate
path in \hyperref[exploratory-case-study-letta-stateful-agent-n3-post-hoc]{§4.5}, not as a row in the C1 / C3 conditions. \\
Supermemory & \texttt{POST\ /v3/memories} & One fact per memory,
\texttt{containerTags=\allowbreak{}\textless{}subject\textgreater{}} & Raw corpus &
10 & \texttt{limit=10} on retrieval. \\
Zep & Graph ingestion via \texttt{zep\_client.\allowbreak{}graph.add} & One fact per
edge & Raw corpus & 10 & Retrieval via
\texttt{client.\allowbreak{}graph.search(user\_id,\ query,\ limit=10)}. \\
Base Layer & Direct into ChromaDB & One fact per vector & n/a (Base
Layer has no native variant) & 10 & MiniLM embeddings, L2 distance,
cosine-like similarity via \texttt{1\ -\ dist\^{}2/2}. \\
\bottomrule
\end{longtable}
\endgroup

All five systems use the same top-k of 10. All five are queried with the
question text as the query. All five feed their retrieval output into
the standard prompt schema (§C.2) as the context block.

\hypertarget{c.7-ingestion-exclusions-and-failure-cases}{%
\subsection{C.7 Ingestion exclusions and failure
cases}\label{c.7-ingestion-exclusions-and-failure-cases}}

\begin{longtable}[]{@{}
  >{\raggedright\arraybackslash}p{(\columnwidth - 4\tabcolsep) * \real{0.33}}
  >{\raggedright\arraybackslash}p{(\columnwidth - 4\tabcolsep) * \real{0.33}}
  >{\raggedright\arraybackslash}p{(\columnwidth - 4\tabcolsep) * \real{0.33}}@{}}
\toprule
Subject / system & Issue & Resolution \\
\midrule
\endhead
Bābur, C9 (raw corpus plus Spec) & 422,772-word source exceeds Haiku
context window. & Excluded from C9. 13 of 14 subjects report C9
numbers. \\
Letta native (all subjects) & Ingestion ceiling on archival passages;
retrieval produces 0.34-0.47 dedup ratio, meaning a top-10 list often
contains 3-5 unique facts. & Reported as-is in \hyperref[memory-system-composition]{§4.4}. Not excluded. \\
Mem0 native & Mem0's \texttt{infer=True} reformulated facts during
native ingestion pilot. & Used \texttt{infer=False} on controlled
configuration to hold input identical. Native variant retains
\texttt{infer=True} (the realistic deployment path). \\
Zep graph bias & Zep graph retrieval surfaces entity-dense chunks over
behavior-dense chunks. & Reported as-is. See
\texttt{docs/PROVIDER\_ISSUES.md}. \\
\bottomrule
\end{longtable}

\hypertarget{c.8-analysis-plan-lock}{%
\subsection{C.8 Analysis plan lock}\label{c.8-analysis-plan-lock}}

The condition matrix was frozen in \texttt{docs/\allowbreak{}ANALYSIS\_\allowbreak{}PLAN\_\allowbreak{}LOCK.\allowbreak{}md}
before scoring. Any condition added after the lock is reported
separately (for example, the Tier 2 3-subject replication and the v2
random-derangement wrong-Spec draws).

\begin{center}\rule{0.5\linewidth}{0.5pt}\end{center}

\hypertarget{appendix-d.-validity-audit-and-score-distributions}{%
\section{Appendix D. Validity Audit and Score
Distributions}\label{appendix-d.-validity-audit-and-score-distributions}}

\hypertarget{d.1-per-subject-5-judge-primary-aggregate-main-gradient}{%
\subsection{D.1 Per-subject 5-judge primary aggregate (main
gradient)}\label{d.1-per-subject-5-judge-primary-aggregate-main-gradient}}

This table reproduces the \hyperref[the-cross-subject-gradient-and-its-per-question-mechanism]{§4.1} cross-subject gradient for reference.
Every number is the 5-judge primary mean (Haiku, Sonnet, Opus, GPT-4o,
GPT-5.4) over the 39-question behavioral-prediction battery per subject
(40 for Franklin).

\begin{longtable}[]{@{}
  >{\raggedright\arraybackslash}p{(\columnwidth - 12\tabcolsep) * \real{0.12}}
  >{\raggedleft\arraybackslash}p{(\columnwidth - 12\tabcolsep) * \real{0.15}}
  >{\raggedleft\arraybackslash}p{(\columnwidth - 12\tabcolsep) * \real{0.15}}
  >{\raggedleft\arraybackslash}p{(\columnwidth - 12\tabcolsep) * \real{0.15}}
  >{\raggedleft\arraybackslash}p{(\columnwidth - 12\tabcolsep) * \real{0.15}}
  >{\raggedleft\arraybackslash}p{(\columnwidth - 12\tabcolsep) * \real{0.15}}
  >{\centering\arraybackslash}p{(\columnwidth - 12\tabcolsep) * \real{0.12}}@{}}
\toprule
Subject & Baseline (C5) & Spec only (C2a) & Facts + Spec (C4a) & Δ\_spec
& Δ\_All Facts + Spec & Anchor crossed \\
\midrule
\endhead
Ebers & 1.02 & 1.54 & 2.07 & +0.52 & +1.05 & yes \\
Sunity Devee & 1.03 & 2.27 & 2.41 & +1.24 & +1.38 & yes \\
Hamerton & 1.26 & 2.63 & 2.77 & +1.37 & +1.51 & yes \\
Fukuzawa & 1.67 & 2.35 & 2.78 & +0.68 & +1.11 & yes \\
Bernal Díaz & 1.70 & 2.27 & 2.48 & +0.57 & +0.78 & partial \\
Bābur & 1.76 & 1.91 & 2.01 & +0.15 & +0.25 & no \\
Seacole & 1.77 & 2.48 & 2.59 & +0.71 & +0.82 & yes \\
Keckley & 1.84 & 2.43 & 2.44 & +0.58 & +0.59 & no \\
Yung Wing & 1.88 & 2.22 & 2.40 & +0.34 & +0.52 & no \\
Zitkala-Ša & 2.34 & 2.03 & 2.02 & −0.31 & −0.32 & no \\
Cellini & 2.38 & 2.54 & 2.53 & +0.16 & +0.15 & no \\
Rousseau & 2.44 & 2.81 & 2.53 & +0.37 & +0.10 & no \\
Augustine & 2.58 & 2.48 & 2.70 & −0.11 & +0.11 & no \\
Equiano & 2.77 & 2.46 & 2.42 & −0.31 & −0.35 & no \\
Franklin (control) & 3.77 & 3.37 & 3.65 & −0.40 & −0.13 & no \\
\bottomrule
\end{longtable}

Raw per-judge files:
\texttt{results/global\_\textless{}subject\textgreater{}/judgments\_v2.json}
and \texttt{*\_judgments\_\textless{}judge\textgreater{}.json}
(per-judge) for the 13 globals. Hamerton:
\texttt{results/hamerton/*\_judgments\_\textless{}judge\textgreater{}.json}.
Franklin:
\texttt{results/\allowbreak{}franklin\_\allowbreak{}legacy\_\allowbreak{}20260411/\allowbreak{}analysis/\allowbreak{}*\_\allowbreak{}judgments.\allowbreak{}json}.

\hypertarget{d.2-per-subject-anchor-crossing-on-the-low-baseline-band}{%
\subsection{D.2 Per-subject anchor-crossing on the low-baseline
band}\label{d.2-per-subject-anchor-crossing-on-the-low-baseline-band}}

Anchor-crossing rate is the fraction of per-question paired (C5, C4a)
responses where the C4a 5-judge primary mean lands in a different
integer rubric anchor than the C5 mean. Definition in \hyperref[score-interpretation]{§3.3.1} and
\texttt{scripts/\allowbreak{}compute\_\allowbreak{}anchor\_\allowbreak{}crossing.\allowbreak{}py}.

Slice-level:

\begin{itemize}
\tightlist
\item
  Total low-baseline questions (9 subjects, 39 Q each): 351
\item
  Upward crossings: 193 (55.0\%)
\item
  Downward crossings: 24 (6.8\%)
\item
  Stayed in anchor: 134 (38.2\%)
\end{itemize}

Per-subject breakdown (5-judge primary, paired C5 vs.~C4a over N=39 per
subject):

\begin{longtable}[]{@{}lrrrr@{}}
\toprule
Subject & Upward & Upward \% & Downward & No crossing \\
\midrule
\endhead
Sunity Devee & 29 & 74.4\% & 0 & 10 \\
Hamerton & 27 & 69.2\% & 0 & 12 \\
Fukuzawa & 26 & 66.7\% & 3 & 10 \\
Bernal Díaz & 23 & 59.0\% & 3 & 13 \\
Seacole & 21 & 53.8\% & 3 & 15 \\
Ebers & 19 & 48.7\% & 0 & 20 \\
Keckley & 19 & 48.7\% & 6 & 14 \\
Yung Wing & 19 & 48.7\% & 5 & 15 \\
Bābur & 10 & 25.6\% & 4 & 25 \\
\textbf{Slice total} & \textbf{193} & \textbf{55.0\%} & \textbf{24} &
\textbf{134} \\
\bottomrule
\end{longtable}

Eight of the nine low-baseline subjects cluster in the 48-74\% upward
range. Bābur is the low-baseline outlier (source corpus 422K words,
partial pretraining exposure); he is the only subject whose
upward-crossing rate falls below 48\%, and his downward-crossing count
(4 of 39) is mid-range. Sunity Devee's 74.4\% upward rate is consistent
with her unusually low C5 baseline of 1.03 noted in \hyperref[the-cross-subject-gradient-and-its-per-question-mechanism]{§4.1}. Per-subject
downward-crossing rates stay at or below 15\% for every low-baseline
subject. Source: \texttt{scripts/\allowbreak{}compute\_\allowbreak{}anchor\_\allowbreak{}crossing.\allowbreak{}py} executed
against
\texttt{results/global\_\textless{}subject\textgreater{}/judgments\_v2.json}
and \texttt{results/hamerton/}.

\hypertarget{d.3-rubric-handling-validity-audit-redirected}{%
\subsection{D.3 Rubric-handling validity audit
(redirected)}\label{d.3-rubric-handling-validity-audit-redirected}}

The full report on rubric-handling limitations (refusal-anchor
ambiguity, length-correlation by condition, per-judge strictness,
per-response-model abstention behavior, memory-system effect on
abstention) is in the public repository at
\texttt{docs/\allowbreak{}supplementary/\allowbreak{}appendix\_\allowbreak{}D\_\allowbreak{}3\_\allowbreak{}rubric\_\allowbreak{}handling\_\allowbreak{}validity\_\allowbreak{}audit.\allowbreak{}md}.

\textbf{Headline results already reported in \hyperref[rubric-handling-limitations-post-hoc-validity-audit-1]{§4.6.7}:} Across 192
responses identified as refusals in the low-baseline band, 82.8\% scored
in the 1.0--1.5 anchor range as expected, but 9.4\% scored at or above
2.0 and 3.1\% at or above 3.0. Mean abstention score: 1.27. The \hyperref[rubric-handling-limitations-post-hoc-validity-audit-1]{§4.6.7}
caveat applies: per-response-model abstention behavior diverges (Sonnet
over-credits abstention at roughly twice Haiku's rate); memory-system
condition effects on the score-1 composition are decomposed in \hyperref[memory-system-composition]{§4.4}.
Audit script: \texttt{scripts/\allowbreak{}audit\_\allowbreak{}low\_\allowbreak{}end\_\allowbreak{}inflation.\allowbreak{}py}. Raw audit
output: \texttt{docs/\allowbreak{}research/\allowbreak{}rubric\_\allowbreak{}handling\_\allowbreak{}validity\_\allowbreak{}full.\allowbreak{}json}.

\hypertarget{d.4-per-judge-score-matrices-redirected}{%
\subsection{D.4 Per-judge score matrices
(redirected)}\label{d.4-per-judge-score-matrices-redirected}}

The full per-subject by per-judge by per-condition mean-score matrix for
all 14 main-study subjects across the 5 gradient conditions (C5, C2a,
C2c, C4, C4a) lives in the public repository at
\texttt{docs/\allowbreak{}supplementary/\allowbreak{}appendix\_\allowbreak{}D\_\allowbreak{}4\_\allowbreak{}per\_\allowbreak{}judge\_\allowbreak{}matrices.\allowbreak{}md}.
Per-subject raw per-judge data is at
\texttt{results/global\_\textless{}subject\textgreater{}/*\_judgments\_\textless{}judge\textgreater{}.json}
(and \texttt{results/hamerton/} for Hamerton).

\textbf{Headline results from the matrices already reported in the
paper:} directional agreement is tight (Spearman ρ 0.86 to 0.93 across
the 5-judge primary panel; \hyperref[inter-judge-agreement]{§3.3.4}) while absolute magnitude varies by
judge (Krippendorff α 0.659 on the 5-judge panel, 0.535 with Gemini
included; \hyperref[calibration]{§3.3.3} and \hyperref[judge-panel-sensitivity-5-judge-primary-vs-7-judge]{§4.6.2}). The Gemini judges show a systematic
+1-point inflation relative to the calibrated five primary judges,
decomposed in \hyperref[calibration]{§3.3.3}.

\hypertarget{d.5-example-verbatim-responses-at-each-rubric-anchor}{%
\subsection{D.5 Example verbatim responses at each rubric
anchor}\label{d.5-example-verbatim-responses-at-each-rubric-anchor}}

Verbatim-response examples at rubric anchors 1-5 for one representative
subject (Hamerton) are in \hyperref[scoring-rubric-with-calibrated-llm-judge-panel]{§3.3} as part of the rubric definition.
Examples at anchor crossings are developed in \hyperref[the-cross-subject-gradient-and-its-per-question-mechanism]{§4.1} (Examples A, B, C on
Ebers, Bernal Díaz, Seacole). Three illustrative paired (C5, C4a)
per-question excerpts for each of the 14 main-study subjects are
collected in Appendix E. The raw response JSON files
(\texttt{results/global\_\textless{}subject\textgreater{}/results\_v2.json})
contain every response verbatim alongside its 5-judge primary score.

\hypertarget{d.6-reasoning-trace-worked-example-sunity-devee-q4}{%
\subsection{D.6 Reasoning trace worked example: Sunity Devee
Q4}\label{d.6-reasoning-trace-worked-example-sunity-devee-q4}}

The conceptual four-step reasoning trace described in \hyperref[traceability-and-reasoning-traces]{§2.3} is walked
here on a single battery item. The example shows how a response in the
C2a condition (Specification only: the model receives the Behavioral
Specification and the question, with no retrieved facts and no source
corpus) decomposes into pattern statements, the facts that imply them,
and the source passages those facts were extracted from.

\begin{quote}
\textbf{Subject:} Sunity Devee

\textbf{Question:} \emph{``When her husband expresses disappointment
about missing an opportunity due to her concerns, does the narrator
typically reconsider her position?''}

\textbf{Held-out ground truth (full passage):} \emph{``The Maharajah was
disappointed, as he was anxious to go and see the old castle, and to
shoot, and I deprived him of both these pleasures.''}

\textbf{Response (C2a, excerpt):} \emph{``Based on the behavioral
specification, the answer is no, not typically, and not in the way the
question assumes. A2 (Spiritual Integrity Over Social Cost) and A5
(Relational Identity) create the relevant dynamic. A5 activates first:
relational identity means his disappointment matters deeply. A2 holds
firm: arguments from social consequence, including spousal
disappointment, will not register as reasons on their own. Rather than
reconsider the position itself, she is more likely to absorb the tension
without resolving it (P3 --- Tension Absorbed, Not Expressed) and
reframe the disappointment within a spiritual logic (A1 --- Divine
Primacy), while grieving the relational cost without reversing the
decision.''}

\textbf{Step 3 --- pattern statements invoked by the response:}

\begin{enumerate}
\def\labelenumi{\arabic{enumi}.}
\tightlist
\item
  \textbf{A5 (Relational Identity)} registers his disappointment as
  real, not dismissible.
\item
  \textbf{A2 (Spiritual Integrity Over Social Cost)} refuses social
  consequence as a reason to reconsider.
\item
  \textbf{P3 (Tension Absorbed, Not Expressed)} holds the conflict
  without resolving it in either direction.
\item
  \textbf{A1 (Divine Primacy)} reframes the disappointment within a
  spiritual logic.
\end{enumerate}

\textbf{Step 4 --- facts grounding the pattern statements:}

\begin{itemize}
\tightlist
\item
  \textbf{F-73:} \emph{``Sunity Devee's mother would never countenance
  anything her conscience told her was wrong.''} (grounds A2)
\item
  \textbf{F-414:} \emph{``Sunity Devee's father believed he acted as a
  public man guided by conscience and divine duty in accepting the
  marriage proposal.''} (corroborates A2 from a different relational
  direction; conscience-as-master-frame pattern reinforced across both
  parents)
\item
  Additional facts grounding A1, A5, and P3 are referenced in the
  specification's anchor and prediction files at
  \texttt{data/\allowbreak{}global\_\allowbreak{}subjects/\allowbreak{}sunity\_\allowbreak{}devee/\allowbreak{}spec/\allowbreak{}anchors.\allowbreak{}md} and
  \texttt{data/\allowbreak{}global\_\allowbreak{}subjects/\allowbreak{}sunity\_\allowbreak{}devee/\allowbreak{}spec/\allowbreak{}predictions.\allowbreak{}md};
  per-fact source-passage excerpts are in the same subject's
  \texttt{facts.json}.
\end{itemize}
\end{quote}

The user can audit any step: read the response, look up each cited
pattern statement by name, look up the facts that ground it, and read
the source passages those facts came from. If a fact misrepresents its
source, correcting the fact propagates through the Specification on
recomposition.

\begin{center}\rule{0.5\linewidth}{0.5pt}\end{center}

\hypertarget{appendix-e.-per-subject-worked-examples}{%
\section{Appendix E. Per-subject worked
examples}\label{appendix-e.-per-subject-worked-examples}}

The 14 per-subject worked examples (one per main-study subject) are in
\texttt{docs/\allowbreak{}supplementary/\allowbreak{}appendix\_\allowbreak{}E\_\allowbreak{}per\_\allowbreak{}subject\_\allowbreak{}worked\_\allowbreak{}examples.\allowbreak{}md}
in the public repository. Each subject has three illustrative paired
(C5, C4a) per-question response excerpts under the 5-judge primary
score, selected by the largest C4a minus C5 panel-mean Δ within each
subject. The excerpts are organized to illustrate the cross-anchor
interpretation rule (\hyperref[score-interpretation]{§3.3.1}) at the subject level and to make concrete
the multi-anchor crossings discussed in \hyperref[per-question-baseline-engagement-and-the-worked-rubric-example]{§4.1.1}. Per-subject baselines
(C5) range from 1.02 (Ebers) to 2.77 (Equiano); per-question Δ\_C4a
values in the supplementary file range from +0.40 to +4.00. Full
response artifacts are at
\texttt{results/\textless{}subject\textgreater{}/results\_v2.json} and
\texttt{results/\allowbreak{}hamerton/\allowbreak{}results.\allowbreak{}json}.

\begin{center}\rule{0.5\linewidth}{0.5pt}\end{center}

\hypertarget{appendix-f.-benchmark-scope-analysis}{%
\section{Appendix F. Benchmark Scope
Analysis}\label{appendix-f.-benchmark-scope-analysis}}

This appendix develops, benchmark by benchmark, the scope differences
summarized in \hyperref[prior-measurement-targets-and-the-gap-representational-accuracy-fills]{§2.1} between prior work on memory and personalization
benchmarks and what this paper measures. The point in each case is the
same: representational accuracy, operationalized as behavioral
prediction on held-out reasoning situations, is not what these
benchmarks evaluate. None of them is wrong on its own axis. None of them
is a substitute for the test in this paper.

\hypertarget{f.1-longmemeval}{%
\subsection{F.1 LongMemEval}\label{f.1-longmemeval}}

\textbf{Reference.} Wu et al., ICLR 2025~\citep{wu2025longmemeval}.

\textbf{Task.} Evaluate long-term memory in chat assistants across
multiple sessions. Five capability dimensions: single-session memory,
multi-session reasoning, temporal reasoning, knowledge updates, and
abstention.

\textbf{Scoring.} Question-answering accuracy, with held-out facts
embedded across session history and queried in a later session. Answers
are compared against ground-truth factual targets drawn from the same
session history the system ingested.

\textbf{Training / test protocol.} Conversation history is ingested; the
system is then queried with fact-recall questions whose answers are
present in the ingested history. The test is whether the memory system
can surface the correct facts at retrieval time.

\textbf{What it measures.} Fact recall across long context windows. A
secondary axis tests whether the system correctly abstains when the
answer is not in the conversation history.

\textbf{What it does not measure.} Whether the memory system's
representation of a specific person captures how that person reasons.
Every LongMemEval target is a fact that was literally said in the
conversation; no target is a held-out behavioral pattern.

\textbf{Published range.} Memory systems reported in the 68\% to 85\%
range depending on provider, model, and benchmark variant (cited in \hyperref[recall-is-not-interpretation.-interpretation-can-be-measured.]{§1.1}
and \hyperref[memory-systems-for-llm-agents]{§2.2}). Specific numbers per system are in the papers and vendor
reports.

\textbf{Relationship to this paper's test.} Orthogonal axis. Our battery
targets held-out behavioral patterns that were never literally said in
the training half of the corpus; every question is backward-designed to
answer only from patterns, not from retrievable content. A system that
ranks at the top of LongMemEval can still sit near the rubric floor on
our battery, and a system that ranks low on LongMemEval (for example,
Base Layer's retrieval substrate) can contribute on our battery through
the Specification rather than through retrieval.

\hypertarget{f.2-personagym}{%
\subsection{F.2 PersonaGym}\label{f.2-personagym}}

\textbf{Reference.} Samuel et al., Findings of EMNLP 2025~\citep{samuel2025personagym}.

\textbf{Task.} Evaluate persona fidelity in conversational agents. Given
a described persona, measure whether the model maintains that persona
across a conversation.

\textbf{Scoring.} Persona-consistency metrics over multi-turn
conversation. LLM-judge evaluation of whether the model's voice, stated
preferences, and surface-level claims remain consistent with the
described persona.

\textbf{Training / test protocol.} A persona is described (occupation,
background, preferences, mannerisms). The model is prompted to roleplay
the persona across a dialogue. Evaluation is whether the dialogue
responses remain internally consistent with the persona description.

\textbf{What it measures.} Persona presentation fidelity. Can the model
stay in character on the described dimensions.

\textbf{What it does not measure.} Whether the model accurately predicts
how the person described by the persona would respond to new situations.
A persona-fidelity system can maintain voice without ever accurately
anticipating decisions. A representationally accurate system can shift
voice (for example, from formal prose to casual conversational register)
while continuing to predict accurately on behavioral questions.

\textbf{Published best-number.} Top PersonaScore of 4.51 ± 0.08 on a 1-5
scale (GPT-4.5), out of 10 evaluated LLMs spanning 200 personas and
10,000 questions; bottom of the range was 3.64 ± 0.57 (Claude 3 Haiku).
Notably, GPT-4.1 and LLaMA-3-8b tied on PersonaScore despite a large
capability gap \citep{samuel2025personagym}. Scoring is on
persona-consistency metrics, not held-out behavioral prediction; these
numbers are not directly comparable to this paper's rubric means on the
1-5 behavioral-prediction scale.

\textbf{Relationship to this paper's test.} Both measure something that
is sometimes called ``personalization,'' but the axes are different.
PersonaGym is surface-presentation consistency; our battery is transfer
of the subject's interpretive patterns to unseen situations. Our rubric
does not credit voice-matching alone; it requires predicting the
held-out behavioral content. A response that maintains voice without
predicting the held-out pattern scores 2 (``wrong prediction'') under
our rubric.

\hypertarget{f.3-alpsbench}{%
\subsection{F.3 AlpsBench}\label{f.3-alpsbench}}

\textbf{Reference.} Xiao et al., 2026~\citep{xiao2026alpsbench}.

\textbf{Task.} Evaluate whether explicit memory mechanisms improve
preference-aligned and emotionally resonant responses in conversational
settings.

\textbf{Scoring.} Preference-alignment scoring and emotional-resonance
scoring on conversational responses, both LLM-judged against reference
targets derived from user preference data.

\textbf{Training / test protocol.} A conversational agent is seeded with
a user's preference history (via an explicit memory mechanism, or as a
baseline without one). The agent responds to new prompts. Responses are
scored on preference-alignment and emotional-resonance metrics.

\textbf{What it measures.} Whether explicit memory makes conversational
responses more aligned with stated preferences and more emotionally
appropriate.

\textbf{What it does not measure.} Whether memory mechanisms enable the
model to predict the user's behavior in unseen reasoning situations.
Preference alignment and behavioral prediction are related but distinct:
a system can match preferences on immediate response choices without
having a representation of the user's reasoning that transfers to
situations the system has never seen.

\textbf{Central finding.} AlpsBench's central empirical result is that
explicit memory mechanisms improve recall but do not guarantee more
preference-aligned or emotionally resonant responses. This is
independently arrived at and complementary to our own finding. They find
the gap in preference alignment; we find it in behavioral prediction.
Both point in the same direction: recall-solving is insufficient for
what memory is ultimately for.

\textbf{Relationship to this paper's test.} Adjacent. Same motivating
intuition (recall improvement does not transfer to downstream behavioral
properties), different downstream property measured.

\hypertarget{f.4-twin-2k}{%
\subsection{F.4 Twin-2K}\label{f.4-twin-2k}}

\textbf{Reference.} Toubia et al., 2025~\citep{toubia2025twin2k}.

\textbf{Task.} Behavioral prediction at scale. 2,058 participants each
answered a large-scale survey, and the system predicts each
participant's responses on held-out survey items given a persona
constructed from their other survey answers.

\textbf{Scoring.} Distance metric on Likert-scale items (numeric
distance between predicted response and actual response, aggregated per
participant and per item).

\textbf{Training / test protocol.} For each participant, a subset of
survey answers is used to author a persona. The persona is served to a
model as context. The model predicts the participant's answer on the
held-out survey items. Distance between predicted and actual response is
the score.

\textbf{What it measures.} Behavioral prediction on survey-response
interpolation. Does a machine-readable transcript of one half of a
participant's survey predict the other half.

\textbf{What it does not measure.} Behavioral prediction on open-ended
reasoning situations. Twin-2K's held-out items are additional Likert
responses from the same survey form; the test is interpolation across a
structured response distribution. Our held-out items are open-ended
behavioral predictions on unseen autobiographical passages; scoring is
via rubric on response content, not distance on a numeric scale.

\textbf{Relationship to this paper's test.} Closest prior work on the
behavioral-prediction axis. Three structural differences remain:

\begin{enumerate}
\def\labelenumi{\arabic{enumi}.}
\tightlist
\item
  \textbf{Task format.} Twin-2K: Likert interpolation. This paper:
  open-ended behavioral prediction with free-text answers.
\item
  \textbf{Persona construction.} Twin-2K: machine-readable transcript of
  the participant's own prior survey responses, served as-is. This
  paper: an authored Behavioral Specification composed of three
  interpretive layers plus a composed brief. The Twin-2K persona is raw
  data; our Specification is compressed interpretation.
\item
  \textbf{Held-out distance.} Twin-2K: the held-out items are drawn from
  the same structured survey instrument. This paper: the held-out items
  are drawn from autobiographical text the representation has never
  seen, in a different form than the training half (different chapters,
  different situations).
\end{enumerate}

Twin-2K measures whether a model can interpolate a person's survey
distribution from other survey responses. Our battery measures whether a
representation of how a person reasons transfers to new situations the
representation has never seen. Both are legitimate tests of behavioral
prediction; neither is a substitute for the other.

An earlier exploratory Base Layer run against Twin-2K's battery produced
positive results on that task format, but we do not report those numbers
as a formal benchmark comparison because the experiment used a prior
iteration of our pipeline, and the task targets are substantively
different (see \hyperref[prior-measurement-targets-and-the-gap-representational-accuracy-fills]{§2.1}).

\textbf{Published best-number.} Top individual-level accuracy of 71.72\%
on held-out survey items using a text-persona representation served to
GPT-4.1-mini \citep{toubia2025twin2k}. Human test-retest
reliability on the same instrument was 81.72\%, putting the top twin at
87.67\% of the human ceiling. Random-guess baseline was 59.17\%.
Aggregate-level replication: the Twin-2K twins reproduced results from 6
of 10 behavioral-economics experiments, with systematic divergences on
medical decision-making and political attitudes. The 71.72\% accuracy is
on Likert interpolation, which is a structurally different task from our
rubric-scored free-text behavioral prediction.

\hypertarget{f.5-locomo}{%
\subsection{F.5 LoCoMo}\label{f.5-locomo}}

\textbf{Reference.} Maharana et al., ACL 2024~\citep{maharana2024locomo}.

\textbf{Task.} Conversational memory quality over long multi-session
dialogues.

\textbf{Scoring.} Fact-recall questions over ingested dialogue history.
Similar scope to LongMemEval but focused on conversational-memory
substrates specifically.

\textbf{Training / test protocol.} A long multi-session conversation is
ingested; the memory system is queried on specific facts from earlier
sessions.

\textbf{What it measures.} Long-dialogue recall accuracy.

\textbf{What it does not measure.} Behavioral reasoning. LoCoMo targets
are literal recalls from session history.

\textbf{Published range.} LoCoMo paper baselines \citep{maharana2024locomo}: GPT-4-turbo 32.1\% overall, GPT-3.5-turbo 22.4\%,
GPT-3.5-turbo-16K 37.8\%, best RAG configuration 41.4\%; human
performance 87.9\%. Memory-system claims on LoCoMo, detailed in \hyperref[memory-systems-for-llm-agents]{§2.2}:
Mem0g variant 68.44 with GPT-4o-mini \citep{chhikara2025mem0}; Mem0 production algorithm 91.6 self-reported with
open-sourced evaluation harness; Letta 74.0 with GPT-4o-mini; earlier
Zep claim of 84 publicly disputed by Mem0 (see \hyperref[memory-systems-for-llm-agents]{§2.2} dispute note). The
methodology disagreement between vendors remains unresolved; \hyperref[memory-systems-for-llm-agents]{§2.2} treats
these single-number comparisons with explicit caution.

\textbf{Relationship to this paper's test.} The benchmark the four
memory systems (Zep, Letta, Mem0, Supermemory) compete on. \hyperref[memory-systems-for-llm-agents]{§2.2} uses
these results as context for the memory-system landscape. Our paper is
orthogonal: we do not evaluate memory systems on LoCoMo; we evaluate
their behavioral-prediction performance on held-out autobiographical
passages with and without the Behavioral Specification added.

\hypertarget{f.6-memos-and-related-systems-level-benchmarks}{%
\subsection{F.6 MemOS and related systems-level
benchmarks}\label{f.6-memos-and-related-systems-level-benchmarks}}

\textbf{Reference.} Systems-level memory benchmarking literature,
including MemOS and adjacent evaluations. See \hyperref[memory-systems-for-llm-agents]{§2.2} for the
memory-systems landscape.

\textbf{Task.} Evaluate memory-layer infrastructure choices (storage
substrate, retrieval algorithm, consistency properties) rather than
memory-quality outcomes.

\textbf{Scoring.} Varies. Typically: retrieval latency, throughput,
consistency guarantees, scalability benchmarks.

\textbf{What it measures.} Infrastructure properties.

\textbf{What it does not measure.} Representational accuracy, persona
fidelity, or preference alignment. Systems-level benchmarks do not
evaluate the quality of the representation the memory layer produces;
they evaluate the mechanics of how that representation is stored and
served.

\textbf{Relationship to this paper's test.} Different layer of the
stack. Our paper evaluates what gets stored and why; systems-level
benchmarks evaluate how well it is stored and served. Both matter for
deployed personal-AI systems. The Specification and the memory-layer
infrastructure compose: our \hyperref[memory-system-composition]{§4.4} Mem0 / Letta / Zep / Supermemory / Base
Layer results show the Specification adding on top of each
infrastructure choice, not replacing it.

\hypertarget{f.7-what-no-prior-benchmark-measures}{%
\subsection{F.7 What no prior benchmark
measures}\label{f.7-what-no-prior-benchmark-measures}}

Pulling the per-benchmark analysis together, the axis that
representational accuracy sits on is not covered by any prior benchmark:

\begin{enumerate}
\def\labelenumi{\arabic{enumi}.}
\tightlist
\item
  \textbf{Test data the system has not seen.} LongMemEval, PersonaGym,
  and LoCoMo target content the system has ingested. Twin-2K's held-out
  items are drawn from the same structured instrument. Our battery's
  held-out passages are from unseen chapters in a different narrative
  register than the training half.
\item
  \textbf{Open-ended behavioral prediction rather than structured-format
  scoring.} Twin-2K is the closest comparison; it is Likert-format
  rather than open-ended.
\item
  \textbf{Representation of how a person reasons, not what they said or
  prefer.} PersonaGym tests voice consistency; AlpsBench tests
  preference alignment; LongMemEval / LoCoMo tests fact recall. None
  tests transfer of interpretive patterns.
\end{enumerate}

This is the gap the paper's battery targets. The battery is not a
replacement for any of the above. It is a test of a different property:
whether a representation of a specific person enables a model that has
never seen the person's held-out reasoning to anticipate it accurately.

\hypertarget{f.8-persona-input-depth-comparison-across-benchmarks}{%
\subsection{F.8 Persona-input depth comparison across
benchmarks}\label{f.8-persona-input-depth-comparison-across-benchmarks}}

The \hyperref[prior-measurement-targets-and-the-gap-representational-accuracy-fills]{§2.1} footnote calls out that
Twin-2K's persona input is much deeper than PersonaGym's one-line
descriptor. This subsection collects persona-input depth across the
benchmarks named in \hyperref[prior-measurement-targets-and-the-gap-representational-accuracy-fills]{§2.1} and Appendix F so the comparison is concrete.
``Persona-input depth'' is the total token volume of the participant- or
subject-specific representation served to the model at inference time,
measured on the input that an evaluated system actually consumes.

\begin{longtable}[]{@{}
  >{\raggedright\arraybackslash}p{(\columnwidth - 6\tabcolsep) * \real{0.23}}
  >{\raggedright\arraybackslash}p{(\columnwidth - 6\tabcolsep) * \real{0.23}}
  >{\raggedleft\arraybackslash}p{(\columnwidth - 6\tabcolsep) * \real{0.31}}
  >{\raggedright\arraybackslash}p{(\columnwidth - 6\tabcolsep) * \real{0.23}}@{}}
\toprule
Benchmark & Persona-input form & Approximate input depth (tokens) &
Notes \\
\midrule
\endhead
\textbf{LongMemEval} (Wu et al., ICLR 2025) & Multi-session conversation
transcript ingested before query & varies, \textasciitilde thousands per
session × 5 sessions & The ``persona'' is the user's accumulated
conversational history. Depth depends on configured session count; the
standard \texttt{\_s} configuration runs \textasciitilde5 ingested
sessions before query. \\
\textbf{LoCoMo} (Maharana et al., ACL 2024) & Long multi-session
dialogue ingested before query & varies, \textasciitilde10K-100K
depending on session count & Similar to LongMemEval in form: ingested
dialogue is the ``persona.'' Single-session and multi-session test
variants. \\
\textbf{PersonaGym} (Samuel et al., Findings of EMNLP 2025) & One-line
descriptor (e.g., \emph{``You are a 45-year-old skeptical accountant
from Toronto''}) & ≈ 20--50 & Surface attributes only; no behavioral
depth. The benchmark is designed to test consistency with a thin
descriptor across multi-turn conversation. \\
\textbf{AlpsBench} (Xiao et al., 2026) & User profile of preferences and
prior emotional-context exchanges & varies, \textasciitilde hundreds to
low thousands & Profile is a concatenation of stated preferences and
selected prior dialogue snippets. Designed to test preference-aligned
response generation rather than reasoning transfer. \\
\textbf{Twin-2K} (Toubia et al., 2025) & Full survey-response transcript
or condensed summary & \texttt{persona\_text} ≈ 32,000;
\texttt{persona\_summary} ≈ 3,750 & Two persona-input variants are
reported in the released code. \texttt{persona\_text} is the
participant's full prior-question-and-answer transcript;
\texttt{persona\_summary} is a model-condensed form. \\
\textbf{Base Layer (this paper)} & Behavioral Specification (anchors /
core / predictions composed into a unified brief) & ≈ 7,000 &
Compressed-interpretation form. Per-subject specification sizes range
from ≈ 4,000 (Hamerton, smallest corpus) to ≈ 9,000 (longest-corpus
subjects). Compression ratios vs.~source corpus span 5× to 80×
(\hyperref[compression-structure-vs.-raw-text]{§4.2}). \\
\bottomrule
\end{longtable}

\textbf{Reading.} The benchmarks span roughly three orders of magnitude
in persona-input depth, from PersonaGym's \textasciitilde50 tokens to
Twin-2K \texttt{persona\_text}'s \textasciitilde32,000. Two structural
points fall out of the comparison.

First, persona-input depth alone does not predict what a benchmark
measures. Twin-2K's full transcript is far deeper than this paper's
specification, but the task is Likert interpolation across the same
survey instrument; what is being measured is whether a model can
complete a participant's response distribution, not whether it captures
their interpretive patterns on situations the representation has never
seen.

Second, the form of the persona input matters more than the size.
PersonaGym's one-line descriptor and Twin-2K's full transcript are at
opposite ends of the depth scale, and both are ``raw'' inputs in
different senses (a descriptor of surface attributes versus a serialized
record of prior responses). The Behavioral Specification at
\textasciitilde7,000 tokens sits in the middle of the depth range but is
structurally distinct: it is an authored compression of how the subject
reasons, not a transcript or a descriptor. The compression-curve
evidence in \hyperref[compression-structure-vs.-raw-text]{§4.2} shows that this structural distinction is what makes
the form behave the way it does in cross-format behavioral prediction.

Sources: footnote-cited papers and the released code or schema where
applicable. PersonaGym descriptor length is from the canonical example
in Samuel et al.~\hyperref[operationalizing-representational-accuracy-via-the-behavioral-specification]{§3.1}; AlpsBench profile composition is from Xiao et
al.~\hyperref[experimental-conditions]{§3.2}; Twin-2K depth values are from \texttt{persona\_text} and
\texttt{persona\_summary} columns in the released dataset; Base Layer
specification depth is per-subject from
\texttt{data/global\_subjects/\textless{}subject\textgreater{}/spec/brief.md}.

\begin{center}\rule{0.5\linewidth}{0.5pt}\end{center}

\hypertarget{appendix-g.-letta-stateful-agent-exploratory-case-study-full}{%
\section{Appendix G. Letta Stateful-Agent: Exploratory Case Study
(full)}\label{appendix-g.-letta-stateful-agent-exploratory-case-study-full}}

This appendix holds the full case study summarized in \hyperref[exploratory-case-study-letta-stateful-agent-n3-post-hoc]{§4.5}. N=3 subjects
(Hamerton, Ebers, Bābur), one Letta version, one response model (Claude
Haiku 4.5), 40 questions per subject. We do not treat the result as a
replication or a headline finding.

\hypertarget{g.1-architectural-difference-letta-memory-block-vs.-behavioral-specification}{%
\subsection{G.1 Architectural difference: Letta memory block
vs.~Behavioral
Specification}\label{g.1-architectural-difference-letta-memory-block-vs.-behavioral-specification}}

The two systems produce structurally different artifacts from the same
source corpus.

\textbf{Letta's memory block} is text the agent has written and
rewritten during ingestion: a mix of verbatim source sentences,
paraphrased restatements, and short synthesis notes the agent generated
as it processed each turn. The block grows with the corpus and is
rewritten in place when it approaches the ingestion ceiling (around 333K
characters per the Letta API). Content shape is whatever the agent
decides during ingestion; size scales with corpus length up to that
limit.

\textbf{A Behavioral Specification} is a fixed-shape document produced
by the five-step pipeline detailed in \hyperref[pipeline-for-the-behavioral-specification]{§3.7}: import the corpus, extract
structured predicates over a 46-predicate controlled vocabulary, embed
each predicate for provenance tracing, author three layers, and compose
into a unified brief. The extracted predicates are constrained
subject-predicate-object triples (for example, \emph{values X, avoids Y,
decides based on Z}). The three layers each play a different role.
\textbf{Anchors} hold 8--12 axiomatic statements per subject distilled
from the predicates (e.g., A1 Intimate Authority, A2 Documented Dignity
in \hyperref[case-study-cross-system-refusal-on-keckley-q21]{§4.4.4}). \textbf{Core} holds the cognitive patterns and value
tensions that govern reasoning under the anchors. \textbf{Predictions}
hold IF/THEN templates that connect the patterns to specific behavioral
situations. The pipeline produces a deterministic full-stack Behavioral
Specification of approximately 7,000 tokens (\textasciitilde37,000
characters) regardless of source corpus length.

\textbf{Trade-off.} Letta retains source-text texture (voice,
vocabulary, syntax) at the cost of compression and mixes source material
with agent-generated synthesis. A Behavioral Specification compresses
aggressively into structured predicates at the cost of source-text
texture, but maintains a deterministic schema the response model can
read consistently across subjects.

\hypertarget{g.2-headline-result-5-judge-primary}{%
\subsection{G.2 Headline result (5-judge
primary)}\label{g.2-headline-result-5-judge-primary}}

\begin{longtable}[]{@{}
  >{\raggedright\arraybackslash}p{(\columnwidth - 14\tabcolsep) * \real{0.10}}
  >{\raggedleft\arraybackslash}p{(\columnwidth - 14\tabcolsep) * \real{0.13}}
  >{\raggedleft\arraybackslash}p{(\columnwidth - 14\tabcolsep) * \real{0.13}}
  >{\raggedleft\arraybackslash}p{(\columnwidth - 14\tabcolsep) * \real{0.13}}
  >{\raggedleft\arraybackslash}p{(\columnwidth - 14\tabcolsep) * \real{0.13}}
  >{\raggedleft\arraybackslash}p{(\columnwidth - 14\tabcolsep) * \real{0.13}}
  >{\raggedleft\arraybackslash}p{(\columnwidth - 14\tabcolsep) * \real{0.13}}
  >{\raggedleft\arraybackslash}p{(\columnwidth - 14\tabcolsep) * \real{0.13}}@{}}
\toprule
Subject & Corpus (words) & Letta score & Spec score (full-stack) & Δ
(Letta − Spec) & Letta block (chars) & Spec (chars) & Letta : Spec
size \\
\midrule
\endhead
Hamerton & 25K & 3.10 & 2.83 & +0.27 & 22.5K & 34.6K & 0.7× \\
Ebers & 48K & 2.76 & 1.56 & +1.21 & 68.4K & 39.7K & 1.7× \\
Bābur & 223K & 2.42 & 2.04 & +0.38 & 335.3K & 37.1K & 9.0× \\
\bottomrule
\end{longtable}

\emph{The Spec score column reports Base Layer's full-stack Behavioral
Specification (anchors + core + predictions + brief), the same artifact
class used in the main-study gradient; per-subject sizes 34.6K / 39.7K /
37.1K characters. Raw data at
\texttt{docs/\allowbreak{}research/\allowbreak{}\_\allowbreak{}letta\_\allowbreak{}rerun/\allowbreak{}fullstack\_\allowbreak{}named/}. Hamerton's
score is the full-stack specification scored with the consistent
short-form judge prompt; Ebers and Bābur are full-stack regenerations on
the Letta battery.}

Letta's self-edited memory block scores higher than Base Layer's
full-stack Behavioral Specification on all 3 subjects, with the gap
largest at the mid-corpus subject (Ebers, Δ +1.21) and smaller at both
endpoints. With three data points we cannot distinguish among the
possible interpretations: a corpus-size band where the self-edited block
is most effective, degradation as the block grows beyond an
architectural sweet spot, or insufficient interpretive content when the
corpus is small. Across all three subjects, both representations land
well above the retrieval-only baseline at matched response model.

\hypertarget{g.3-the-block-grows-with-the-corpus-then-breaks-down}{%
\subsection{G.3 The block grows with the corpus, then breaks
down}\label{g.3-the-block-grows-with-the-corpus-then-breaks-down}}

Letta's block scales with corpus length: 22.5K characters at Hamerton
(smaller than the Spec's 37K), 68.4K at Ebers, 335.3K at Bābur (about 9×
the Spec). The Bābur block hits Letta's ingestion ceiling near 333K
characters, and the agent starts rewriting content it has already
written rather than compressing it. The result is heavy duplication at
the ceiling: 25.4\% of Bābur's sentences are verbatim repeats, and
56.1\% have a near-paraphrase elsewhere in the block. Hamerton and Ebers
show 0\% and minor duplication respectively; the architectural ceiling
produces both exact and near-paraphrase repeats only when the corpus is
large enough to push against it.\footnote{Semantic-similarity
  duplication via
  \texttt{scripts/\allowbreak{}analyze\_\allowbreak{}letta\_\allowbreak{}semantic\_\allowbreak{}duplication.\allowbreak{}py}
  (MiniLM-L6-v2 sentence embeddings). At Bābur: 56.1\% of sentences have
  a near-paraphrase at cosine ≥ 0.85; 35.2\% at ≥ 0.95. Ebers: 3.3\% at
  ≥ 0.85, 0.5\% at ≥ 0.95. Hamerton: 0\% at any threshold above 0.80.
  Full per-threshold data at
  \texttt{docs/\allowbreak{}research/\allowbreak{}letta\_\allowbreak{}semantic\_\allowbreak{}duplication\_\allowbreak{}20260501.\allowbreak{}json}.}

\hypertarget{g.4-what-explains-lettas-lift}{%
\subsection{G.4 What explains Letta's
lift}\label{g.4-what-explains-lettas-lift}}

Two things, neither of which is surface-text leakage. The leakage
hypothesis was tested directly: the held-out battery was authored from
the same source autobiographies that Letta ingested, so verbatim phrases
could in principle leak from block to response. They do not. Five-word
verbatim overlap with the held-out passage is 0\% on every one of 119
questions for both Letta and Spec. The score-delta correlation with
surface-text similarity is essentially zero (Pearson −0.046).\footnote{3-gram
  overlap is tiny on both sides (Letta 0.05--0.08\%, Spec 0.02--0.06\%),
  dominated by proper-name sequences. Pre-registered strict and relaxed
  leakage signatures: 0 of 119 questions. Full analysis at
  \texttt{docs/\allowbreak{}research/\allowbreak{}letta\_\allowbreak{}vs\_\allowbreak{}spec\_\allowbreak{}leakage\_\allowbreak{}analysis\_\allowbreak{}20260507.\allowbreak{}md}.}
What does explain Letta's lift is that its block keeps the corpus's
named entities (people, places, institutions) that the Spec compresses
away by design, and Letta speaks in committed predictions where the Spec
hedges when its behavioral scaffold cannot reach a specific question.
When Letta's committed answer matches the held-out, it scores high; when
it doesn't, the Spec's hedge at least lands closer to the right
direction. Ebers q30 is the canonical entity-grounding case: the
held-out turns on Frederick William IV at the Keilhau Institute; Letta
names him and scores 4.0; the Spec reaches for behavioral anchors that
are structurally adjacent but referentially different and scores 1.8.

\hypertarget{g.5-pattern-4-when-confidence-backfires}{%
\subsection{G.5 Pattern 4: when confidence
backfires}\label{g.5-pattern-4-when-confidence-backfires}}

The same mechanism that lifts Letta's score on entity-rich questions
hurts it on principle-driven questions. Where the held-out resolution
turns on an underlying principle the entity-rich text does not surface,
Letta commits confidently to the wrong direction; the Spec's hedge keeps
the answer closer to the held-out. Three Hamerton decision questions
illustrate the inversion at the cleanest scale: q27 (self-publish
poetry, Spec +2.4), q29 (Lancashire exhibition, Spec +2.0), q51 (distant
school vs.~dying guardian, Spec +3.0). Pattern 4 is the
Letta-architecture analog of \hyperref[where-the-spec-helps-where-it-hurts-and-which-question-types-route-to-each]{§4.4.3} Pattern 3 (Spec-induced refusal):
each architecture has a structural failure mode born of its
representational choice. Letta's commit-without-value-anchor produces
confident inversions on principle-driven questions; the Spec's
value-anchor-without-context produces principled refusals the rubric
penalizes. Letta also abstains more often than the Spec at the two
larger subjects (Bābur 17.9\% vs 0\%; Ebers 10.3\% vs 5.1\%) and less at
the small-corpus Hamerton (7.7\% vs 10.3\%); the Bābur spike co-occurs
with the duplication ceiling, suggesting the agent recognizes block
degradation and refuses to over-extend. Whether Letta's
content-confidence advantage comes from the corpus content the block
retained (real interpretive work) or from the agent's general tendency
to commit regardless of grounding is the open question; a
named-entity-vs-axiom-grounding ablation is flagged in \hyperref[stateful-agent-implementations-and-temporal-drift-tracking]{§7.5}. Per-subject
abstention decomposition at
\texttt{docs/\allowbreak{}reviews/\allowbreak{}letta\_\allowbreak{}vs\_\allowbreak{}spec\_\allowbreak{}abstention\_\allowbreak{}20260507.\allowbreak{}md}.

\hypertarget{g.6-caveats-and-naming-asymmetry}{%
\subsection{G.6 Caveats and naming
asymmetry}\label{g.6-caveats-and-naming-asymmetry}}

The \hyperref[exploratory-case-study-letta-stateful-agent-n3-post-hoc]{§4.5} / Appendix G main result table compares Letta's named,
self-edited block against Base Layer's full-stack Behavioral
Specification (anchors + core + predictions + brief), the same artifact
class used in the main-study gradient. Letta ingests the named corpus;
Base Layer's pipeline anonymizes the corpus during extraction by design,
so the named-vs-anonymized comparison is asymmetric on referential
vocabulary. A named-entity-controlled Base Layer variant is flagged as
future work in \hyperref[stateful-agent-implementations-and-temporal-drift-tracking]{§7.5}. Multi-subject replication across the full
14-subject gradient is the highest-priority external falsification of
the Letta architectural-ceiling finding.

\hypertarget{g.7-exploratory-stacking-the-letta-block-with-the-behavioral-specification}{%
\subsection{G.7 Exploratory: stacking the Letta block with the
Behavioral
Specification}\label{g.7-exploratory-stacking-the-letta-block-with-the-behavioral-specification}}

The \hyperref[exploratory-case-study-letta-stateful-agent-n3-post-hoc]{§4.5} comparison treats Letta's self-edited block and the Behavioral
Specification as alternatives. A follow-up probe served both at once,
concatenated into a single Haiku 4.5 system prompt, on the same three
subjects and the same Letta battery. The question is whether the two
interpretive representations conflict or compose.

\begin{longtable}[]{@{}lrrr@{}}
\toprule
Subject & Stacked (block + Spec) & Letta block alone & BL full-stack
Spec alone \\
\midrule
\endhead
Hamerton & 3.17 & 3.10 & 2.83 \\
Ebers & 2.84 & 2.76 & 1.56 \\
Bābur & 2.52 & 2.42 & 2.04 \\
\bottomrule
\end{longtable}

Stacking lands above both representations alone on all three subjects:
it beats the Letta block by +0.07 / +0.08 / +0.10 and the full-stack
Specification by +0.34 / +1.28 / +0.48. The longer combined context does
not dilute either signal, including on Bābur, whose Letta block alone is
already \textasciitilde335K characters with heavy duplication. The two
representations are complementary rather than redundant.

The uplift over the Letta block alone is small, and the comparison is
not panel-matched. The stacked condition was scored by a 4-judge panel
(Haiku, Sonnet, Opus, GPT-4o) because GPT-5.4 returned errors on every
call, the same failure mode the main Letta rerun encountered and
patched; the comparison columns are 5-judge primary means. The
stacked-vs-Letta gap is therefore within what a panel-composition
difference could account for. The defensible read is that stacking does
not hurt and modestly helps, not that it is clearly additive. This is an
exploratory N=3 post-hoc probe, not a headline finding; a GPT-5.4
backfill would make it a panel-matched 5-judge result. Raw data and
scripts:
\texttt{docs/\allowbreak{}research/\allowbreak{}v12\_\allowbreak{}1\_\allowbreak{}letta\_\allowbreak{}stacked\_\allowbreak{}experiment\_\allowbreak{}20260513.\allowbreak{}md},
\texttt{data/\allowbreak{}experiments/\allowbreak{}memory\_\allowbreak{}systems/\allowbreak{}run\_\allowbreak{}letta\_\allowbreak{}stacked\_\allowbreak{}spec.\allowbreak{}py},
and \texttt{judge\_\allowbreak{}letta\_\allowbreak{}stacked\_\allowbreak{}spec.\allowbreak{}py}.

\hypertarget{appendix-h.-glossary}{%
\section{Appendix H. Glossary}\label{appendix-h.-glossary}}

Defined terms used as terms of art throughout the paper. Entries that
share substantive meaning are cross-referenced; for each named artifact,
\textbf{the acceptable short forms and capitalization conventions are
explicit}.

\begin{description}
\tightlist
\item[5-judge primary panel]
The locked judge aggregation for headline numbers. Aggregation rule:
per-judge per-question score → per-judge per-subject mean → panel mean
across \{Haiku 4.5, Sonnet 4.6, Opus 4.6, GPT-4o, GPT-5.4\}. See \hyperref[calibration]{§3.3.3}.
\item[7-judge sensitivity panel]
The 5-judge primary plus Gemini 2.5 Flash and Gemini 2.5 Pro, reported
as a sensitivity check. See \hyperref[calibration]{§3.3.3}.
\item[All Facts (C4)]
The condition where the response model is served the full extracted fact
set with no Behavioral Specification. The fact-only baseline for
isolating Spec contribution from fact contribution. See \hyperref[experimental-conditions]{§3.2},
\textbf{Fact set}.
\item[Anchors / Core / Predictions]
The three layered artifacts comprising a Behavioral Specification.
Anchors: short axiomatic claims about how the subject reasons. Core:
connects anchors into coherent reasoning patterns. Predictions: derives
forward-looking decisions from the core. The unified brief sits above
all three as the composed integration. See \hyperref[pipeline-for-the-behavioral-specification]{§3.7}, \textbf{Unified brief}.
\item[Anchor crossing]
A change in 5-judge primary mean that moves the score across an integer
rubric anchor (1, 2, 3, 4, or 5) between two conditions. Distinguished
from a within-anchor shift, which stays inside a single integer band and
is a weaker categorical claim. The strongest single-question signal in
the rubric. See \hyperref[score-interpretation]{§3.3.1}, \textbf{Categorical shift}, \textbf{Cross-anchor
interpretation rule}, \textbf{Multi-anchor crossing}.
\item[Behavioral prediction]
The operational test for representational accuracy. Given a situation
drawn from text the model has never seen, the model generates how the
subject would respond; the response is scored against the subject's own
verbatim response on a 1-5 interpretive rubric. See \hyperref[recall-is-not-interpretation.-interpretation-can-be-measured.]{§1.1}, \hyperref[scoring-rubric-with-calibrated-llm-judge-panel]{§3.3}.
\item[Behavioral Specification]
A static document of approximately 7,000 tokens that extracts and
encodes a person's behavioral patterns. Composed of three layered
artifacts (anchors, core, predictions) plus a unified brief. Layered
above memory-system retrieval as an interpretive structure.
\textbf{Acceptable short forms (after first introduction):}
\emph{Specification}, \emph{Spec}, \emph{the Spec}.
\textbf{Capitalization rule:} when referring to the artifact, always
Title Case (Behavioral Specification) or capitalized (Specification,
Spec). Lowercase \emph{specification} or \emph{spec} should not be used
to refer to the artifact in this paper. See \hyperref[recall-is-not-interpretation.-interpretation-can-be-measured.]{§1.1}, \hyperref[pipeline-for-the-behavioral-specification]{§3.7}, \textbf{Unified
brief}, \textbf{Interpretive layer}.
\item[Categorical shift]
A change in the model's response that crosses one or more integer rubric
anchors when conditions change, marking a qualitative change in the kind
of response produced (e.g., from refusal at anchor 1 to substantive
prediction at anchor 4). The unit of effect this paper emphasizes
alongside raw rubric-point gain. See \textbf{Anchor crossing},
\textbf{Cross-anchor interpretation rule}.
\item[Compression / Compression ratio]
The ratio of the Behavioral Specification's predictive performance to
that of the full source corpus, divided by the ratio of their context
costs. The Specification recovers roughly 75\% of corpus predictive
performance at \textasciitilde25× less context. See \hyperref[compression-structure-vs.-raw-text]{§4.2}, \hyperref[what-we-found]{§1.3}.
\item[Condition codes (C1-C9)]
Letter codes for the experimental conditions used throughout the paper.
\textbf{C5:} No-Context Baseline. \textbf{C2a:} Specification only, no
facts. \textbf{C2c:} Wrong-Spec (a mismatched Behavioral Specification
served in place of the correct one). \textbf{C4:} All Facts only, no
Specification. \textbf{C4a:} Facts + Specification (the full pipeline).
\textbf{C3:} Retrieval + Specification (memory-system layering).
\textbf{C1:} Retrieval only. \textbf{C9:} Raw corpus + Specification.
Full table of condition definitions in \hyperref[experimental-conditions]{§3.2}.
\item[Cross-anchor interpretation rule]
A fractional delta between two conditions that crosses an integer rubric
anchor (1, 2, 3, 4, or 5) reflects a categorical shift in the underlying
response distribution. A delta that stays inside a single integer anchor
is a within-category shift and a weaker claim. See \hyperref[score-interpretation]{§3.3.1},
\textbf{Anchor crossing}, \textbf{Multi-anchor crossing}.
\item[Evidentiary bar]
An implicit threshold the response model appears to apply before
committing to a behavioral prediction: where the served context supplies
grounding evidence that clears the bar, the model produces a substantive
answer; where it does not, the model abstains. Visible in the
cross-system patterns in \hyperref[where-the-spec-helps-where-it-hurts-and-which-question-types-route-to-each]{§4.4.3} and \hyperref[case-study-cross-system-refusal-on-keckley-q21]{§4.4.4}. The matched Behavioral
Specification can both supply the interpretive scaffolding that clears
the bar (Pattern 1, \hyperref[where-the-spec-helps-where-it-hurts-and-which-question-types-route-to-each]{§4.4.3}) and raise the bar in a subject-consistent
way (Pattern 3 and the Keckley case study in \hyperref[case-study-cross-system-refusal-on-keckley-q21]{§4.4.4}). See \hyperref[what-we-found]{§1.3}, \hyperref[where-the-spec-helps-where-it-hurts-and-which-question-types-route-to-each]{§4.4.3},
\hyperref[case-study-cross-system-refusal-on-keckley-q21]{§4.4.4}.
\item[Fact set]
The extracted structured triples (subject-predicate-object) representing
a person's documented facts, used as the input to the C4 / All Facts
condition. The extraction step of the pipeline produces this artifact
from the raw corpus. See \hyperref[pipeline-for-the-behavioral-specification]{§3.7}, \textbf{All Facts}.
\item[Held-out passage]
A passage drawn from the holdout half of a subject's corpus (the half
the model has never seen during Specification authoring). Held-out
passages serve as the verbatim ground-truth text against which
response-model predictions are scored. The 50/50 train/holdout split is
per-subject. See \hyperref[scoring-rubric-with-calibrated-llm-judge-panel]{§3.3}, \textbf{Behavioral prediction}.
\item[Hedging / hedging reduction]
A response in which the model declines to commit to a substantive
prediction, either with explicit refusal language (``I cannot confirm'')
or with low-confidence qualification. The matched Behavioral
Specification collapses baseline hedging from approximately 41\% to near
zero under the correct-Spec conditions; the wrong-Spec condition
produces hedging at approximately 60\%. See \hyperref[mechanism-correct-content-not-format]{§4.3}, \textbf{Refusal
(abstention)}.
\item[High-baseline subject]
A subject whose No-Context Baseline score (C5) is above 3.0 on the 1-5
rubric, indicating substantial pretraining representation of that
person. Franklin (C5 = 3.77) is the high-baseline reference in this
paper; he is not included in any N=14 inferential statistic. Operational
threshold: C5 \textgreater{} 3.0. See \hyperref[the-gradient-at-the-high-baseline-end-franklin-reference]{§4.1.2}, \hyperref[pretraining-coverage-variance-high-vs-low-baseline]{§3.4.1},
\textbf{Low-baseline subject}, \textbf{Mid-baseline subject}.
\item[Interpretation]
In this paper, the human-side property: the way a specific person
processes facts and experiences into judgments, decisions, and
reactions. The property the Behavioral Specification is designed to
mirror. See \hyperref[recall-is-not-interpretation.-interpretation-can-be-measured.]{§1.1}, \textbf{Interpretive frame}, \textbf{Interpretive
layer}.
\item[Interpretive frame]
The cognitive lens through which a specific person interprets
situations. The human-side property. Distinguished from
\textbf{Interpretive layer}, which is the AI-side artifact (the
Behavioral Specification) that mirrors a person's interpretive frame.
\item[Interpretive layer]
The Behavioral Specification's structural function: an external, encoded
interpretation that sits between facts and the response model and
supplies the interpretive structure that facts alone do not carry. Used
throughout the paper as both a structural and operational term. See
\hyperref[recall-is-not-interpretation.-interpretation-can-be-measured.]{§1.1}, \hyperref[discussion]{§5}, \textbf{Behavioral Specification}.
\item[Interpretive structure]
The arrangement of behavioral anchors, core patterns, and predictions
that encodes a person's interpretive frame in the Specification
document. The internal organization of the \textbf{Interpretive layer}.
\item[Interpretive supply / over-theorization / principled refusal
(Patterns 1 / 2 / 3)]
Three patterns of Specification-retrieval interaction observed in
\hyperref[where-the-spec-helps-where-it-hurts-and-which-question-types-route-to-each]{§4.4.3}. \textbf{Pattern 1 (interpretive supply):} retrieval
underdetermines the answer; the Specification supplies interpretive
scaffolding and the response improves. \textbf{Pattern 2
(over-theorization):} retrieval already determines the answer; the
Specification adds incorrect generalization and the response degrades.
\textbf{Pattern 3 (principled refusal):} the Specification's axioms
trigger refusal where retrieval would have produced a substantive
answer. The aggregate Δ on any memory system is the balance of these
three patterns across its question battery. A fourth pattern,
\textbf{Pattern 4 (confidence backfire)}, is specific to the Letta
stateful-agent architecture and is developed in Appendix G.5.
\item[Low-baseline subject]
A subject whose No-Context Baseline score (C5) is at or below 2.0 on the
1-5 rubric, indicating that the response model has insignificant
pretraining representation of the person. Nine of the 14 main-study
subjects are low-baseline. The \textbf{population of relevance} for AI
personalization. See \hyperref[subjects]{§3.4}, \hyperref[the-cross-subject-gradient-and-its-per-question-mechanism]{§4.1}.
\item[Memory-system layering]
Serving the Behavioral Specification as additional context on top of a
commercial memory system's retrieved output (Mem0, Letta, Zep,
Supermemory). The Specification produces a net-positive aggregate Δ on
three of four commercial systems tested, with per-question patterns
varying by retrieval architecture. See \hyperref[memory-system-composition]{§4.4}, \textbf{Interpretive supply
/ over-theorization / principled refusal}.
\item[Mid-baseline subject]
A subject whose No-Context Baseline score (C5) is between 2.0 and 3.0 on
the 1-5 rubric. Per-subject Δ estimates for mid-baseline subjects carry
sign uncertainty within the pipeline run-to-run variance band; see \hyperref[pipeline-and-specification-stability]{§6.3}
for the Augustine sign-flip disclosure. See \textbf{Low-baseline
subject}, \textbf{High-baseline subject}.
\item[Multi-anchor crossing]
A single question whose 5-judge primary mean shifts across two or more
integer rubric anchors when the condition changes. Crossings can span
two anchors (e.g., 1 → 3) or, more rarely, three anchors (e.g., 1 → 4 or
2 → 5). The strongest categorical signal the rubric detects. See \hyperref[score-interpretation]{§3.3.1},
\hyperref[per-question-improvement-rate]{§4.2.1}, \textbf{Anchor crossing}, \textbf{Categorical shift}.
\item[No-Context Baseline (C5)]
The condition in which the response model is given no external
information about the subject. Serves as the observable proxy for
pretraining coverage and as the per-subject baseline against which all
Spec-effect Δ's are measured. See \hyperref[experimental-conditions]{§3.2}, \hyperref[subjects]{§3.4}, \textbf{Condition codes}.
\item[Per-subject C4a mean (≈2.44)]
The cross-subject mean of per-subject C4a scores on the 14-subject main
study with the current question mix. Used as a math note inside the
gradient explanation (\hyperref[the-cross-subject-gradient-and-its-per-question-mechanism]{§4.1}) to clarify why the change-score slope
(−0.96) is partly mechanical: the level-regression slope of C4a on C5 is
+0.04 (essentially flat). NOT a per-response ceiling --- per-question
variance is wide (\hyperref[per-question-baseline-engagement-and-the-worked-rubric-example]{§4.1.1} multi-anchor crossings range across the 1-5
scale). See \hyperref[the-cross-subject-gradient-and-its-per-question-mechanism]{§4.1}, \hyperref[battery-composition-sensitivity]{§4.6.3}, Appendix B.7.
\item[Personalization (this paper's sense)]
Representing the interpretive layer that sits beneath stated preferences
and biographical facts: how a specific person organizes experience, what
they treat as evidence, what reasoning patterns they apply across new
situations. Distinguished from surface-level responsiveness to stated
preferences (dietary restrictions, communication style) or stored facts
about the user (location, occupation, history), which are downstream
artifacts of the interpretive layer rather than the layer itself. See \hyperref[prior-work-industry-benchmarks-the-fifth-target]{§2}
lede.
\item[Population of relevance]
The population the paper's findings are designed to address:
low-baseline subjects whose interpretive patterns are not already
represented in pretraining. The paper argues this is the typical AI
user, since most users' reasoning patterns are not in any public
training corpus regardless of how much demographic information about
them is. See \hyperref[what-this-implies]{§1.4}, \hyperref[subjects]{§3.4}, \hyperref[why-the-gradient-is-the-load-bearing-finding]{§5.2}, \textbf{Low-baseline subject}.
\item[Refusal (abstention)]
A response in which the model declines to predict because the available
context does not support a prediction. Distinct from a substantively
wrong prediction; under the current rubric both score at the lowest
anchor (1). Detection in the validity-audit script uses phrase patterns
(``no specific information,'' ``I cannot confirm,'' ``would need
additional context''). See \hyperref[rubric-handling-limitations-post-hoc-validity-audit]{§3.3.6}, \textbf{Hedging}.
\item[Representational accuracy]
The AI-side property: how faithfully a model's internal representation
of a specific person captures that person's interpretive patterns.
Operationalized via behavioral prediction on held-out reasoning
situations. See \hyperref[recall-is-not-interpretation.-interpretation-can-be-measured.]{§1.1}, \hyperref[operationalizing-representational-accuracy-via-the-behavioral-specification]{§3.1}, \textbf{Behavioral prediction}.
\item[Specification-effect claim]
When a Behavioral Specification is served to the response model as
context, the model's responses shift in the direction of the subject's
demonstrated behavioral patterns, and that shift registers as a measured
increase in representational accuracy against held-out passages from the
same subject. The claim is directional, not a claim of new model
capability or absolute correctness. See \hyperref[inter-judge-agreement]{§3.3.4}.
\item[Tier 1 / Tier 2]
Tier 1 is the main study: Haiku 4.5 response model across all 14
subjects, every condition, Haiku-generated batteries. Tier 2 is the
cross-provider directional probe: Sonnet 4.6 and Gemini 2.5 Pro response
models on 3 subjects (Ebers, Yung Wing, Zitkala-Ša) with
GPT-5.4-regenerated batteries. See \hyperref[response-models]{§3.6}, \hyperref[cross-provider-response-generation-tier-2-replication]{§4.6.1}.
\item[Unified brief]
The composed prose document that integrates the three Behavioral
Specification layers (anchors, core, predictions) into a continuous
synthesis. Approximately 5,000 words; serves as both the integration
layer that weaves the three layers together and a coherent first pass
for human readers. The Behavioral Specification served to the response
model in C2a / C4a / C3 conditions is the three layers plus the unified
brief together. See \hyperref[pipeline-for-the-behavioral-specification]{§3.7}, \textbf{Behavioral Specification},
\textbf{Anchors / Core / Predictions}.
\item[Wrong-Spec control]
A deliberately mismatched Behavioral Specification served in place of
the correct one. Two variants: \textbf{v1} (adversarial fixed
derangement maximizing cultural and temporal distance; aggregate Δ
−0.25) and \textbf{v2} (seed-fixed random derangement; aggregate Δ
+0.15). See \hyperref[what-we-found]{§1.3}, \hyperref[experimental-conditions]{§3.2}, \hyperref[mechanism-correct-content-not-format]{§4.3}.
\end{description}

\end{document}